\documentclass{article} 
\usepackage{iclr2025_conference,times}

\usepackage{amsmath,amsfonts,bm}

\def\eqref#1{equation~\ref{#1}}

\def\1{\bm{1}}

\DeclareMathAlphabet{\mathsfit}{\encodingdefault}{\sfdefault}{m}{sl}
\SetMathAlphabet{\mathsfit}{bold}{\encodingdefault}{\sfdefault}{bx}{n}

\usepackage{hyperref}
\usepackage{url}

\usepackage{times}
\usepackage{latexsym}

\usepackage[T1]{fontenc}

\usepackage[utf8]{inputenc}

\usepackage{microtype}

\usepackage{inconsolata}

\usepackage{graphicx}
\usepackage{float}
\usepackage{amsmath,gensymb}
\usepackage{bbm}
\usepackage{subcaption}

\usepackage{multirow}
\usepackage{tabularx}

\usepackage{wrapfig}
\usepackage{xspace}

\usepackage[shortlabels]{enumitem}
\setlist[itemize]{leftmargin=*}
\setlist[enumerate]{leftmargin=*}
\usepackage[normalem]{ulem}

\usepackage{booktabs}
\usepackage{soul}
\usepackage{todonotes}

\definecolor{tticblue}{RGB}{0, 94, 184}
\newcommand{\dataset}{\texttt{COMFORT}\xspace}
\newcommand{\datasetSYMM}{\texttt{COMFORT-BALL}\xspace}
\newcommand{\datasetFOR}{\texttt{COMFORT-CAR}\xspace}

\usepackage{array}
\newcolumntype{C}[1]{>{\centering\arraybackslash}m{#1}}
\newcolumntype{R}{>{\raggedleft\arraybackslash}p{1cm}}

\newcommand{\prefer}[1]{\textcolor{black}{\bf\uline{#1}}}
\newcommand{\sota}[1]{\textcolor{black}{\bf{#1}}}
\newcommand{\poor}[1]{\textcolor{red}{#1}}

\usepackage{booktabs}
\usepackage{rotating}

\usepackage{epigraph}
\newrobustcmd*{\citefirstlastauthor}{\AtNextCite{\DeclareNameAlias{labelname}{given-family}}\citeauthor}
\AtBeginDocument{}

\title{\vspace{-20pt}Do Vision-Language Models Represent Space and How? Evaluating Spatial Frame of Reference under Ambiguities}

\author{%
  Zheyuan Zhang\textsuperscript{\rm 1}\thanks{Authors contributed equally to this work.}
  \qquad Fengyuan Hu\textsuperscript{\rm 1}\footnotemark[1]
  \qquad Jayjun Lee\textsuperscript{\rm 1}\footnotemark[1] \\
  {\bf Freda Shi\textsuperscript{\rm 2,3}
  \qquad Parisa Kordjamshidi\textsuperscript{\rm 4}
  \qquad Joyce Chai\textsuperscript{\rm 1}
  \qquad Ziqiao Ma\textsuperscript{\rm 1}} \\
  \textsuperscript{\rm 1}University of Michigan
  \; \textsuperscript{\rm 2}University of Waterloo \\
  \textsuperscript{\rm 3}Vector Institute, Canada CIFAR AI Chair
  \; \textsuperscript{\rm 4}Michigan State University \\
  \texttt{\url{https://spatial-comfort.github.io/}}
}

\iclrfinalcopy

\begin{document}

\maketitle

\vspace*{-10pt}
\begin{abstract}

\vspace{-10pt}
Spatial expressions in situated communication can be ambiguous, as their meanings vary depending on the frames of reference (FoR) adopted by speakers and listeners. 
While spatial language understanding and reasoning by vision-language models (VLMs) have gained increasing attention, potential ambiguities in these models are still under-explored.
To address this issue, we present the \underline{CO}nsistent \underline{M}ultilingual \underline{F}rame \underline{O}f \underline{R}eference \underline{T}est (\dataset), an evaluation protocol to systematically assess the spatial reasoning capabilities of VLMs.
We evaluate nine state-of-the-art VLMs using \dataset.
Despite showing some alignment with English conventions in resolving ambiguities, our experiments reveal significant shortcomings of VLMs: notably, the models (1) exhibit poor robustness and consistency, (2) lack the flexibility to accommodate multiple FoRs, and (3) fail to adhere to language-specific or culture-specific conventions in cross-lingual tests, as English tends to dominate other languages. 
With a growing effort to align vision-language models with human cognitive intuitions, we call for more attention to the ambiguous nature and cross-cultural diversity of spatial reasoning.

\end{abstract}
\vspace*{-12pt}
\section{Introduction}

The recent success of large language models has sparked breakthroughs in multi-modalities, leading to the development of many vision-language models \citep[VLMs;][\textit{inter alia}]{chen2023pali,gpt4o,reid2024gemini}.
With some benchmarks developed to evaluate the downstream performance of these models \citep{liu2023mmbench,yue2024mmmu}, there has been growing excitement around evaluations and analyses inspired by human cognitive capabilities such as referential grounding \citep{ma2023world}, compositional reasoning \citep{ma2023crepe}, visual illusions \citep{zhang2023grounding,guan2024hallusionbench}, and theory of mind \citep{jin2024mmtomqa}.
One direction among them that captures significant attention is spatial language understanding and reasoning, leading to several benchmarks \citep{mirzaee-etal-2021-spartqa,mirzaee-kordjamshidi-2022-transfer,kamath2023whatsup, liu2023visualspatial} and enhanced models \citep{chen2024spatialvlm, cheng2024spatialrgpt,premsri2024neurosymbolictrainingreasoningspatial}. 

Indeed, spatial cognition is a crucial part of human cognitive capability, developed since infancy and continuing through the elementary school years~\citep{tommasi2012psychology,vasilyeva2012development}. 
Language is closely intertwined with spatial cognition, with each contributing to the acquisition of the other \citep{hayward1995spatial,regier2001grounding,pyers2010evidence,pruden2011children,gentner2013spatial}.
While spatial language and non-linguistic spatial representations in memory are closely correlated and share foundational properties, they are, to some extent, divergent---spatial conventions are not consistently preserved across different languages or tasks, and humans demonstrate flexibility in using multiple coordinate systems for both non-linguistic reasoning and linguistic expressions \citep{munnich2001spatial,shusterman2016frames}.
Thus, spatial language is inherently ambiguous, and as we quote:

\begin{figure}[!h]
    \vspace{-10pt}
    \setlength{\epigraphwidth}{0.9\linewidth}
    \epigraph{Languages just do turn out to use fundamentally different semantic parameters in their categorization of spatial relations---different coordinate systems, different principles for constructing such coordinate systems, yielding different categorizations of `same' and `different' across spatial scenes.}{\textit{Stephen C. Levinson (\citeyear{levinson2003space})}}
    \vspace{-15pt}
\end{figure}

In situated communication, even a simple spatial expression like ``the basketball to the right of the car'' may have multiple interpretations.
People may use different \textit{frames of reference}~\citep[FoR;][\textit{inter alia}]{levinson1996frames,frank1998formal} to resolve ambiguity about the underlying coordinate system, as illustrated in Figure~\ref{fig:ambiguity}a.
The diversity of conventions across languages and cultures further complicates this ambiguity---different languages employ different conventions in choosing one FoR among multiple competing options.
As shown in Figure~\ref{fig:ambiguity}b, speakers may project themselves onto the ball or consider an imaginary listener facing them \citep{shusterman2016frames}.
These ambiguities are not easily resolvable based solely on linguistic expressions \citep{tenbrink2004identifying,liu2010ambiguities}.

Our main research question is not new: \textit{Do vision-language models represent space, and how?}
Several benchmarks \citep{kamath2023whatsup,liu2023visualspatial} have been developed for this purpose, consisting of text-image pairs where objects may or may not follow certain spatial relations.
However, the aforementioned spatial ambiguities remain largely under-explored when studying VLM-based spatial language understanding and reasoning.
We emphasize that FoRs are crucial to studying spatial cognition across modalities, as they provide a foundational framework for understanding how spatial relationships are perceived, interpreted, and communicated \citep{levinson2003space}.

To fill this gap, we present \underline{CO}nsistent \underline{M}ultilingual \underline{F}rame \underline{O}f \underline{R}eference \underline{T}est (\dataset), a framework that systematically evaluates the spatial reasoning capabilities of VLMs, emphasizing consistency in understanding ambiguous and disambiguated spatial expressions.
\dataset introduces (1) a set of spatial reasoning tasks instantiated by synthetic 3D images and corresponding text describing spatial relations and (2) metrics to evaluate the robustness and consistency of the model responses.
We extend the setup to multilingual settings by evaluating models in 109 languages across 170 regions worldwide.
We find that VLMs show alignment with English conventions in spatial language understanding when resolving ambiguities. 
However, they (1) are still far from achieving robustness and consistency, (2) lack the flexibility to accommodate multiple FoRs, and (3) fail to adhere to linguistic and cultural conventions in cross-lingual tests, as English tends to dominate other languages.
With a growing effort to align vision-language models with human cognition, we highlight the ambiguous nature of spatial language and call for increased attention to cross-cultural diversity in spatial reasoning.

\vspace*{-10pt}
\section{Background and Related Work}

\vspace*{-7pt}
\subsection{Spatial Language and Spatial Representation}

\begin{figure*}[t]
    \centering
    \includegraphics[width=\textwidth]{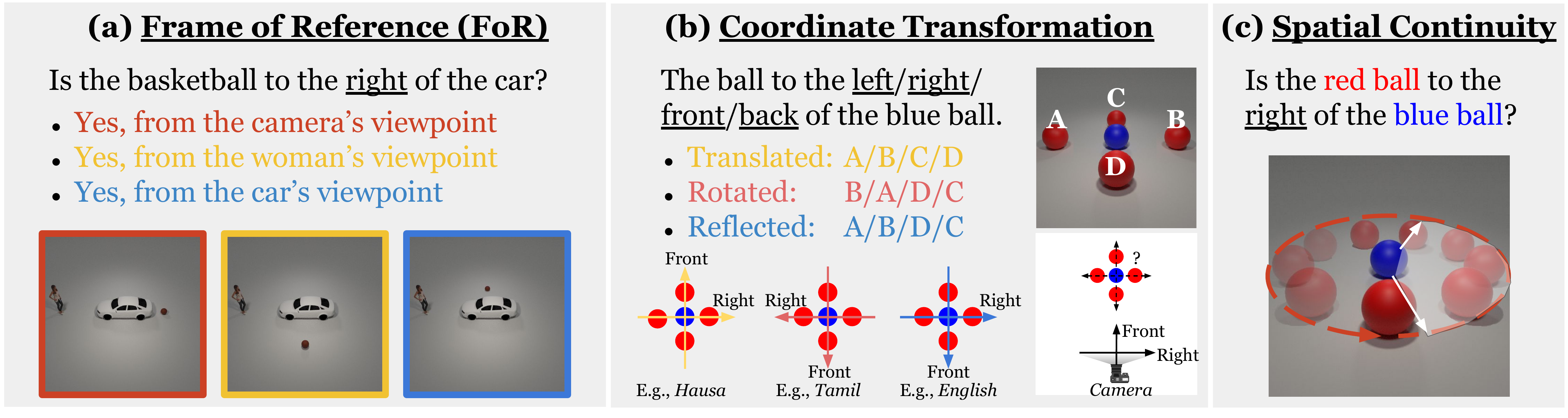}
    \vspace{-20pt}
    \caption{In situated communication, spatial language understanding and reasoning are often ambiguous, leading to varying interpretations among people from different cultural backgrounds. Specifically: (a) different frames of reference can result in different interpretations of the same spatial expression; (b) speakers of different languages may use distinct coordinate frames for non-fronted reference objects; and (c) spatial relations extend beyond exact axes to include acceptable regions. \vspace{-15pt}}
    \label{fig:ambiguity}
\end{figure*}

Some projective terms, such as the English words \textit{front}, \textit{back}, \textit{right}, and \textit{left}, convey meanings of spatial relations \citep{eschenbach2005contextual}. 
These terms articulate the spatial relation between two entities within a designated \textit{frame of reference} (FoR), often involving one entity as the reference object (\textit{relatum}/\textit{ground}) and another target object (\textit{referent}/\textit{figure}) that is positioned relative to the relatum along a specific axis/direction \citep{levinson1996frames,frank1998formal}. 
In situated communication, speech act participants (e.g., an \textit{addressee}) may also be considered \citep{danziger2010deixis}.
To determine acceptable uses of various spatial relations, existing theories suggest that people fit \textit{spatial templates}, which are centered on the relatum and aligned with the FoR \citep{logan1996computational}, to parse out \textit{regions of acceptability} of certain directions \citep{franklin1995parsing,carlson1997influence}.

\begin{figure*}[t]
\centering
\begin{minipage}{0.6\textwidth}
\scalebox{0.825}{
\setlength{\tabcolsep}{4pt}
\renewcommand{\arraystretch}{0.8}
\hspace*{-10pt}
\begin{tabular}{lll}
\toprule
\textbf{Origin} & \textbf{Frame of Reference} & \textbf{Example} \small{(English)}  \\
\midrule
\begin{tabular}[l]{@{}l@{}}
Camera \\ (Preferred)
\end{tabular} &
\begin{tabular}[l]{@{}l@{}}
Egocentric \\ Relative FoR
\end{tabular}
& \begin{tabular}[l]{@{}l@{}}
(From the \underline{camera}'s viewpoint,) \\ 
the ball is \textbf{behind} the car.
\end{tabular} \\
\midrule
Addressee & 
\begin{tabular}[l]{@{}l@{}}
Addressee-Centered\\ Relative FoR
\end{tabular}
& \begin{tabular}[l]{@{}l@{}}
(From the \underline{woman}'s viewpoint,) \\ 
the ball is to the \textbf{left} of the car.
\end{tabular} \\ 
\midrule
Reference & 
\begin{tabular}[l]{@{}l@{}} 
Object-Centered\\ Intrinsic FoR
\end{tabular} 
& 
\begin{tabular}[l]{@{}l@{}}
(From the \underline{car}'s viewpoint,) \\ 
the ball is to the \textbf{right} of the car.
\end{tabular} \\
\bottomrule
\end{tabular}}
\end{minipage}
\begin{minipage}{0.185\textwidth}
\centering
\includegraphics[width=1.0\textwidth]{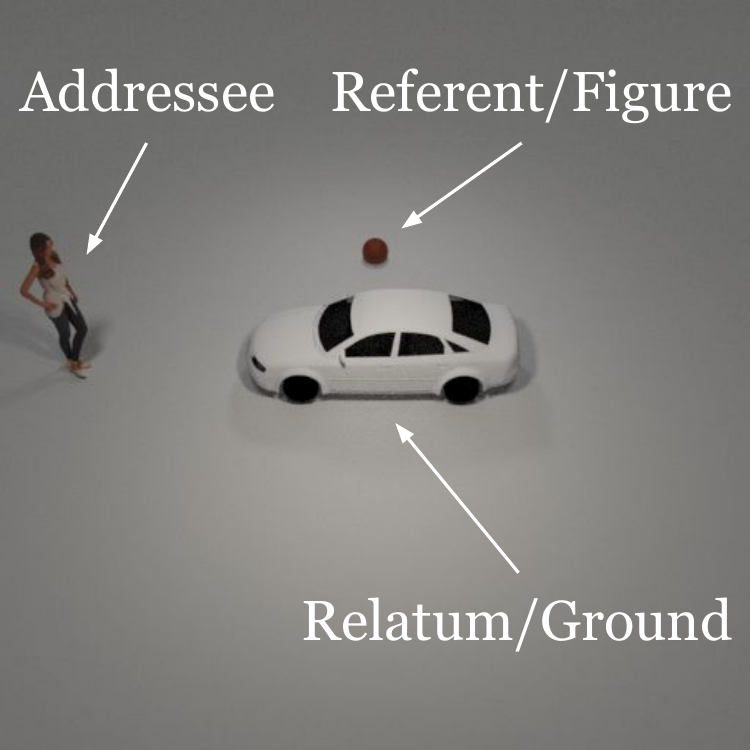}
\end{minipage}
\hspace*{1pt}
\begin{minipage}{0.18\textwidth}
\centering
\includegraphics[width=1.0\textwidth]{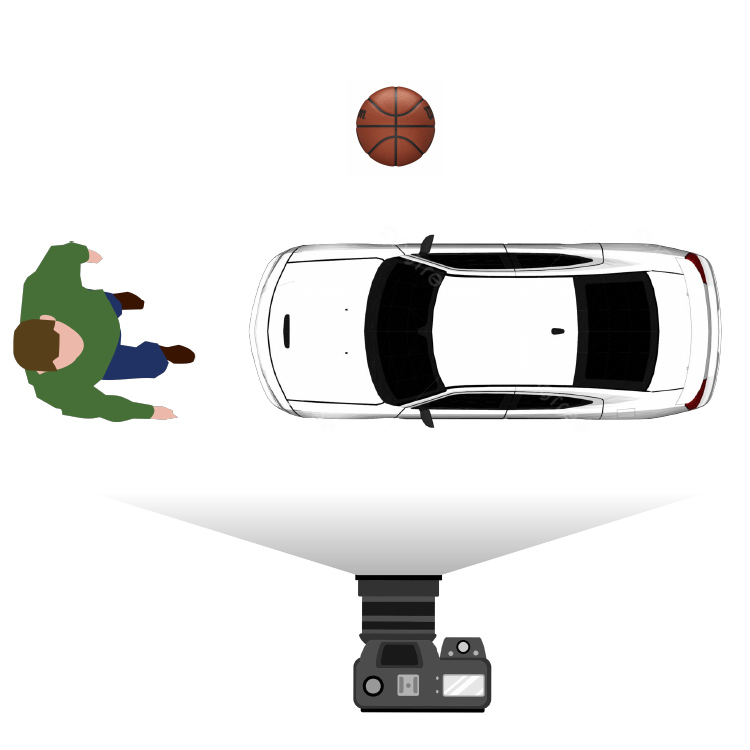}
\end{minipage}
\vspace{-5pt}
\caption{An illustrative example of how a frame of reference (FoR) specifies the reference system when describing the spatial relation between a target object (i.e., the ball) and a reference object (i.e., the car). When the FoR is not explicitly specified, English prefers an egocentric relative FoR, i.e., ``the ball is behind the car.'' We study FoRs that lead to ambiguity~\citep{liu2010ambiguities}. \vspace{-15pt}}
\label{fig:perspective_prompts_and_image}
\end{figure*}

\noindent\textbf{Ambiguities in frame of reference.}
The choice of perspectives may lead to different FoRs, where \citet{levinson2003space} has identified three main types of FoR: \textit{absolute}, \textit{intrinsic}, and \textit{relative}.
The absolute FoR uses cardinal directions, such as \textit{north} and \textit{south}, as fixed bearings. 
The intrinsic FoR aligns the origin with the relatum, describing the referent's position relative to the relatum's inherent orientation. 
The relative FoR positions a \textit{viewer} (egocentric or addressee) as the origin, focusing on the observer's intrinsic perspective. 
\citet{liu2010ambiguities} have highlighted the ambiguities in situated communication among three variations of intrinsic and relative FoRs (Figure~\ref{fig:perspective_prompts_and_image}): the \textit{egocentric relative}, the \textit{addressee-centered relative}, and the \textit{object-centered intrinsic} FoRs.\footnote{We exclude the absolute FoR from our study as it introduces little ambiguity \citep{liu2010ambiguities}.}
When not specified, these FoRs are not easily distinguishable based solely on their linguistic expressions \citep{tenbrink2004identifying}.
To resolve the ambiguity, individuals from diverse linguistic and cultural backgrounds adopt different preferences and conventions in choosing FoRs \citep{majid2004can,o2011spatial,bohnemeyer2014cultural,bender2020being,ogelo2024spatial}.

\noindent\textbf{Ambiguities in relative FoRs.}
The variations of relative FoRs form another source of ambiguity. 
After putting the origin of the coordination system on the viewer, multiple strategies specifying how to transform the axes can be considered (Figure~\ref{fig:ambiguity}b).
Different languages adopt different transformation conventions to relative FoRs \citep{levinson2003space,shusterman2016frames}, including: 
(1) \textit{translated} projection (e.g., Hausa) where the coordinate frame of the speaker is directly applied, (2) \textit{rotated} projection (e.g., Tamil), where the coordinate frame of the speaker is transformed with a 180-degree rotation, and (3) \textit{reflected} projection (e.g., English), where only the front-back axis is reversed.

\vspace*{-3pt}
\subsection{Spatial Understanding in Vision-Language Models}

Large language models (LLMs) have exhibited strong adaptability that extends beyond text, encompassing 2D and 3D vision \citep{tsimpoukelli2021multimodal,alayrac2022flamingo,yang20243d}, their affordances in the physical embodiment \citep{driess2023palm,qian2024affordancellm,yuan2024robopoint}, and various other modalities \citep{yu2024crema}.
Especially, a variety of vision-language models (VLM) have been developed by visual instruction tuning on paired text-image data \citep{dai2023instructblip,liu2023llava,dong2024internlm}.
With supervised fine-tuning using entity-phrase mappings in text-image pairs, grounded VLMs have been developed for fine-grained vision-language understanding at both the region \citep{chen2023shikra,bai2023qwen,you2023ferret,peng2024grounding} and pixel level \citep{lai2023lisa,xia2023gsva,rasheed2023glamm,zhang2024groundhog}.

Spatial understanding is known to be challenging even for state-of-the-art VLMs and is receiving increasing attention \citep{achiam2023gpt}.
In addition to using explicit spatial language understanding modules \citep{rajabi2024grounded}, recent works such as SpatialVLM \citep{chen2024spatialvlm} and SpatialRGPT \citep{cheng2024spatialrgpt} improve spatial reasoning in VLMs by leveraging 3D VQA or scene graph data for supervised fine-tuning. 
Several benchmarks have also been developed to evaluate spatial reasoning in VLMs from various perspectives \citep{liu2023visualspatial,cheng2024spatialrgpt,kamath2023whatsup}. 
Still, these benchmarks overlook ambiguities related to the FoR, lack spatial continuity, and have not proposed metrics to evaluate the robustness and consistency of spatial reasoning.

\vspace*{-5pt}
\section{Consistent Multilingual Frame of Reference Test (\dataset)}
\label{sec:data}

\begin{figure}[!t]
    \centering
    \begin{subfigure}[t]{.48\textwidth}
        \centering
        \includegraphics[width=1\textwidth]{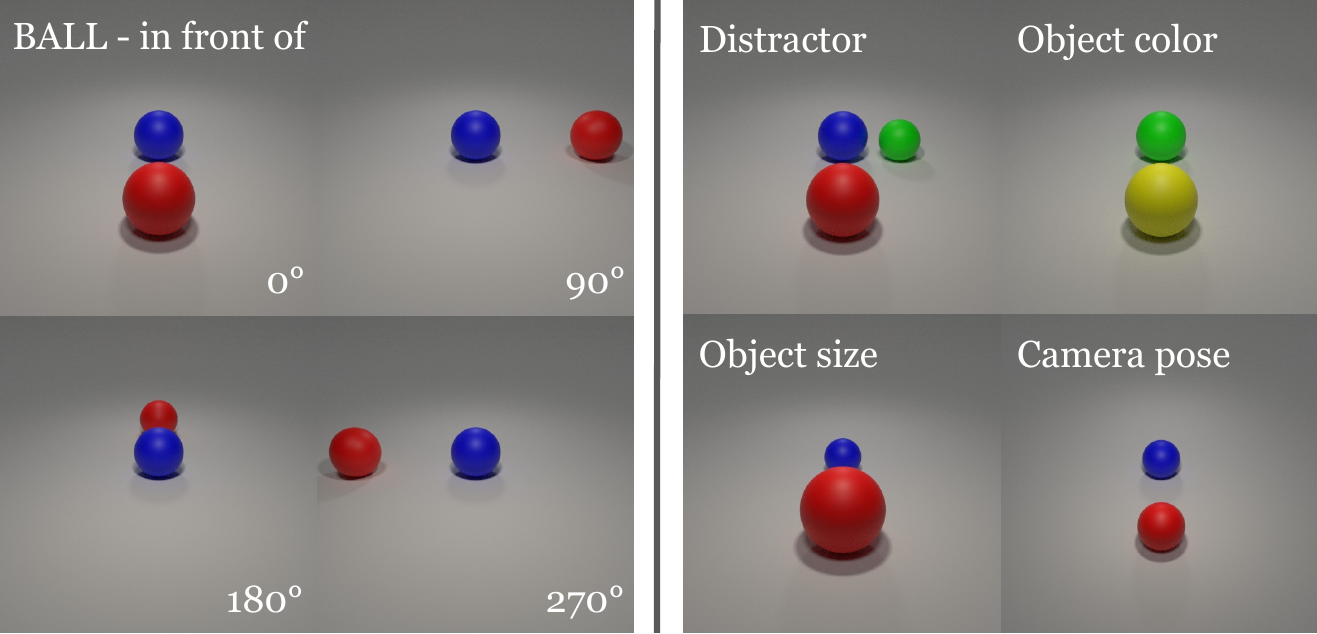}
        \vspace*{-11pt}
        \caption{Sample images from \datasetSYMM. The 4 images on the left are selected every 90\textdegree~interval along the rotational path out of 36 images. The 4 images on the right illustrate variations with a distractor, different object colors, sizes, or camera poses.}
    \label{fig:data_symm}
    \end{subfigure}
    ~
    \begin{subfigure}[t]{.48\textwidth}
        \centering
        \includegraphics[width=1\textwidth]{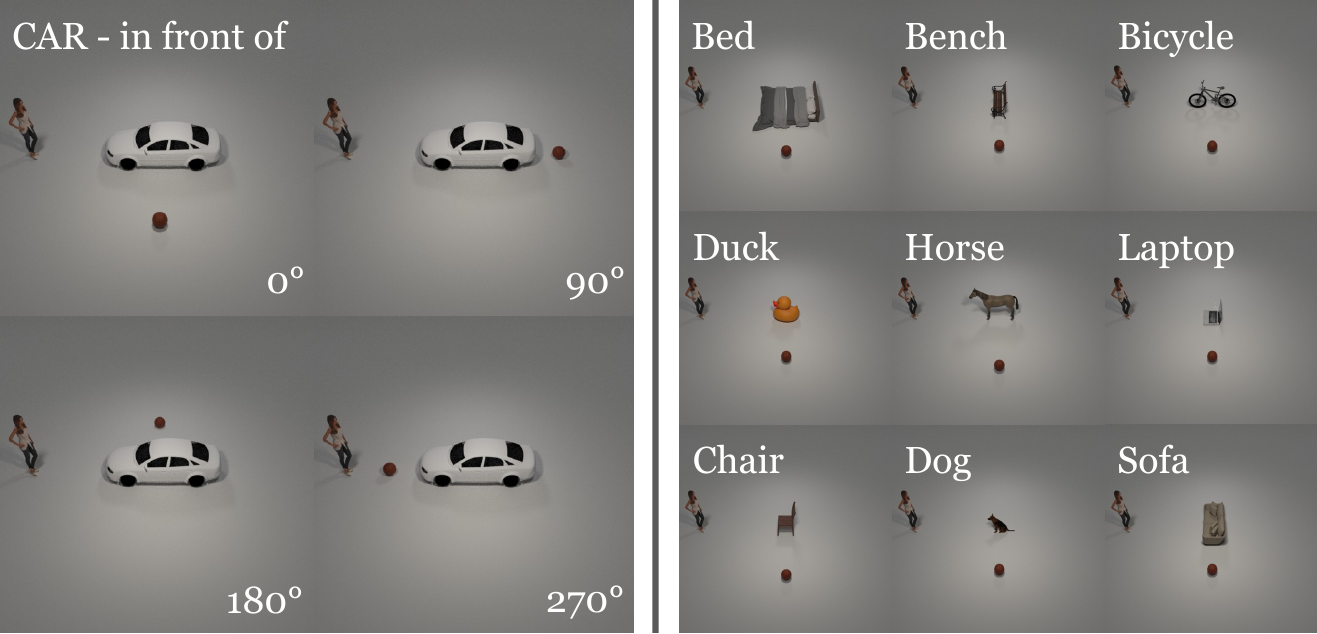}
        \caption{Sample images from \datasetFOR. The 4 images on the left are selected every 90\textdegree~interval along the rotational path out of 36 images. The 9 images on the right are sample images of each variation with different relatum objects.}
    \label{fig:data_for}
    \end{subfigure}
    \vspace*{-5pt}
    \caption{Examples from the \datasetSYMM~and \datasetFOR~datasets. \vspace*{-15pt}}
    \label{fig:trial}
\end{figure}

We introduce the COnsistent Multilingual Frame Of Reference Test (\dataset), a new evaluation protocol with dataset, tasks, and comprehensive metrics, to study VLM behaviors in spatial language reasoning with FoR-related ambiguity. 
This protocol accommodates spatial continuity and various ambiguities, drawing insights from several well-defined metrics to assess performance and prediction consistency. 
Given our primary focus on the analytical inquiry of models' linguistic competence (i.e., spatial knowledge encoded in the latent representations) rather than performance (i.e., behavioral evaluation) only \citep{warstadt2022artificial,saxon2024benchmarks},\footnote{Here, we use the terms \textit{competence} and \textit{performance} analogically to \citet{chomsky1965aspects}.} we additionally develop better evaluation and consistency metrics to deepen our understanding of model capabilities.

\vspace*{-5pt}
\subsection{Task Formulation}
Following the setups in object hallucination evaluation \citep{li2023evaluating,chen2024multi}, we formulate the task as a spatial relation inference problem.
In this task, a VLM $\mathcal{M}$ is presented with an RGB image $x_\textrm{img}$ and a textual question $x_\textrm{query}$.
The image shows the egocentric perception of a scene $s\in \mathcal{S}$, where $\mathcal{S}$ is the set of possible scenes in which the referent moves along a rotational trajectory with a constant radius from the relatum.
In contrast to fixing the referents on the standard canonical axes, this setup better mirrors the spatial continuity in common real-world scenarios.
A language prompt (Table~\ref{tab:prediction_notation}) queries whether a spatial relation $r\in \mathcal{R}$ is satisfied by a referent-relatum pair in the image under FoR $f \in \mathcal{F}$ (Figure~\ref{fig:perspective_prompts_and_image}) in language $\ell \in \mathcal{L}$.
This work also examines models using queries with no FoR specified; therefore, a test case in \dataset is defined as a 4-tuple in $\mathcal{S} \times \mathcal{R} \times \left(\mathcal{F}\cup \{\emptyset\}\right) \times \mathcal{L}$.
While there are many spatial relations in daily languages, we primarily focus on four canonical directions; that is, the considered relation set $\mathcal{R} = \{\textit{to the left of}, \textit{to the right of}, \textit{in front of}, \textit{behind}\}$.
\dataset covers $|\mathcal{L}|=109$ languages worldwide; however, we use English as an example to describe the data synthesis and evaluation processes for simplicity and clarity,  and refer readers to Appendix~\ref{sec:appendix_data} for more details. 

\vspace*{-5pt}
\subsection{Scene Setup}

We render the scenes into images using Blender \citep{blender2018blender}. 
Each scene consists of a referent and a relatum.
The referent follows a rotational trajectory with a constant radius from the relatum to implement spatial continuity. 
Starting from the canonical front direction, we move the referent with a uniform step of $10\degree$, totaling up to 36 images per scene. 
In \dataset, there are configurations determined by whether the relatum has an intrinsic semantic front:
\vspace*{-5pt}
\begin{itemize}[leftmargin=*, topsep=0pt]
    \setlength\itemsep{-0.25em}
    \item \textbf{\datasetSYMM}: When the relatum is non-fronted (e.g., Figure~\ref{fig:ambiguity}b), we focus on the ambiguity of FoR conventions associated with different languages.
    The split involves an observer's egocentric perception of a referent (e.g., a red ball) and a non-fronted relatum (e.g., a blue ball). 
    We further randomize the dataset with object-level (colors, sizes, and shapes) and scene-level variations (camera positions and distractors) to consider more diverse yet reasonable settings (Figure~\ref{fig:data_symm}).
    \item \textbf{\datasetFOR}: When the relatum is fronted (e.g., Figure~\ref{fig:ambiguity}a), multiple FoRs can be explicitly adopted to interpret the scene.
    A \datasetFOR image, therefore, involves the egocentric perception of a referent, a fronted relatum, and an additional human addressee.
    One can interpret the spatial relations using either the \underline{C}amera, \underline{A}ddressee, or \underline{R}elatum (C/A/R) as the origin to resolve the reference frame ambiguity.
    \datasetFOR has a set of 10 realistic objects in a typical household or outdoor scene, including \textit{horse}, \textit{car}, \textit{bench}, \textit{laptop}, \textit{rubber duck}, \textit{chair}, \textit{dog}, \textit{sofa}, \textit{bed}, and \textit{bicycle}, all of which have a clear semantic front.
    We use a basketball as the referent and vary the relatum.
    In addition to these objects, we include a human addressee in the scene.
    To disentangle different FoRs as much as possible, we let the addressee face right, and let the relatum face either left or right in the rendered images from the rendering camera's perspective (Figure~\ref{fig:data_for}).
\end{itemize}

\vspace*{-5pt}
\subsection{Language Query Setup}

Given a pair of referent \texttt{[A]} and a relatum \texttt{[B]}, together with a spatial relation of interest, the query is posed as ``Is \texttt{[A]} \texttt{[relation]} \texttt{[B]}?''
Depending on whether or not and which FoR is specified, the query is appended after four different perspective prompts (Table~\ref{tab:prediction_notation}): no perspective (nop), camera perspective (cam), addressee perspective (add), and relatum perspective (rel).
We only query from the camera egocentric perspective (cam) for \datasetSYMM, focusing on the ambiguity introduced by variations of the relative FoR.
For \datasetFOR, we use all four possible language prompts to study how ambiguity in the reference system is resolved.
Overall, for English, the above data generation pipeline leads to 720 test cases in \datasetSYMM, and 57.6k test cases in \datasetFOR.
The same method for dataset synthesis can be generalized to any other language; however, for computational efficiency, we only include the scenes corresponding to the four most prototypical directions (i.e., left, right, front, and back) in our multilingual analysis. 

\vspace*{-5pt}
\subsection{Metrics}

\begin{figure*}[t]
\centering
\begin{minipage}{0.4\textwidth}
    \setlength{\tabcolsep}{1pt}
    \renewcommand{\arraystretch}{0.9}
    \centering
    \vspace*{-5pt}
    \scalebox{0.875}{
    \begin{tabular}{C{0.9cm}C{5.2cm}}
        \toprule
        \textbf{Origin} & \textbf{Prompt Template} \\
        \midrule
        \texttt{nop} & Is \texttt{[A]} \texttt{[relation]} \texttt{[B]}? \\
        \midrule
        \texttt{cam} & From the camera's viewpoint,\newline is \texttt{[A]} \texttt{[relation]} \texttt{[B]}? \\
        \midrule
        \texttt{add} & From the \texttt{[addressee]}'s viewpoint,\newline is \texttt{[A]} \texttt{[relation]} \texttt{[B]}? \\
        \midrule
        \texttt{rel} & From the \texttt{[relatum]}'s viewpoint,\newline is \texttt{[A]} \texttt{[relation]} \texttt{[B]}? \\
        \bottomrule
    \end{tabular}}    
    \vspace*{3pt}
    \captionof{table}{The origins of each coordinate system and the corresponding prompt templates for querying the FoR given a referent-relation-relatum triple.}
    \label{tab:prediction_notation}
\end{minipage}
\hspace{5pt}
\begin{minipage}{0.225\textwidth}
\centering
    \vspace*{-5pt}
    \begin{subfigure}[t]{1.0\textwidth}
        \centering
        \includegraphics[width=\linewidth]{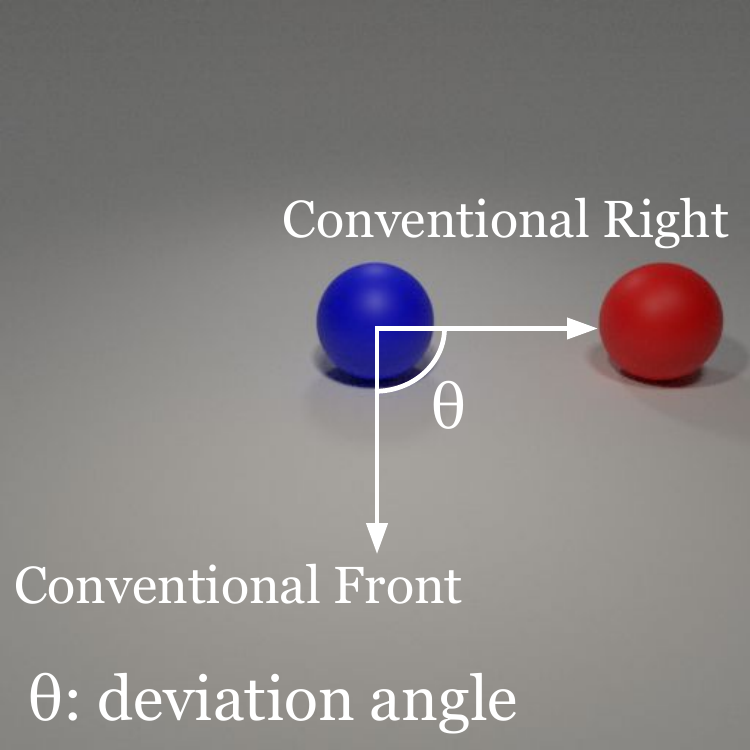}
    \end{subfigure}
    \vspace*{-12pt}
    \caption{A red ball with a deviation angle $\theta = 90\degree$ relative to the conventional front (English) of the blue ball.}
    \label{fig:metric-angle}
\end{minipage}
\hspace{5pt}
\begin{minipage}{0.325\textwidth}
\centering
    \vspace*{-5pt}
    \begin{subfigure}[t]{0.98\textwidth}
        \centering
        \includegraphics[width=\linewidth]{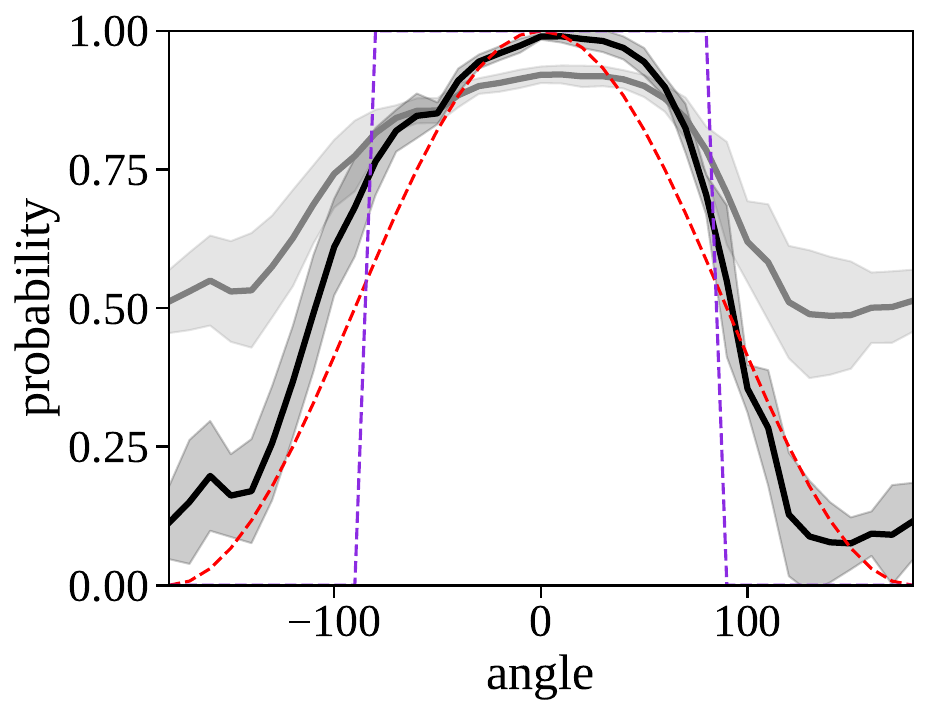}
    \end{subfigure}
    \vspace*{-10pt}
    \caption{The raw probability $p(\theta)$ in gray, normalized probability $\widehat{p}(\theta)$ in black, and two reference probability $\lambda^\textrm{hemi}(\theta)$ and $\lambda^\textrm{cos}(\theta)$ in purple and red.}
    \label{fig:metric-curve}
\end{minipage}
\vspace{-15pt}
\end{figure*}

Quantitatively assessing the spatial understanding and reasoning capabilities of models is challenging for two reasons.
First, the physical world is continuous, and spatial relations may extend beyond the precise canonical sagittal and lateral axes.
As noted by \citet{carlson1997influence}, there exists \textit{regions of acceptability} where, for instance, an object slightly to the front-left might still be considered being ``in front.''
Second, language models are biased towards affirmative responses \citep{dentella2023systematic}.
However, the intermediate representations may be sensitive to variations in input and, to some extent, align with human perceptions of spatial cues.
Based on these concerns and findings, we introduce multiple metrics to evaluate the models' competence to enable more nuanced analyses in addition to accuracy that measures performance.

Unless further clarified, we adopt a right-handed coordinate system with the thumb pointing upwards when describing angles. 
We define the \textit{deviation angle} $\theta\in (-180\degree, 180\degree]$ as the angular displacement from the canonical direction $r$ to the vector connecting the relatum and target.
For example, in Figure~\ref{fig:metric-angle}, the deviation angle of canonical right from canonical front is $\theta=90\degree$. 
Following \citet{carlson1997influence}, we define the acceptable region for a spatial relation $r$ as the 180-degree hemisphere centered at the corresponding canonical direction.
For a VLM $\mathcal{M}$ and a test case indexed by $i$, we let $P_i(\textit{response}; \mathcal{M})$ denote the probability of $\textit{response}\in\{\text{Yes}, \text{No}\}$ assigned by $\mathcal{M}$, and abbreviate it as $P_i(\textit{response})$ if there is no confusion.

\noindent \textbf{Accuracy.} Given a spatial relation $r$ in the textual prompt, we assess whether the assigned response probabilities correspond to whether the referent lies within the acceptable region defined by the relatum and $r$.
In line with recent work \citep{hu2023prompting,wang2025logical}, we define the local probability of the model responding `Yes' by $p_i = P_i(\text{Yes})/\left[P_i(\text{Yes}) + P_i(\text{No})\right]$. 
We consider the inference correct if (1) the scene falls into the acceptability region and $p_i > 0.5$ or (2) the scene falls out of the acceptability region and $p_i \leq 0.5$.

\noindent \textbf{Region Parsing Error.}
To mitigate the known bias towards affirmative answers, where the expectation of $p_i$ is generally higher than $0.5$, we normalize it across all image-prompt pairs, resulting in the normalized probability 
\vspace*{-12pt}
\begin{align*}
    \widehat{p}_i := \frac{p_i - \min_jp_j }{\max_jp_j -\min_j p_j}. 
\end{align*}

\vspace*{-3pt}
We adopt the root mean square error (RMSE) between the normalized acceptance probability $\widehat{p}$ and reference probability threshold $\lambda^\textrm{ref}$ that represents the actual regions of acceptability,
\begin{align*}
    \varepsilon^\textrm{ref} = \sqrt{\frac{1}{n}\sum_{i=1}^{n}{(\widehat{p}_i - \lambda^\textrm{ref})^2}},
\end{align*}
where $\lambda^{\textrm{ref}}$ denotes the reference of the assigned probability, analogically to ground-truth labels in machine learning terms. 

We introduce two analytically and geometrically motivated proposals defining $\lambda^{\textrm{ref}}$, $\lambda^\textrm{hemi}$ and $\lambda^\textrm{cos}$, based on hemispheres and cosine of angles, respectively. 
First, the hemisphere-based reference $\lambda^\textrm{hemi}$ is defined as
\vspace*{-5pt}
\begin{align*}
    \lambda^\textrm{hemi}(\theta) :=
    \begin{cases} 
    1 & \text{if } \theta \in (-90\degree, 90\degree) \\
    0 & \text{if } \theta \in (-180\degree, -90\degree] \cup [90\degree, 180\degree].
    \end{cases}
    \vspace*{-10pt}
\end{align*}
Here, $\theta = 0\degree$ corresponds to the most prototypical spatial relation, and $\theta = 180 \degree$ corresponds to the opposite. 
Intuitively, $\lambda^\textrm{hemi} = 1$ denotes the test case falls into the acceptable region defined by the textual prompt, and otherwise not. 

The second reference is derived from the cosine of the deviation angle. 
Matching the range of the cosine function to that of probability, i.e., $[0, 1]$, we define the cosine-based reference $\lambda^\textrm{cos}(\theta)$ by 
\begin{align*}
    \lambda^\textrm{cos}(\theta) := [\cos(\theta) + 1] / 2.
\end{align*}

\vspace*{-3pt}
Figure~\ref{fig:metric-curve} shows an example of the vanilla probability curve $p(\theta)$ from LLaVA-v1.5-7B \citep{liu2023llava}, normalized probability curve $\widehat{p}(\theta)$, and two reference curves $\lambda^\textrm{hemi}(\theta)$ and $\lambda^\textrm{cos}(\theta)$.
We report both $\varepsilon^\textrm{hemi}(\theta)$ and $\varepsilon^\textrm{cos}(\theta)$ in experiments.
We also note that in human spatial cognition, the regions of acceptability are neither mutually exclusive 90\textdegree~quadrants nor overlapping 180\textdegree~hemispheres, as they vary across individuals and depend on the situational context~\citep{franklin1995parsing}.
We discuss the limitations of this design in Section~\ref{sec:limitation}.

\begin{figure}
    \centering
    \vspace*{-10pt}
    \begin{subfigure}[t]{.535\textwidth}
        \centering
        \includegraphics[width=\textwidth]{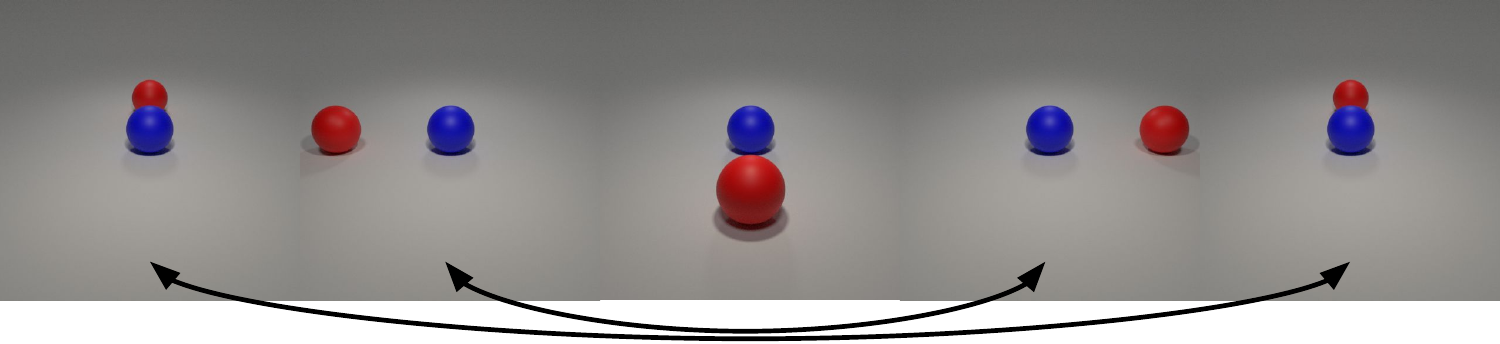}
    \vspace*{-15pt}
    \caption{An illustration of the spatial symmetry with respect to the (conventional) front. As the red ball rotates around the blue ball, spatial symmetry consistency ensures that each symmetric pair, with different deviation angles \(\theta\) and \(-\theta\), has the same probability of being identified as the front.}
    \label{fig:metric-sym}
    \end{subfigure}
    ~
    \begin{subfigure}[t]{.44\textwidth}
        \centering
        \includegraphics[width=\linewidth]{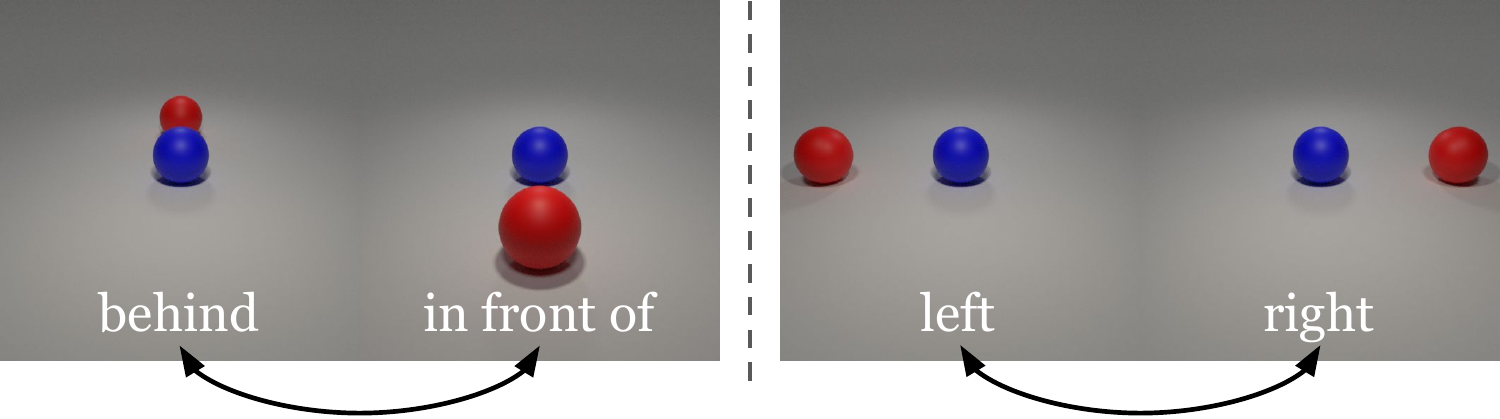}
    \vspace*{-15pt}
    \caption{Antonyms for spatial opposite consistency, e.g., When evaluating if the red ball is to the left of the blue ball, spatial opposite consistency ensures the probability of accepting a sample as left equals the probability of identifying it as not right.}
    \label{fig:metric-opp}
    \end{subfigure}
    \vspace*{-8pt}
    \caption{Illustrations for the consistency metrics defined in \dataset. \vspace*{-12pt}}
    \label{fig:consistency}
\end{figure}

\subsection{Robustness Metrics}

\noindent \textbf{Standard deviation.}
In \dataset, some images depict variations of the same scene, sharing identical spatial relations between the referent and the object but differing in terms of object colors, sizes, or distractors.
When the spatial relation and the query text remain unchanged, an ideal model should have consistent predictions for all variations.
To measure the robustness of the model prediction, we report the average standard deviation of the predicted probability $\widehat{p}_i$ across all deviation angles $\sigma:= \textrm{avg}_\theta{\sigma(\theta)}$.

\paragraph{Prediction noise.}
Since our data is generated through interpolation, ideally, if a model well understands spatial relations, the probability curve with respect to the deviation angle should be low-frequency (i.e., smooth) rather than high-frequency (i.e., noisy).
Therefore, we measure the noise by the RMSE, denoted by $\eta$, between the predicted probability and a Butterworth Low Pass Filter \citep[LPF;][]{butterworth1930theory}:
\begin{align*}
    \eta :=
    \sqrt{\frac{1}{n}\sum_{i=1}^{n}{[\widehat{p}_i - \textrm{LPF}(\widehat{p}_i)]^2}}.
\end{align*}
A smaller value of $\eta$ indicates that the probability curve is smoother, which is more desirable.

\vspace*{-8pt}
\subsection{Consistency Metrics}

\noindent\textbf{Spatial symmetric consistency.}
A critical aspect of consistent spatial reasoning is geometric symmetry.
As our tested target object rotates around the relatum in a circular path that is spatially symmetric, we expect the probabilities of an ideal VLM to consistently reflect geometric symmetry (Figure~\ref{fig:metric-sym}).
For a pair of test cases, indexed by $i$ and $j$, that have the same configurations but opposite deviation angles, i.e., $\theta_i + \theta_j = 0\degree$, we define the symmetry consistency:
\begin{align*}
    c^\textrm{sym} := 
    \sqrt{\frac{2}{n-2}\sum_{i,j} (\widehat{p}_i - \widehat{p}_j)^2}.
\end{align*}
\vspace*{-10pt}

\noindent\textbf{Spatial opposite consistency.}
Similarly, we expect the probabilities of an ideal VLM to consistently reflect geometric opposition (Figure~\ref{fig:metric-opp}).
For example, the probability that a sample is accepted by the spatial relation ``to the left'' should be identical to the probability that it is rejected by ``to the right.''
For a pair of opposite spatial relation $r, \textrm{opp}(r)\in \mathcal{R}$ with the same configurations including the deviation angles $\theta_i$, the opposition consistency is given as:
\begin{align*}
    c^\textrm{opp} := 
    \sqrt{\frac{1}{n}\sum_{i=0}^n (\widehat{p}_i^r + \widehat{p}_i^{\textrm{opp}(r)} - 1)^2}.
\end{align*}
\vspace*{-15pt}
\vspace*{-5pt}
\section{Empirical Experiments and Main Findings}

The \dataset framework enables us to investigate whether the internal representations of vision-language models encode spatial relations, and if they do, which underlying coordinate systems these representations capture.
This can further be broken down into two research questions: 
(1) When presented with an ambiguous spatial expression, do VLMs follow conventions and exhibit specific preferred FoRs (and the coordinate transformation in relative FoRs) to resolve the ambiguity?
(2) How effectively can VLMs adopt different FoRs, when perspectives are explicitly specified to disambiguate spatial expressions paired with visual scenes? 

In principle, \dataset~can be applied to all VLMs, whether multilingual or monolingual. 
We note that most existing open-source VLMs are English-based language models; therefore, we begin our experiments on English conventions, where both \textit{relative} and \textit{intrinsic} FoRs are available, but there is a conventional preference for a \textit{relative} FoR combined with a \textit{reflected} coordinate transformation in the relative FoR \citep[see][Table 5.4]{levinson2003space}.
We further extend our setup to multilingual settings by evaluating models in 109 languages across 170 regions worldwide.
To cover a variety of VLMs with different capabilities and training approaches, we evaluate the following models:
\vspace*{-3pt}
\begin{itemize}[leftmargin=*,topsep=0pt]
    \setlength\itemsep{0pt}
    \item VLMs build from supervised instruction fine-tuning: InstructBLIP-7B/13B \citep{dai2023instructblip}, LLaVA-v1.5-7B/13B \citep{liu2023llava}, InternLM-XComposer2-7B \citep{dong2024internlm};
    \item VLMs with both supervised fine-tuning and reinforcement learning alignment: MiniCPM-Llama3-V-v2.5-8B \citep{hu2024minicpm,yu2024rlaif};
    \item Mechanistically grounded VLMs: GLaMM-7B \citep{rasheed2023glamm};
    \item Multilingual VLMs: mBLIP-BLOOMZ-7B \citep{geigle2024mblip} and GPT-4o \citep{gpt4o}.
\end{itemize}

\begin{wrapfigure}{!tr}{0.5\textwidth}
    \centering  
    \vspace*{-10pt}
    \begin{subfigure}[t]{.23\textwidth}
        \centering
        \includegraphics[width=1.05\textwidth]{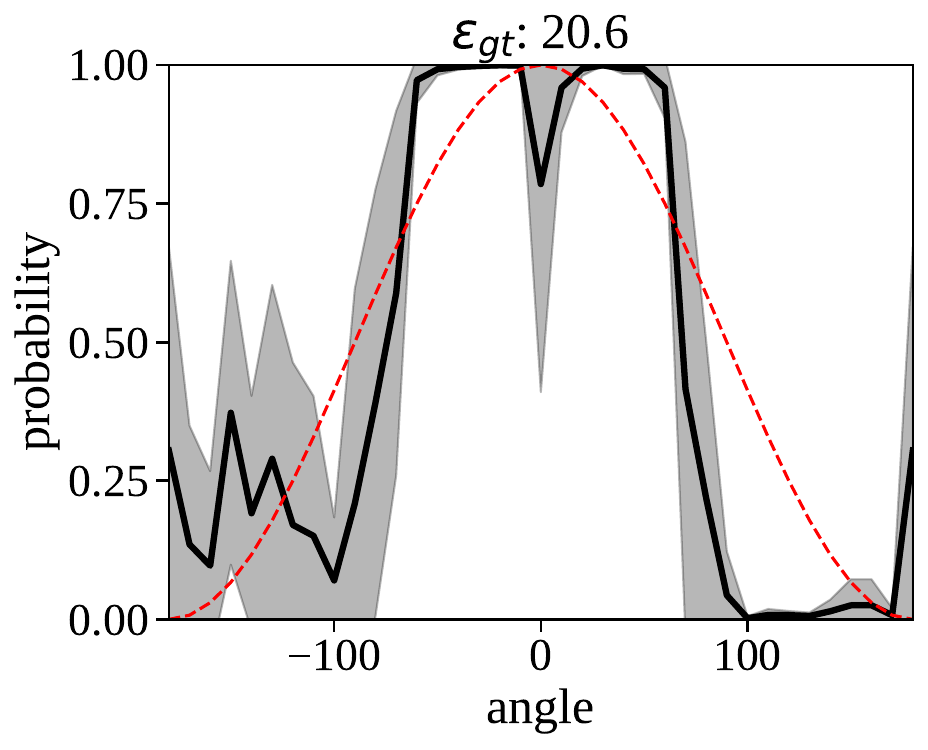}
        \vspace*{-15pt}
        \caption{\texttt{Behind} in GPT-4o.}
    \end{subfigure}
    ~
    \begin{subfigure}[t]{.23\textwidth}
        \centering
        \includegraphics[width=1.05\textwidth]{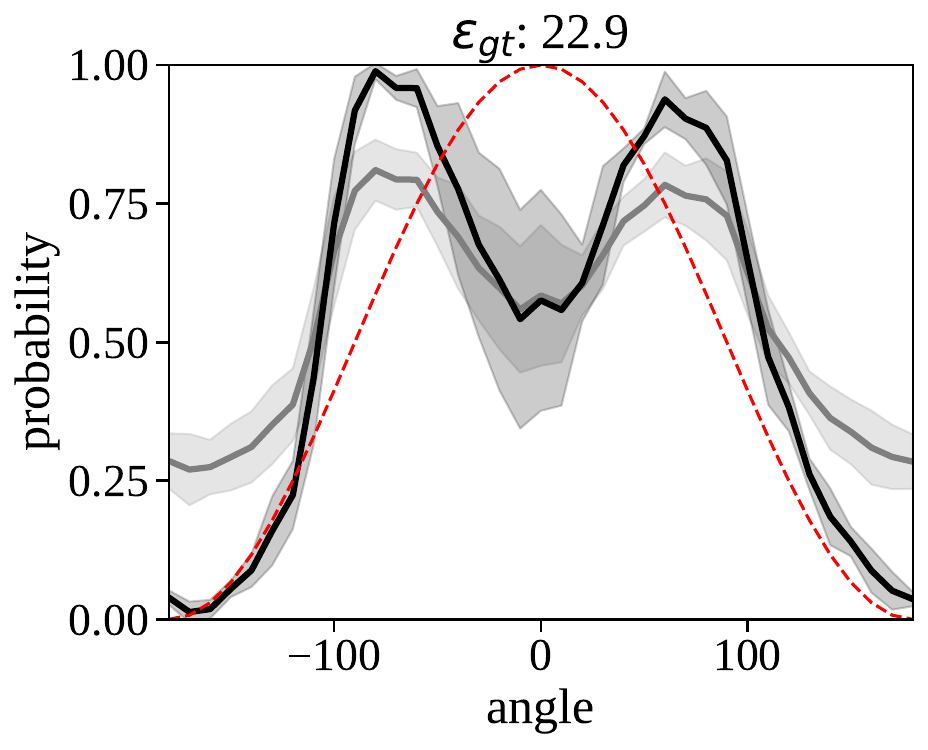}
        \vspace*{-15pt}
        \caption{\texttt{Right} in LLaVA-13B.}
    \end{subfigure}   
    \vspace*{-8pt}
    \caption{At $\theta=0$, some models show sensitivity to multiple conventions. \vspace*{-10pt}}
    \label{fig:transformation-convention}
\end{wrapfigure}
\begin{table}[!t]
    \scalebox{0.885}{
    \begingroup
    \setlength{\tabcolsep}{1pt}
    \renewcommand{\arraystretch}{0.9}
    \hspace*{-9pt}
    \begin{tabular}{lRRRRRRRRRRRc}
    \toprule
    \multirow{2}{*}{Model} & \multicolumn{2}{c}{Back $\varepsilon^\textrm{cos}$ ($\downarrow$)} & \multicolumn{2}{c}{Front $\varepsilon^\textrm{cos}$ ($\downarrow$)} & \multicolumn{2}{c}{Left $\varepsilon^\textrm{cos}$ ($\downarrow$)} & \multicolumn{2}{c}{Right $\varepsilon^\textrm{cos}$ ($\downarrow$)} & \multicolumn{3}{c}{Aggregated}   & Preferred \\
    \cmidrule(r){2-3}\cmidrule(r){4-5}\cmidrule(r){6-7}\cmidrule(r){8-9}\cmidrule(r){10-12}
    & Same                & \prefer{Rev.}                 & Same                 & \prefer{Rev.}                 & \prefer{Same}                & Rev.                 & \prefer{Same}                 & Rev.                 & Tran. & Rot. & \prefer{\prefer{Ref.}} &                   \\
    \cmidrule(r){1-1}\cmidrule(r){2-3}\cmidrule(r){4-5}\cmidrule(r){6-7}\cmidrule(r){8-9}\cmidrule(r){10-12}\cmidrule(r){13-13}
    InstructBLIP-7B      & 45.6                             & \sota{39.0}                   & \sota{31.6}                    & 52.0                             & \sota{37.2}                        & 48.0                        & 47.5                             & \sota{37.8}                    & \sota{40.5}       & 44.2    & 43.9          & -         \\
    InstructBLIP-13B     & \sota{40.9}                    & 45.5                            & 46.0                             & \sota{37.4}                    & \sota{43.4}                        & 44.9                        & 45.6                    & \sota{41.6}                             & 44.0       & \sota{42.3}    & 43.0          & -         \\
    mBLIP-BLOOMZ                & \sota{51.2}                    & 53.7                            & 51.2                             & \sota{47.9}                    & \sota{52.4}                        & 53.5                        & 54.6                    & \sota{46.8}                             & 52.3       & \sota{50.5}    & 52.1          & -         \\
    GLaMM                & 58.3                             & \sota{33.3}                   & 43.9                             & \sota{42.9}                    & \sota{38.3}                        & 51.8                        & \sota{17.3}                    & 63.7                             & 39.5       & 47.9    & \sota{33.0} & \prefer{Ref.}      \\
    LLaVA-1.5-7B         & 54.0                             & \sota{32.9}                   & 59.1                             & \sota{24.8}                    & \sota{11.9}                        & 70.0                        & \sota{13.0}                    & 68.5                             & 34.5       & 49.0    & \sota{20.7} & \prefer{Ref.}      \\
    LLaVA-1.5-13B        & 61.8                             & \sota{19.2}                   & 56.0                             & \sota{27.7}                    & \sota{31.7}                        & 61.8                        & \sota{24.3}                    & 64.3                             & 43.4       & 43.2    & \sota{25.7} & \prefer{Ref.}      \\
    XComposer2           & 73.2                             & \sota{17.9}                   & 74.5                             & \sota{20.7}                    & \sota{20.1}                        & 80.9                        & \sota{21.3}                    & 81.1                             & 47.3       & 50.1    & \sota{20.0} & \prefer{Ref.}      \\
    MiniCPM-V            & 70.9                             & \sota{21.9}                   & 64.3                             & \sota{26.9}                    & \sota{19.7}                        & 74.1                        & \sota{21.1}                    & 73.3                             & 44.0       & 49.1    & \sota{22.4} & \prefer{Ref.}      \\
    GPT-4o               & 75.7                             & \sota{28.2}                   & 73.6                             & \sota{32.0}                    & \sota{24.3}                        & 80.8                        & \sota{25.1}                    & 80.8                             & 49.7       & 55.5    & \sota{27.4} & \prefer{Ref.}      \\
    \bottomrule
    \end{tabular}
\endgroup}
\vspace*{-8pt}
\caption{Preferred coordinate transformation mapping from the egocentric viewer (camera) to the relatum in the relative FoR. 
The cosine region parsing errors $\varepsilon^\textrm{cos}$ are computed against both the \uline{Same} and \uline{Rev}ersed directions relative to the egocentric viewer's coordinate system. For example, native English speakers typically prefer a \uline{Ref}lected transformation, which maintains the lateral (left/right) axis but reverses the sagittal (front/back) axis relative to the viewer (Figure~\ref{fig:ambiguity}). We determine the preferred transformation based on the aggregated performance, with ``\texttt{-}'' for no significant preference. \vspace*{-15pt}}
\label{tab:transform-main}
\end{table}

\begin{table}[!t]
    \scalebox{0.85}{
    \begingroup
    \setlength{\tabcolsep}{2.5pt}
    \renewcommand{\arraystretch}{1.0}
    \hspace*{-9pt}
    \begin{tabular}{lrrrrrrrrrrrrrrrc}
    \toprule
    \multirow{2}{*}{Model} & \multicolumn{3}{c}{Back $\varepsilon^\textrm{cos}$ ($\downarrow$)} & \multicolumn{3}{c}{Front $\varepsilon^\textrm{cos}$ ($\downarrow$)} & \multicolumn{3}{c}{Left $\varepsilon^\textrm{cos}$ ($\downarrow$)} & \multicolumn{3}{c}{Right $\varepsilon^\textrm{cos}$ ($\downarrow$)} & \multicolumn{3}{c}{Aggregated} & \multirow{2}{*}{Prefer} \\
    \cmidrule(r){2-4}\cmidrule(r){5-7}\cmidrule(r){8-10}\cmidrule(r){11-13}\cmidrule(r){14-16}
     & \prefer{Ego.}         & Int.          & Add.         & \prefer{Ego.}         & Int.          & Add.          & \prefer{Ego.}         & Int.          & Add.         & \prefer{Ego.}         & Int.          & Add.          & \prefer{Ego.}     & Int.      & Add.      &                                \\
    \cmidrule(r){1-1}\cmidrule(r){2-4}\cmidrule(r){5-7}\cmidrule(r){8-10}\cmidrule(r){11-13}\cmidrule(r){14-16}\cmidrule(r){17-17}
    InstructBLIP-7B      & 41.0                 & \sota{38.6}        & \sota{38.6}        & \sota{40.9}         & 46.9                 & 46.9                 & 45.6                 & \sota{32.5}        & 51.9                 & 39.6                  & 51.2                 & \sota{31.8}        & \sota{41.8} & 42.3          & 42.3          & -                              \\
    InstructBLIP-13B     & \sota{32.9}        & 34.4                 & 34.4                 & 52.5                  & \sota{48.5}        & \sota{48.5}        & 47.8                 & 56.2                 & \sota{27.8}        & 40.6                  & \sota{27.6}        & 56.6                 & 43.5          & \sota{41.7} & 41.8          & -                              \\
    mBLIP-BLOOMZ         & \sota{52.2}        & 53.2                 & 53.2                 & 45.3                  & \sota{44.6}        & \sota{44.6}        & 47.8                 & \sota{47.6}        & 48.1                 & 45.4                  & 48.4                 & \sota{42.4}        & 47.7          & 48.4          & \sota{47.1} & -                              \\
    GLaMM                & \sota{28.0}        & 49.1                 & 49.1                 & \sota{30.0}         & 40.2                 & 40.2                 & \sota{14.0}        & 56.8                 & 41.5                 & \sota{13.7}         & 53.0                 & 46.6                 & \sota{21.4} & 49.8          & 44.4          & \prefer{Ego.}                           \\
    LLaVA-1.5-7B         & \sota{20.9}        & 43.0                 & 43.0                 & 34.5                  & \sota{32.6}        & \sota{32.6}        & \sota{13.4}        & 53.5                 & 47.4                 & \sota{14.3}         & 53.6                 & 49.3                 & \sota{20.8} & 45.7          & 43.1          & \prefer{Ego.}                           \\
    LLaVA-1.5-13B        & \sota{31.9}        & 38.8                 & 38.8                 & \sota{24.8}         & 57.1                 & 57.1                 & \sota{11.7}        & 51.1                 & 51.1                 & \sota{27.5}         & 57.4                 & 48.7                 & \sota{24.0} & 51.1          & 48.9          & \prefer{Ego.}                           \\
    XComposer2           & \sota{12.7}        & 49.3                 & 49.3                 & \sota{15.2}         & 48.3                 & 48.3                 & \sota{18.8}        & 61.2                 & 53.7                 & \sota{16.5}         & 58.4                 & 54.5                 & \sota{15.8} & 54.3          & 51.4          & \prefer{Ego.}                           \\
    MiniCPM-V            & \sota{34.2}        & 40.7                 & 40.7                 & \sota{35.5}         & 53.4                 & 53.4                 & \sota{18.0}        & 53.9                 & 58.4                 & \sota{19.0}         & 58.1                 & 52.7                 & \sota{26.7} & 51.5          & 51.3          & \prefer{Ego.}                           \\
    GPT-4o               & 38.3        & \sota{36.7}                 & \sota{36.7}                 & \sota{43.1}         & 50.2                 & 50.2                 & \sota{34.7}        & 59.3                 & 56.5                 & \sota{24.3}         & 57.3                 & 61.7                 & \sota{35.1} & 50.9          & 51.3          & \prefer{Ego.}                          \\
    \bottomrule
    \end{tabular}
\endgroup}
\vspace*{-8pt}
\caption{Preferred frame of reference in VLMs. Models' Cosine Region Parsing Errors $\varepsilon^\textrm{cos}$ are computed against the \uline{Int}rinsic FoR (relatum origin), \uline{Ego}centric relative FoR (camera origin), and \uline{Add}ressee-centric relative FoR (addressee origin). English typically prefers an egocentric relative FoR. We determine the preferred FoR based on the aggregated performance, with ``\texttt{-}'' indicating no significant preference.\vspace*{-15pt}}
\label{tab:preferred_for}
\end{table}

\vspace*{-5pt}
\subsection{Most VLMs Prefer Reflected Coordinate Transformation Convention}
\label{sec:transform}

In this section, we address the research question: 
\textbf{do VLMs have a preferred coordinate transformation convention, and if so, what is it?}
Experiments are conducted on \datasetSYMM using the camera perspective prompt (\texttt{cam}) that explicitly specifies an egocentric relative FoR (Table~\ref{tab:transform-main}).
Table~\ref{tab:transform-full} in the appendix shows the complete evaluation including $\varepsilon^{\textrm{hemi}}$ and $\varepsilon^{\textrm{cos}}$. 

We observe that almost all VLMs demonstrate a clear preference over the \textit{reflected} transformation similar to English, except the BLIP series.
Still, some models are also affected by the ambiguity of multiple transformation conventions. 
With the textual prompting specifying a relation, at $\theta=0$, GPT-4o and LLaVA-1.5-13B show a sharp drop of performance and a significant variance for \texttt{behind} and \texttt{right}, respectively (Figure~\ref{fig:transformation-convention}), indicating that some models are sensitive to other transformations.

\vspace*{-5pt}
\subsection{Most VLMs Prefer Egocentric Relative Frame of Reference}
\label{sec:for}

We now attempt to answer: 
\textbf{do VLMs have a preferred frame of reference, and if so, what is it?}
We conduct our study on \datasetFOR using the no perspective prompt (\texttt{nop}) that deliberately leaves the FoR ambiguous.
When calculating the performance with respect to relative FoRs (either egocentric or addressee-centered), we assume a reflected coordinate transformation convention.
Table~\ref{tab:preferred_for}
shows the results of preferred FoR in English measured by the region parsing error $\varepsilon^\textrm{cos}$, and Table~\ref{tab:for-full} in the appendix shows the complete evaluation including both $\varepsilon^{\textrm{hemi}}$ and $\varepsilon^{\textrm{cos}}$. 
Almost all VLMs demonstrate a significant preference towards the \textit{egocentric relative} FoR similar to English, again, except for the BLIP series.
Additionally, the models' performances are inconsistent across different spatial relations---models generally perform better in the lateral directions (left and right) than the sagittal ones (front and behind), even in competitive industry models like GPT-4o. 
For instance, GLaMM does not show a very strong preference when resolving ambiguities along the sagittal axes, but it demonstrates a significant preference when resolving lateral ambiguity.

\begin{table}[!t]
    \scalebox{0.87}{
    \begingroup
    \setlength{\tabcolsep}{2.7pt}
    \renewcommand{\arraystretch}{0.1}
    \hspace*{-9pt}
    \begin{tabular}{lllllllll}
    \toprule
    \multirow{2}{*}{Model}  & \multicolumn{2}{c}{Egocentric}                                       & \multicolumn{2}{c}{Intrinsic}                                      & \multicolumn{2}{c}{Addressee}                                    & \multicolumn{2}{c}{Aggregated}                                   \\
    \cmidrule(r){2-3}\cmidrule(r){4-5}\cmidrule(r){6-7}\cmidrule(r){8-9}
    \multicolumn{1}{l}{} & Acc\% $(\uparrow)$ & $\varepsilon^{\textrm{cos}}_{\times10^2} (\downarrow)$ & Acc\% $(\uparrow)$ & $\varepsilon^{\textrm{cos}}_{\times10^2} (\downarrow)$ & Acc\% $(\uparrow)$ & $\varepsilon^{\textrm{cos}}_{\times10^2} (\downarrow)$ & Acc\% $(\uparrow)$ & $\varepsilon^{\textrm{cos}}_{\times10^2} (\downarrow)$ \\
    \cmidrule(r){1-1}\cmidrule(r){2-3}\cmidrule(r){4-5}\cmidrule(r){6-7}\cmidrule(r){8-9}
    InstructBLIP-7B      & 47.2$_{(+0.0)}$          & 43.5$_{(+1.7)}$                             & 47.2$_{(+0.0)}$          & 42.3$_{(+0.0)}$                             & 47.2$_{(+0.0)}$          & 43.6$_{(+1.3)}$                             & 47.2$_{(+0.0)}$          & 43.1$_{(+1.0)}$                             \\
    InstructBLIP-13B     & 47.2$_{(+0.0)}$          & 43.8$_{(+0.3)}$                             & 47.2$_{(+0.0)}$          & 43.2$_{(+1.5)}$                             & 47.2$_{(+0.0)}$          & 42.9$_{(+1.1)}$                             & 47.2$_{(+0.0)}$          & 43.3$_{(+1.0)}$                             \\
    mBLIP-BLOOMZ         & 51.9$_{(-0.9)}$          & 55.4$_{(+7.7)}$                             & 49.8$_{(-3.0)}$          & 54.2$_{(+5.8)}$                             & 49.6$_{(-3.2)}$          & 55.8$_{(+8.7)}$                             & 50.4$_{(-2.4)}$          & 55.1$_{(+7.4)}$                             \\
    GLaMM                & 47.2$_{(-10.6)}$         & 23.3$_{(-0.7)}$                             & 47.2$_{(+0.8)}$          & \textbf{44.2$_{(-6.9)}$}                    & 47.2$_{(-2.8)}$          & 42.8$_{(-6.1)}$                             & 47.2$_{(-4.2)}$          & 36.8$_{(-4.6)}$                             \\
    LLaVA-1.5-7B         & 55.2$_{(-2.6)}$          & \textbf{18.4$_{(-3.0)}$}                    & 48.3$_{(+4.7)}$          & 45.7$_{(-4.1)}$                             & 48.2$_{(-5.0)}$          & 43.4$_{(-1.0)}$                             & 50.6$_{(-1.0)}$          & \textbf{35.8$_{(-2.7)}$}                    \\
    LLaVA-1.5-13B        & 51.6$_{(-15.0)}$         & 23.9$_{(+3.1)}$                             & 47.3$_{(+0.8)}$          & 45.0$_{(-0.7)}$                             & 47.5$_{(-3.8)}$          & \textbf{38.9$_{(-4.2)}$}                    & 48.8$_{(-6.0)}$          & 35.9$_{(-0.6)}$                             \\
    XComposer2           & \textbf{85.6$_{(-7.0)}$} & 18.8$_{(+3.0)}$                             & 51.0$_{(+0.5)}$          & 51.0$_{(-3.3)}$                             & \textbf{53.2$_{(-0.6)}$} & 49.8$_{(-1.6)}$                             & \textbf{63.3$_{(-2.4)}$} & 39.9$_{(-0.6)}$                             \\
    MiniCPM-V            & 72.4$_{(-4.8)}$          & 24.6$_{(-2.1)}$                             & 49.9$_{(-2.6)}$          & 47.8$_{(-3.7)}$                             & 52.9$_{(-0.5)}$          & 45.1$_{(-6.2)}$                             & 58.4$_{(-2.6)}$          & 39.2$_{(-4.0)}$                             \\
    GPT-4o               & 78.3$_{(+4.6)}$          & 28.1$_{(-7.0)}$                             & \textbf{53.4$_{(-1.9)}$} & 44.6$_{(-6.3)}$                             & 49.1$_{(-5.7)}$          & 44.9$_{(-6.4)}$                             & 60.3$_{(-1.0)}$          & 39.2$_{(-6.6)}$                            \\
    \bottomrule
    \end{tabular}
\endgroup}
\vspace*{-8pt}
\caption{The accuracy and cosine region parsing errors of VLMs when explicitly prompted to follow each frame of reference are provided (\texttt{cam}/\texttt{rel}/\texttt{add}). The values in parentheses indicate the performance change relative to the scenario with no perspective (\texttt{nop}) prompting. \vspace*{-8pt}}
\label{tab:perspective-main}
\end{table}

\begin{table}[!t]
    \scalebox{0.875}{
    \begingroup
    \setlength{\tabcolsep}{2.2pt}
    \renewcommand{\arraystretch}{0.9}
    \hspace*{-9pt}
    \begin{tabular}{lrrrrrrrrrrrrrrrr}
    \toprule
    \multirow{2}{*}{Model} & \multicolumn{2}{c}{Obj F1 ($\uparrow$)} & \multicolumn{2}{c}{Acc\% ($\uparrow$)} & \multicolumn{2}{c}{$\varepsilon^\textrm{cos}_{\times10^2}$ ($\downarrow$)} & \multicolumn{2}{c}{$\varepsilon^\textrm{hemi}_{\times10^2}$ ($\downarrow$)} & \multicolumn{2}{c}{$\sigma_{\times10^2}$ ($\downarrow$)} & \multicolumn{2}{c}{$\eta_{\times10^2}$ ($\downarrow$)} & \multicolumn{2}{c}{$c^\textrm{sym}_{\times10^2}$ ($\downarrow$)} & \multicolumn{2}{c}{$c^\textrm{opp}_{\times10^2}$ ($\downarrow$)} \\
    \cmidrule(r){2-3}\cmidrule(r){4-5}\cmidrule(r){6-7}\cmidrule(r){8-9}\cmidrule(r){10-11}\cmidrule(r){12-13}\cmidrule(r){14-15}\cmidrule(r){16-17}
     & \texttt{BALL} & \texttt{CAR} & \texttt{BALL} & \texttt{CAR} & \texttt{BALL} & \texttt{CAR} & \texttt{BALL} & \texttt{CAR} & \texttt{BALL} & \texttt{CAR} & \texttt{BALL} & \texttt{CAR} & \texttt{BALL} & \texttt{CAR} & \texttt{BALL} & \texttt{CAR} \\
    \cmidrule(r){1-1}\cmidrule(r){2-3}\cmidrule(r){4-5}\cmidrule(r){6-7}\cmidrule(r){8-9}\cmidrule(r){10-11}\cmidrule(r){12-13}\cmidrule(r){14-15}\cmidrule(r){16-17}
    InstructBLIP-7B      & \poor{66.7}            & \poor{66.7}            & 47.2             & 47.2             & 43.9                       & 43.5                       & 57.8                       & 56.4                       & 26.7                & 30.5                & 48.4               & 43.4               & 17.2                       & 16.9                       & 16.6                      & 22.6                      \\
    InstructBLIP-13B     & \poor{67.3}            & \poor{50.3}            & 47.2             & 47.2             & 43.0                       & 43.8                       & 55.5                       & 55.9                       & 27.1                & 36.8                & 48.2               & 46.4               & 17.3                       & 17.0                       & 21.0                      & 21.9                      \\
    mBLIP-BLOOMZ         & 99.1            & \poor{33.3}            & 47.5             & 51.9             & 52.1                       & 55.4                       & 62.1                       & 65.6                       & 43.7                & 48.6                & 54.1               & 60.7               & 29.1                       & 30.1                       & 33.8                      & 42.0                      \\
    GLaMM                & 100.0           & 99.8            & 47.2             & 47.2             & 33.0                       & 23.3                       & 45.2                       & 37.6                       & 29.9                & 23.4                & 45.0               & 28.4               & 10.1                       & 9.4                        & 13.7                      & 14.6                      \\
    LLaVA-1.5-7B         & 100.0           & 88.6            & 63.2             & 55.2             & 20.7                       & \sota{18.4}                       & 33.7                       & 32.5                       & 25.2                & 20.0                & 23.5               & \sota{21.8}               & \sota{5.8}                        & \sota{5.4}                        & \sota{8.3}                       & \sota{10.7}                      \\
    LLaVA-1.5-13B        & 100.0           & 98.6            & 55.3             & 51.6             & 25.7                       & 23.8                       & 37.6                       & 37.1                       & 19.3                & 20.8                & 24.9               & 29.9               & 7.0                        & 5.8                        & 9.3                       & 10.8                      \\
    XComposer2           & 100.0           & 95.3            & \sota{92.4}             & \sota{85.6}             & \sota{20.0}                       & 18.8                       & \sota{21.1}                       & \sota{26.3}                       & \sota{19.2}                & \sota{15.3}                & \sota{13.7}               & 22.9               & 9.0                        & 6.5                        & 10.5                      & 12.0                      \\
    MiniCPM-V            & \poor{66.8}            & 81.5            & 81.0             & 72.4             & 22.4                       & 24.6                       & 32.8                       & 35.8                       & 19.2                & \sota{19.2}                & 29.8               & 22.7               & 10.1                       & 9.2                        & 12.4                      & 14.9                      \\
    GPT-4o               & 100.0           & 94.5            & 89.2             & 78.3             & 27.4                       & 28.1                       & 27.5                       & 35.0                       & 20.9                & 24.0                & 43.1               & 38.8               & 14.1                       & 13.3                       & 14.2                      & 16.7                     \\
    \cmidrule(r){1-1}\cmidrule(r){2-3}\cmidrule(r){4-5}\cmidrule(r){6-7}\cmidrule(r){8-9}\cmidrule(r){10-11}\cmidrule(r){12-13}\cmidrule(r){14-15}\cmidrule(r){16-17}
    Random (30 trials)         & \multicolumn{2}{c}{50.0}                    & \multicolumn{2}{c}{50.9}                    & \multicolumn{2}{c}{46.3}                                & \multicolumn{2}{c}{58.7}                                & \multicolumn{2}{c}{28.3}                    & \multicolumn{2}{c}{26.6}                    & \multicolumn{2}{c}{42.5}                                & \multicolumn{2}{c}{44.2}                              \\
    Always ``Yes'' & \multicolumn{2}{c}{50.0}                    & \multicolumn{2}{c}{47.2}                    & \multicolumn{2}{c}{61.2}                                & \multicolumn{2}{c}{68.7}                                & \multicolumn{2}{c}{0.0}                     & \multicolumn{2}{c}{0.0}                     & \multicolumn{2}{c}{0.0}                                 & \multicolumn{2}{c}{100.0}                              \\   
    \bottomrule
    \end{tabular}
\endgroup}
\vspace*{-8pt}
\caption{A comprehensive evaluation of VLMs in egocentric relative FoR with reflected transformation, using an explicit camera perspective (\texttt{cam}) prompt, is conducted. The metrics considered include object hallucination (F1-score), accuracy (Acc), region parsing error ($\varepsilon$), prediction noise ($\eta$), standard deviation ($\sigma$), and consistency ($c$). \vspace*{-15pt}}
\label{tab:consistency}
\end{table}

\vspace*{-5pt}
\subsection{VLMs Fail to Adopt Alternative Frames of Reference Flexibly}
\label{sec:perspective}
We now address the research question: 
\textbf{can VLMs adopt different FoRs when perspectives are explicitly specified to disambiguate spatial expressions?}
We again use \datasetFOR; however, instead of using the no-perspective prompt (\texttt{nop}), we require VLMs to follow one FoR by explicitly specifying the perspective (\texttt{cam}/\texttt{rel}/\texttt{add}) in the textual prompts (Table~\ref{tab:prediction_notation}).
Table~\ref{tab:perspective-main} shows the results in accuracy and $\varepsilon^{\textrm{cos}}$ and the performance compared to when no perspective is specified, and Table~\ref{tab:perspective-full} in the appendix gives the complete evaluation.
We find that all models, including the strong ones like GPT-4o and InternLM-XComposer2, show close-to-chance performance (50\% accuracy) when being prompted to use the intrinsic or addressee-centered relative FoRs. 
Compared to the same probing setup without a perspective specified (\texttt{nop}), we find generally marginal improvements in region parsing error ($\varepsilon$), whereas the accuracy decreases.
Overall, the results indicate that while VLMs can comprehend scenes using egocentric relative FoR, they struggle to adapt flexibly to alternative FoRs. 

\vspace*{-5pt}
\subsection{Spatial Representations in VLMs Are Not Robust and Consistent}
\label{sec:consistency}
In this section, we further ask: 
\textbf{are spatial representations in VLMs robust and consistent?}
The considered metrics include accuracy (Acc), region parsing error ($\varepsilon$), prediction noise ($\eta$), standard deviation ($\sigma$), and consistency ($c$) as defined in Section~\ref{sec:data}.
One commonly considered possibility that VLMs underperform is that they suffer from \textit{object hallucination}, where they misperceive objects in the scenes \citep{li2023evaluating,chen2024multi}.
Following the object probing setups, we prompt the VLMs to inquire about the presence of an existing object and a non-existing object in the scene, and compute the F1-score (Table~\ref{tab:consistency}).
We find that the BLIP models suffer from severe object hallucinations, which contribute to their underperformance in the previous evaluation. 
Many VLMs, despite showing decent performance metrics in terms of spatial understanding and reasoning accuracy, demonstrate a lack of robustness and consistency. 
For example, the spatial opposite consistency ($c^{\textrm{opp}}$) of GPT-4o is not significantly better than 30 random trials.
In contrast, VLMs that have undergone supervised fine-tuning on spatial relation tasks have a more robust and consistent spatial representation. 
For instance, InternLM-XComposer2 and MiniCPM-V (on the \datasetSYMM task, with no object hallucinations) show improved performance.
On the other hand, although GLaMM is mechanistically grounded to objects and exhibits minimal object hallucination, its spatial understanding capability is poor. 
This suggests that improving visual entity grounding helps in recognizing individual objects but does not automatically translate to better spatial understanding between multiple objects.

\vspace*{-5pt}
\subsection{A Cross-lingual and Cross-cultural Evaluation of Frame of Reference}
\label{sec:multilingual}

\begin{figure*}[t]
\centering
\begin{minipage}{0.45\textwidth}
\scalebox{0.85}{
\setlength{\tabcolsep}{4pt}
\renewcommand{\arraystretch}{0.8}
\hspace*{-10pt}
\begin{tabular}{ccccc}
\toprule
\multicolumn{2}{c}{\textbf{Language}}     & \textbf{English} & \textbf{Tamil} & \textbf{Hausa} \\
\cmidrule(r){1-2}\cmidrule(r){3-5}
\multicolumn{2}{c}{Intrinsic}    & 50.9    & 52.0  & 54.0  \\
\cmidrule(r){1-2}\cmidrule(r){3-5}
\multirow{3}{*}{Ego-Rel} & Ref.  & \uline{\sota{35.8}}    & \sota{40.4}  & \sota{41.0}  \\
                         & Rot.  & 57.3    & \uline{55.2}  & 56.1  \\
                         & Tran. & 53.7    & 51.1  & \uline{53.0}  \\
\cmidrule(r){1-2}\cmidrule(r){3-5}
\multirow{3}{*}{Add-Rel} & Ref.  & 58.8    & 52.2  & 52.8  \\
                         & Rot.  & 51.3    & 52.9  & 55.3  \\
                         & Tran. & 56.1    & 56.1  & 56.1  \\
\cmidrule(r){1-2}\cmidrule(r){3-5}
\multicolumn{2}{c}{GPT-4o Prefer}   & Ego-Ref.    & Ego-Ref.  & Ego-Ref.  \\
\multicolumn{2}{c}{Human Prefer}    & Ego-Ref.    & Ego-Rot.  & Ego-Trans.  \\
\bottomrule
\end{tabular}}
\end{minipage}
\hspace*{5pt}
\begin{minipage}{0.5\textwidth}
\centering
\includegraphics[width=1.0\textwidth]{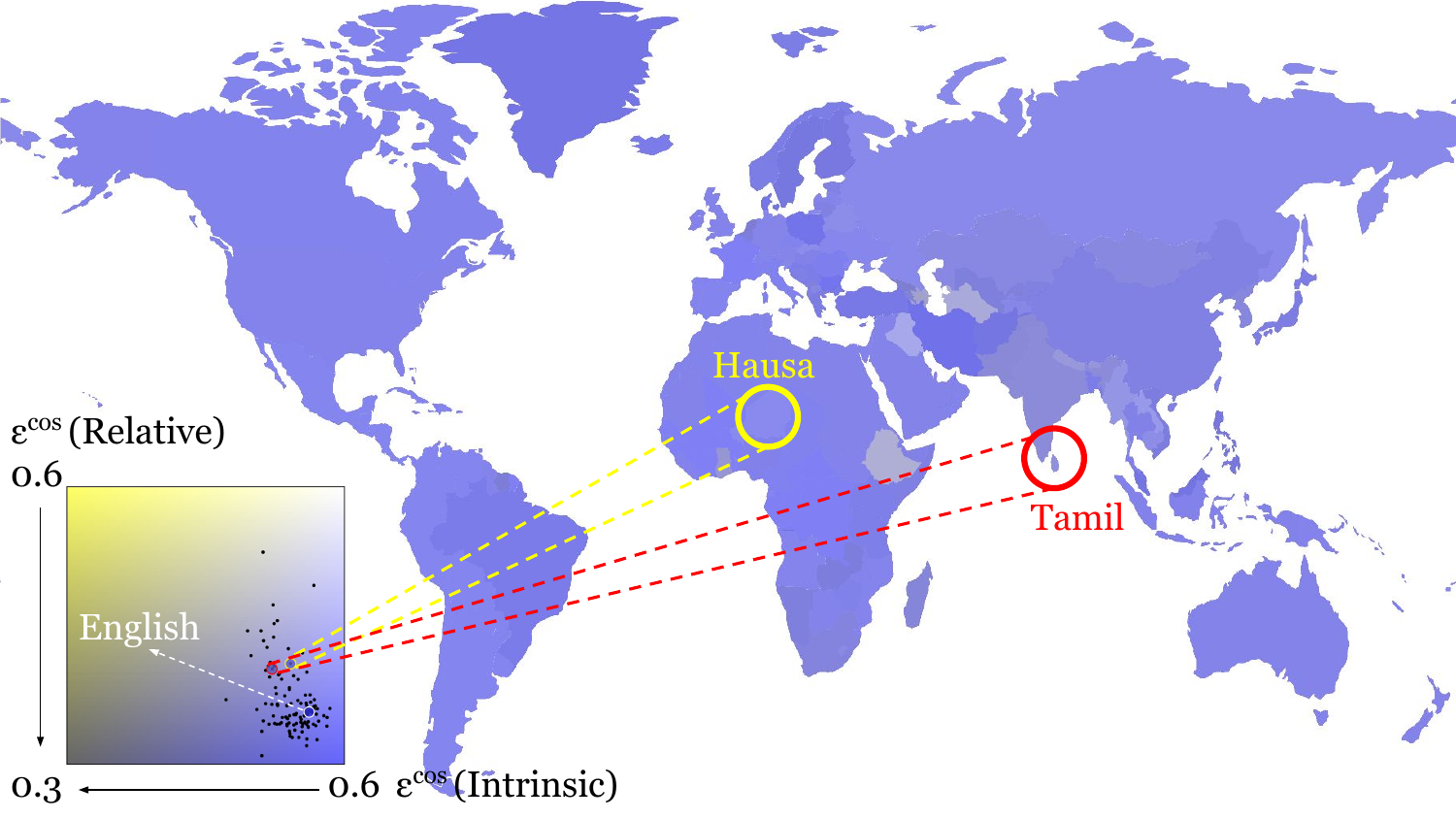}
\end{minipage}
\vspace{-8pt}
\caption{GPT-4o's preferences for intrinsic FoR over the relative FoR across regions. The plot is based on the top three spoken languages in each region, as reported by The World Factbook~\citep{central2009world}, and averages the cosine parsing error ($\varepsilon^{\textrm{cos}},\;\downarrow$), weighted by the speaking population. We present a quantitative comparison of English, Tamil, and Hausa, with the best-performing FoR marked in bold and the convention preferred by human speakers underlined. \vspace{-15pt}}
\label{fig:world_map}
\end{figure*}

All previous experiments are centered around English; however, individuals from diverse multilingual and cultural backgrounds may adopt different preferences and conventions to select their FoR in resolving ambiguities~\citep{majid2004can,o2011spatial,bohnemeyer2014cultural,ogelo2024spatial}.
Our next research question naturally arises: 
\textbf{Do multilingual VLMs faithfully follow the preferences and conventions (associated with different languages) to select the FoR?}
To extend the study of preferred FoR from English to a multilingual setting, we evaluate 109 languages worldwide to investigate whether each language shows a preferred FoR.
We translate the English prompts into the target languages using the Google Cloud Translate API.
Given that the open-source language models either lack strong multilingual capabilities or underperform in previous evaluations, we study this problem on the GPT-4o model \citep{gpt4o}. 
We follow the setup similar to Section~\ref{sec:for}, but only evaluate the images corresponding to the four canonical directions using the \texttt{nop} prompt.
For each language, we compute $\varepsilon^{\textrm{cos}}$ for each FoR and coordinate transformation.
Figure~\ref{fig:world_map} presents a visualization of the world map, displaying the preference of each region for using the (object-centered) intrinsic FoR over the relative FoR, where the latter corresponds to a low $\varepsilon^{\textrm{cos}}$ value.
Table~\ref{tab:preferred_for_multilingual} summarizes the results for all languages tested.

Nearly all tested languages demonstrate a preference towards the relative FoR, except several underrepresented languages, such as Konkani, Kurdish, and Amharic, which exhibit near-random performance without a significant preference.
In Figure~\ref{fig:world_map}, we present a classic comparison between English, Tamil, and Hausa similar to that of \citet{levinson2003space}, with the best-performing FoR marked in bold, and the preferred convention by humans underlined. 
Although human speakers of these languages have different preferred coordinate transformation conventions, the English convention of reflected projection is observed for both Tamil and Hausa. 
Although, for example, Hausa permits an English-like interpretation of front-back relations, this interpretation is generally less favored and may confuse Hausa speakers \citep{hill1982up}.
This raises concerns that English may dominate the FoR preference conventions of other languages in multilingual VLMs.

\vspace*{-8pt}
\section{Discussions}
\vspace*{-5pt}

\subsection{Do vision-language models represent space and how?}
It is insufficient to answer this question by simply querying the model with text-image pairs and comparing the output with a fixed ground truth. 
We must, at least, query the models with awareness of the ambiguity in FoRs, which is essential in determining how the scenes in the physical world are mapped to spatial expressions \citep{levinson2003space}.
Our experiments confirm that many VLMs are equipped with reasonable spatial representations through vision-language training alone; in particular, most VLMs clearly prefer the egocentric relative FoR with reflected projection, aligning with English conventions. 
However, our results also show these representations lack robustness and consistency in a continuous space. 
Similar experimental setups can yield widely varying performance across different spatial relations---for example, GPT-4o shows minimal preferences for the egocentric relative FoR along the sagittal axis but a significant preference along the lateral one (Table~\ref{tab:preferred_for}). 
As a result, VLMs demonstrate unsatisfactory consistency in their spatial performance (Table~\ref{tab:consistency}).
Future work is necessary to improve the consistency and robustness of spatial representations in these models.

Our research also aligns with the growing trend of grounding linguistic analysis in rich modalities representing the real world \citep{chai2018language,shi-2024-learning}.
Language is not text in isolation; its meaning is significantly enriched when grounded in real-world contexts. 
For example, the ambiguity of spatial terms, as discussed in this paper, becomes meaningful only when combined with FoRs, and these FoRs are much more intuitively illustrated when visual cues are available.
We advocate for future work that extends linguistic analysis to more grounded settings.

\subsection{Perspective taking as a prerequisite of human-like spatial reasoning}
Most languages support multiple FoRs.\footnote{Some languages, in very rare cases, have only one available spatial frame of reference. For example, Guugu Yimithirr exclusively uses the absolute FoR~\citep{levinson2003space}.}
The ability to understand and reason about space from a non-egocentric perspective is an important foundation of the Theory of Mind, a basic building block of our situated communication skill that allows us to infer others' mental states \citep{ma2023towards}.
One of our key findings is that VLMs still struggle to adopt alternative FoRs flexibly, even when provided with explicit perspective-taking instructions (Section~\ref{sec:perspective}).
We hypothesize that this phenomenon may come from a reporting bias in the image-text datasets available on the internet---it is natural to take the reflected relative FoR to view images presented on a screen, but this does not always apply in real-world applications. 
To address this issue, we suggest future work extend the current 2D VLMs to the 3D domain, by considering camera poses and multiview data \citep{yang20243d,hao2024ego3dt} for training.

\subsection{Cross-cultural conventions in cross-lingual spatial understanding}
The conventions for resolving spatial ambiguities are not uniform, as individuals from diverse linguistic and cultural backgrounds select their FoR differently. 
Cultural conventions can even be transmitted as individuals are exposed to other languages.
Interestingly, \citet{bohnemeyer2014cultural} found that among native speakers of Indigenous languages (with various preferences in FoRs), those more proficient in Spanish tend to use the reflected relative FoR (Spanish convention) more in their native language. 
This phenomenon has led to their Linguistic Transmission Hypothesis: ``Using any language or linguistic variety, independently of its structures, may facilitate the acquisition of cultural practices of nonlinguistic cognition shared among the speakers of the language.''
Analogously, our experiment raises important concerns that English may dominate the FoR preference conventions of other languages in multilingual VLMs.
This is not surprising, as current training recipes for multilingual multimodal language models heavily rely on machine-translated captions \citep{chen2023pali,geigle2024mblip}.
However, this practice can be problematic.
For instance, Hausa prefers an interpretation where the ``front'' aligns with the English concept of ``back,'' \citep{hill1982up}, where this approach may lead to English conventions overshadowing those of other languages. 
At a high level, this issue is not limited to spatial reasoning---for example, \citet{shi2023language} have demonstrated that English is always the best chain-of-thought language for math reasoning with multilingual LLMs, no matter what language is used for the problem description. 
To enable similar linguistic transmission in AI models, exposure to naturally generated multilingual data is crucial \citep{romero2024cvqa,nguyen2024multilingual}.

\vspace*{-5pt}
\subsection{Limitations and Future Work}
\label{sec:limitation}

\noindent
\textbf{The acceptance regions.}
Using cosine and hemisphere as acceptance regions is analytical but might not capture some human cognitive biases.
In reality, regional angles might not be uniformly distributed per relation, nor are they exactly 90 degrees. 
These angles vary across individuals and cultures \citep{franklin1995parsing}.

\noindent
\textbf{Spatial relations.}
This work primarily focuses on the most basic types of spatial relations (front-back and left-right).
However, many other relations exist, such as \textit{away from} and \textit{near} \citep{logan1996computational,liu2023visualspatial}. 
Additionally, not all languages possess terms for ``left,'' ``right,'' ``front,'' and ``back.'' Some languages, like Guugu Yimithirr, use only absolute frames of reference instead \citep{levinson2003space}.

\noindent
\textbf{Camera angle and occlusion.}
Currently, there is minimal occlusion, and the camera angle is high. 
Languages may differ in whether the speaker emphasizes these factors, such as the preference to use ``behind'' in cases of occlusion \citep{levinson2003space}.

\noindent
\textbf{Pragmatic aspect of spatial cognition.}
Many conversational and pragmatic aspects of spatial cognition are simplified in this work, such as F-Formation \citep{kendon2010spacing} and human-robot interaction \citep{liu2010ambiguities}. 
For example, in human-robot interaction settings, users prefer an addressee-centered frame of reference to facilitate the robot's comprehension of referents \citep{moratz2006spatial}.

\noindent
\textbf{Multilingual prompts.}
In this work, we used machine-translated text to construct the multilingual portion of the dataset. 
Although we verified data quality through back translation, incorporating human annotations in the future would be a valuable future step.
\section*{Acknowledgments}

This work was supported in part by NSF IIS-1949634, NSF SES-2128623, NSERC RGPIN-2024-04395, ONR N00014-23-1-2417, and the Microsoft Accelerate Foundation Models Research (AFMR) grant program.
Ziqiao Ma is supported in part by the Weinberg Cognitive Science Fellowship.
Any opinions, findings, conclusions, or recommendations expressed in this material are those of the authors and do not necessarily reflect the views of these funding agencies.
The authors would like to thank Yinpei Dai, Run Peng, Jung-Chun Liu, and Xuejun Zhang for proofreading and feedback.

\bibliographystyle{iclr2025_conference}
\bibliography{references}

\appendix
\clearpage

\section{Dataset and Metric Details}
\label{sec:appendix_data}

\subsection{Dataset Configurations}

The entire data generation pipeline produces 720 English test cases in \datasetSYMM, and 57.6k English test cases in \datasetFOR.
For \datasetSYMM: 1 object combination $\times$ 5 variants $\times$ 4 relations $\times$ 36 angles = 720 test cases. 
For \datasetFOR: 20 object combinations $\times$ 5 variants $\times$ 4 relations $\times$ 36 angles $\times$ 4 prompts = 57,600 test cases.
The table below lists all possible variants and configurations for the dataset, and we describe our dataset configuration in detail as follows.

\begin{table}[H]
    \setlength{\tabcolsep}{2pt}
    \renewcommand{\arraystretch}{0.9}
    \centering
    \vspace*{-5pt}
    \scalebox{0.875}{
    \begin{tabular}{l p{12.5cm}}
        \toprule
        \textbf{Test Case Setup} & \textbf{Possible Variants} \\
        \cmidrule(r){1-1}\cmidrule(r){2-2}
        \multirow{5}{*}{Scene $\mathcal{S}$} & \datasetSYMM: \textbf{Relatum}: \textit{red ball}; \textbf{Referent}: \textit{blue ball}; 36 samples uniformly collected along a rotational path.  \\
        \cmidrule(r){2-2}
                                             & \datasetFOR: \textbf{Relatum}: \textit{basketball}; \textbf{Referent}: \textit{horse}, \textit{car}, \textit{bench}, \textit{laptop}, \textit{rubber duck}, \textit{chair}, \textit{dog}, \textit{sofa}, \textit{bed}, \textit{bicycle}; \textbf{Addressee}: woman; 36 samples uniformly collected along a rotational path.  \\
        \cmidrule(r){1-1}\cmidrule(r){2-2}
       Spatial Relation $\mathcal{R}$ & \textit{to the left of}, \textit{to the right of}, \textit{in front of}, \textit{behind}  \\
        \cmidrule(r){1-1}\cmidrule(r){2-2}
       Frame of Reference $\mathcal{F}$ & \textit{egocentric relative,} \textit{addressee-centered relative}, \textit{object-centered intrinsic}  \\
        \cmidrule(r){1-1}\cmidrule(r){2-2}
       Language $\mathcal{L}$ & See Table~\ref{tab:preferred_for_multilingual}.  \\
        \bottomrule
    \end{tabular}}    
    \vspace*{-5pt}
    \caption{A test case in \dataset is defined as a 4-tuple in $\mathcal{S} \times \mathcal{R} \times \left(\mathcal{F}\cup \{\emptyset\}\right) \times \mathcal{L}$. This table enumerates all possible variants and configurations of the dataset. \vspace{-12pt}}
    \label{tab:data_config}
\end{table}

\subsection{List of Evaluated Languages}
We started with 132 candidate languages supported by Google Translate API.\footnote{\url{https://cloud.google.com/translate}}
We removed 23 languages from our multilingual evaluation due to their failure to adhere to instructions for generating ``yes'' and ``no'' predictions, or because they did not pass the back-translation test for quality control: Aymara, Bambara, Croatian, Dhivehi, Dogri, Ewe, Guarani, Hmong, Kyrgyz, Luganda, Malayalam, Meiteilon (Manipuri), Mizo, Odia (Oriya), Punjabi, Quechua, Samoan, Tatar, Telugu, Tigrinya, Uyghur, Xhosa, Yoruba.

\subsection{Visualizations of Region Parsing Error}

\begin{figure}[H]
  \centering
  \begin{subfigure}[b]{0.19\textwidth}
    \centering
    \includegraphics[width=\textwidth]{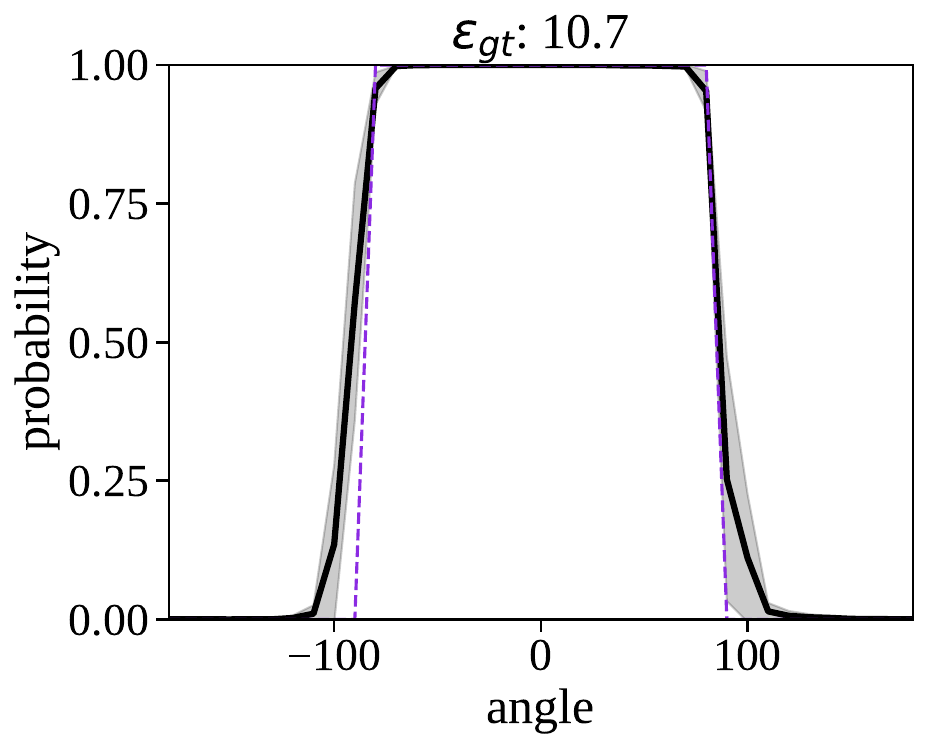}
    \caption{$\varepsilon^{\textrm{hemi}}=10.7$}
  \end{subfigure}
  \begin{subfigure}[b]{0.19\textwidth}
    \centering
    \includegraphics[width=\textwidth]{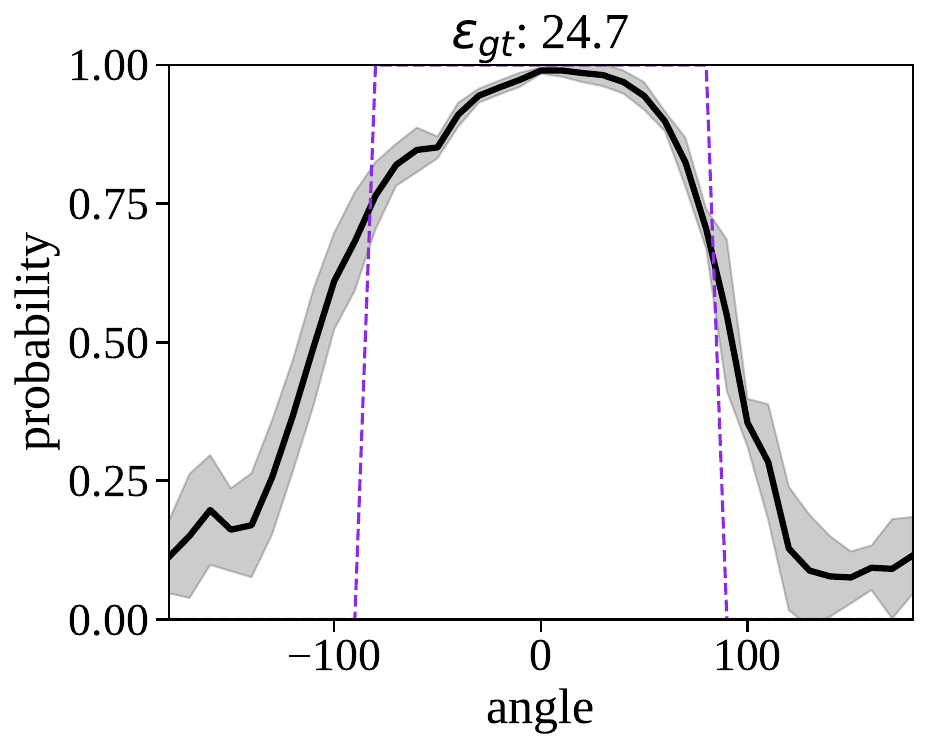}
    \caption{$\varepsilon^{\textrm{hemi}}=24.7$}
  \end{subfigure}
  \begin{subfigure}[b]{0.19\textwidth}
    \centering
    \includegraphics[width=\textwidth]{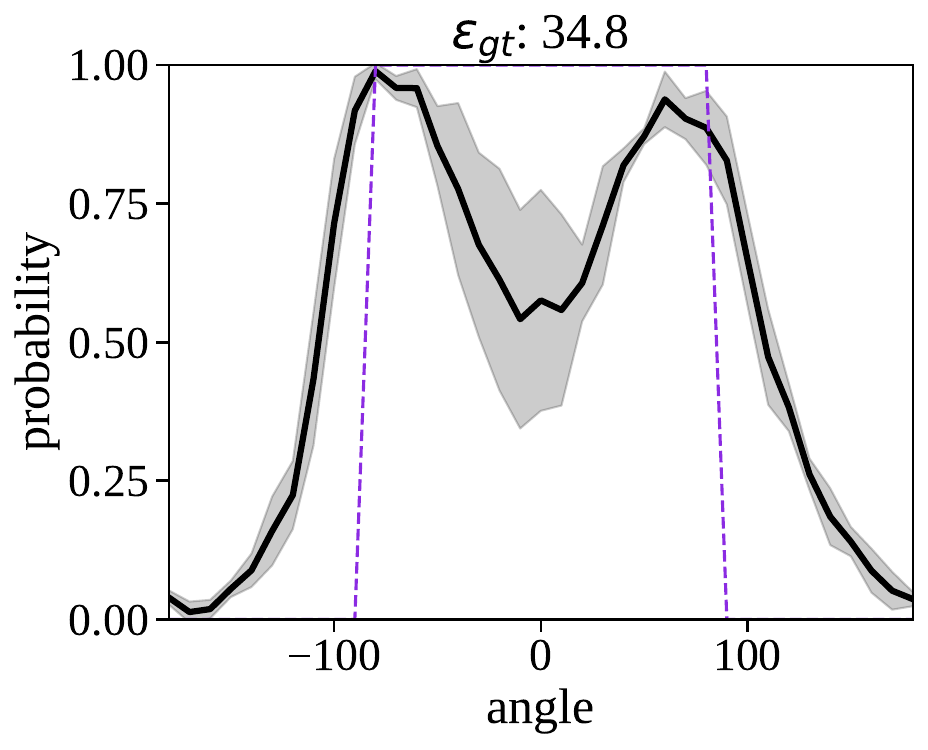}
    \caption{$\varepsilon^{\textrm{hemi}}=34.8$}
  \end{subfigure}
  \begin{subfigure}[b]{0.19\textwidth}
    \centering
    \includegraphics[width=\textwidth]{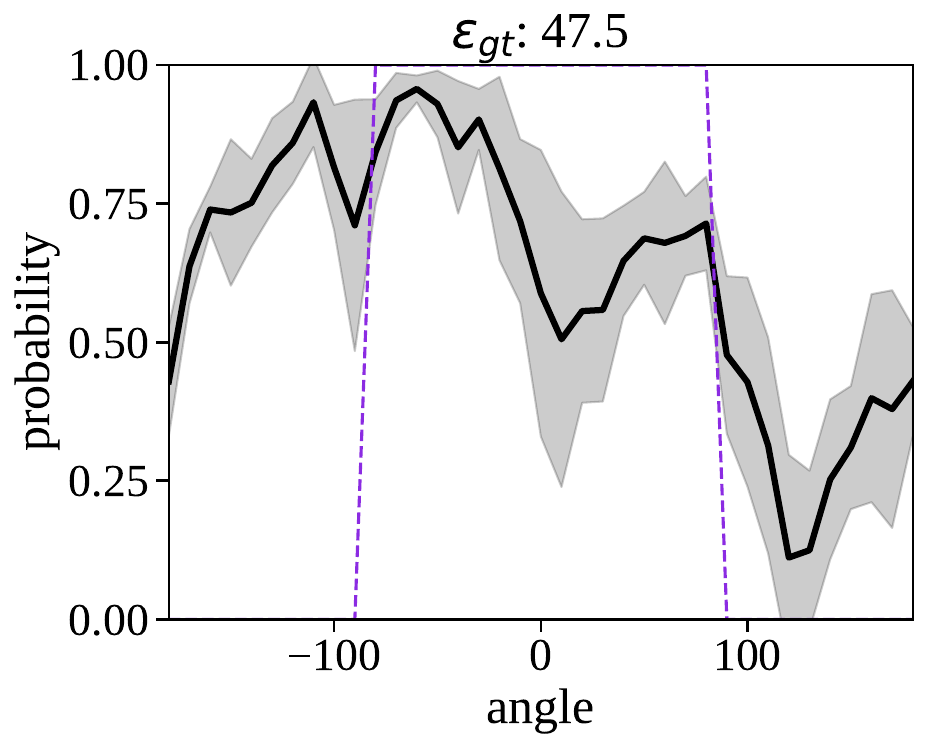}
    \caption{$\varepsilon^{\textrm{hemi}}=47.2$}
  \end{subfigure}
  \begin{subfigure}[b]{0.19\textwidth}
    \centering
    \includegraphics[width=\textwidth]{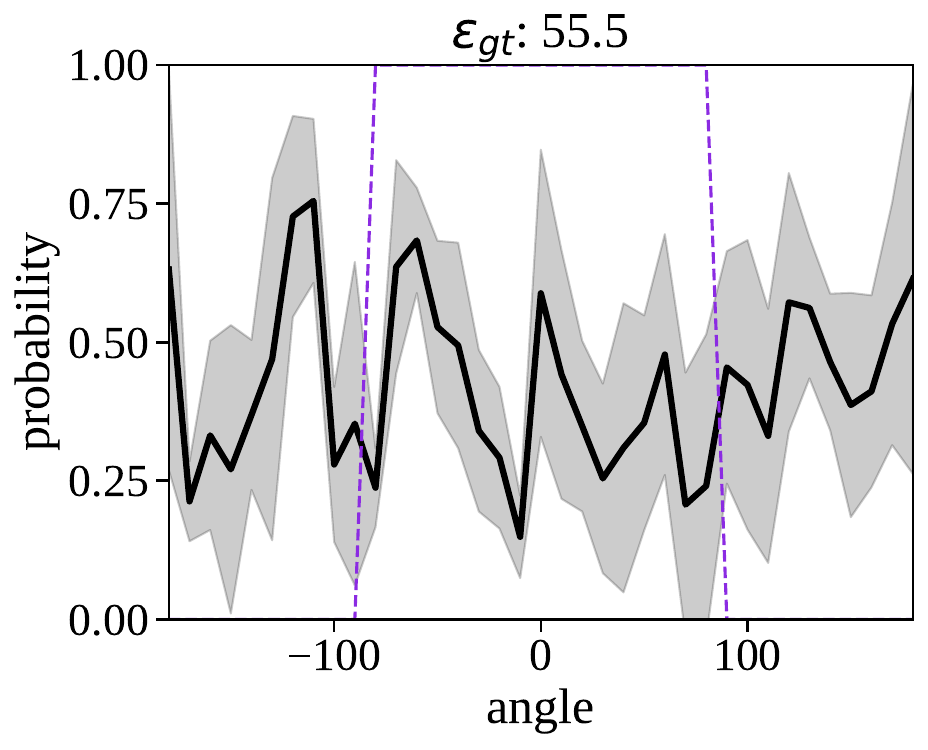}
    \caption{$\varepsilon^{\textrm{hemi}}=55.5$}
  \end{subfigure}

  \vspace{0.2cm}
  
  \begin{subfigure}[b]{0.19\textwidth}
    \centering
    \includegraphics[width=\textwidth]{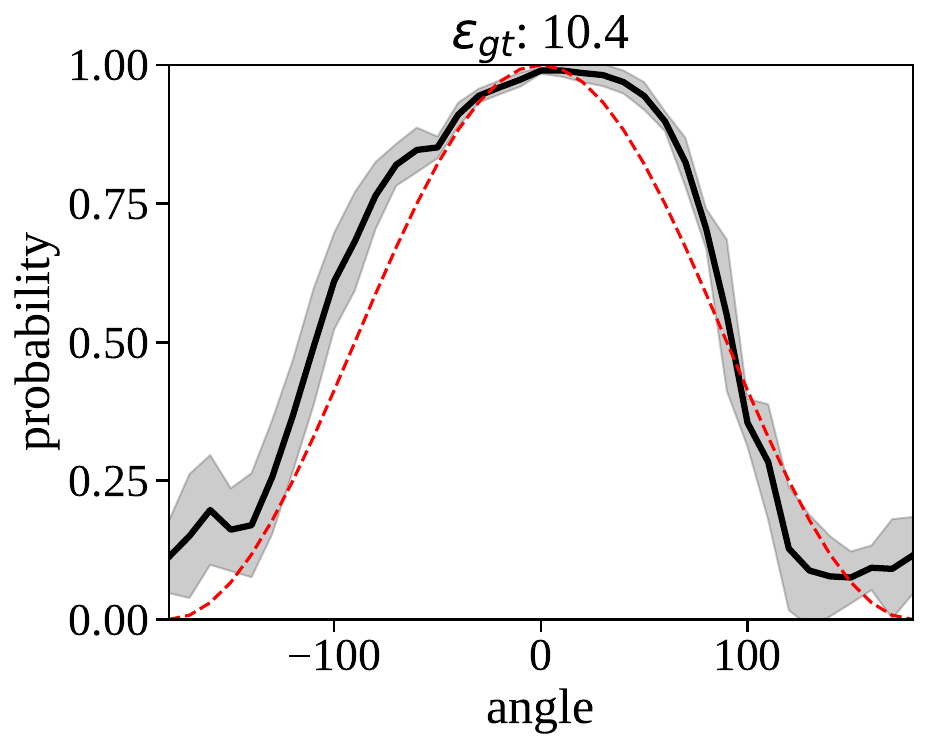}
    \caption{$\varepsilon^{\textrm{cos}}=10.4$}
  \end{subfigure}
  \begin{subfigure}[b]{0.19\textwidth}
    \centering
    \includegraphics[width=\textwidth]{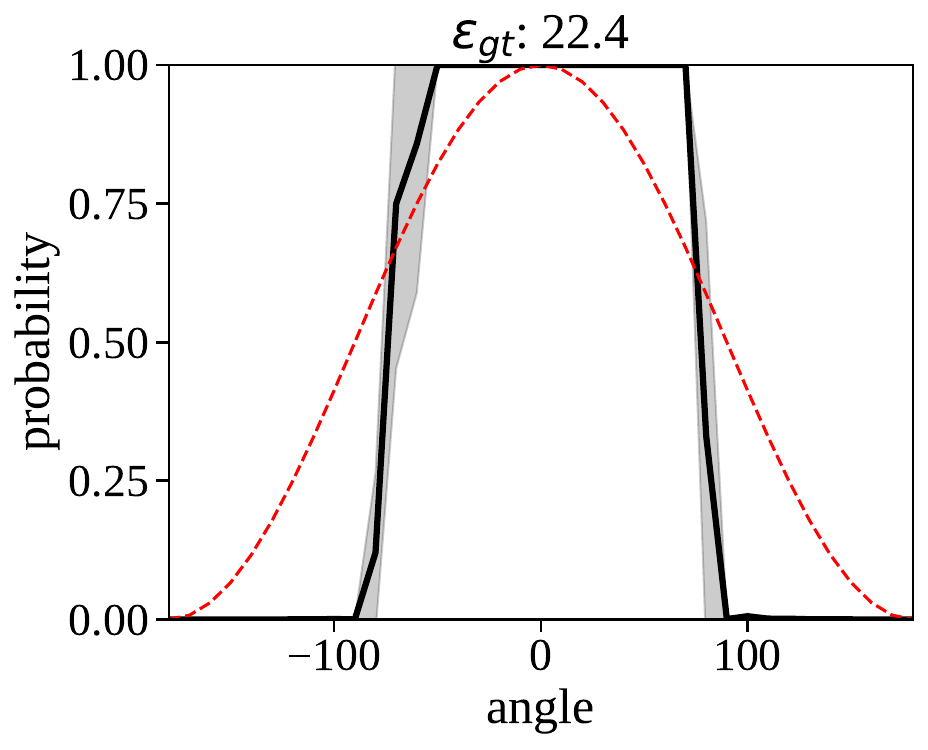}
    \caption{$\varepsilon^{\textrm{cos}}=22.4$}
  \end{subfigure}
  \begin{subfigure}[b]{0.19\textwidth}
    \centering
    \includegraphics[width=\textwidth]{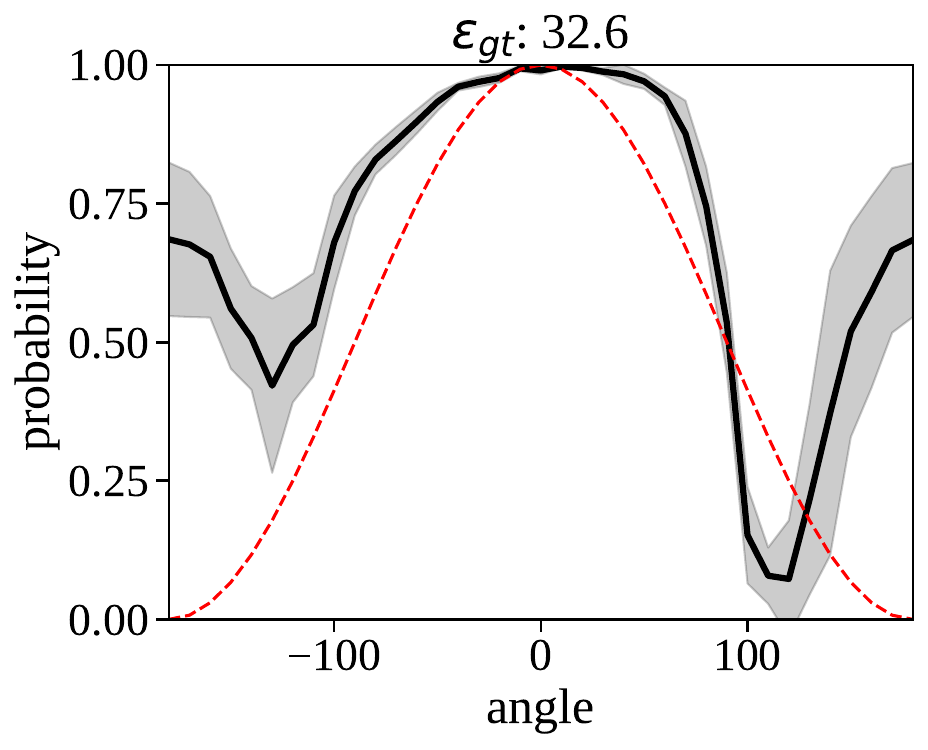}
    \caption{$\varepsilon^{\textrm{cos}}=32.6$}
  \end{subfigure}
  \begin{subfigure}[b]{0.19\textwidth}
    \centering
    \includegraphics[width=\textwidth]{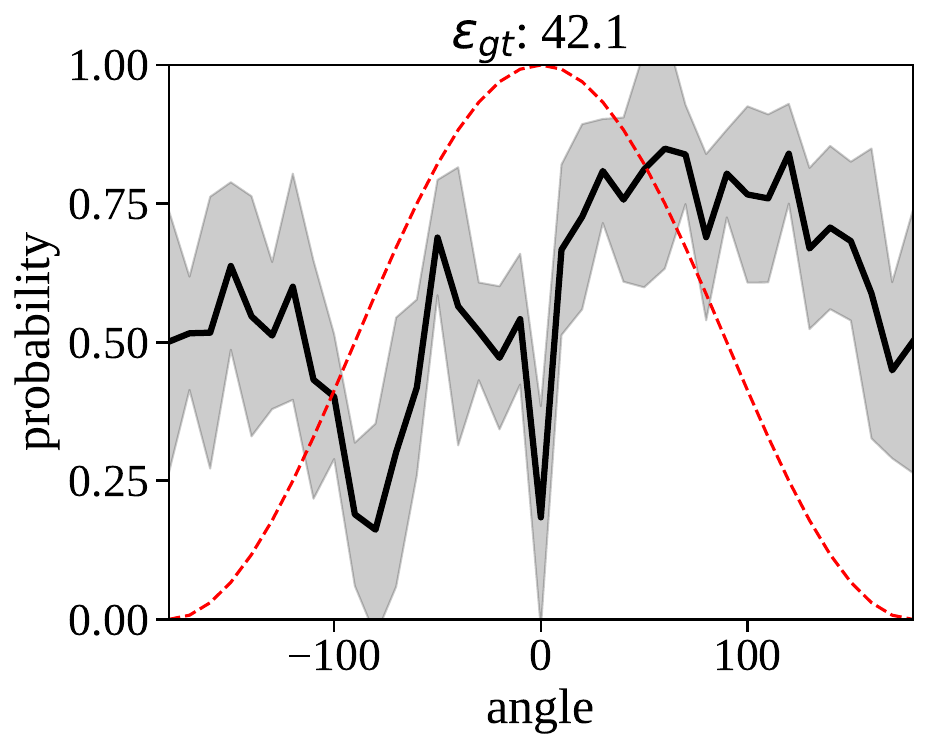}
    \caption{$\varepsilon^{\textrm{cos}}=42.1$}
  \end{subfigure}
  \begin{subfigure}[b]{0.19\textwidth}
    \centering
    \includegraphics[width=\textwidth]{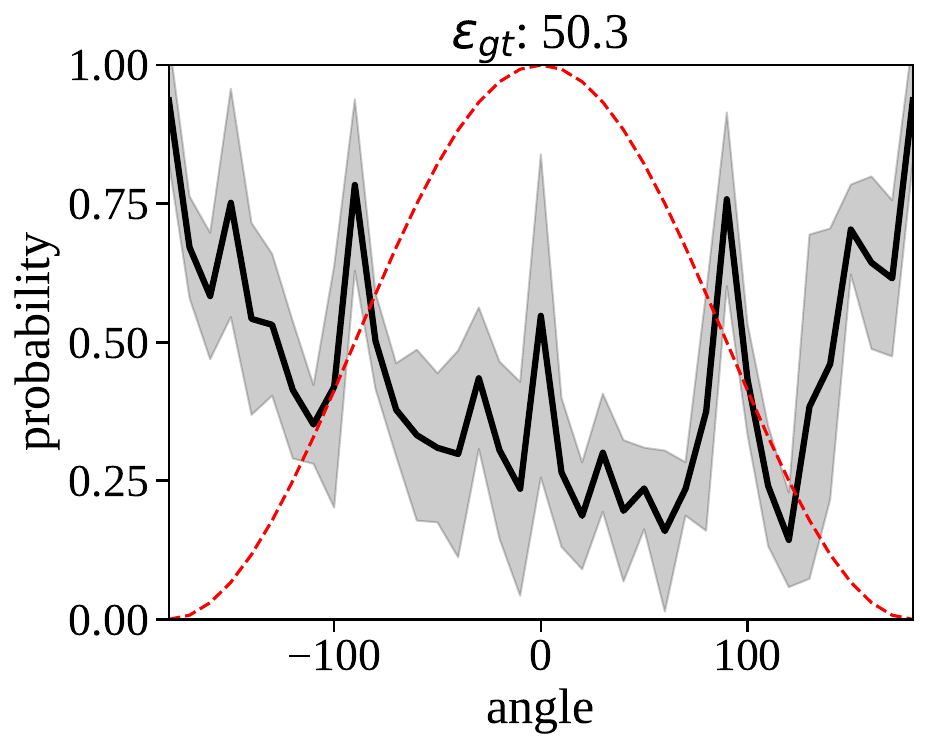}
    \caption{$\varepsilon^{\textrm{cos}}=50.3$}
  \end{subfigure}
  \vspace{-5pt}
  \caption{$\varepsilon$ visualization: (a-e) correspond to $\varepsilon^{\textrm{hemi}}$, and (f-j) correspond to $\varepsilon^{\textrm{cos}}$. \vspace{-15pt}}
  \label{fig:epsilon}
\end{figure}

\section{Addendum to Results}

\begin{table}[!t]
    \scalebox{0.875}{
    \begingroup
    \setlength{\tabcolsep}{1.3pt}
    \renewcommand{\arraystretch}{0.8}
    \hspace*{-8pt}
\begin{tabular}{lRRRRRRRRRRRR}
    \toprule
    \multicolumn{1}{l}{} & \multicolumn{6}{c}{\texttt{Back}}                                                                                                                                                                                              & \multicolumn{6}{c}{\texttt{Front}}                                                                                                                                                                                                 \\
    \cmidrule(r){2-7}\cmidrule(r){8-13}
    \multicolumn{1}{l}{} & \multicolumn{3}{c}{Same}                                                                                  & \multicolumn{3}{c}{Reversed}                                                                              & \multicolumn{3}{c}{Same}                                                                                  & \multicolumn{3}{c}{Reversed}                                                                                  \\
    \cmidrule(r){2-4}\cmidrule(r){5-7}\cmidrule(r){8-10}\cmidrule(r){11-13}
    \multicolumn{1}{l}{} & Acc\% & $\varepsilon^{\textrm{hemi}}_{\times 10^2}$ & $\varepsilon^{\textrm{cos}}_{\times 10^2}$ & Acc\% & $\varepsilon^{\textrm{hemi}}_{\times 10^2}$ & $\varepsilon^{\textrm{cos}}_{\times 10^2}$ & Acc\% & $\varepsilon^{\textrm{hemi}}_{\times 10^2}$ & $\varepsilon^{\textrm{cos}}_{\times 10^2}$ & Acc\%     & $\varepsilon^{\textrm{hemi}}_{\times 10^2}$ & $\varepsilon^{\textrm{cos}}_{\times 10^2}$ \\
\bottomrule
    \end{tabular}
\endgroup}
\vspace*{-5pt}
\caption{The full results for testing the preferred coordinate transformation mapping from the viewer to the relatum in the relative frame of reference. \vspace*{-15pt}}
\label{tab:transform-full}
\end{table}
\begin{table}[!t]
    \scalebox{0.8}{
    \begingroup
    \setlength{\tabcolsep}{0.7pt}
    \renewcommand{\arraystretch}{0.8}
    \hspace*{-15pt}
    \begin{tabular}{lrrrrrrrrrrrrrrrrrr}
    \toprule
    \multicolumn{1}{l}{} & \multicolumn{9}{c}{\texttt{Back}}                                                                                                                                                                                                                                                                                                          & \multicolumn{9}{c}{\texttt{Front}}                                                                                                                                                                                                                                                                                                         \\
    \cmidrule(r){2-10}\cmidrule(r){11-19}
    \multicolumn{1}{l}{} & \multicolumn{3}{c}{Egocentric}                                                                                & \multicolumn{3}{c}{Intrinsic}                                                                               & \multicolumn{3}{c}{Addressee}                                                                             & \multicolumn{3}{c}{Egocentric}                                                                                & \multicolumn{3}{c}{Intrinsic}                                                                               & \multicolumn{3}{c}{Addressee}                                                                             \\
\bottomrule
    \end{tabular}
\endgroup}
\vspace*{-10pt}
\caption{The full results for testing the preferred frame of reference in VLMs. \vspace*{-15pt}}
\label{tab:for-full}
\end{table}
\begin{table}[!t]
    \scalebox{0.8}{
    \begingroup
    \setlength{\tabcolsep}{0.7pt}
    \renewcommand{\arraystretch}{0.8}
    \hspace*{-15pt}
    \begin{tabular}{lrrrrrrrrrrrrrrrrrr}
    \toprule
    \multicolumn{1}{l}{} & \multicolumn{9}{c}{\texttt{Back}}                                                                                                                                                                                                                                                                                                          & \multicolumn{9}{c}{\texttt{Front}}                                                                                                                                                                                                                                                                                                         \\
    \cmidrule(r){2-10}\cmidrule(r){11-19}
    \multicolumn{1}{l}{} & \multicolumn{3}{c}{Egocentric}                                                                                & \multicolumn{3}{c}{Intrinsic}                                                                               & \multicolumn{3}{c}{Addressee}                                                                             & \multicolumn{3}{c}{Egocentric}                                                                                & \multicolumn{3}{c}{Intrinsic}                                                                               & \multicolumn{3}{c}{Addressee}                                                                             \\
\bottomrule
    \end{tabular}
\endgroup}
\vspace*{-10pt}
\caption{The full results for benchmarking perspective-taking performance in VLMs. \vspace*{-15pt}}
\label{tab:perspective-full}
\end{table}
\begin{table}[!t]
    \scalebox{0.8}{
    \begingroup
    \setlength{\tabcolsep}{2pt}
    \renewcommand{\arraystretch}{0.9}
    \hspace*{-25pt}
    \begin{tabular}{rlcccccccrlccccccc}
    \toprule
    \multirow{2}{*}{Code} & \multirow{2}{*}{Language} & \multirow{2}{*}{Intrinsic} & \multicolumn{3}{c}{Egocentric} & \multicolumn{3}{c}{Addressee} & \multirow{2}{*}{Code} & \multirow{2}{*}{Language} & \multirow{2}{*}{Intrinsic} & \multicolumn{3}{c}{Egocentric} & \multicolumn{3}{c}{Addressee} \\
    \cmidrule(r){4-6}\cmidrule(r){7-9}\cmidrule(r){13-15}\cmidrule(r){16-18}
                          &                           &                            & Ref.     & Rot.     & Tran.    & Ref.     & Rot.    & Tran.    &                       &                           &                            & Ref.     & Rot.     & Tran.    & Ref.     & Rot.    & Tran.    \\
    \cmidrule(r){1-2}\cmidrule(r){3-3}\cmidrule(r){4-6}\cmidrule(r){7-9}\cmidrule(r){10-11}\cmidrule(r){12-12}\cmidrule(r){13-15}\cmidrule(r){16-18}
    af                    & Afrikaans                 & 50.9                       & 33.7     & 57.8     & 49.2     & 56.6     & 55.5    & 57.5     & ku                    & Kurdish                   & 56.5                       & 49.5                 & 54.4                 & 53.1                 & 53.1                 & 55.2                 & 54.0                 \\
    ak                    & Akan                      & 51.8                       & 39.6     & 52.2     & 48.8     & 50.4     & 50.6    & 53.8     & la                    & Latin                     & 52.2                       & 43.9                 & 49.8                 & 55.5                 & 55.1                 & 50.8                 & 56.6                 \\
    am                    & Amharic                   & 52.1                       & 47.4     & 60.7     & 50.9     & 56.8     & 54.2    & 57.6     & lb                    & Luxembourgish             & 54.7                       & 35.6                 & 57.6                 & 50.3                 & 58.6                 & 53.0                 & 59.9                 \\
\bottomrule
    \end{tabular}
\endgroup}
\vspace*{-5pt}
\caption{The full results for the cross-lingual and cross-cultural evaluation of the preferred frame of reference in VLMs.\vspace*{-15pt}}
\label{tab:preferred_for_multilingual}
\end{table}

\begin{figure*}[ht!]
  \centering
  \vspace*{-30pt}
  \makebox[\textwidth][c]{
  \begin{tabular}{ccccc}
  & \hphantom{aa} Left & \hphantom{aa} Right & \hphantom{aa} Front & \hphantom{aa} Back \\
  \multirow{-8}{*}{\rotatebox[origin=c]{90}{\scriptsize InstructBLIP7B}} & \includegraphics[width=0.2\linewidth]{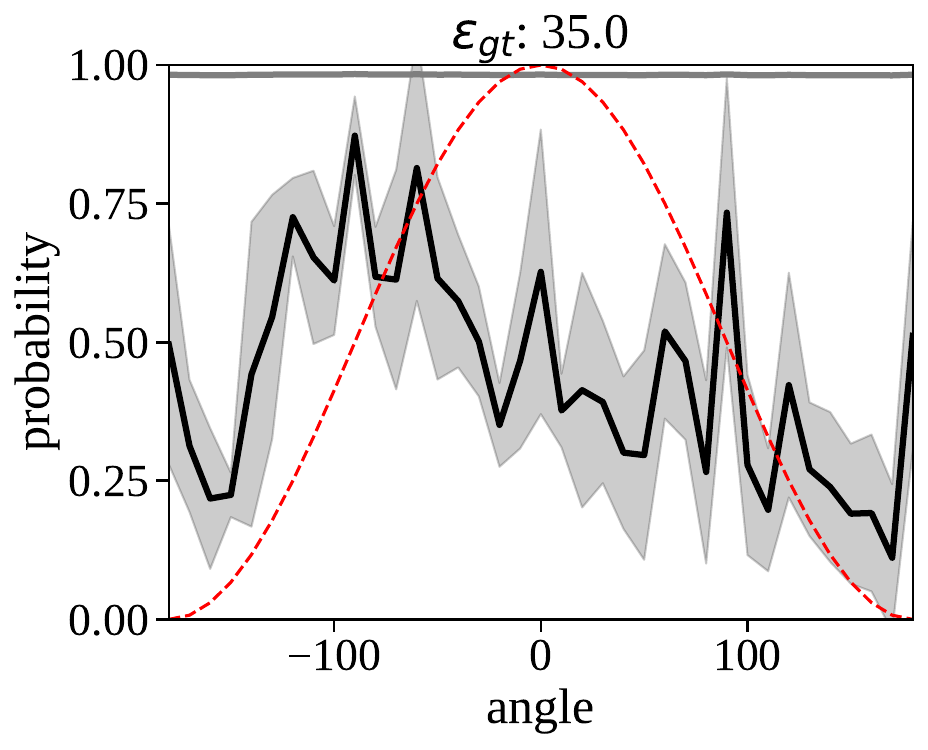} & \includegraphics[width=0.2\linewidth]{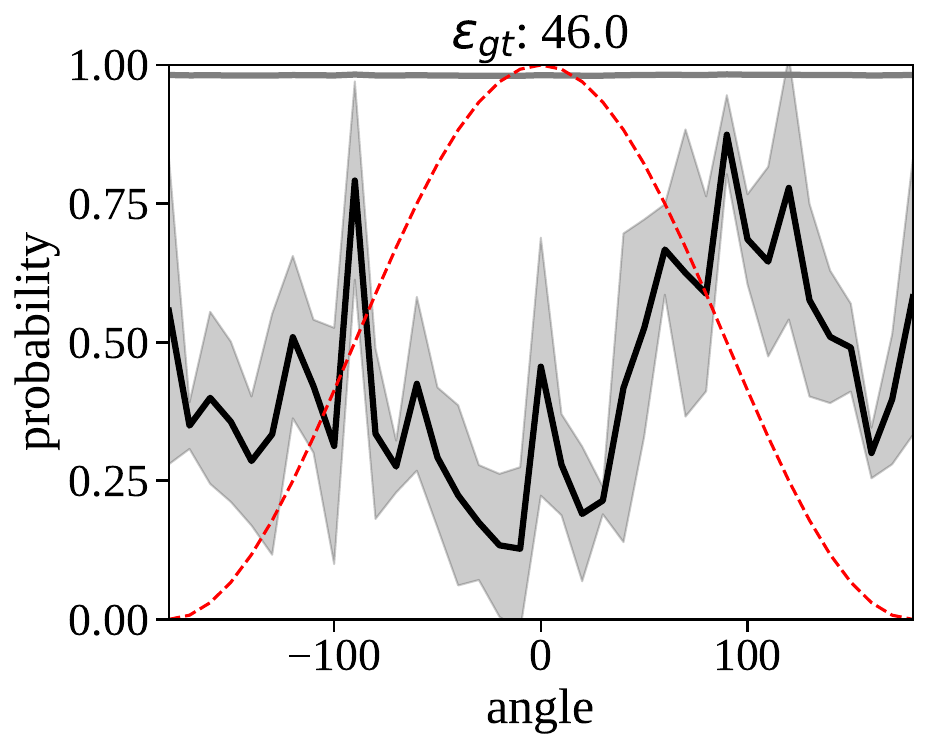} & \includegraphics[width=0.2\linewidth]{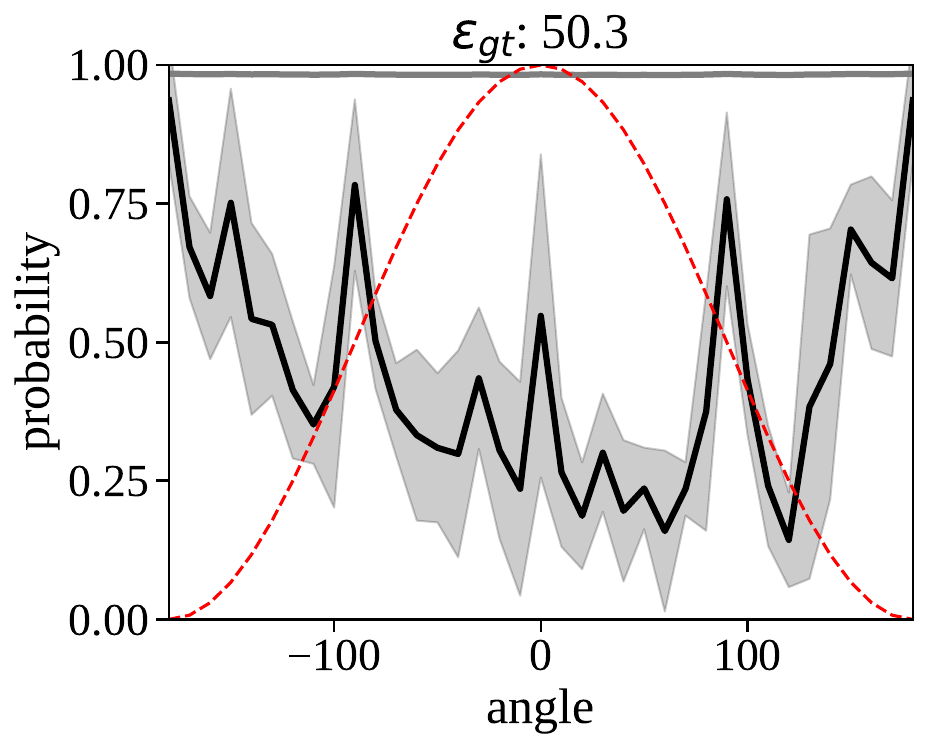} & \includegraphics[width=0.2\linewidth]{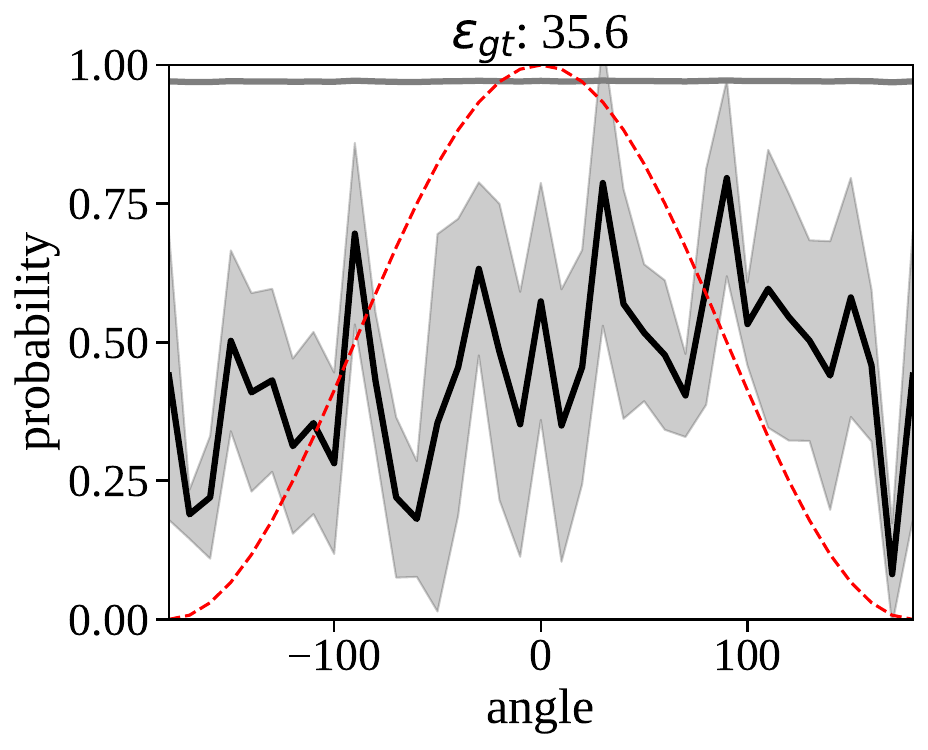} \\
\multirow{-8}{*}{\rotatebox[origin=c]{90}{\scriptsize InstructBLIP13B}} & \includegraphics[width=0.2\linewidth]{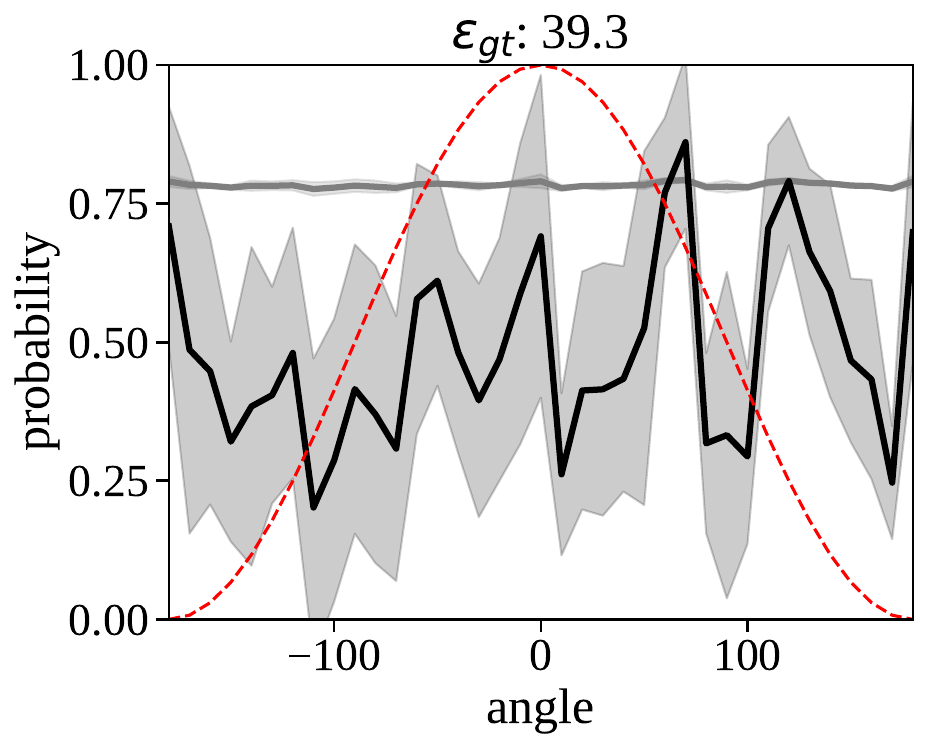} & \includegraphics[width=0.2\linewidth]{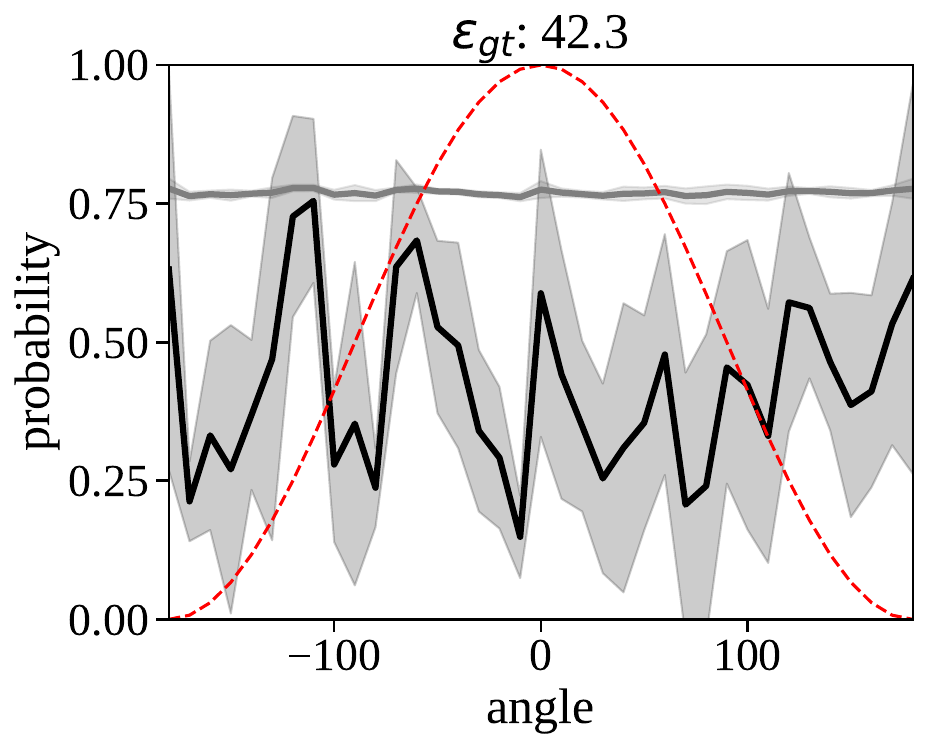} & \includegraphics[width=0.2\linewidth]{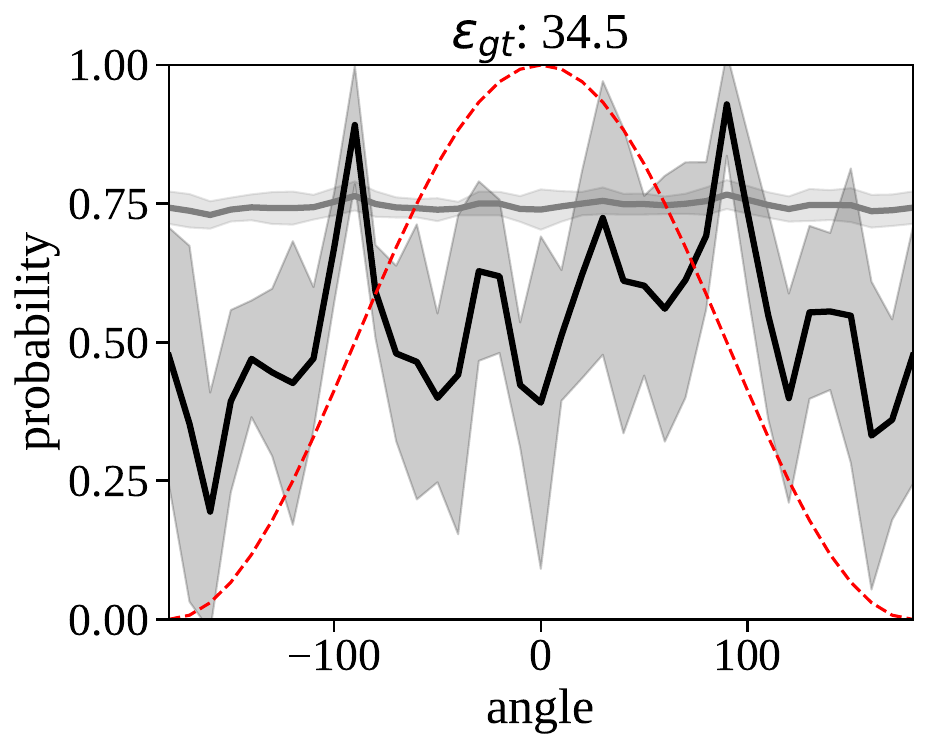} & \includegraphics[width=0.2\linewidth]{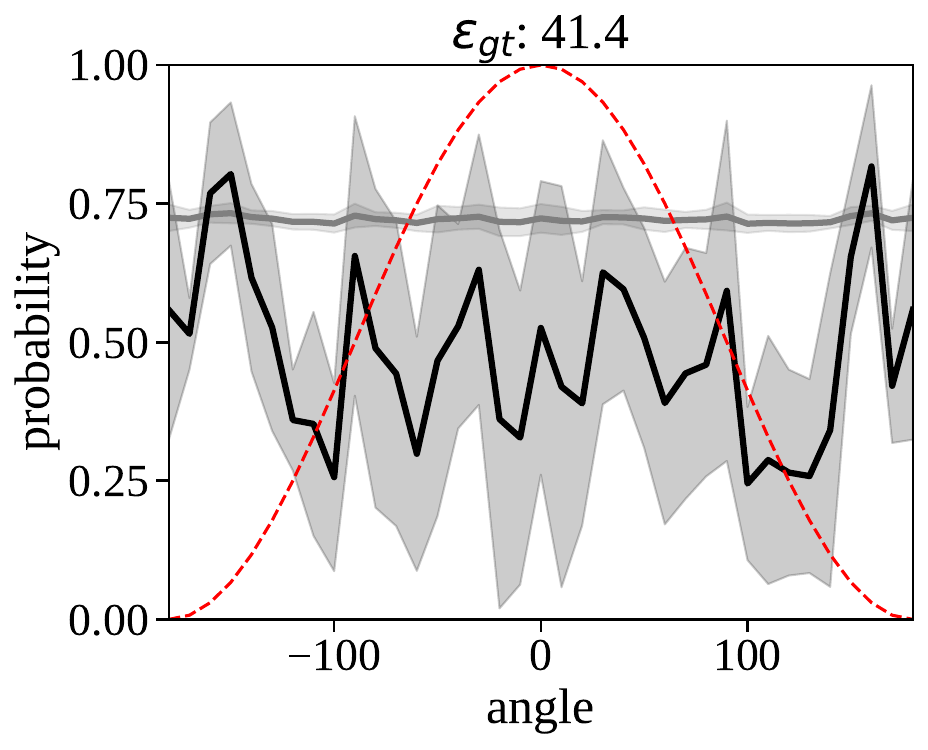} \\
\multirow{-8}{*}{\rotatebox[origin=c]{90}{\scriptsize MBLIPBLOOMZ7B}} & \includegraphics[width=0.2\linewidth]{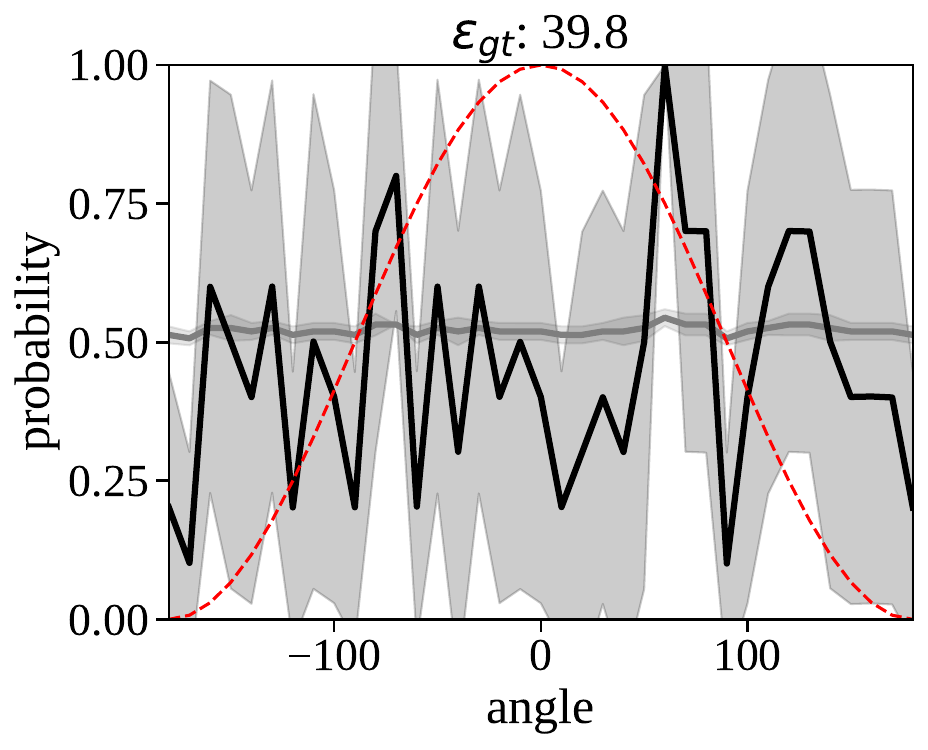} & \includegraphics[width=0.2\linewidth]{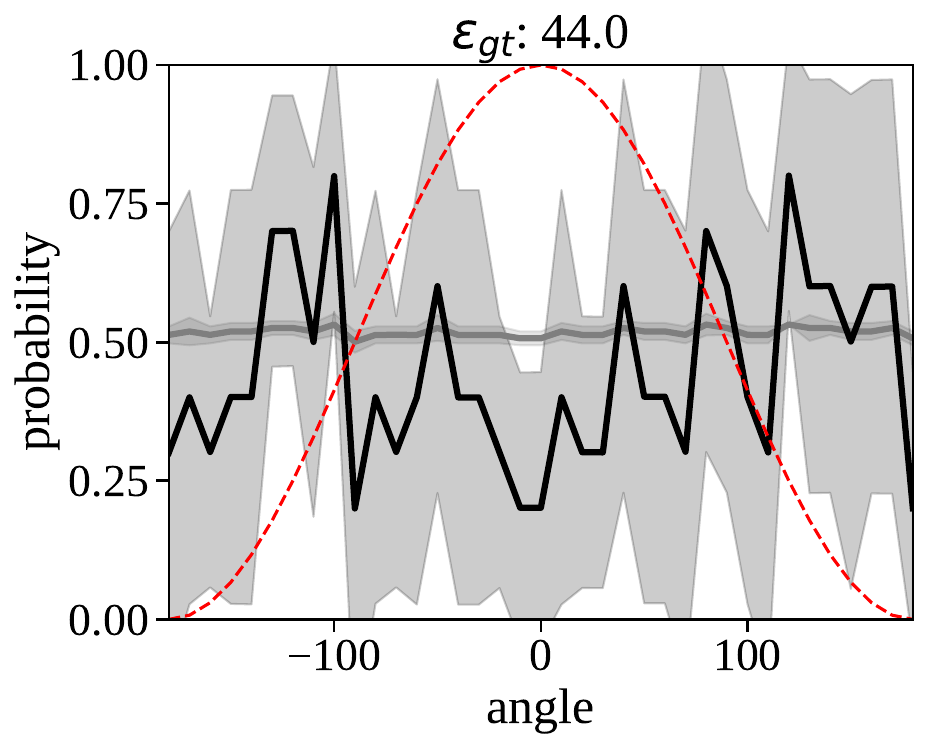} & \includegraphics[width=0.2\linewidth]{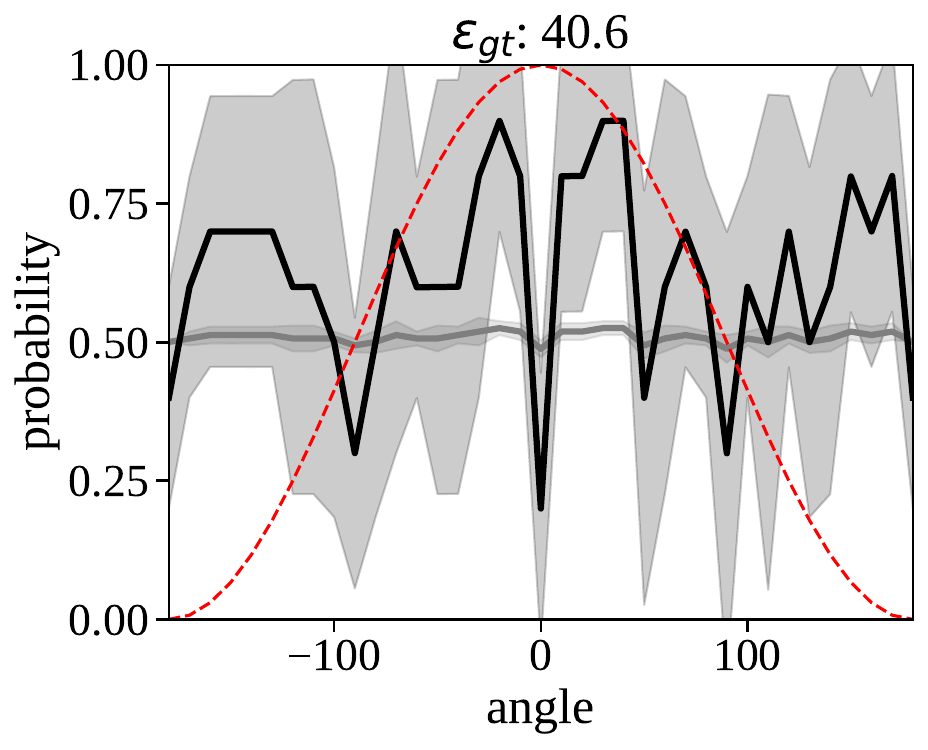} & \includegraphics[width=0.2\linewidth]{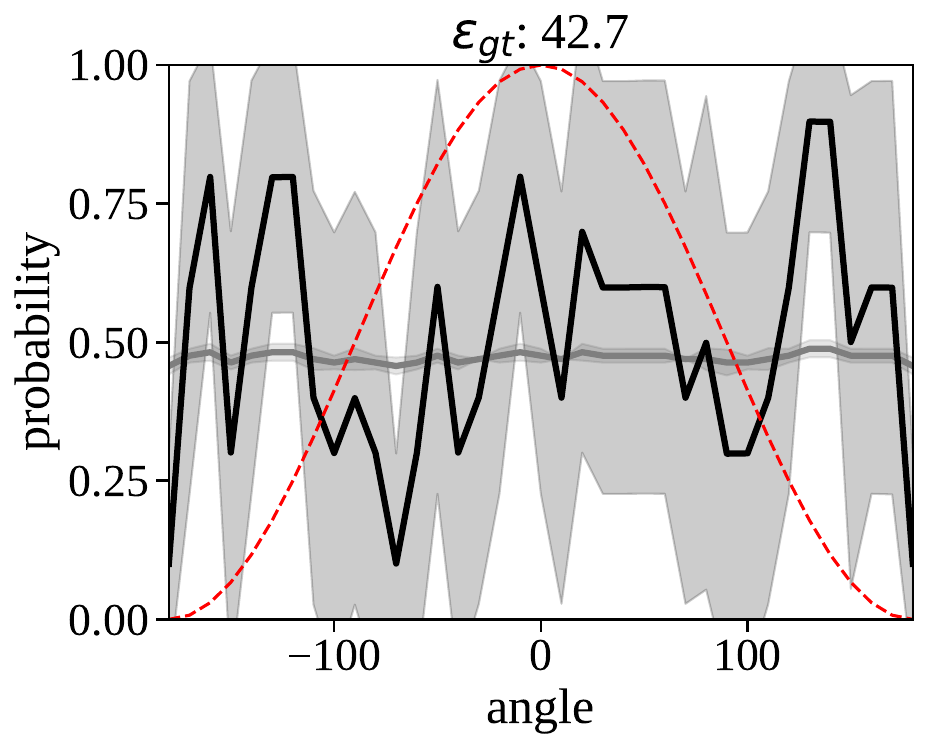} \\
\multirow{-8}{*}{\rotatebox[origin=c]{90}{\scriptsize GLaMM}} & \includegraphics[width=0.2\linewidth]{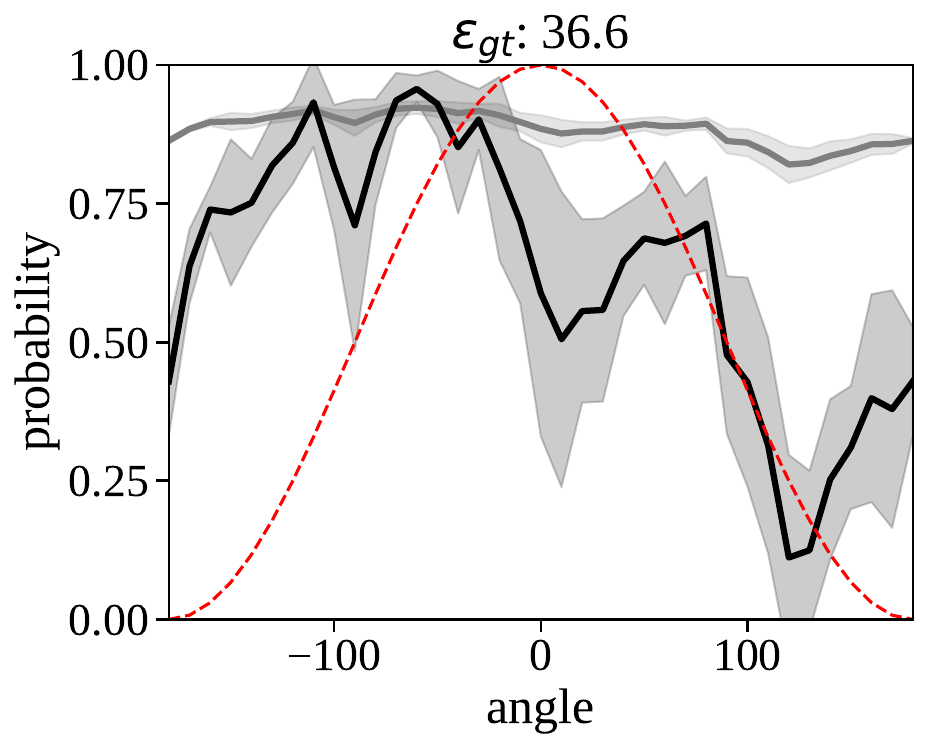} & \includegraphics[width=0.2\linewidth]{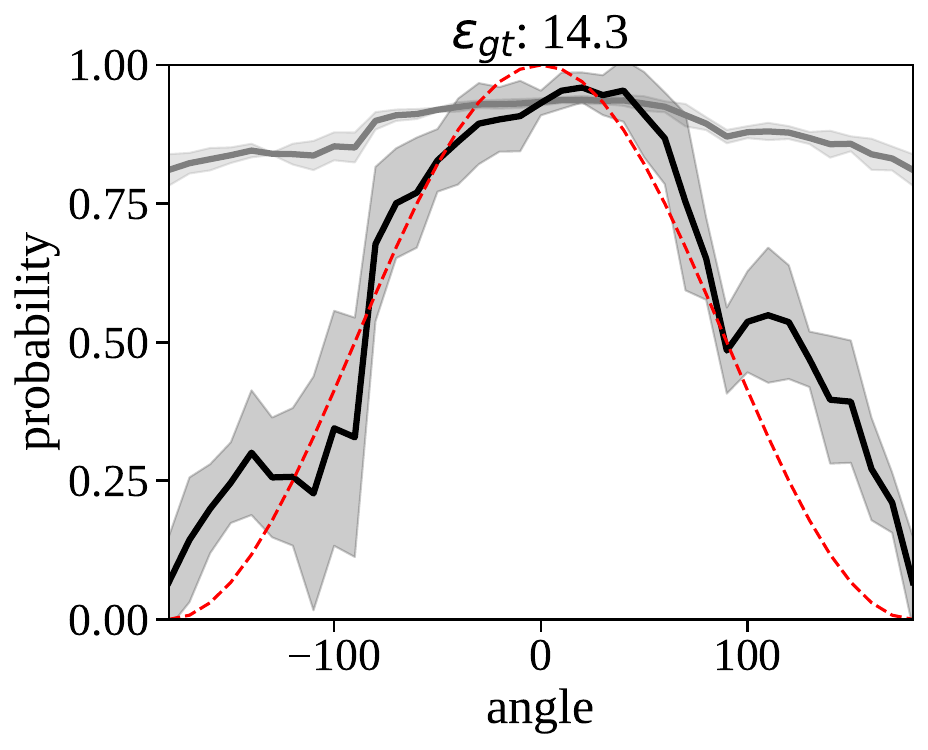} & \includegraphics[width=0.2\linewidth]{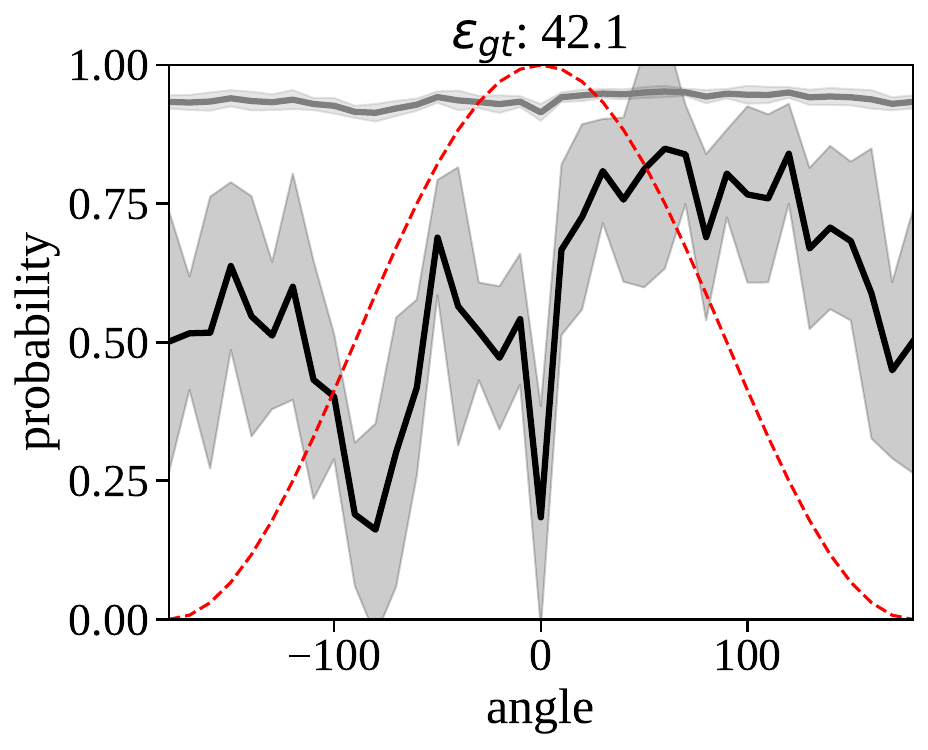} & \includegraphics[width=0.2\linewidth]{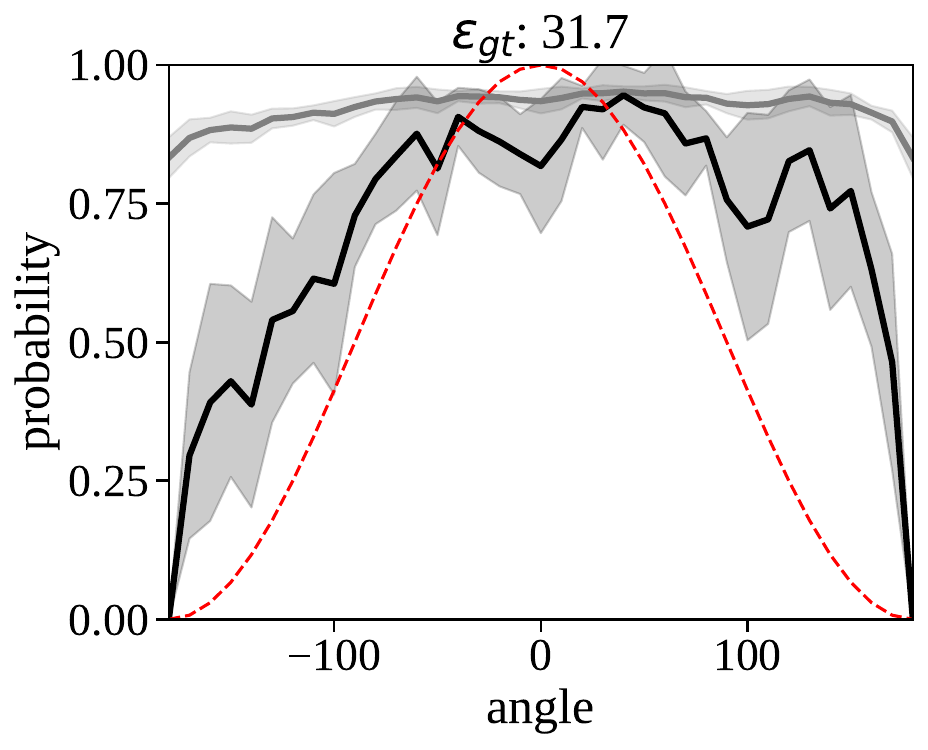} \\
\end{tabular}
 }
\vspace*{-5pt}
\caption{All prediction plots for each model on \texttt{COMFORT-BALL} using the camera perspective prompt (\texttt{cam}). The raw probability $p(\theta)$ in gray, normalized probability $\widehat{p}(\theta)$ in black, and the reference probability $p_\textrm{cos}(\theta)$ of \texttt{cam} in red.}
\label{fig:ball-cam-all}
\end{figure*}
\begin{figure*}[ht!]
  \centering
  \vspace*{-30pt}
  \makebox[\textwidth][c]{
  \begin{tabular}{ccccc}
  & \hphantom{aa} Left & \hphantom{aa} Right & \hphantom{aa} Front & \hphantom{aa} Back \\
\multirow{-8}{*}{\rotatebox[origin=c]{90}{\scriptsize InstructBLIP7B}} & \includegraphics[width=0.2\linewidth]{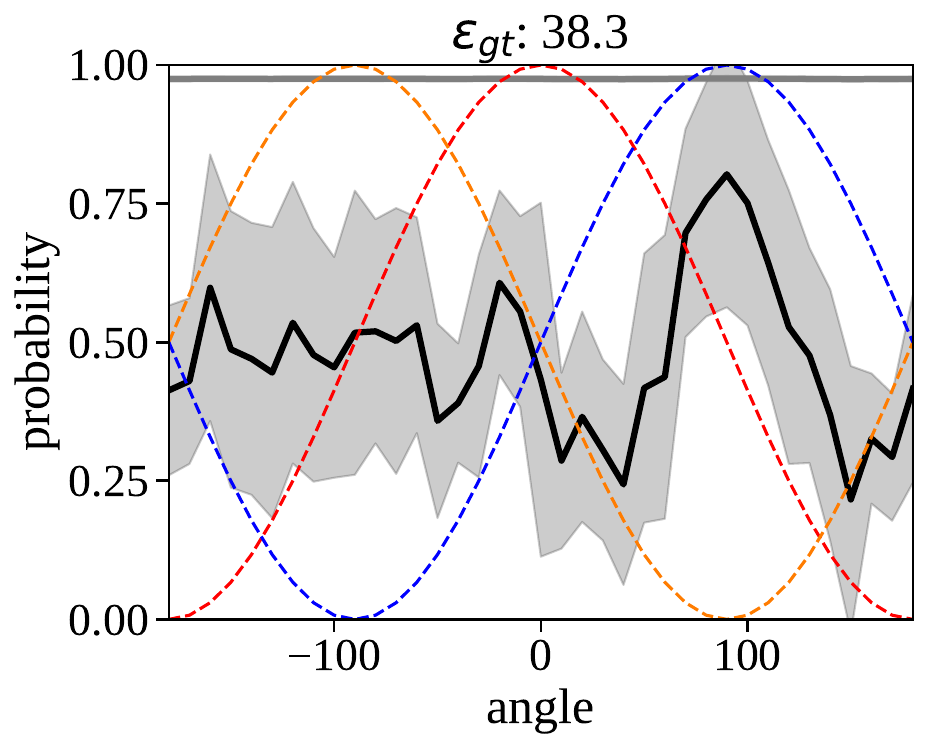} & \includegraphics[width=0.2\linewidth]{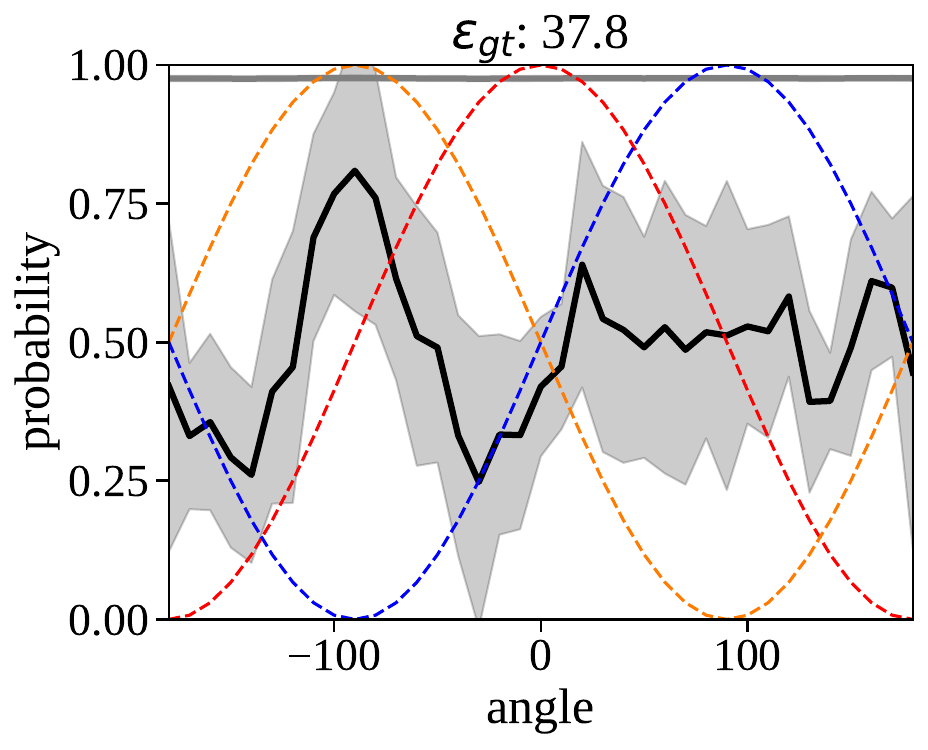} & \includegraphics[width=0.2\linewidth]{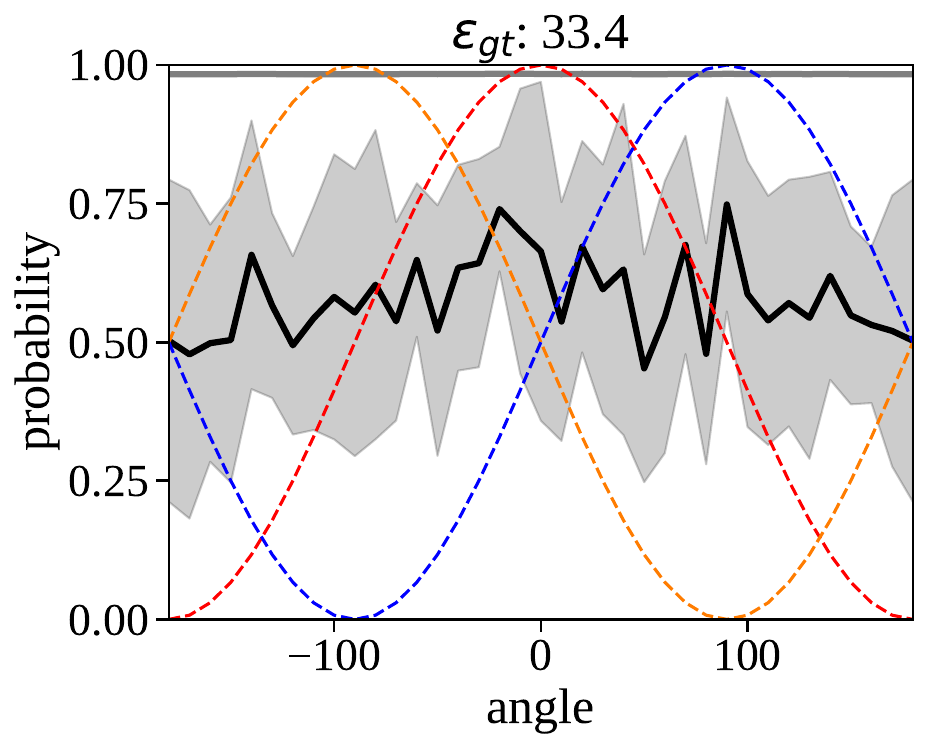} & \includegraphics[width=0.2\linewidth]{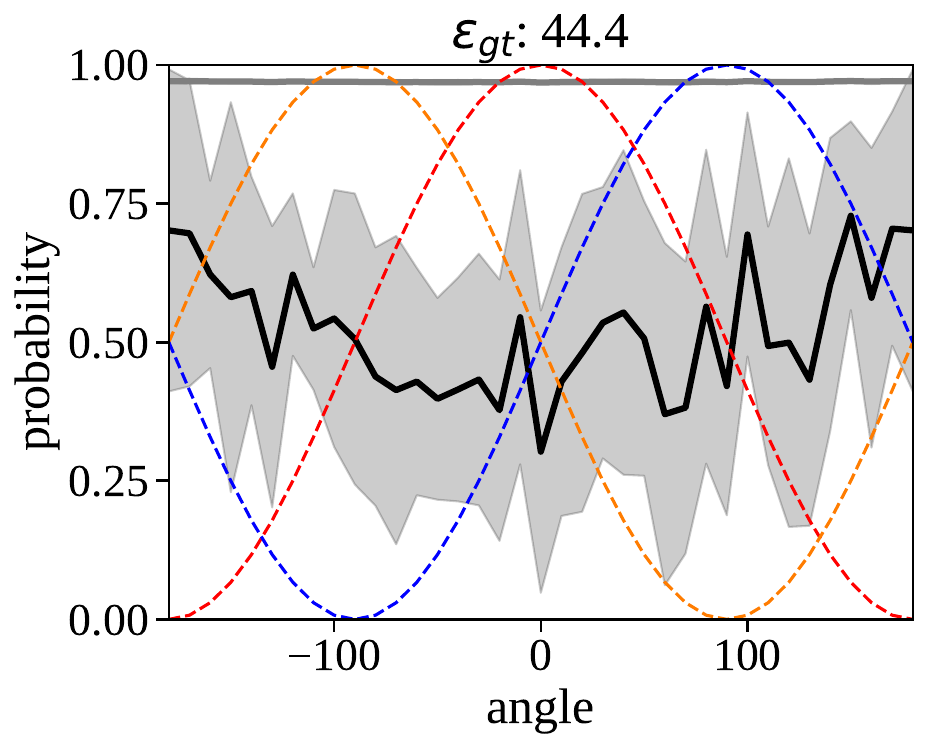} \\
\multirow{-8}{*}{\rotatebox[origin=c]{90}{\scriptsize InstructBLIP13B}} & \includegraphics[width=0.2\linewidth]{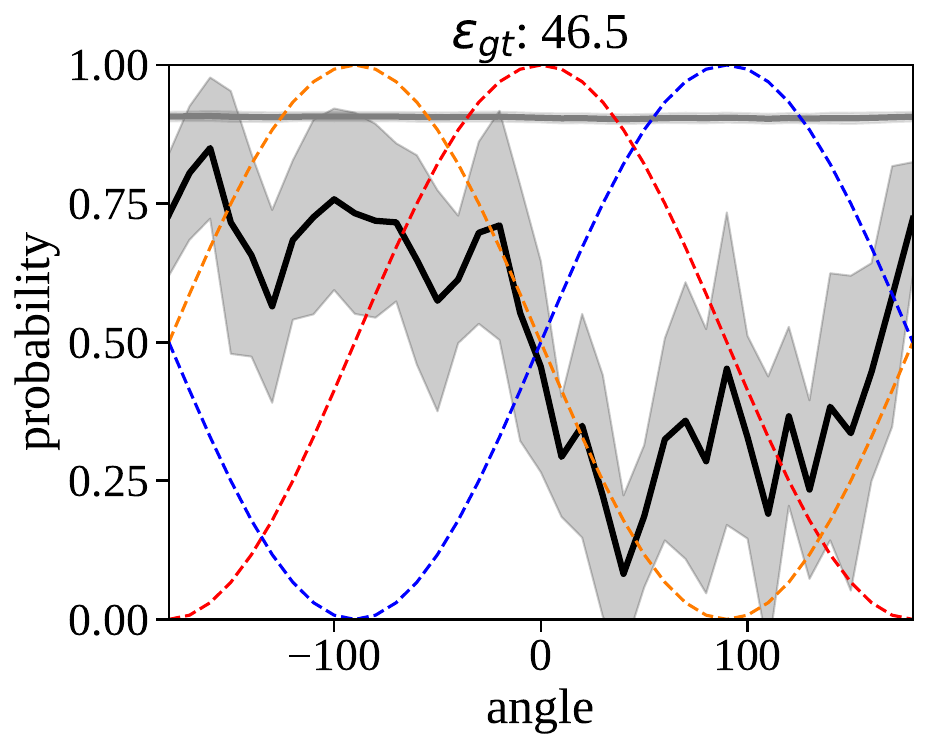} & \includegraphics[width=0.2\linewidth]{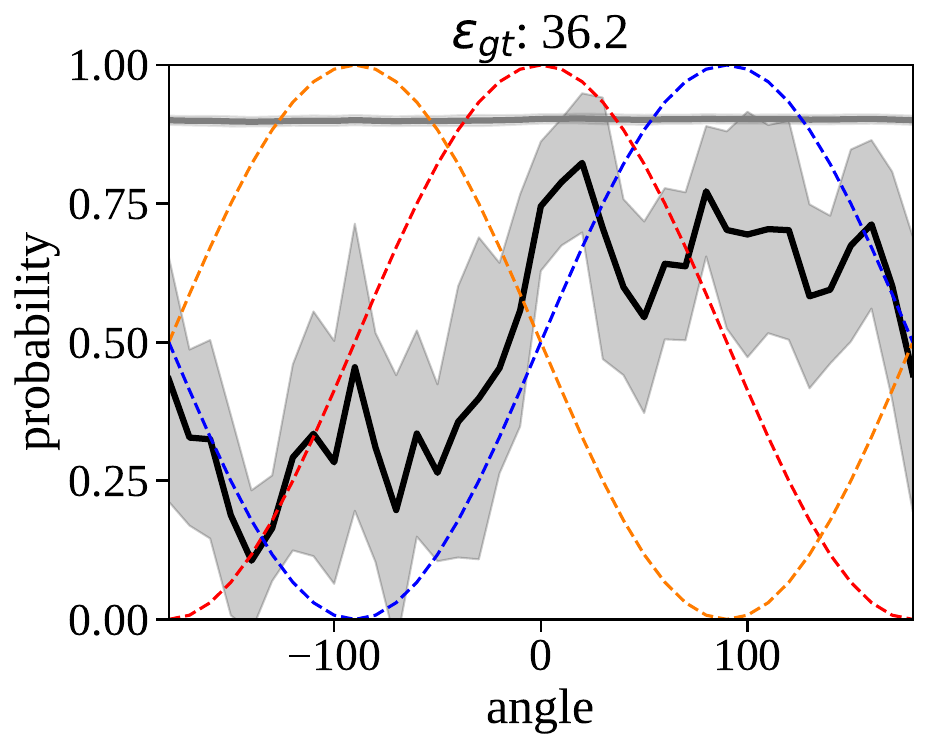} & \includegraphics[width=0.2\linewidth]{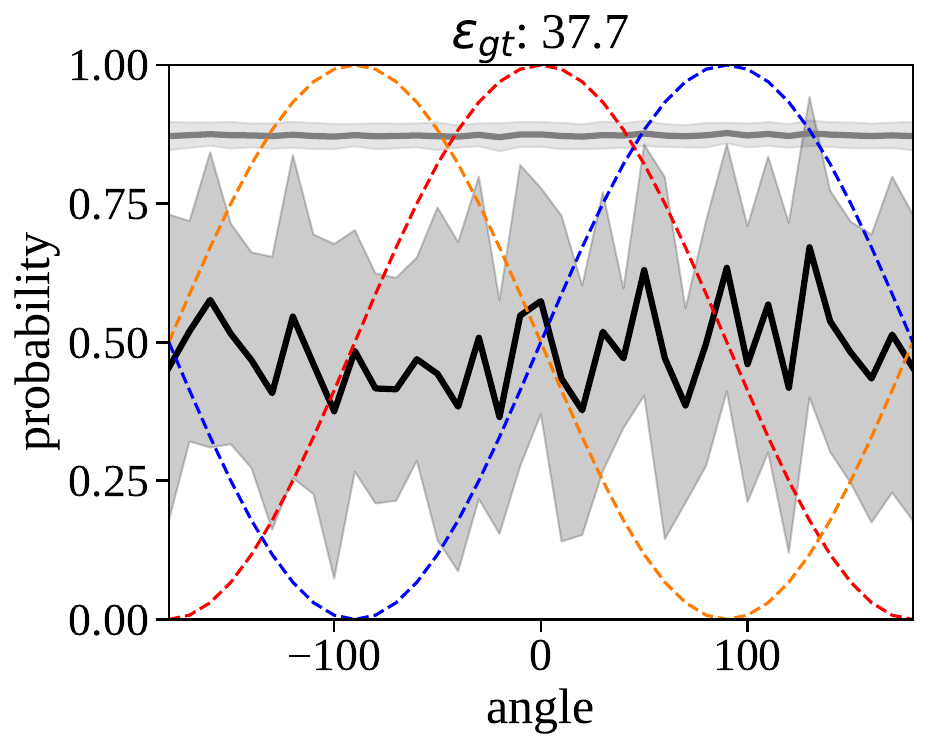} & \includegraphics[width=0.2\linewidth]{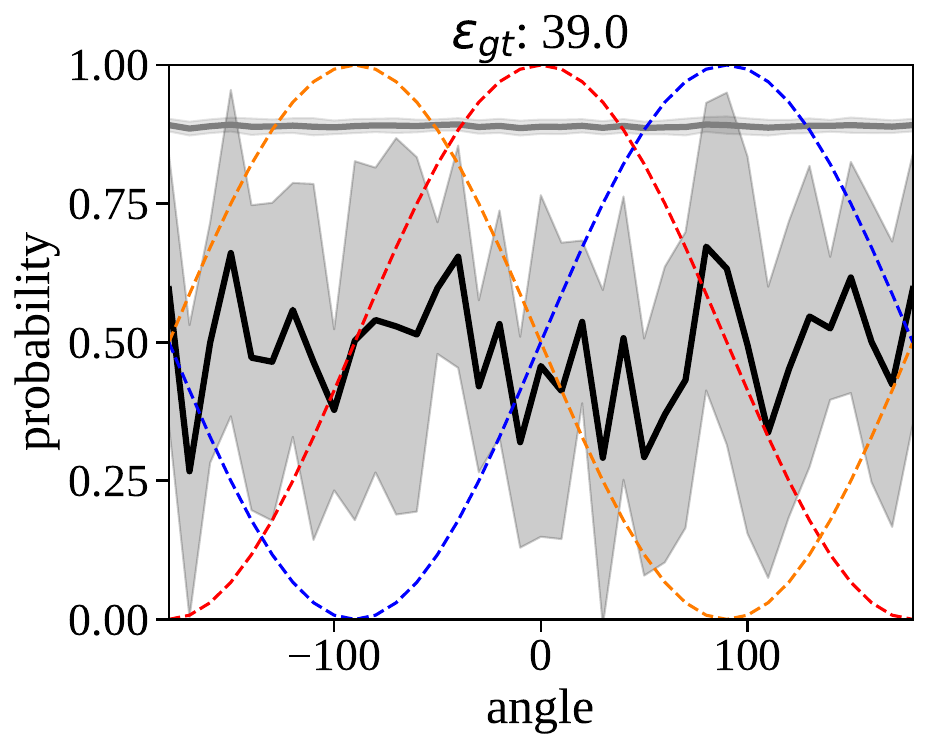} \\
\multirow{-8}{*}{\rotatebox[origin=c]{90}{\scriptsize MBLIPBLOOMZ7B}} & \includegraphics[width=0.2\linewidth]{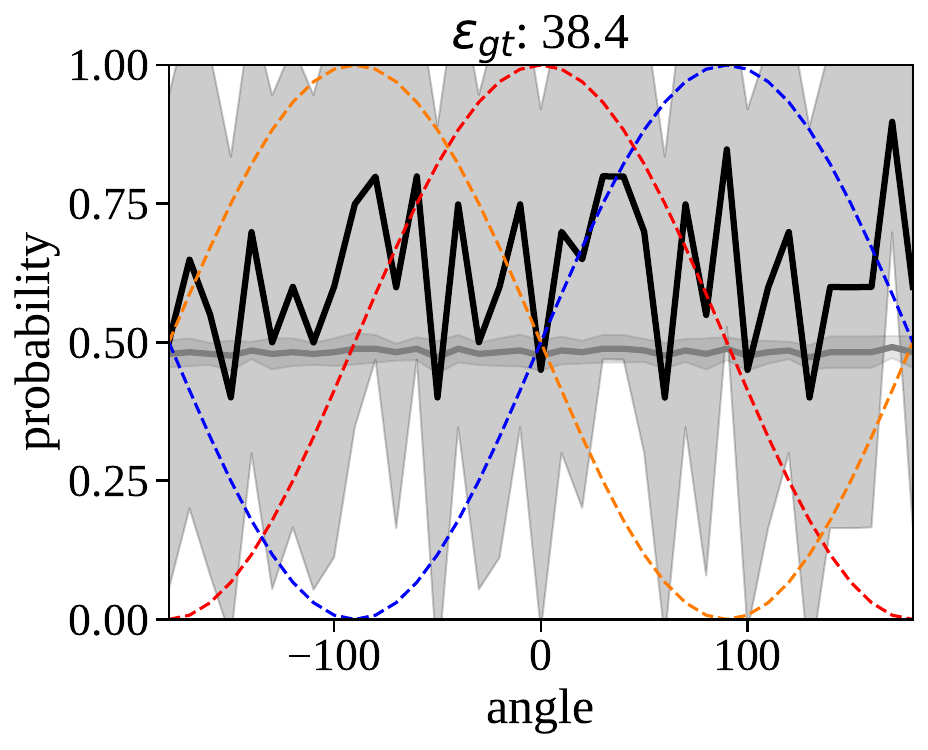} & \includegraphics[width=0.2\linewidth]{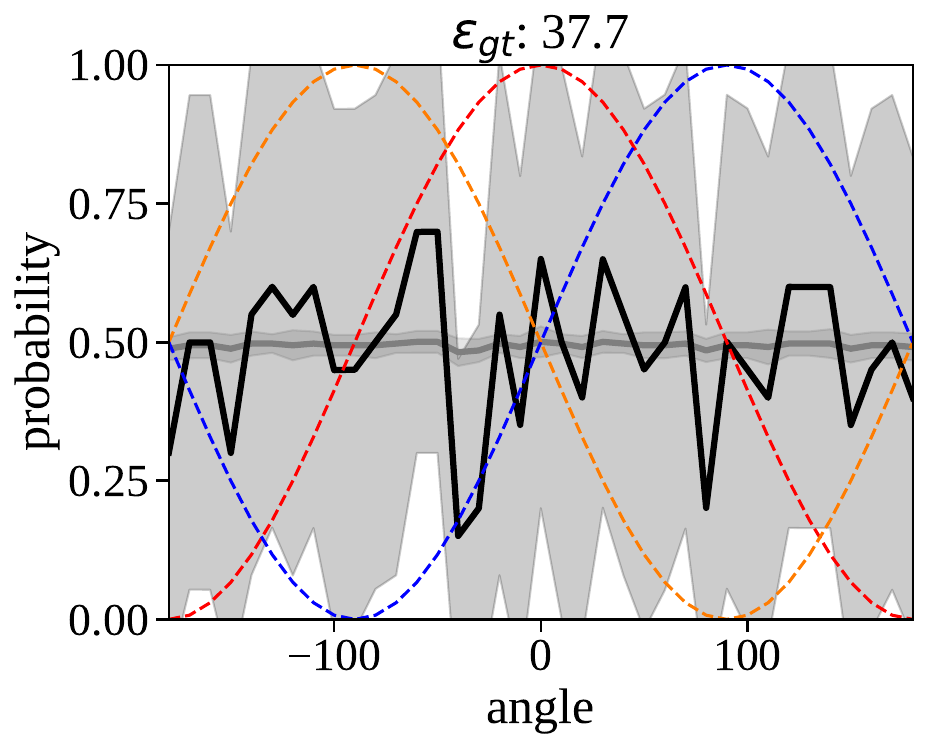} & \includegraphics[width=0.2\linewidth]{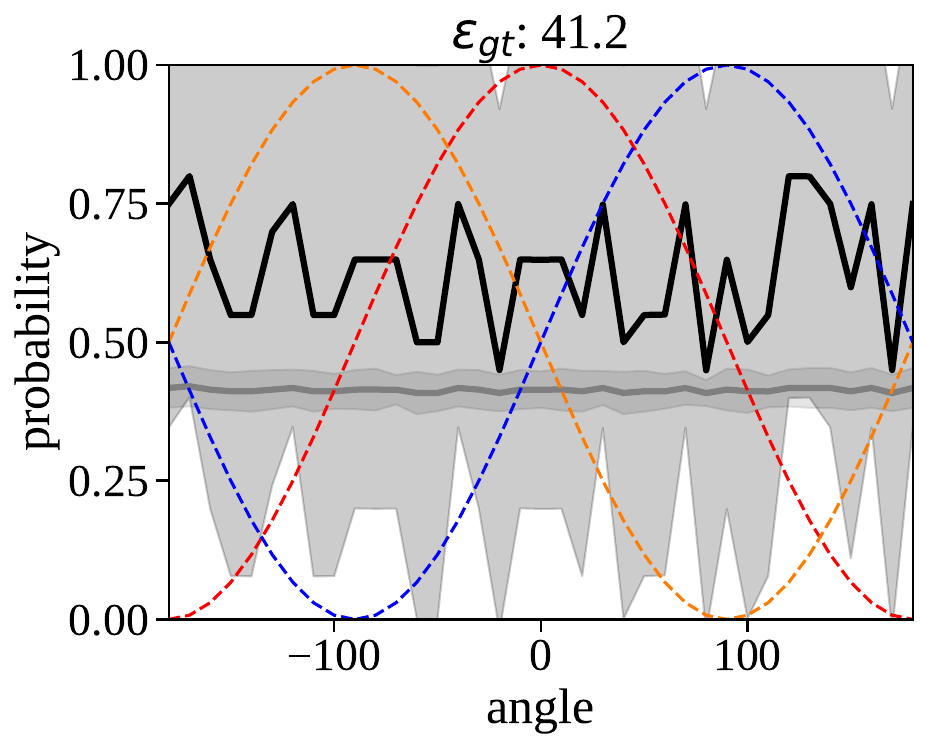} & \includegraphics[width=0.2\linewidth]{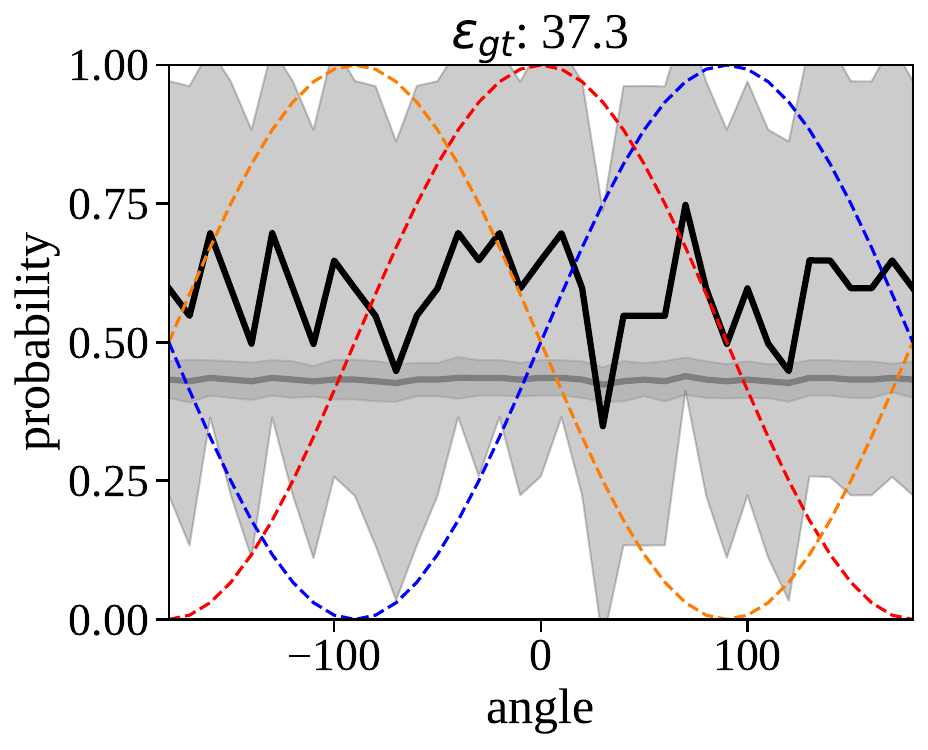} \\
\end{tabular}
 }
\vspace*{-5pt}
\caption{All prediction plots for each model on \texttt{COMFORT-CAR} using the camera perspective prompt (\texttt{cam}). The raw probability $p(\theta)$ in gray, normalized probability $\widehat{p}(\theta)$ in black, and the reference probabilities $p_\textrm{cos}(\theta)$ of \texttt{cam} in red, \texttt{add} in orange, \texttt{rel} in blue. To avoid overlapping reference probabilities of \texttt{add} and \texttt{rel}, we use plots on \texttt{COMFORT-CAR} with relatum facing left for left and right relations and \texttt{COMFORT-CAR} with relatum facing right for front and behind relations.}
\label{fig:car-cam-all}
\end{figure*}
\begin{figure*}[ht!]
  \centering
  \vspace*{-30pt}
  \makebox[\textwidth][c]{
  \begin{tabular}{ccccc}
  & \hphantom{aa} Left & \hphantom{aa} Right & \hphantom{aa} Front & \hphantom{aa} Back \\
\multirow{-8}{*}{\rotatebox[origin=c]{90}{\scriptsize InstructBLIP7B}} & \includegraphics[width=0.2\linewidth]{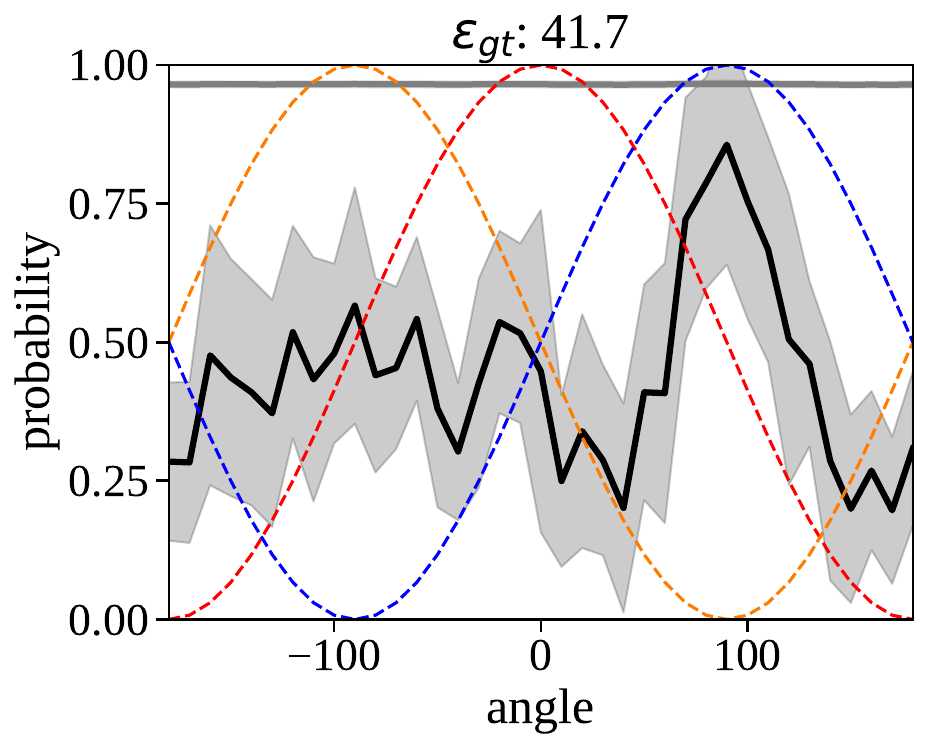} & \includegraphics[width=0.2\linewidth]{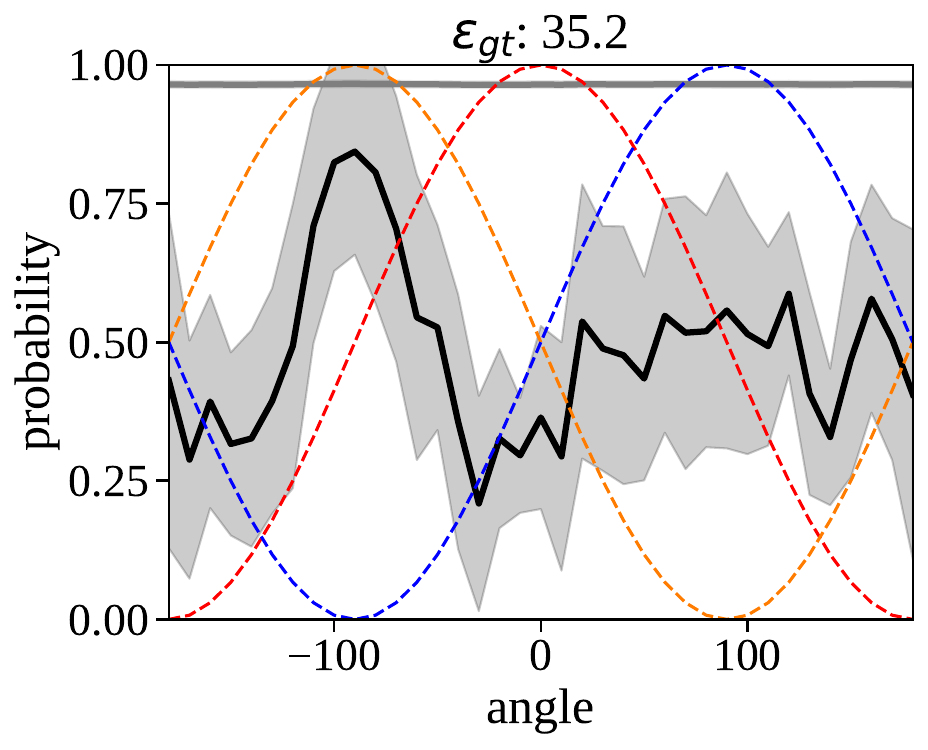} & \includegraphics[width=0.2\linewidth]{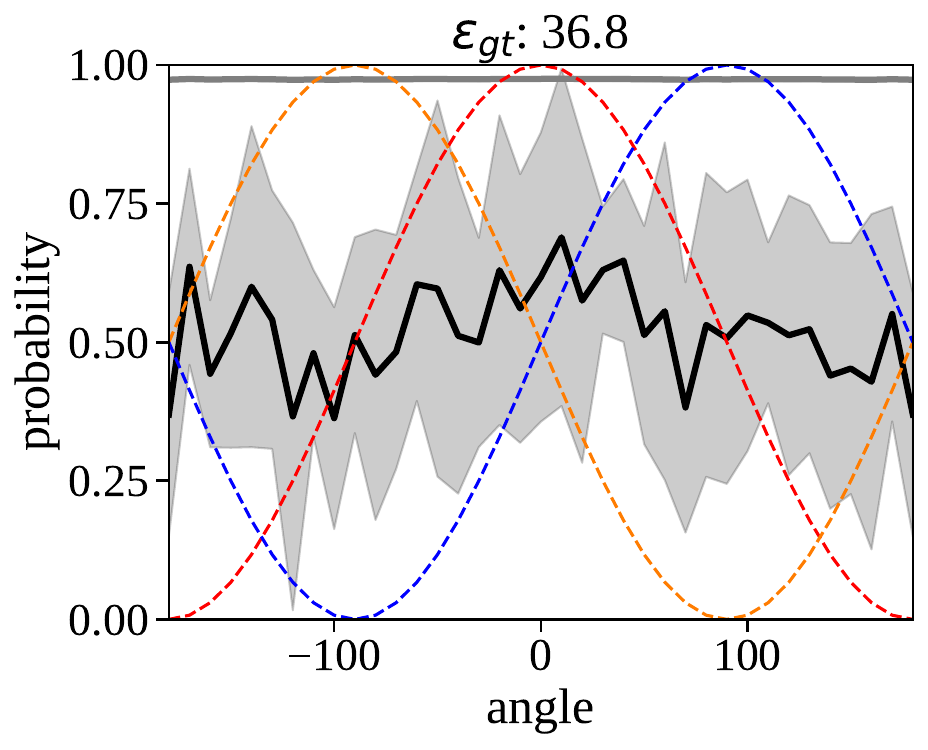} & \includegraphics[width=0.2\linewidth]{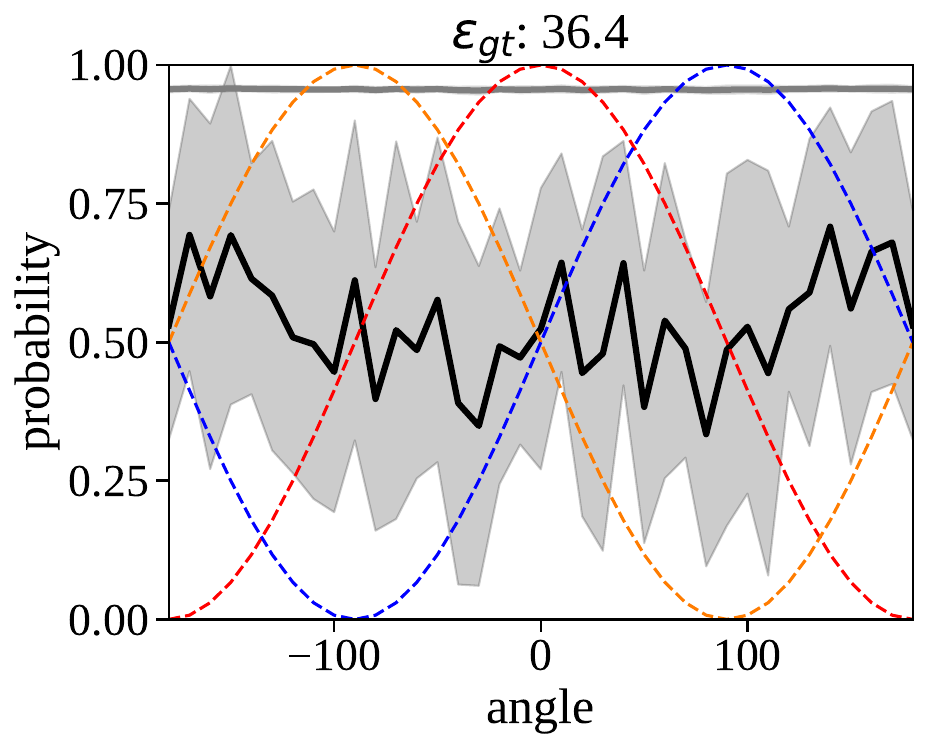} \\
\multirow{-8}{*}{\rotatebox[origin=c]{90}{\scriptsize InstructBLIP13B}} & \includegraphics[width=0.2\linewidth]{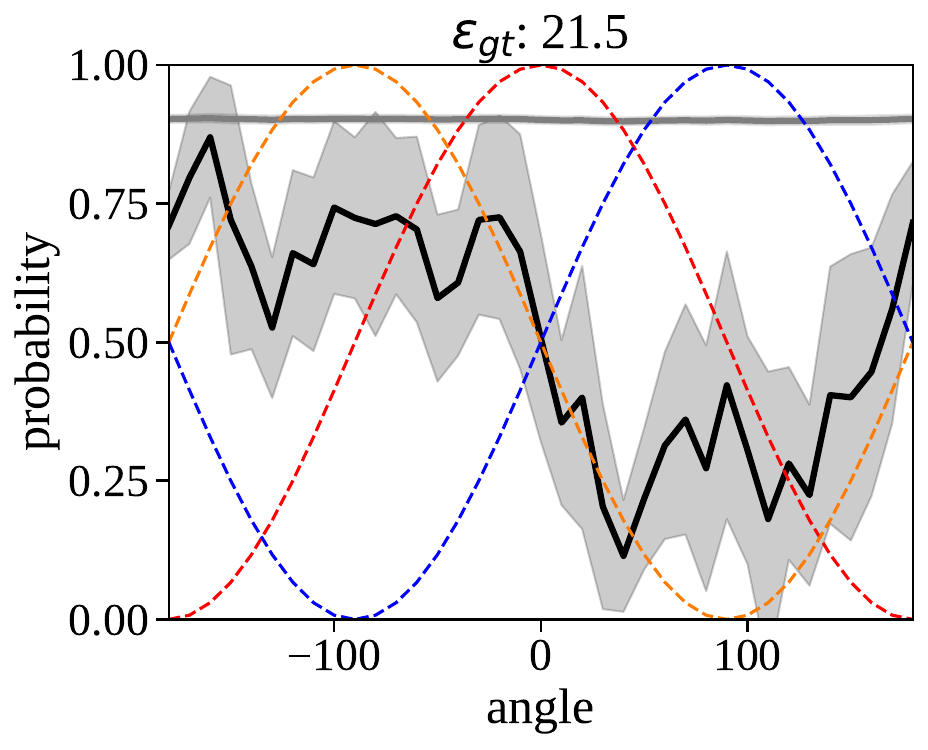} & \includegraphics[width=0.2\linewidth]{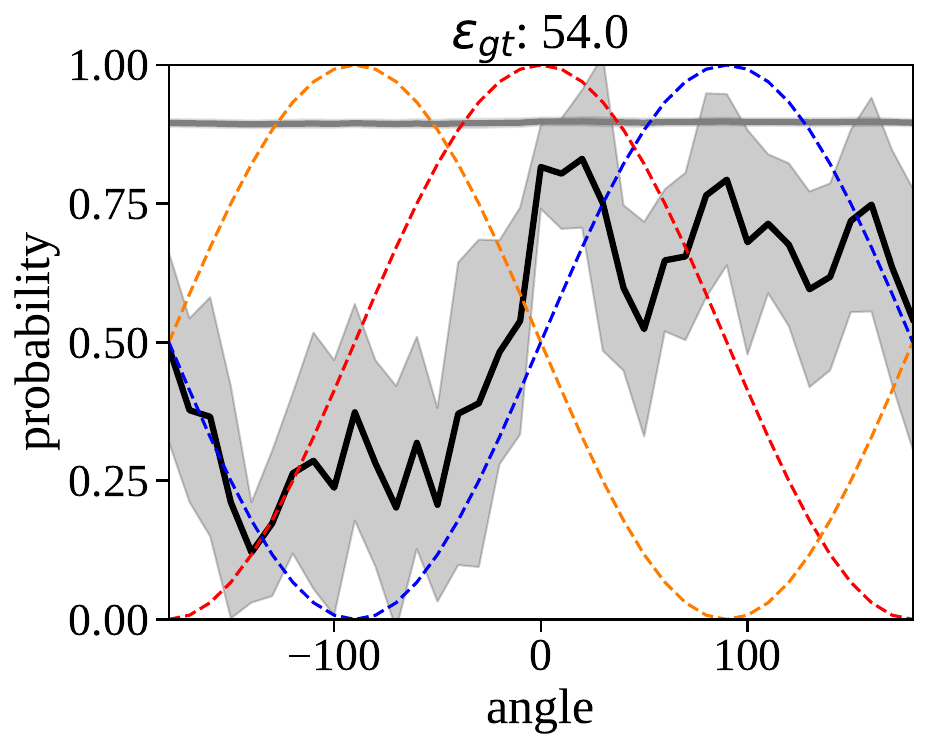} & \includegraphics[width=0.2\linewidth]{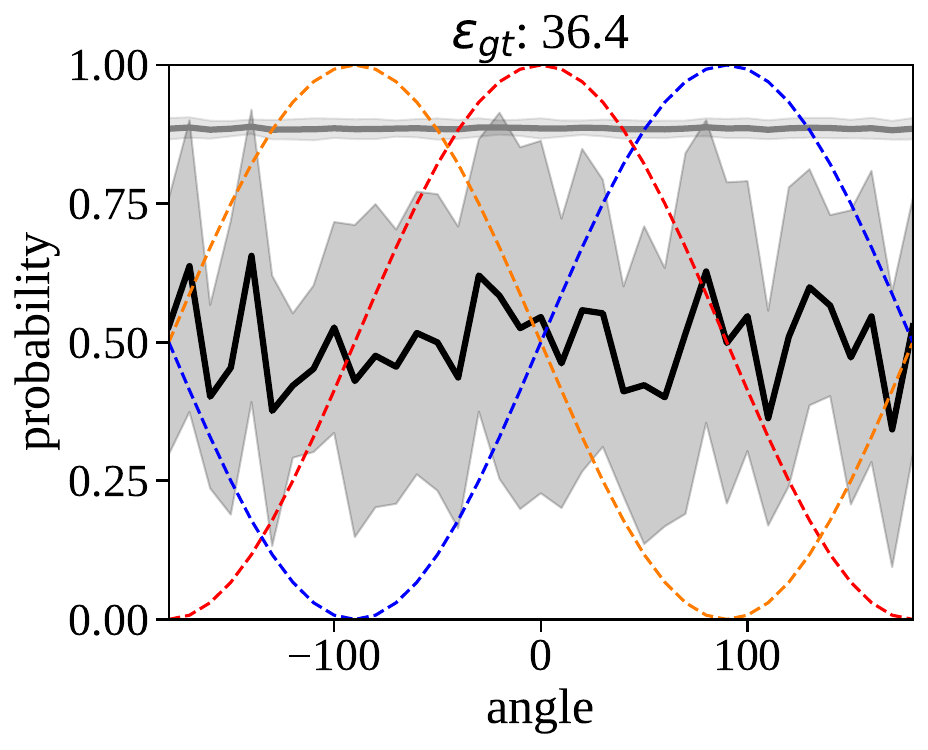} & \includegraphics[width=0.2\linewidth]{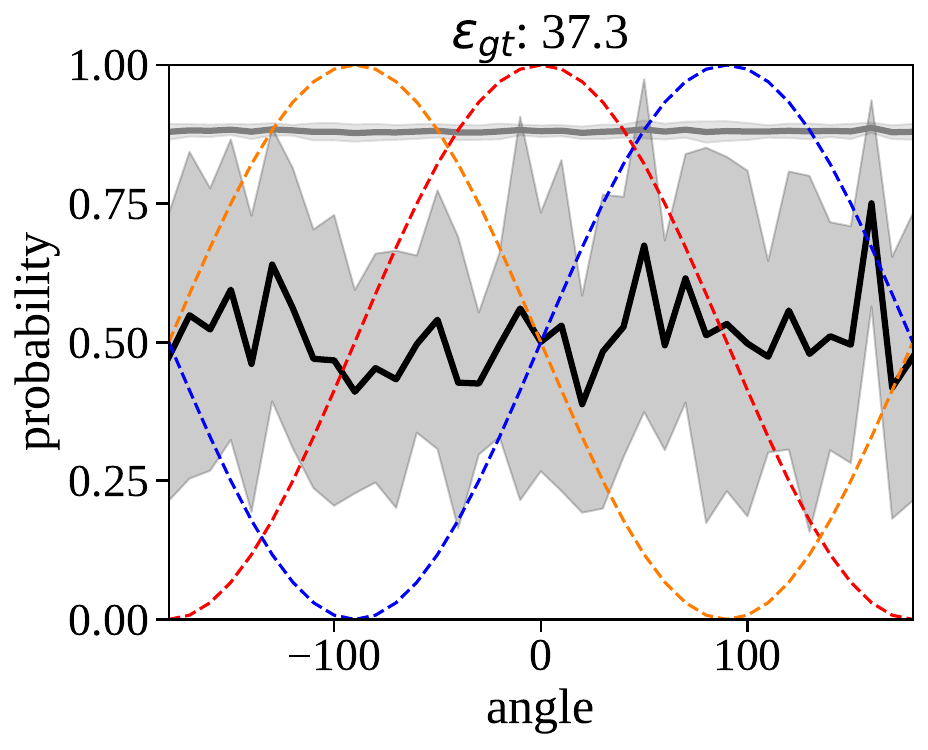} \\
\multirow{-8}{*}{\rotatebox[origin=c]{90}{\scriptsize MBLIPBLOOMZ7B}} & \includegraphics[width=0.2\linewidth]{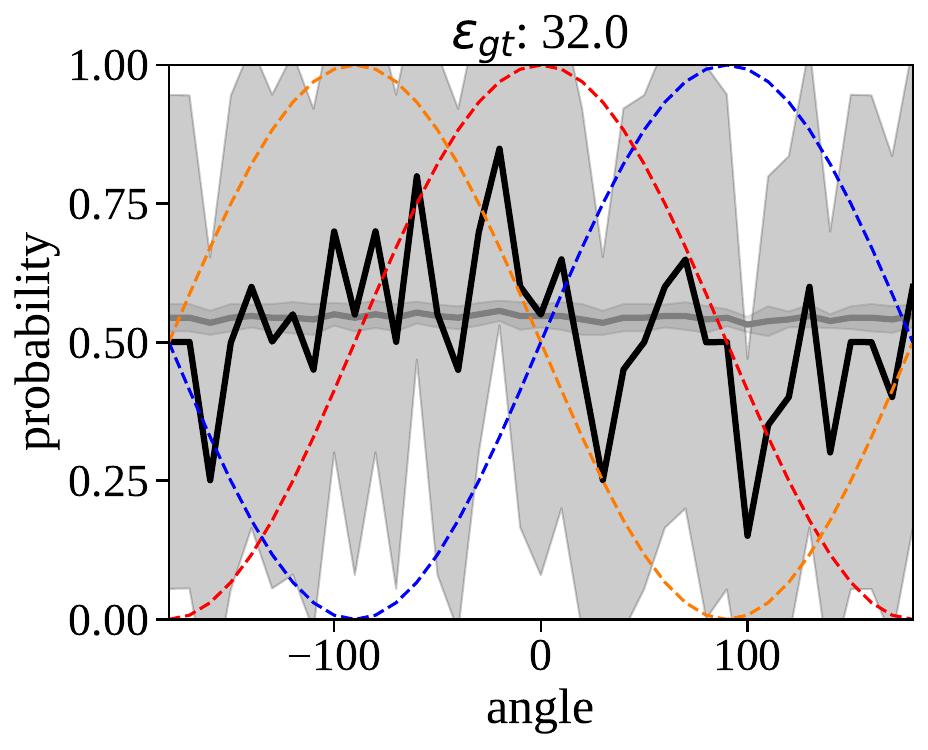} & \includegraphics[width=0.2\linewidth]{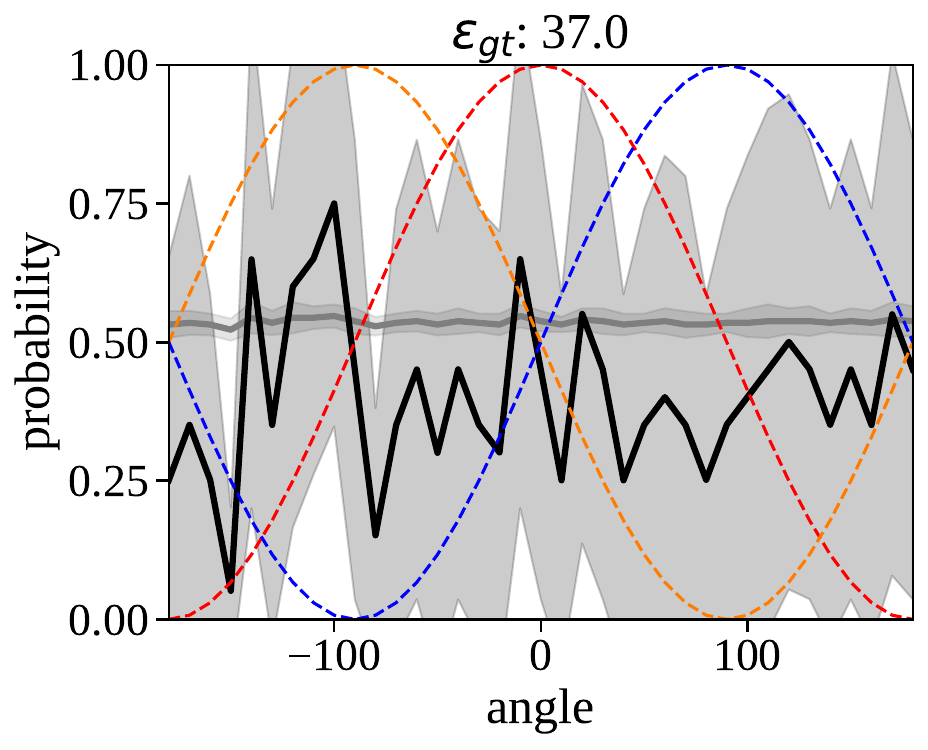} & \includegraphics[width=0.2\linewidth]{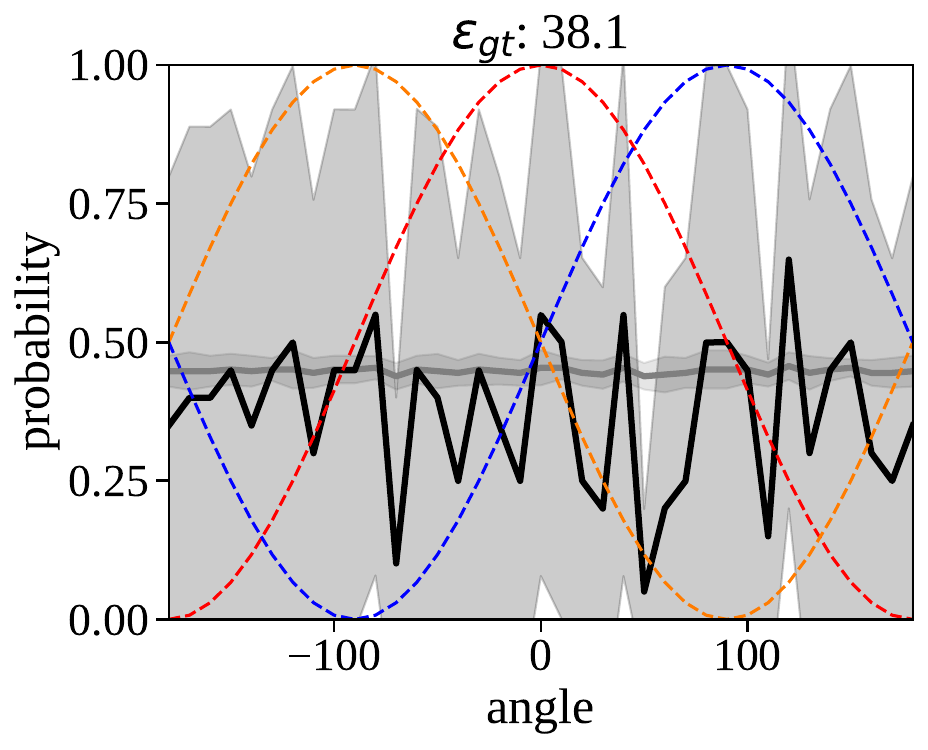} & \includegraphics[width=0.2\linewidth]{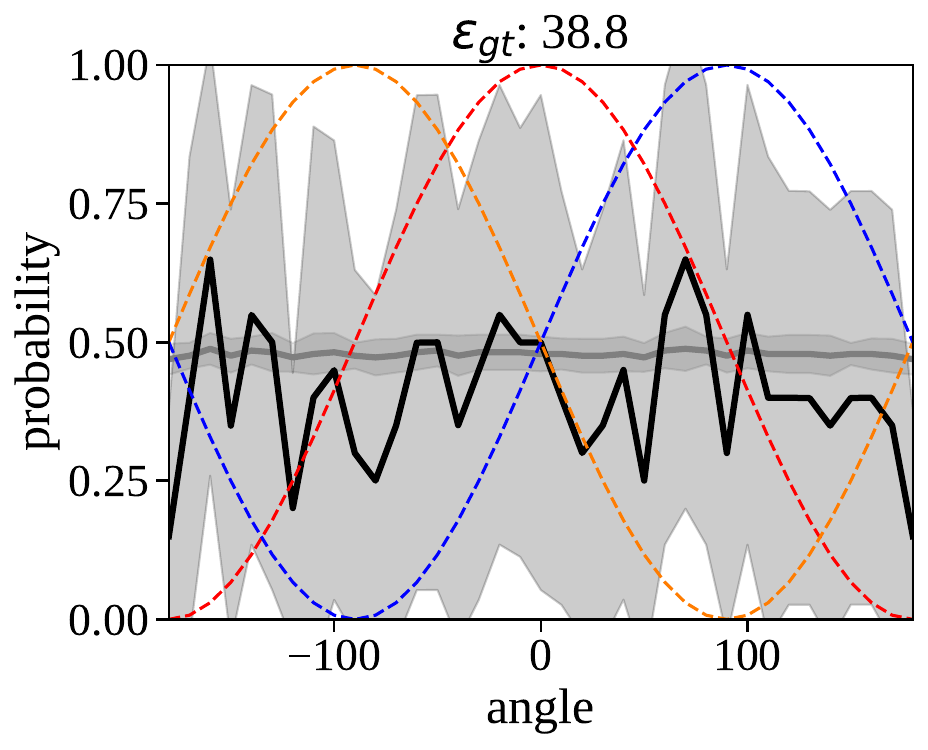} \\
\multirow{-8}{*}{\rotatebox[origin=c]{90}{\scriptsize GLaMM}} & \includegraphics[width=0.2\linewidth]{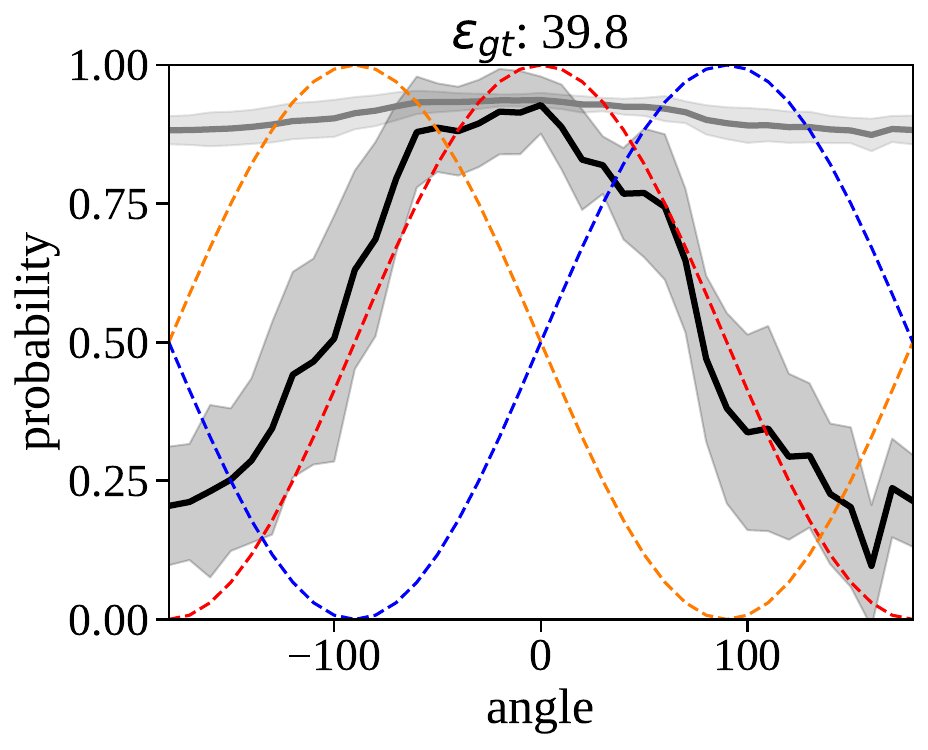} & \includegraphics[width=0.2\linewidth]{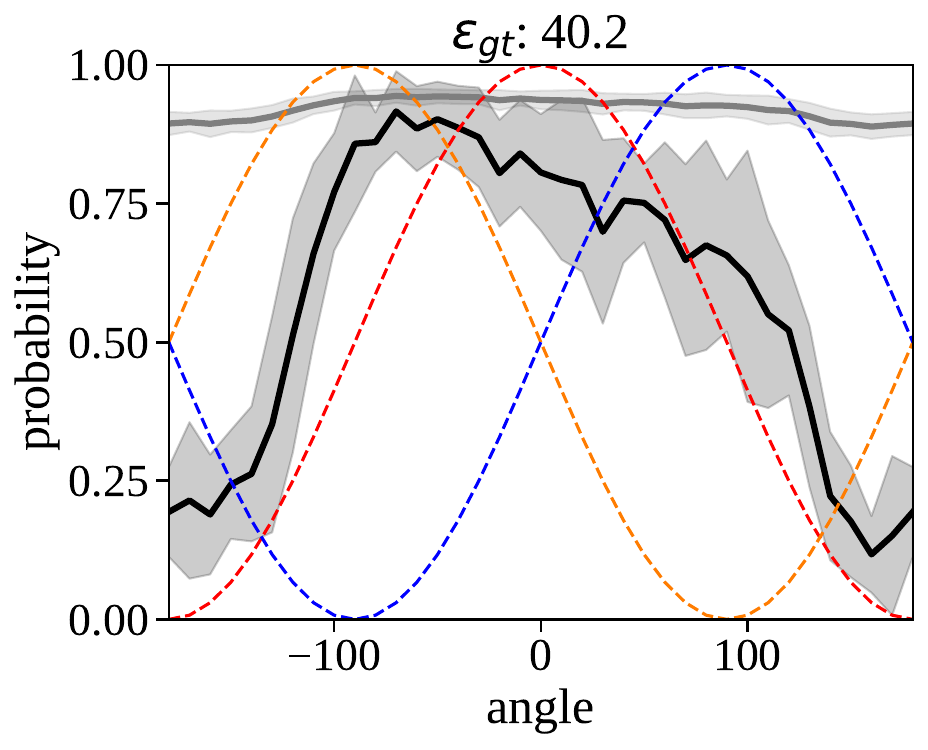} & \includegraphics[width=0.2\linewidth]{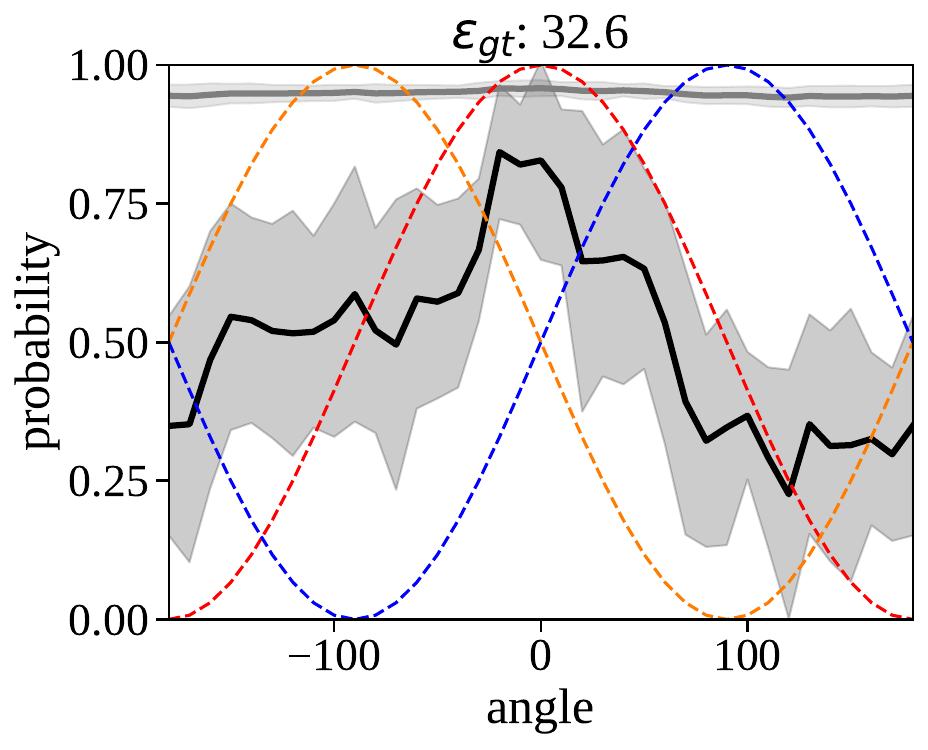} & \includegraphics[width=0.2\linewidth]{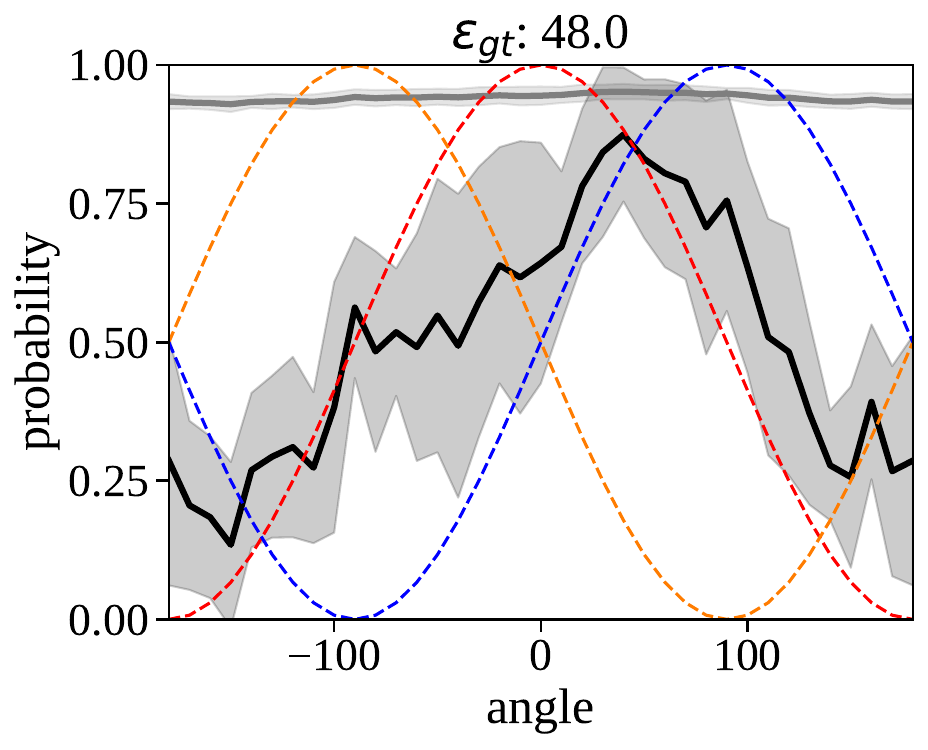} \\
\multirow{-8}{*}{\rotatebox[origin=c]{90}{\scriptsize LLaVA1.57B}} & \includegraphics[width=0.2\linewidth]{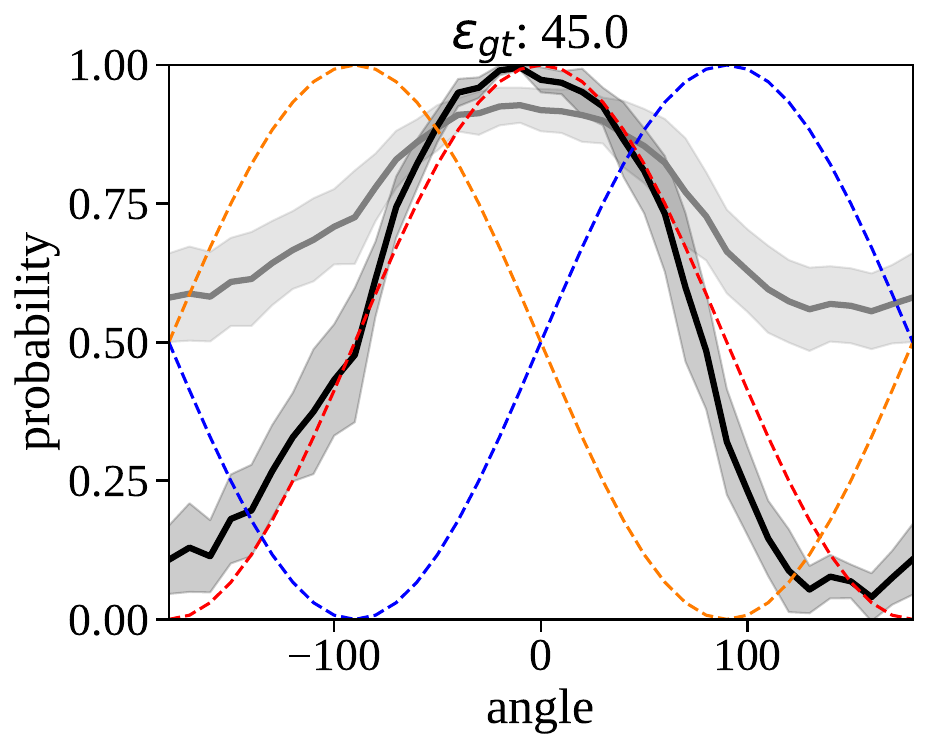} & \includegraphics[width=0.2\linewidth]{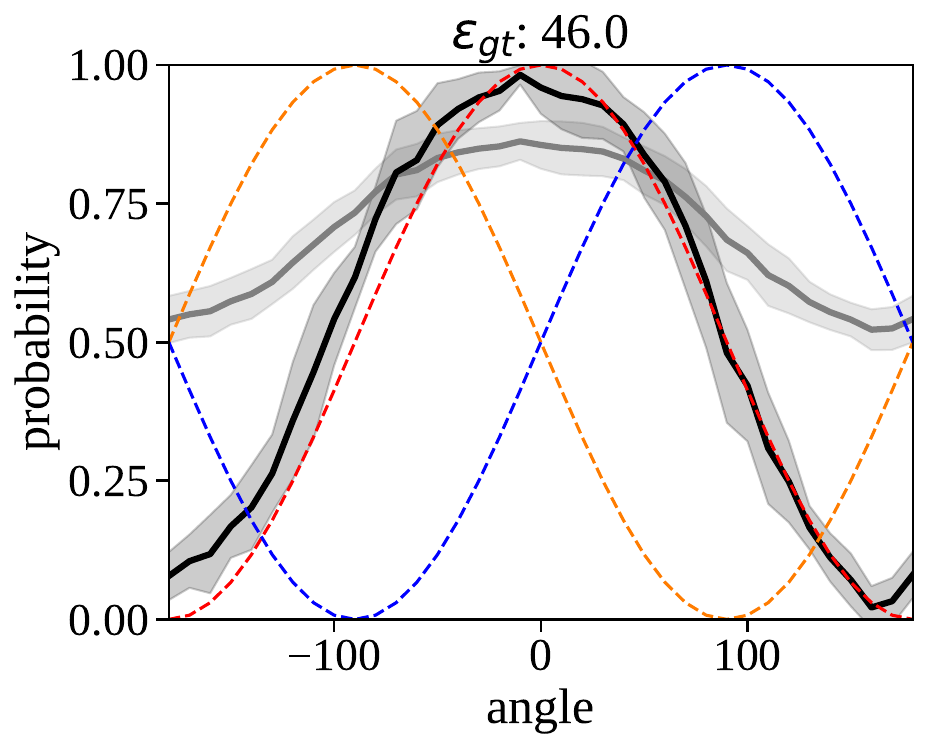} & \includegraphics[width=0.2\linewidth]{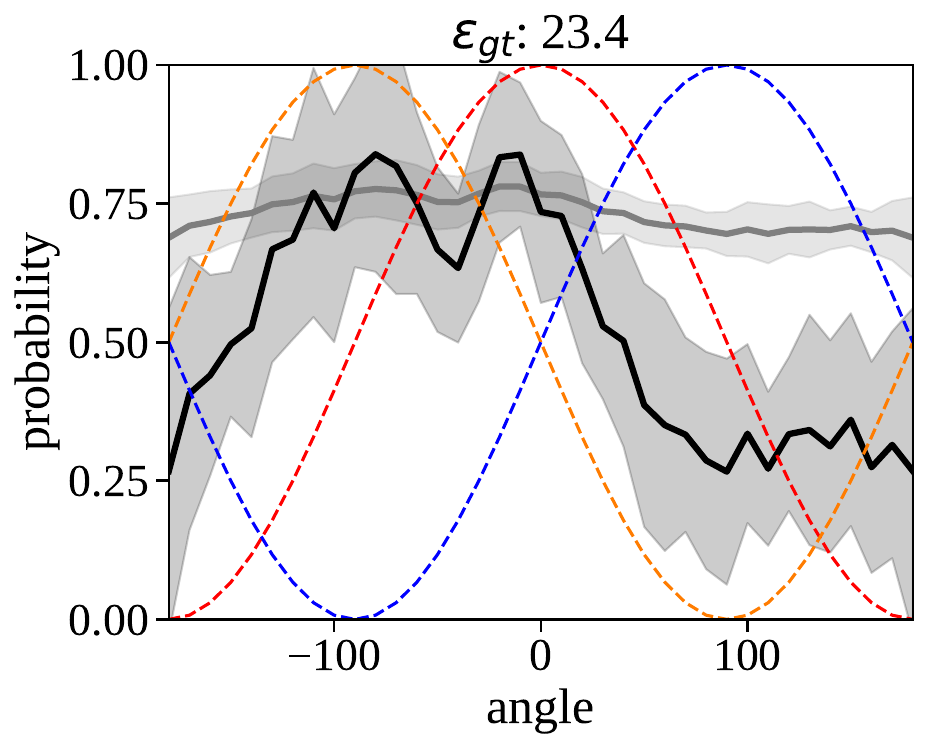} & \includegraphics[width=0.2\linewidth]{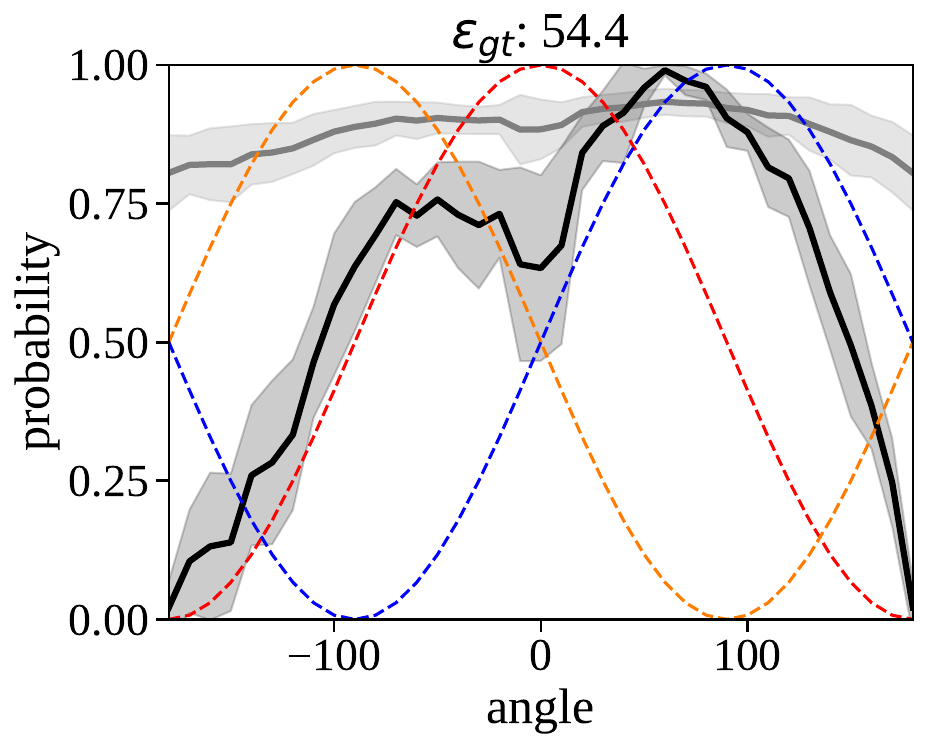} \\
\multirow{-8}{*}{\rotatebox[origin=c]{90}{\scriptsize LLaVA1.513B}} & \includegraphics[width=0.2\linewidth]{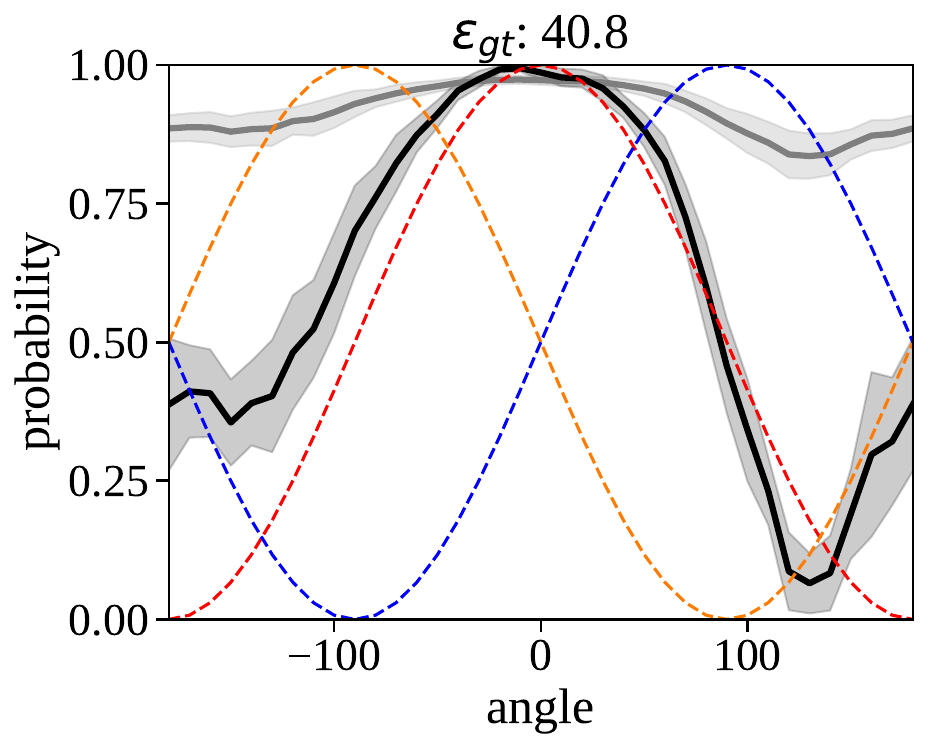} & \includegraphics[width=0.2\linewidth]{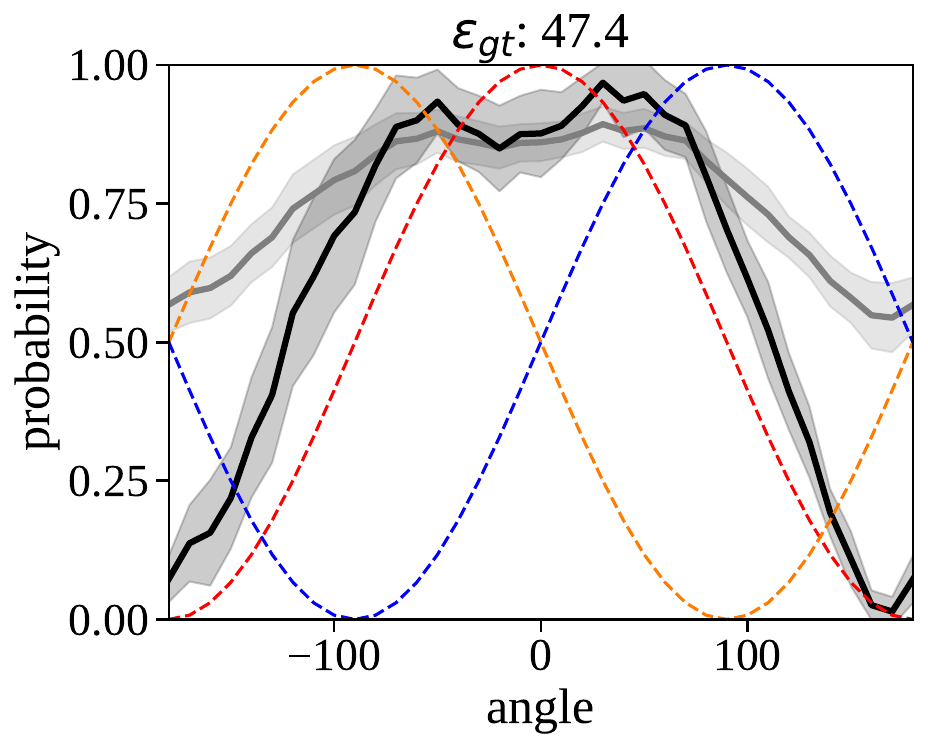} & \includegraphics[width=0.2\linewidth]{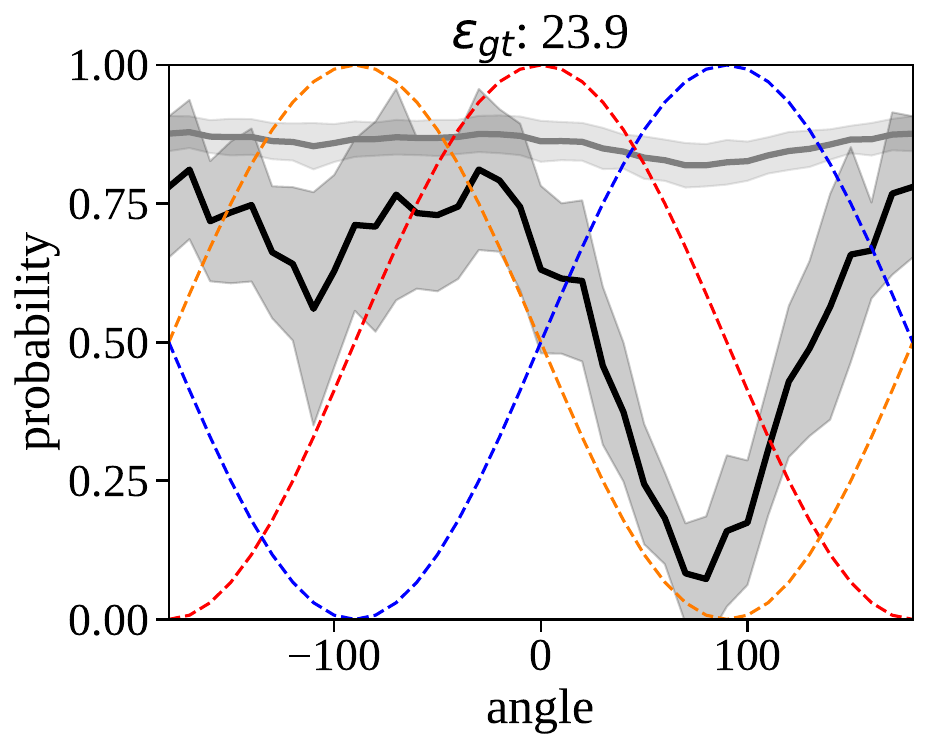} & \includegraphics[width=0.2\linewidth]{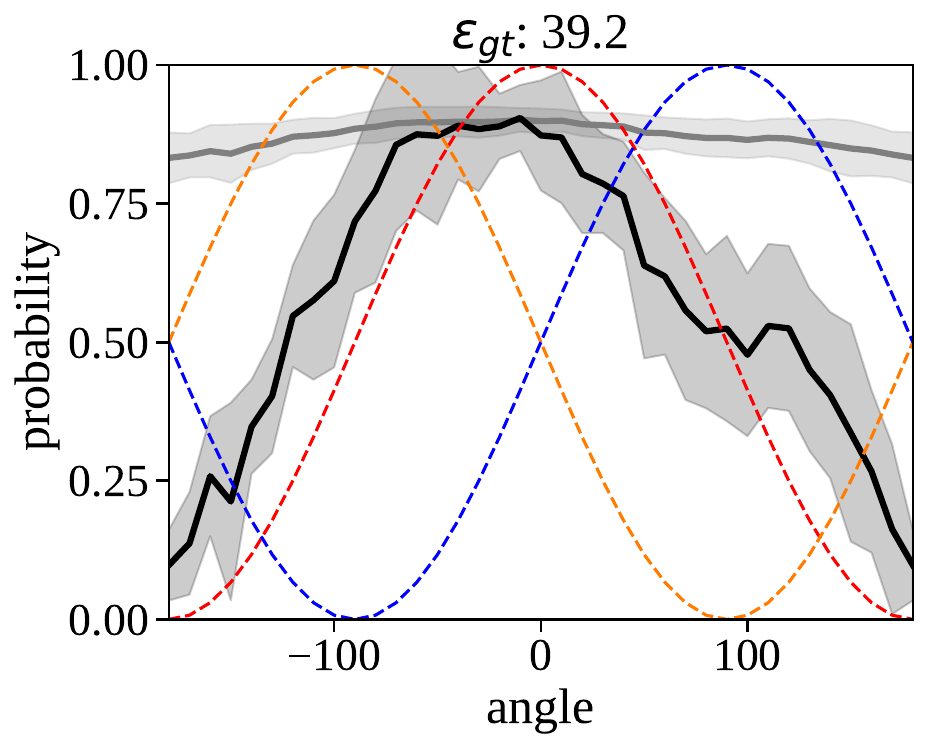} \\
\multirow{-8}{*}{\rotatebox[origin=c]{90}{\scriptsize XComposer2}} & \includegraphics[width=0.2\linewidth]{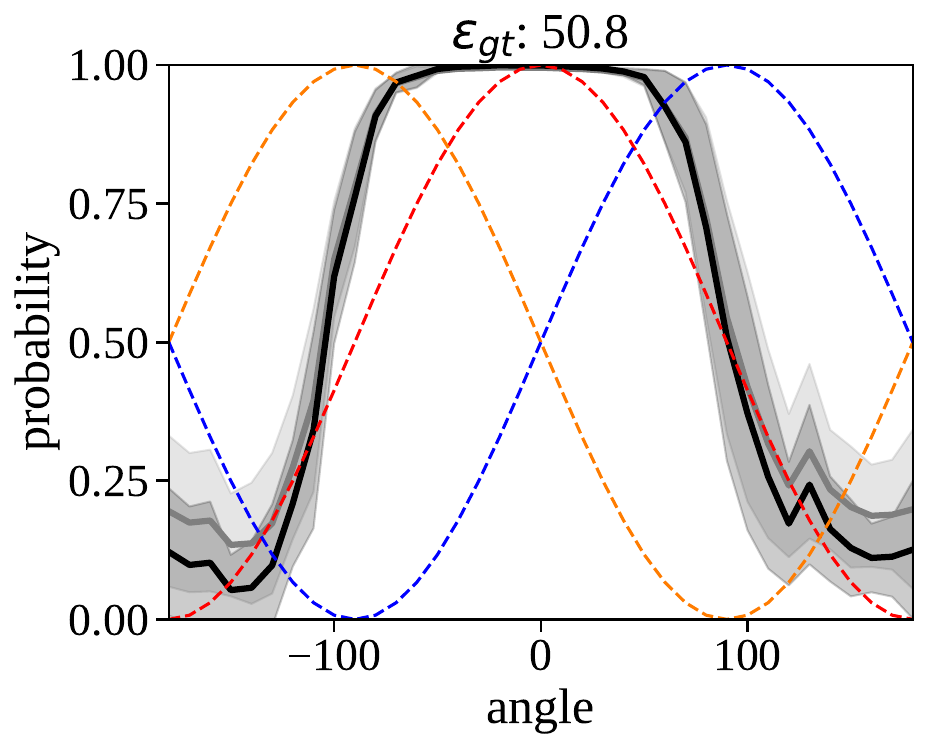} & \includegraphics[width=0.2\linewidth]{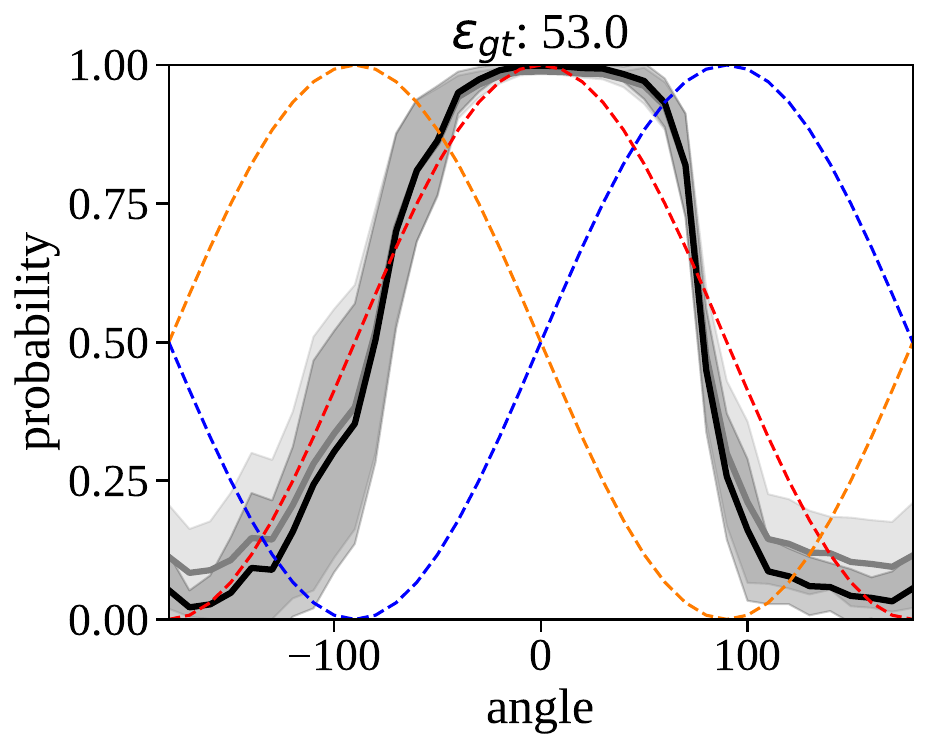} & \includegraphics[width=0.2\linewidth]{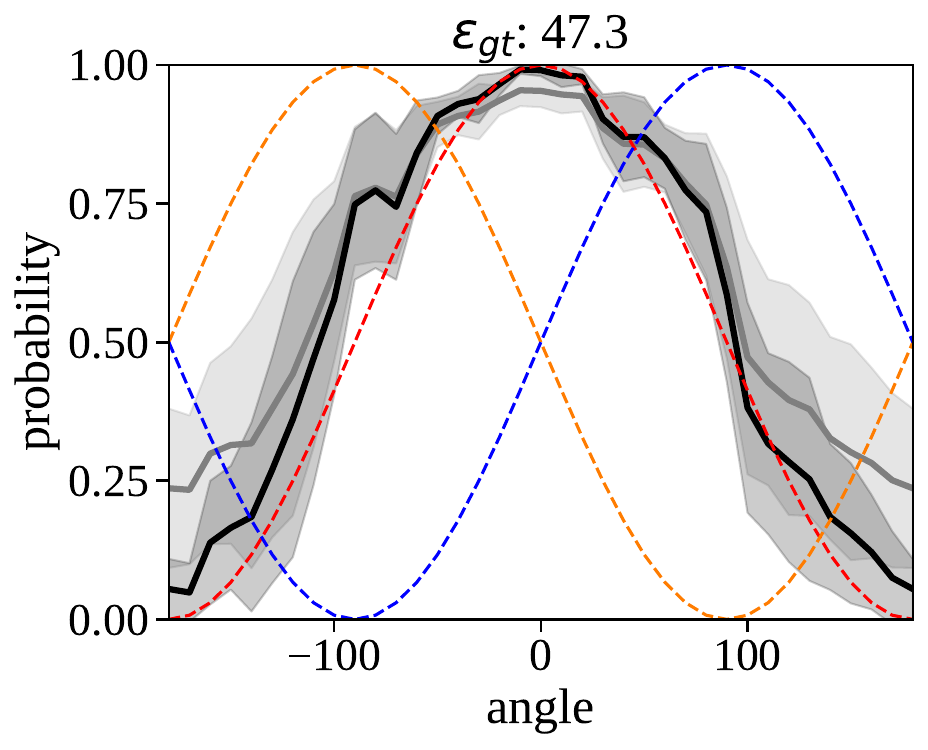} & \includegraphics[width=0.2\linewidth]{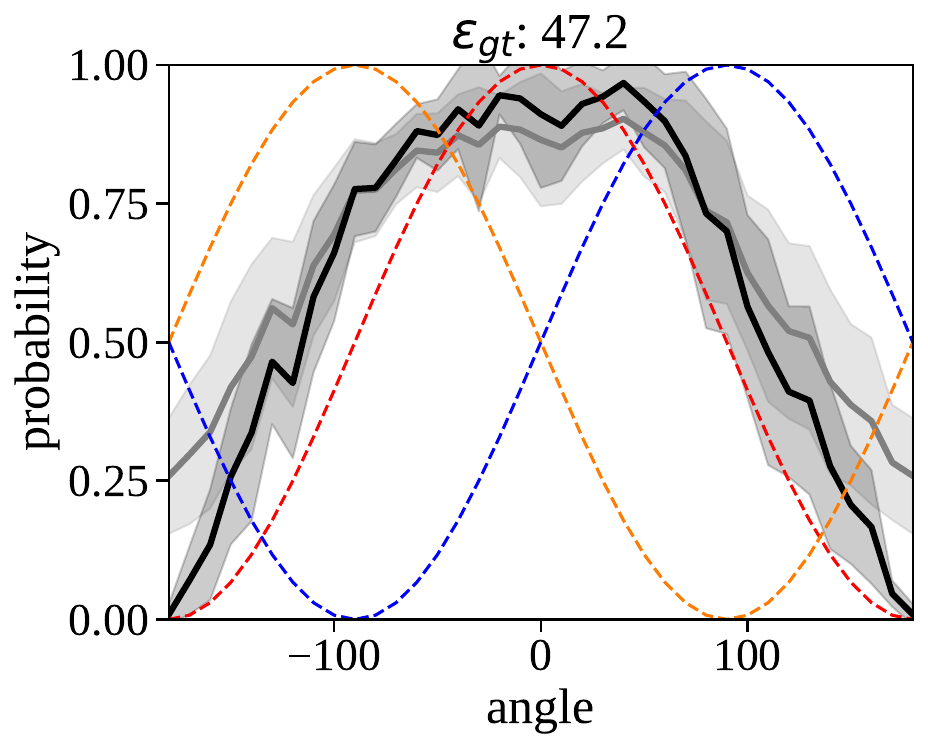} \\
\multirow{-8}{*}{\rotatebox[origin=c]{90}{\scriptsize MiniCPM}} & \includegraphics[width=0.2\linewidth]{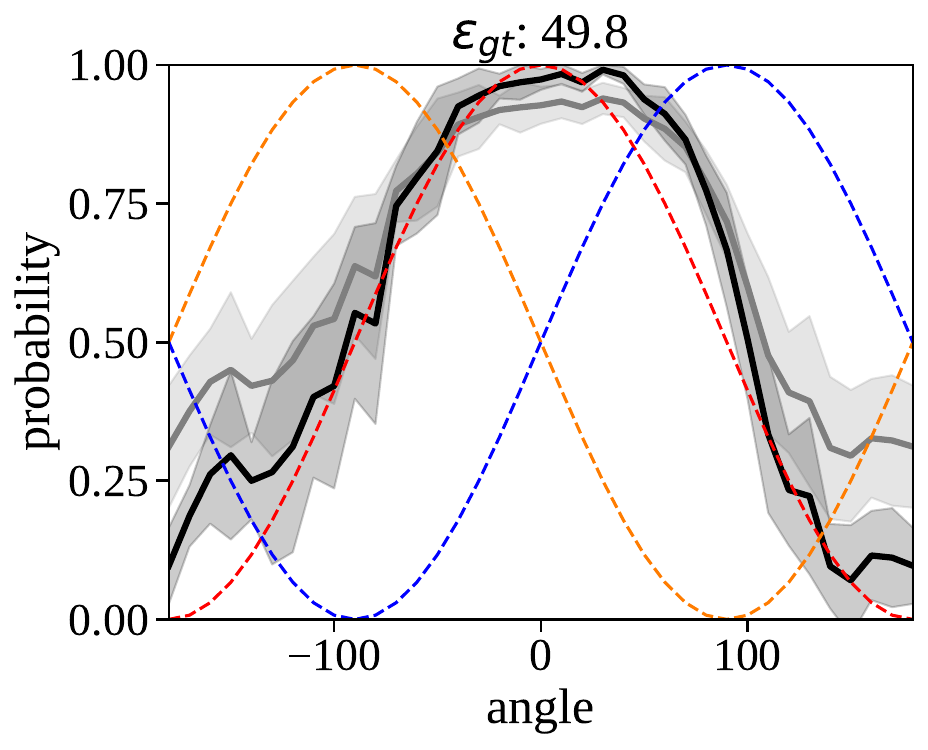} & \includegraphics[width=0.2\linewidth]{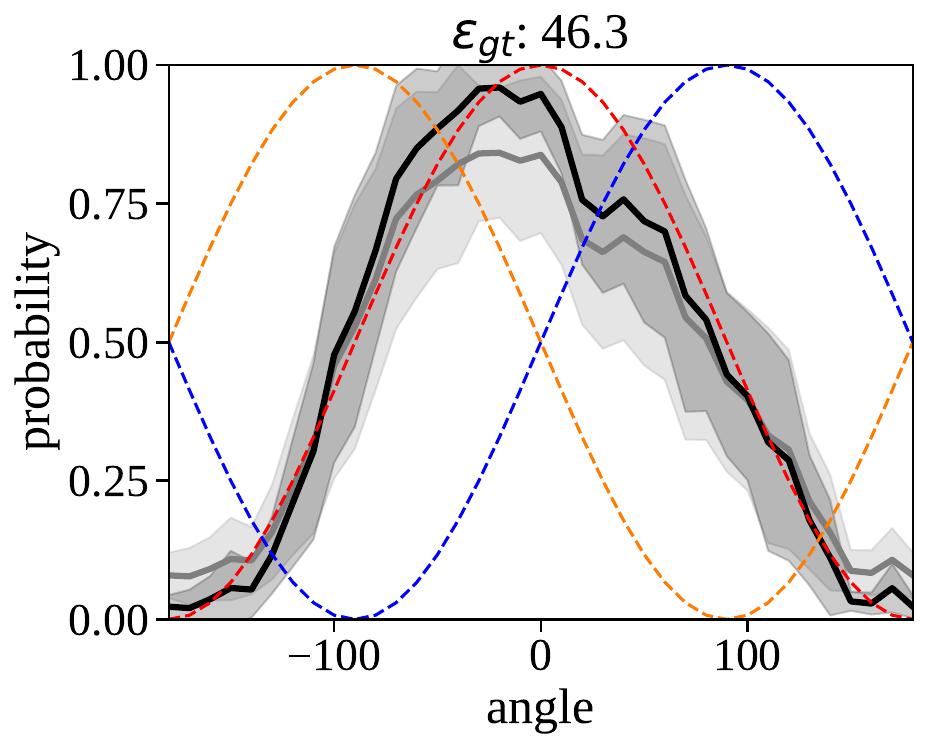} & \includegraphics[width=0.2\linewidth]{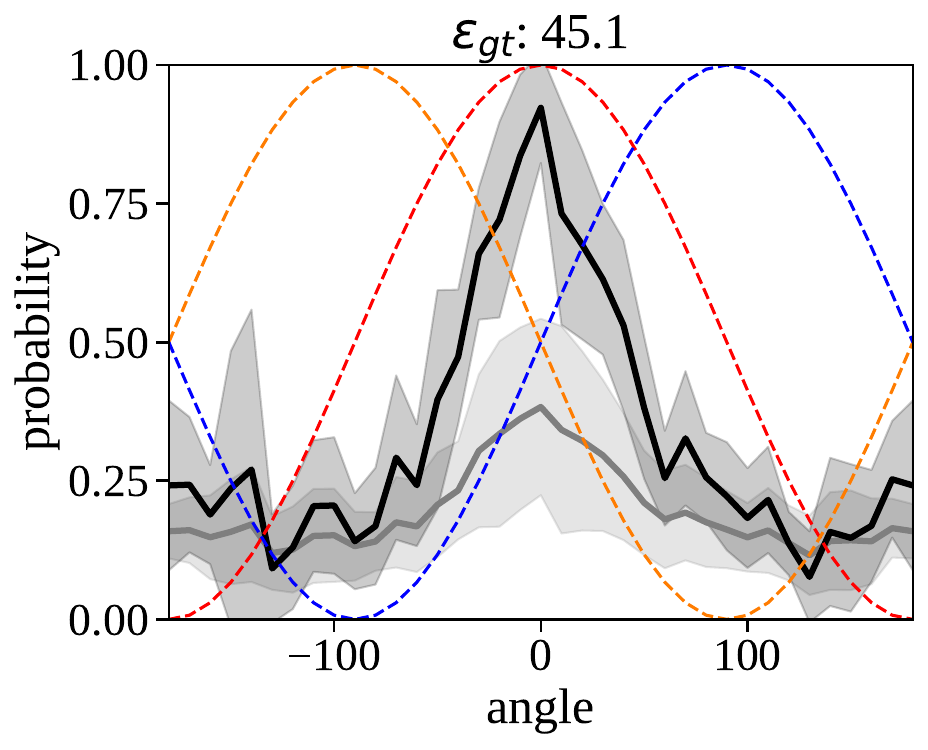} & \includegraphics[width=0.2\linewidth]{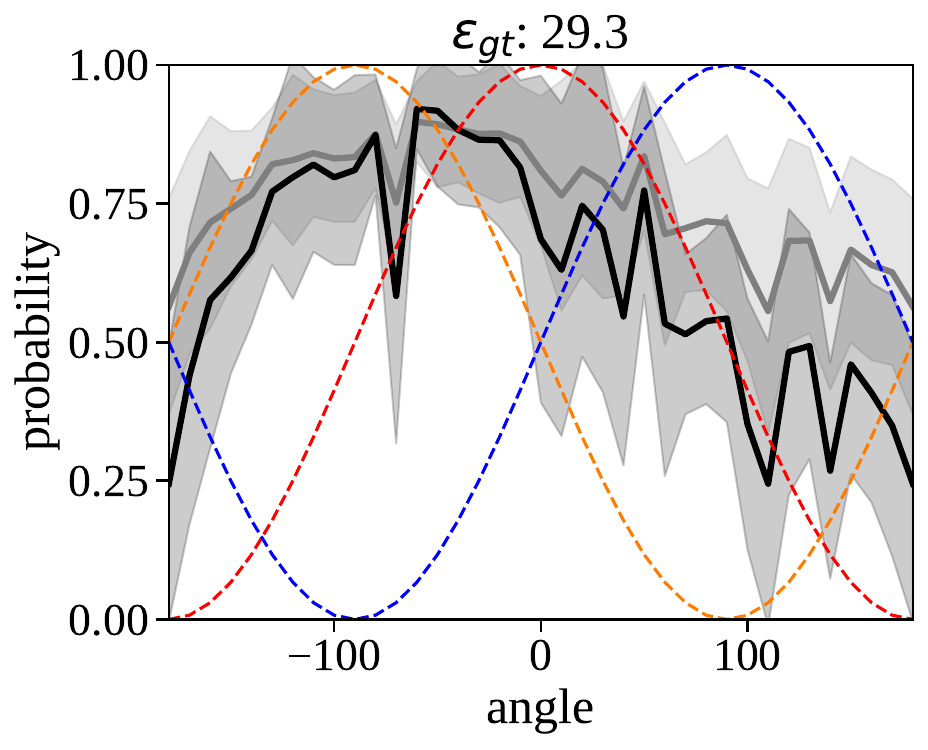} \\
\multirow{-8}{*}{\rotatebox[origin=c]{90}{\scriptsize GPT4o}} & \includegraphics[width=0.2\linewidth]{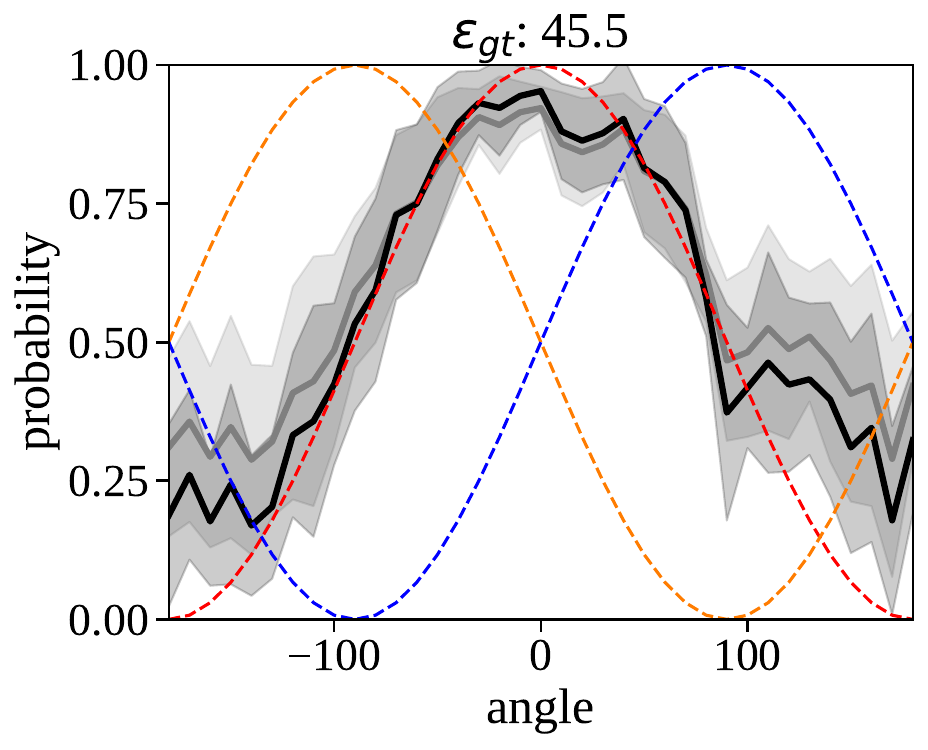} & \includegraphics[width=0.2\linewidth]{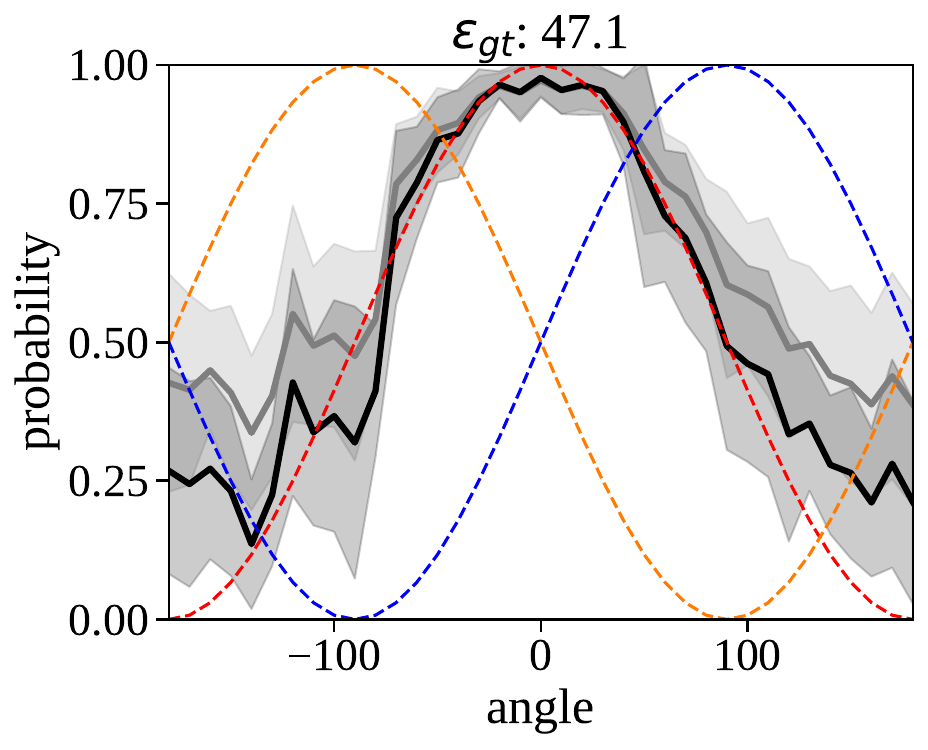} & \includegraphics[width=0.2\linewidth]{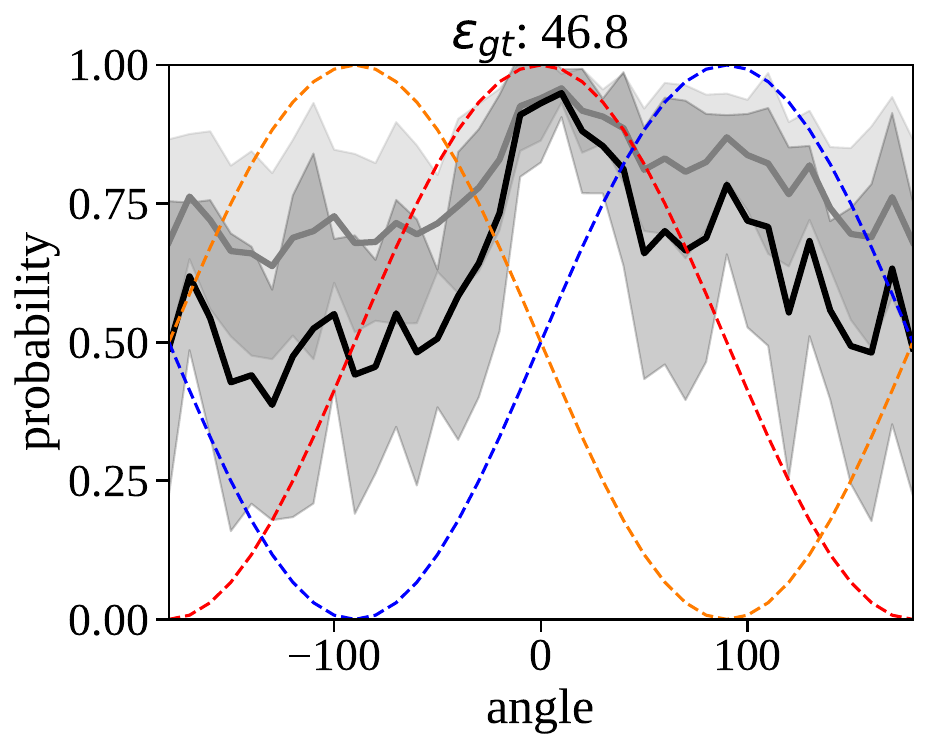} & \includegraphics[width=0.2\linewidth]{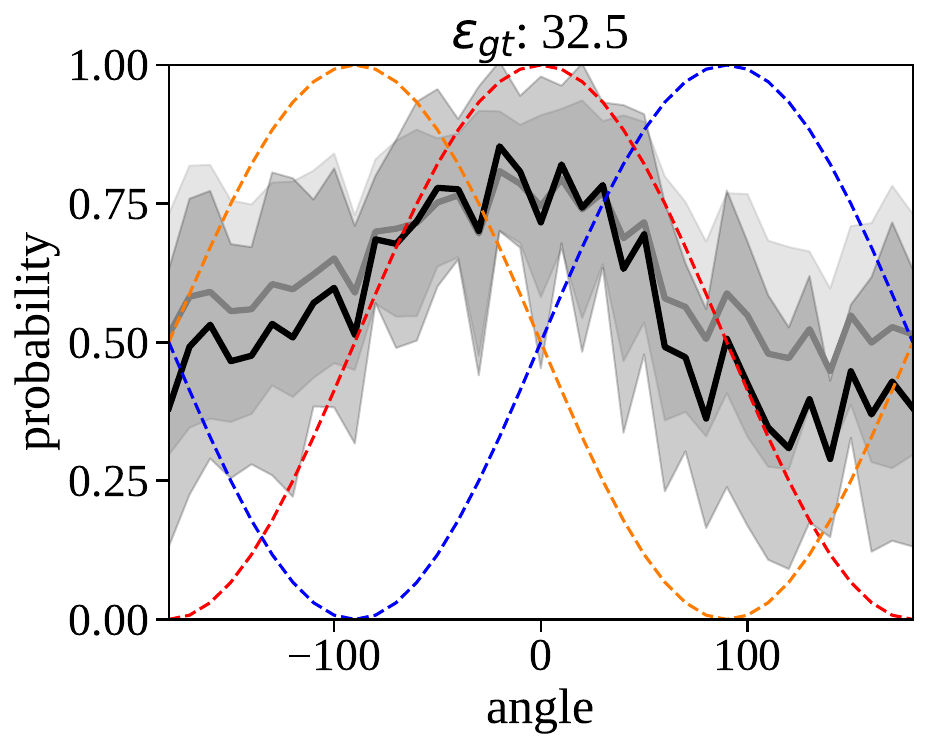} \\
  \end{tabular}
 }
\vspace*{-5pt}
\caption{All prediction plots for each model on \texttt{COMFORT-CAR} using the addressee perspective prompt (\texttt{add}). The raw probability $p(\theta)$ in gray, normalized probability $\widehat{p}(\theta)$ in black, and the reference probabilities $p_\textrm{cos}(\theta)$ of \texttt{cam} in red, \texttt{add} in orange, \texttt{rel} in blue. To avoid overlapping reference probabilities of \texttt{add} and \texttt{rel}, we use plots on \texttt{COMFORT-CAR} with relatum facing left for left and right relations and \texttt{COMFORT-CAR} with relatum facing right for front and behind relations.}
\label{fig:car-add-all}
\end{figure*}
\begin{figure*}[ht!]
  \centering
  \vspace*{-30pt}
  \makebox[\textwidth][c]{
  \begin{tabular}{ccccc}
  & \hphantom{aa} Left & \hphantom{aa} Right & \hphantom{aa} Front & \hphantom{aa} Back \\
\multirow{-8}{*}{\rotatebox[origin=c]{90}{\scriptsize InstructBLIP7B}} & \includegraphics[width=0.2\linewidth]{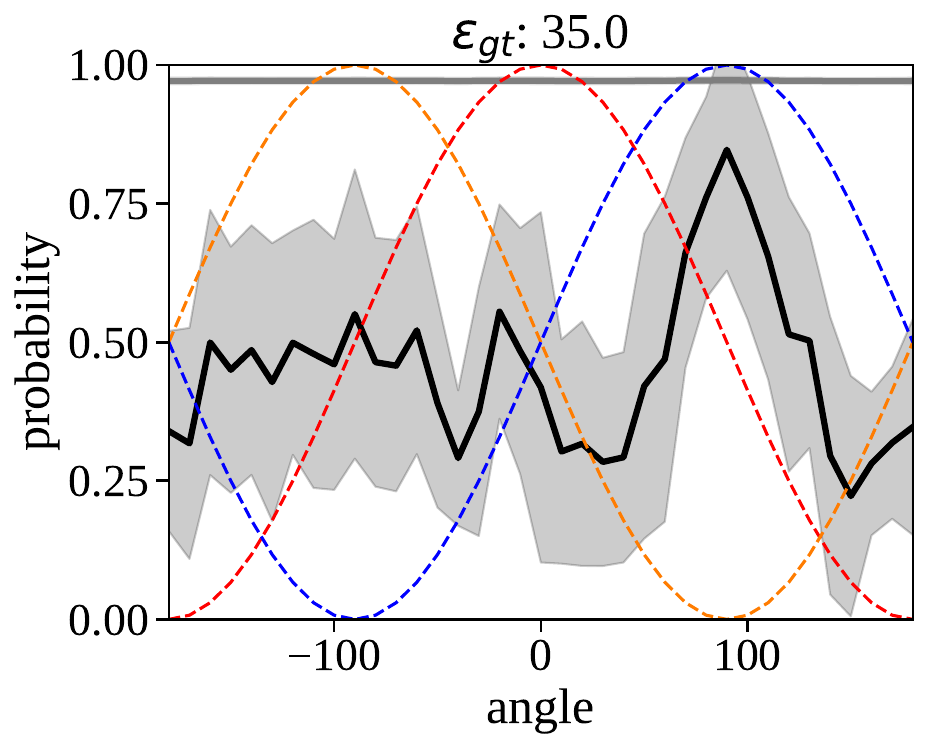} & \includegraphics[width=0.2\linewidth]{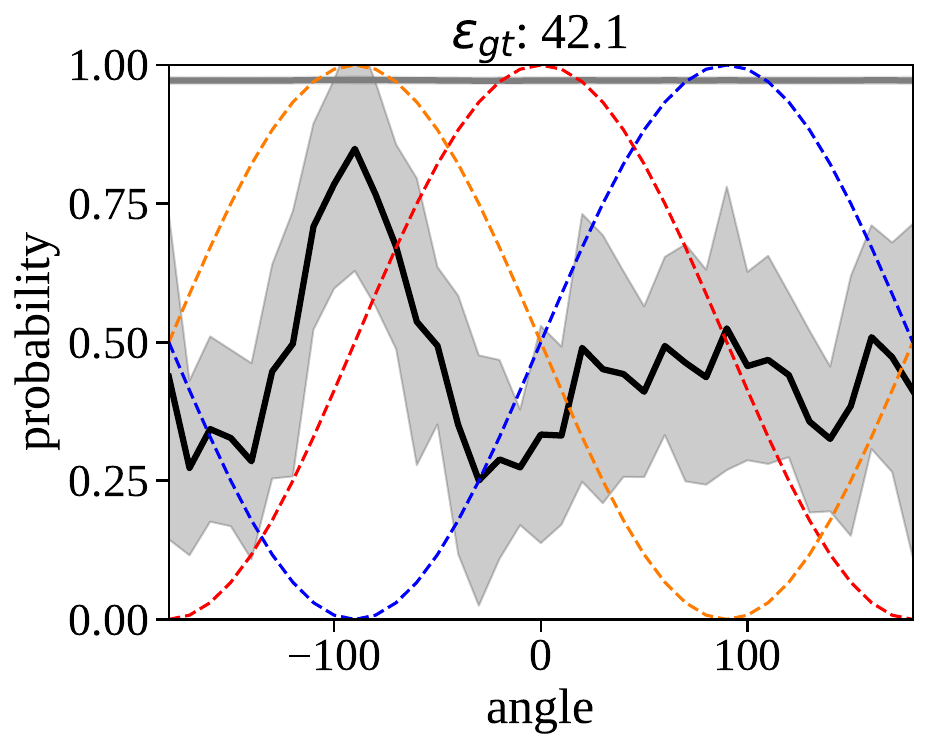} & \includegraphics[width=0.2\linewidth]{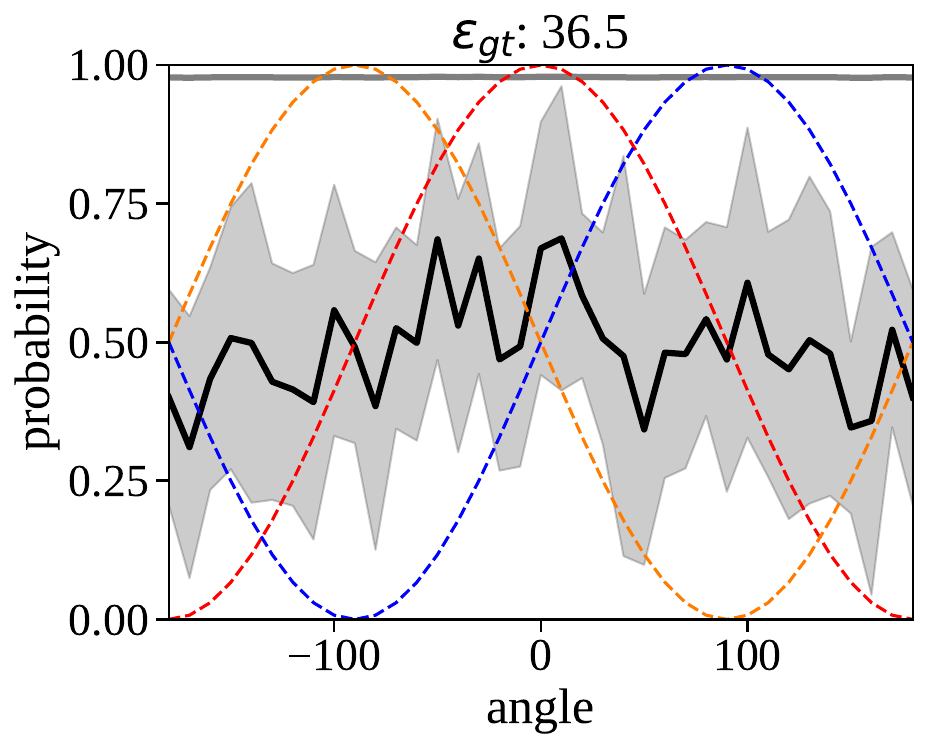} & \includegraphics[width=0.2\linewidth]{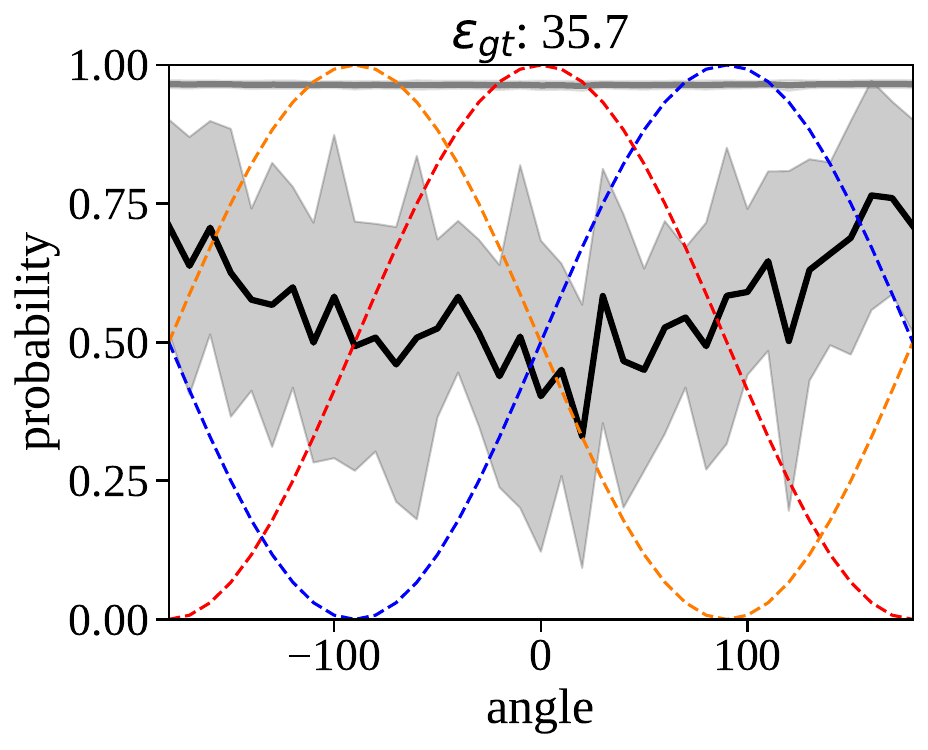} \\
\multirow{-8}{*}{\rotatebox[origin=c]{90}{\scriptsize InstructBLIP13B}} & \includegraphics[width=0.2\linewidth]{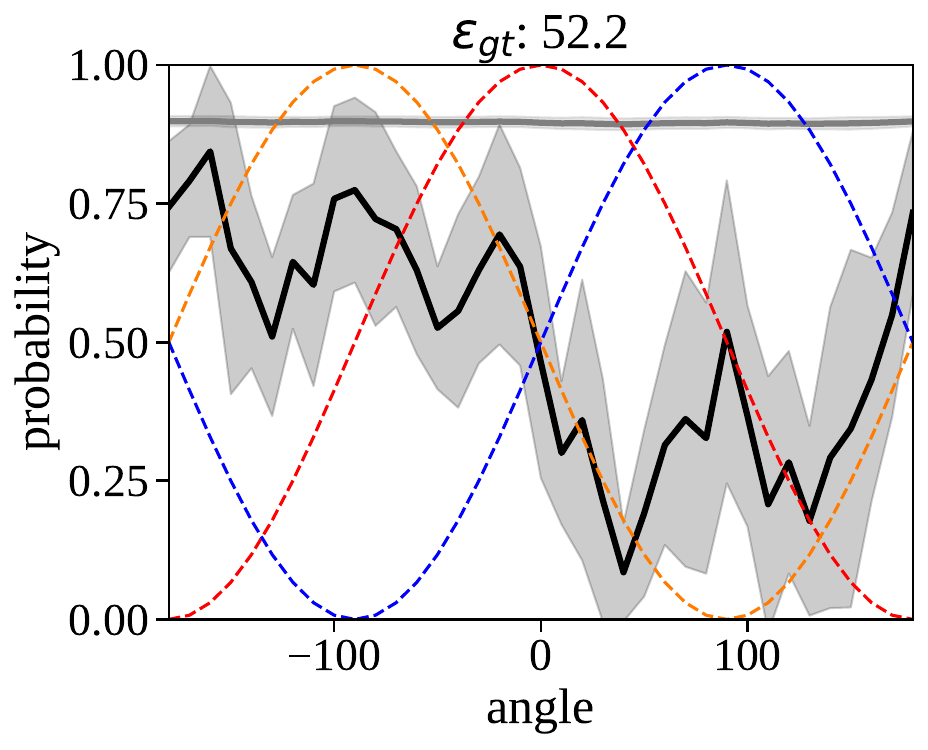} & \includegraphics[width=0.2\linewidth]{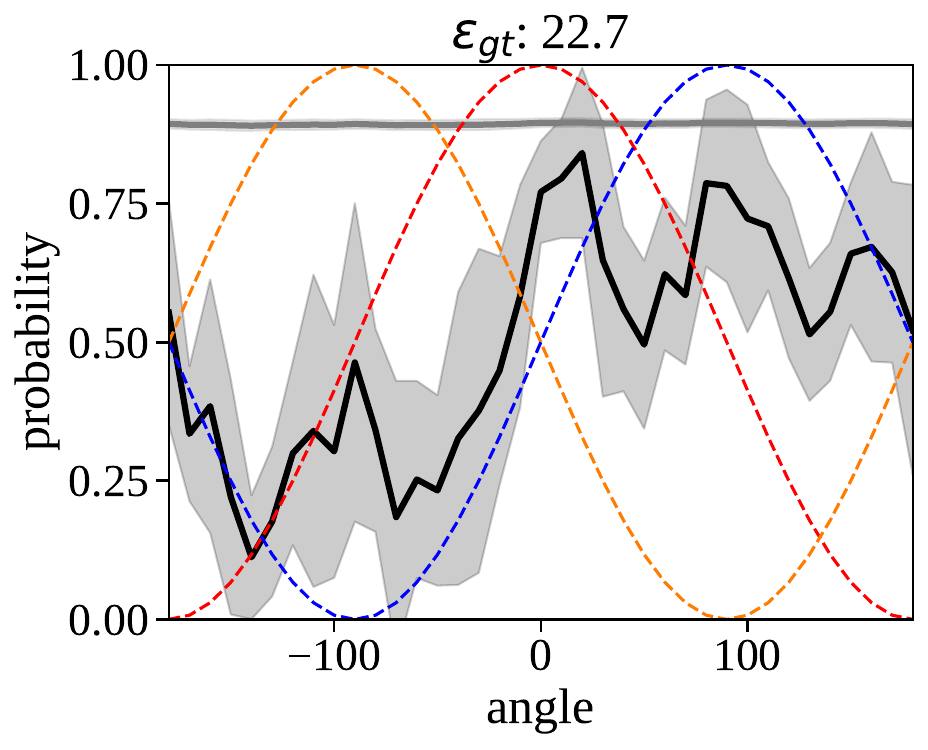} & \includegraphics[width=0.2\linewidth]{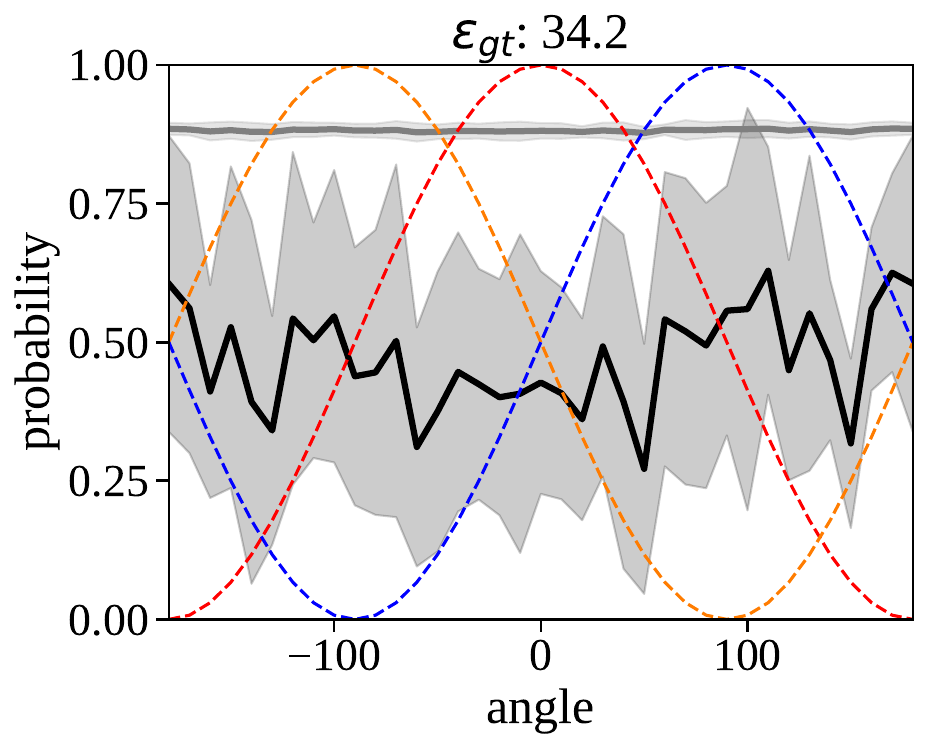} & \includegraphics[width=0.2\linewidth]{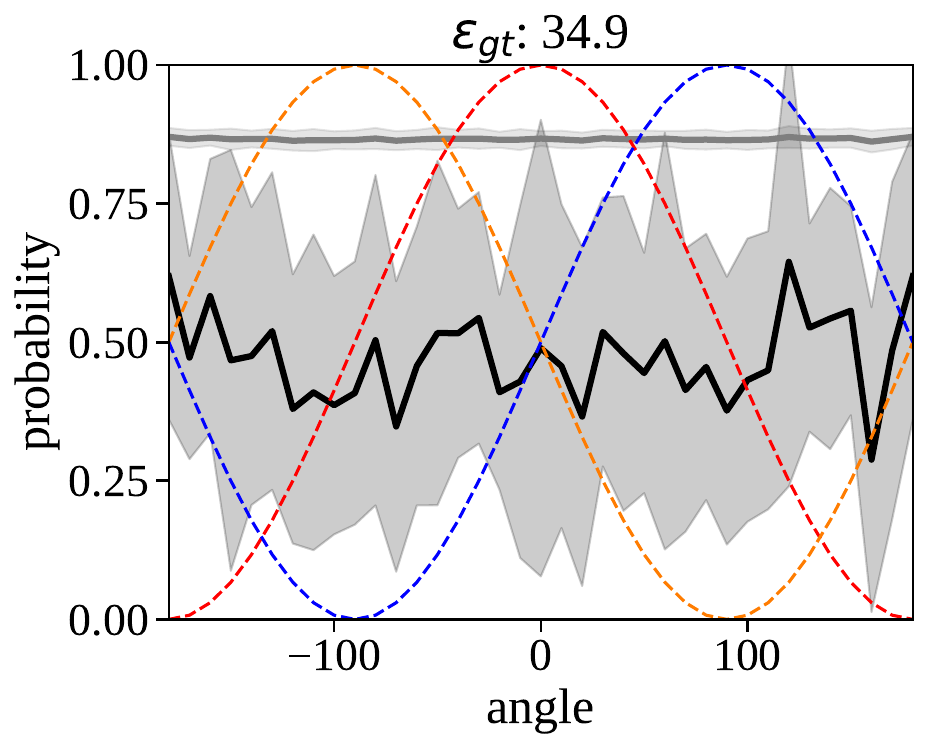} \\
\multirow{-8}{*}{\rotatebox[origin=c]{90}{\scriptsize MBLIPBLOOMZ7B}} & \includegraphics[width=0.2\linewidth]{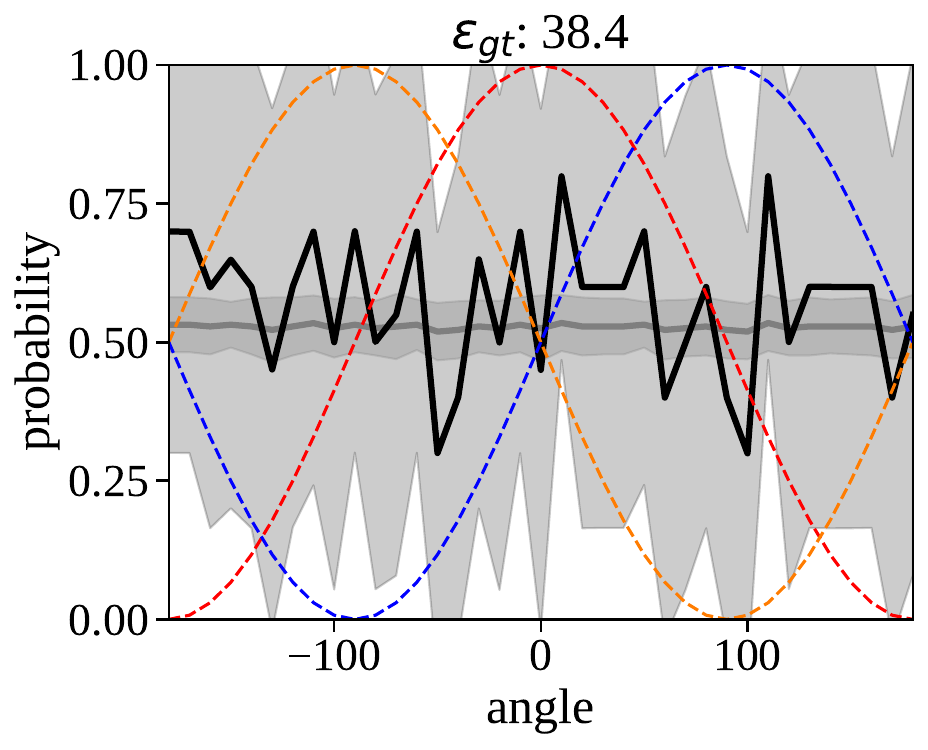} & \includegraphics[width=0.2\linewidth]{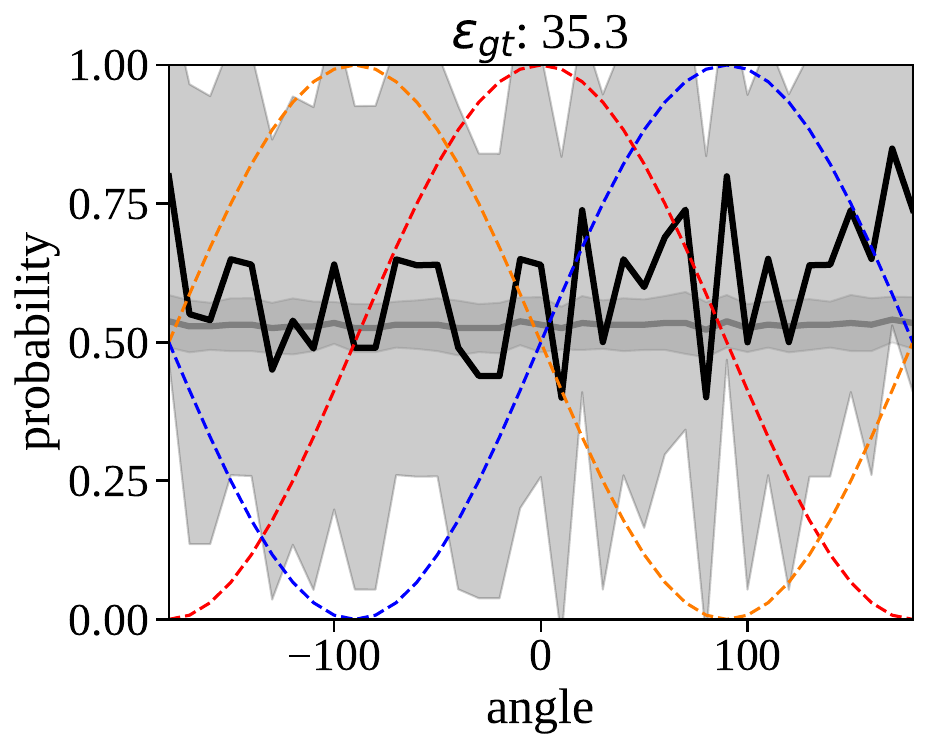} & \includegraphics[width=0.2\linewidth]{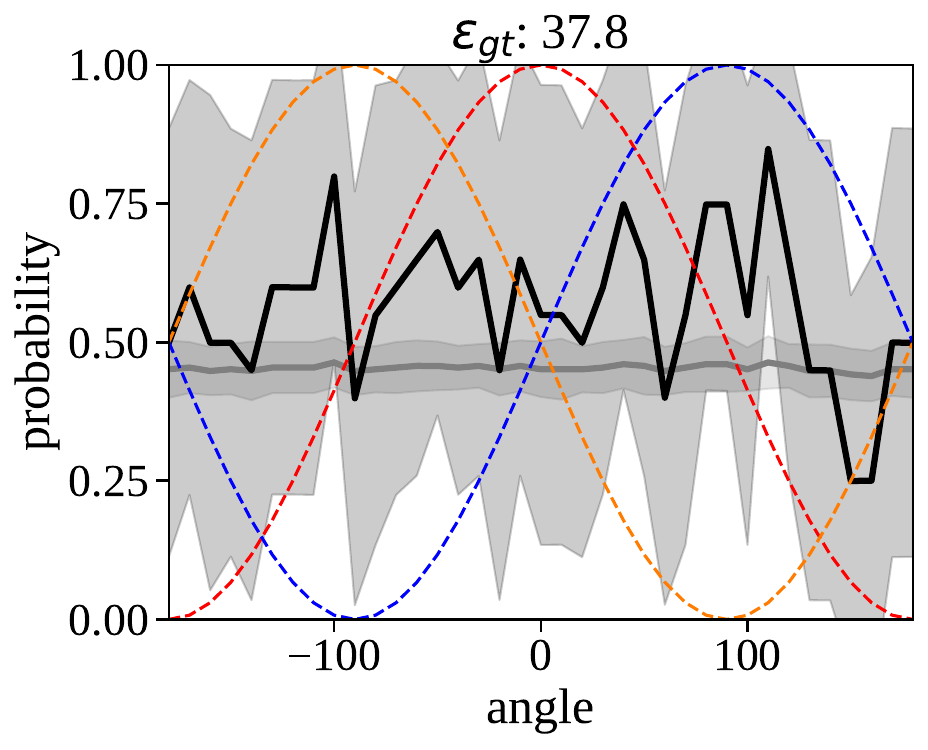} & \includegraphics[width=0.2\linewidth]{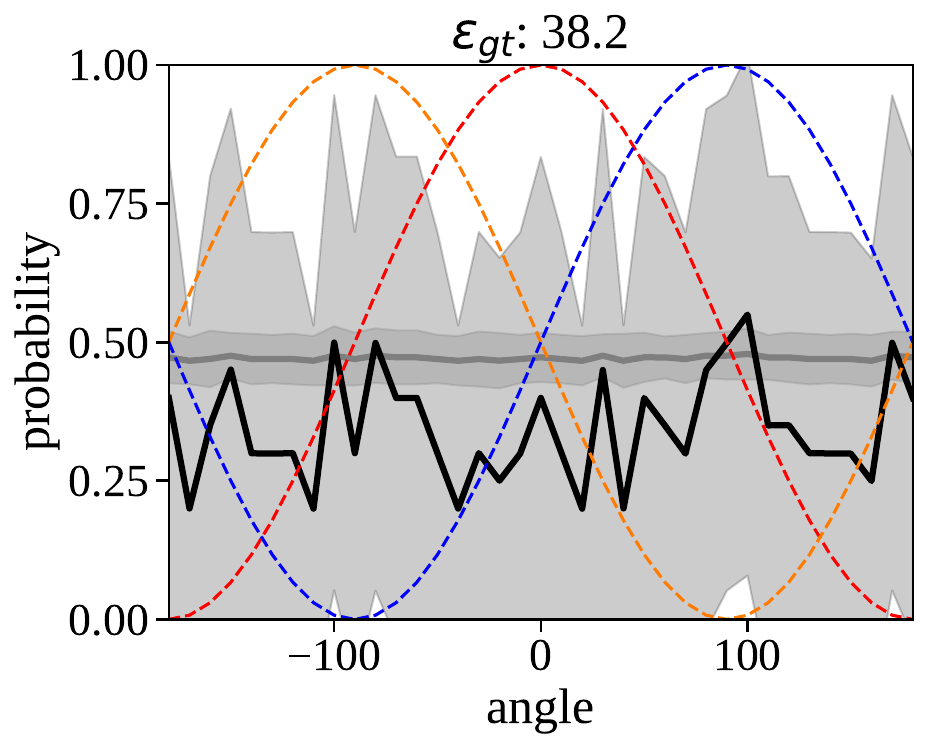} \\
\multirow{-8}{*}{\rotatebox[origin=c]{90}{\scriptsize GLaMM}} & \includegraphics[width=0.2\linewidth]{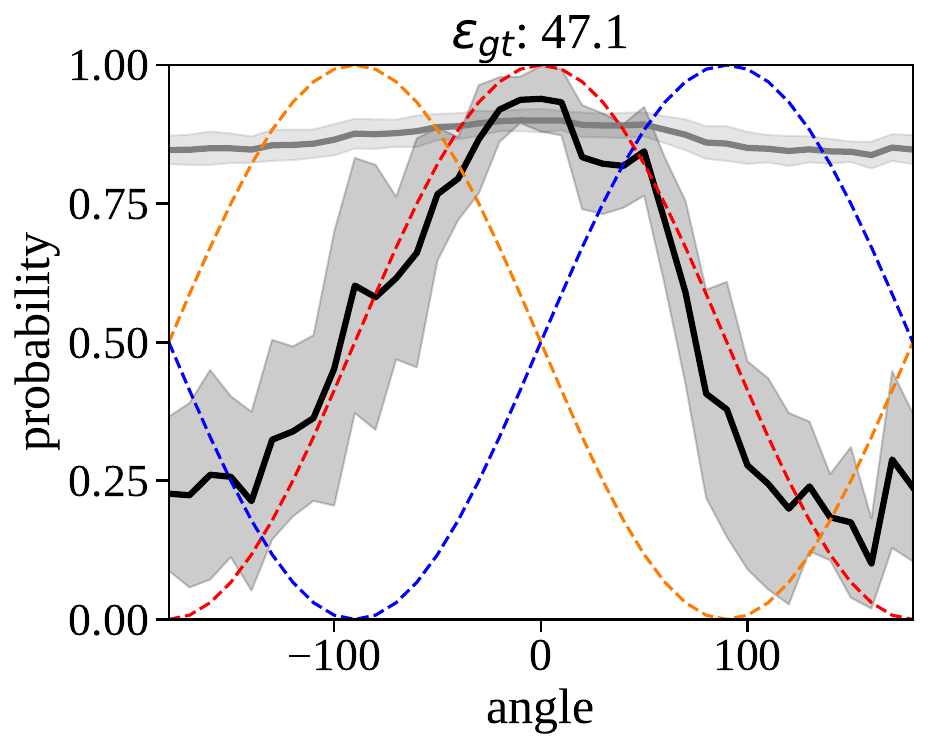} & \includegraphics[width=0.2\linewidth]{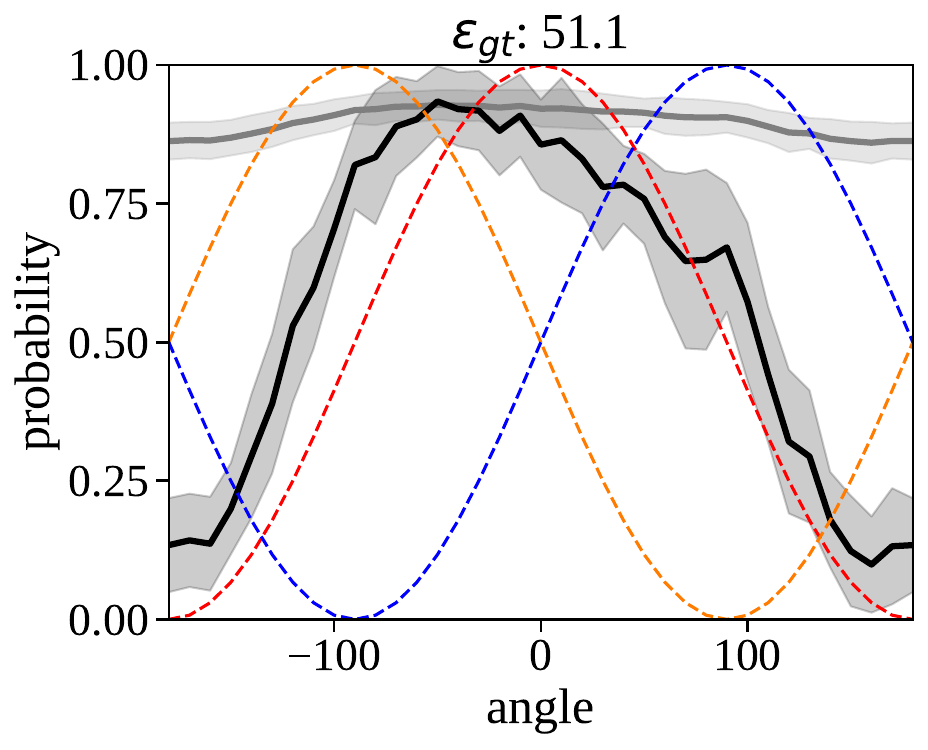} & \includegraphics[width=0.2\linewidth]{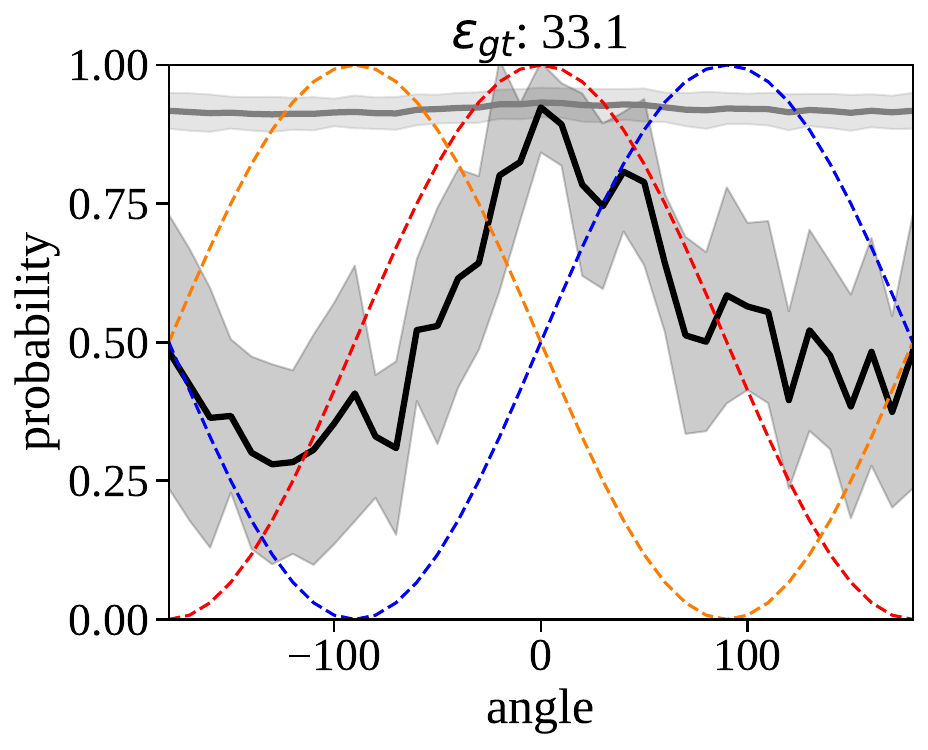} & \includegraphics[width=0.2\linewidth]{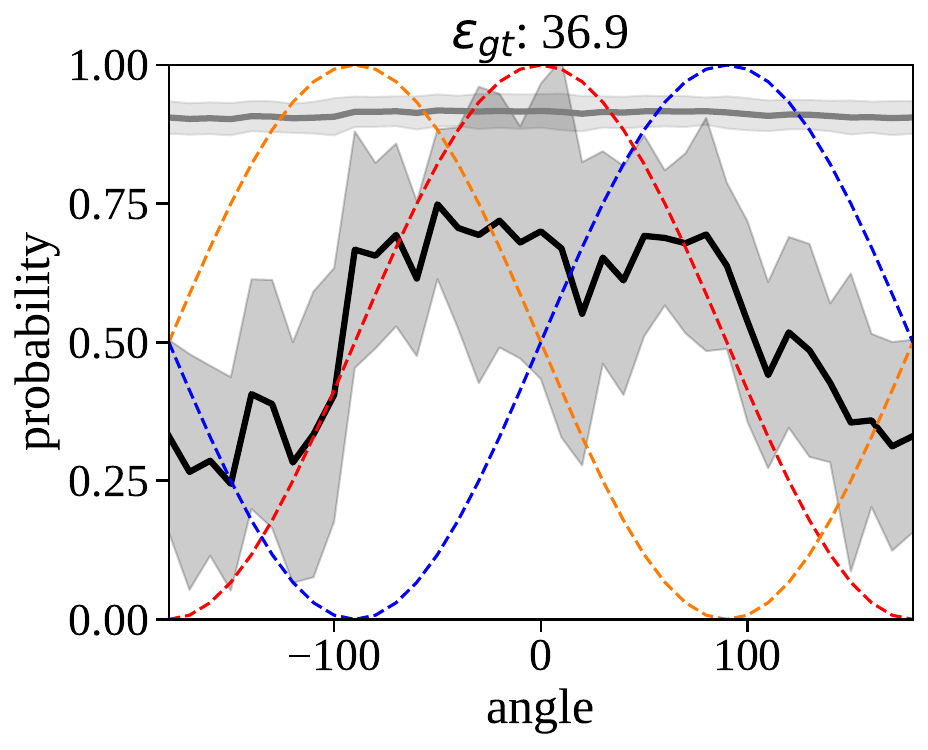} \\
\multirow{-8}{*}{\rotatebox[origin=c]{90}{\scriptsize LLaVA1.57B}} & \includegraphics[width=0.2\linewidth]{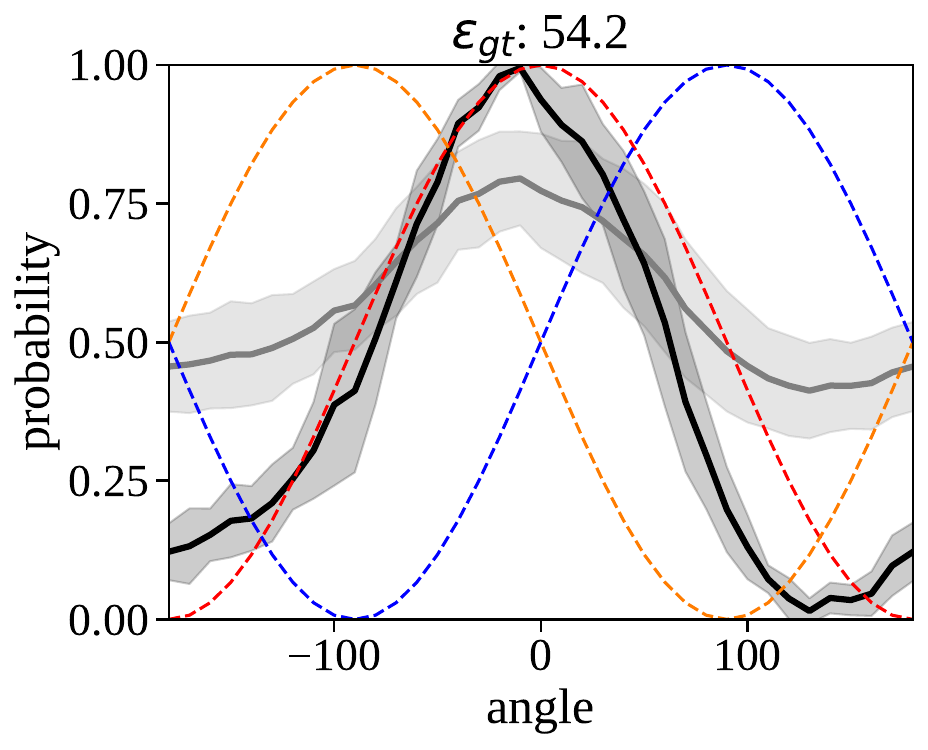} & \includegraphics[width=0.2\linewidth]{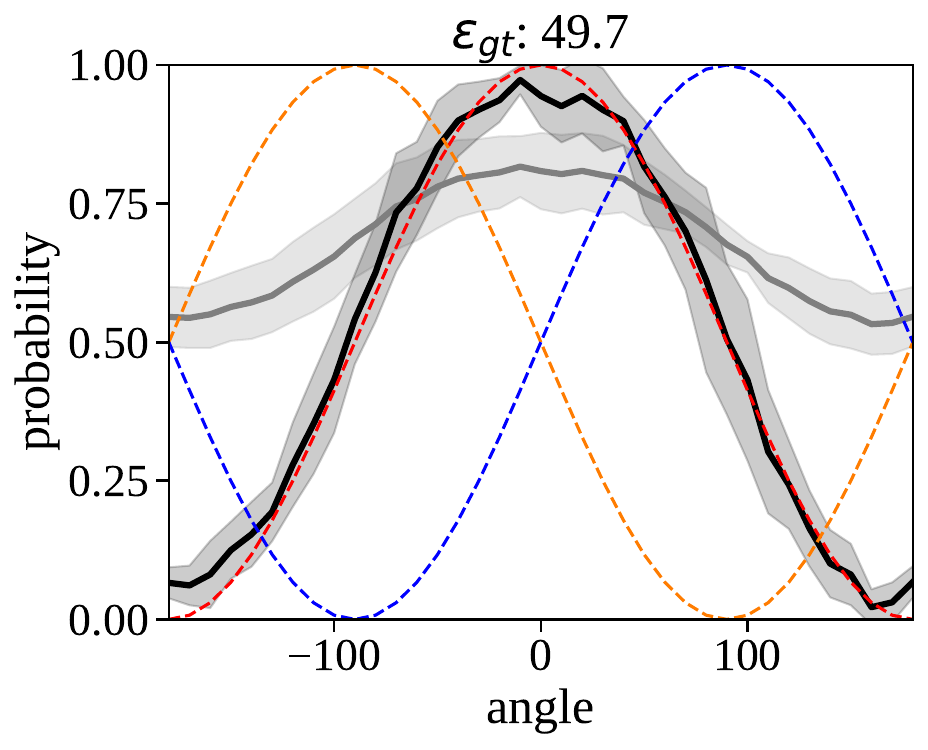} & \includegraphics[width=0.2\linewidth]{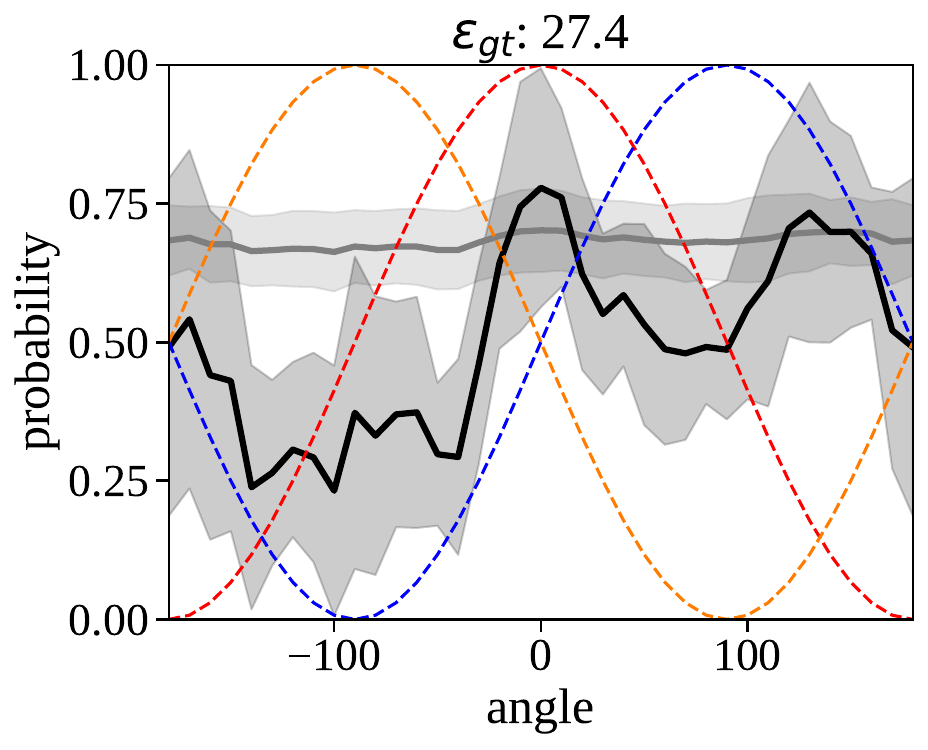} & \includegraphics[width=0.2\linewidth]{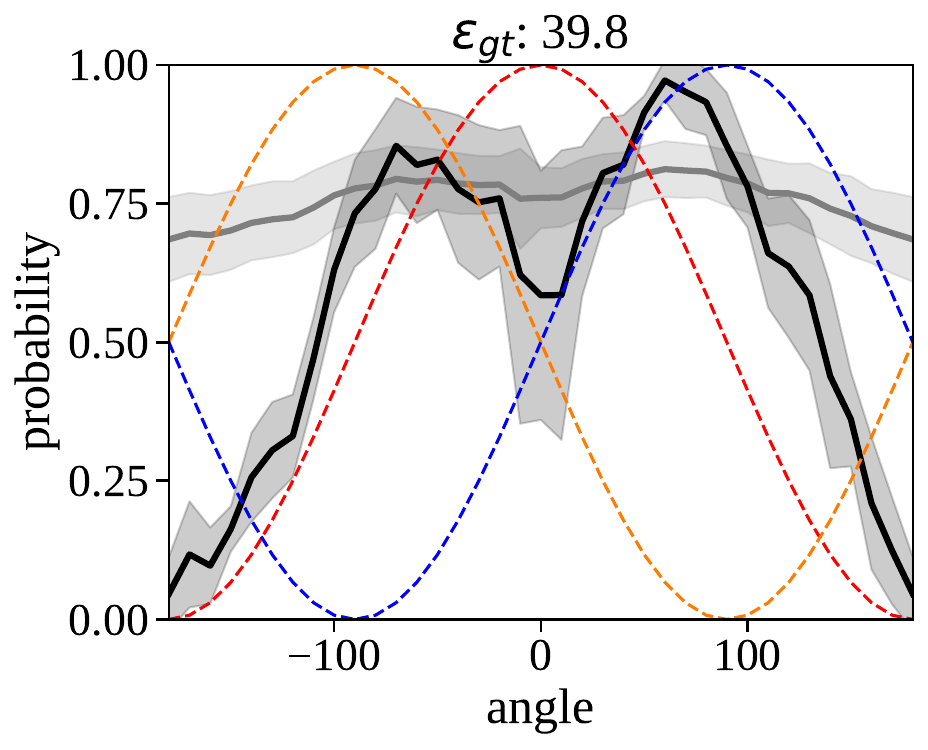} \\
\multirow{-8}{*}{\rotatebox[origin=c]{90}{\scriptsize LLaVA1.513B}} & \includegraphics[width=0.2\linewidth]{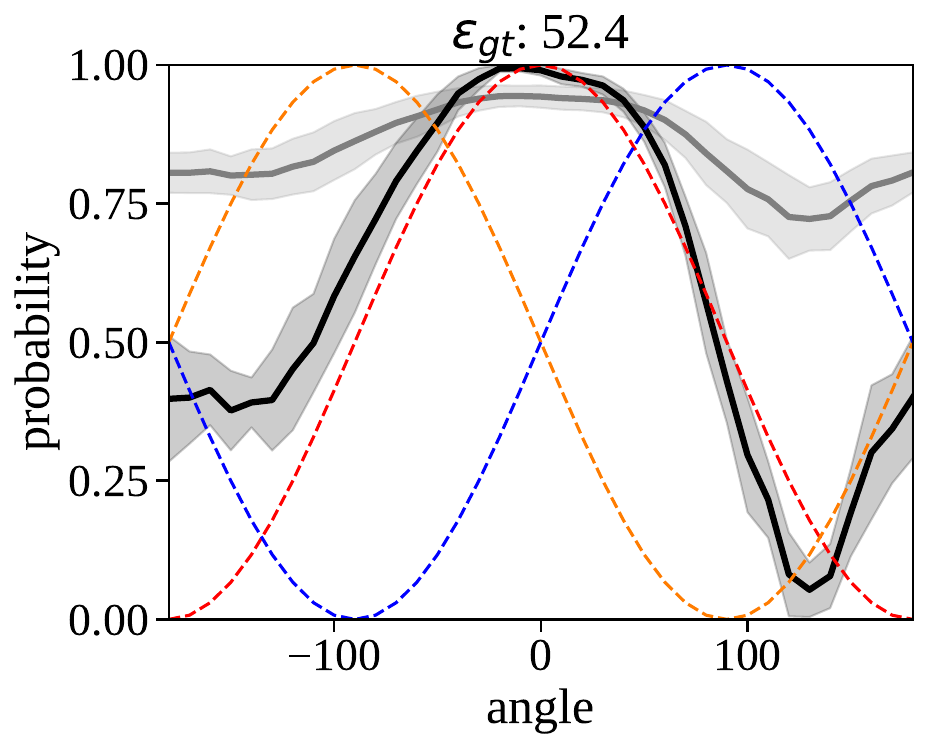} & \includegraphics[width=0.2\linewidth]{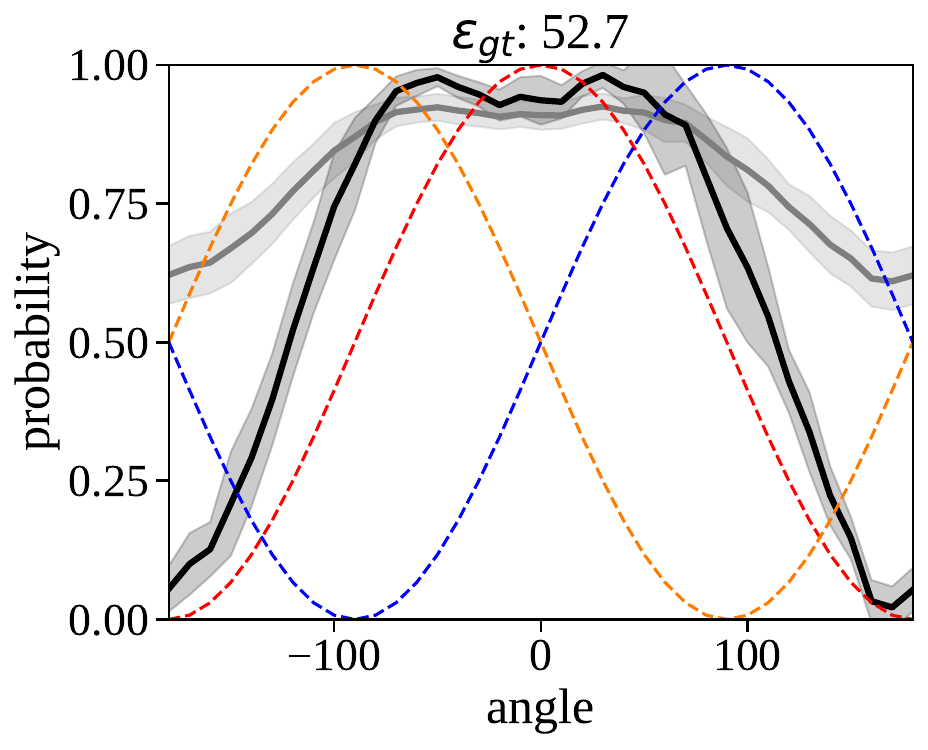} & \includegraphics[width=0.2\linewidth]{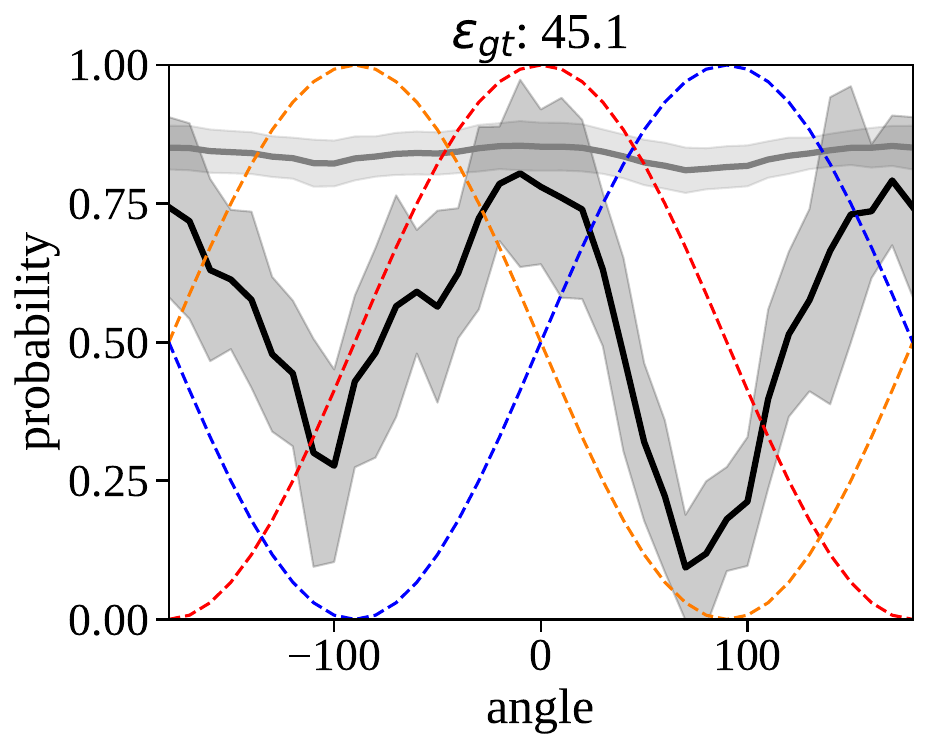} & \includegraphics[width=0.2\linewidth]{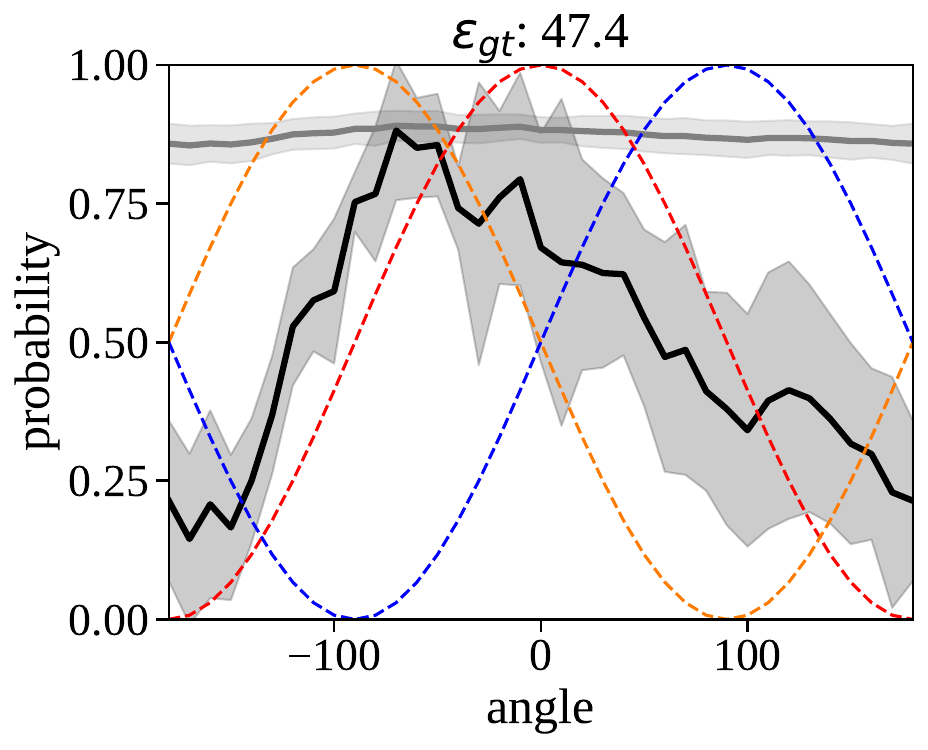} \\
\multirow{-8}{*}{\rotatebox[origin=c]{90}{\scriptsize XComposer2}} & \includegraphics[width=0.2\linewidth]{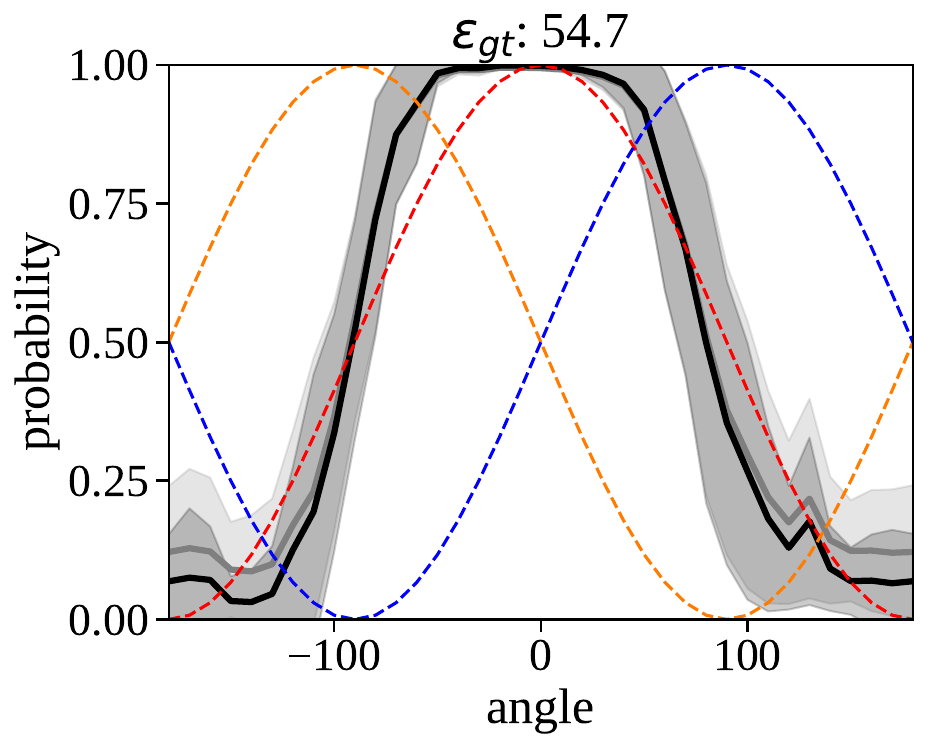} & \includegraphics[width=0.2\linewidth]{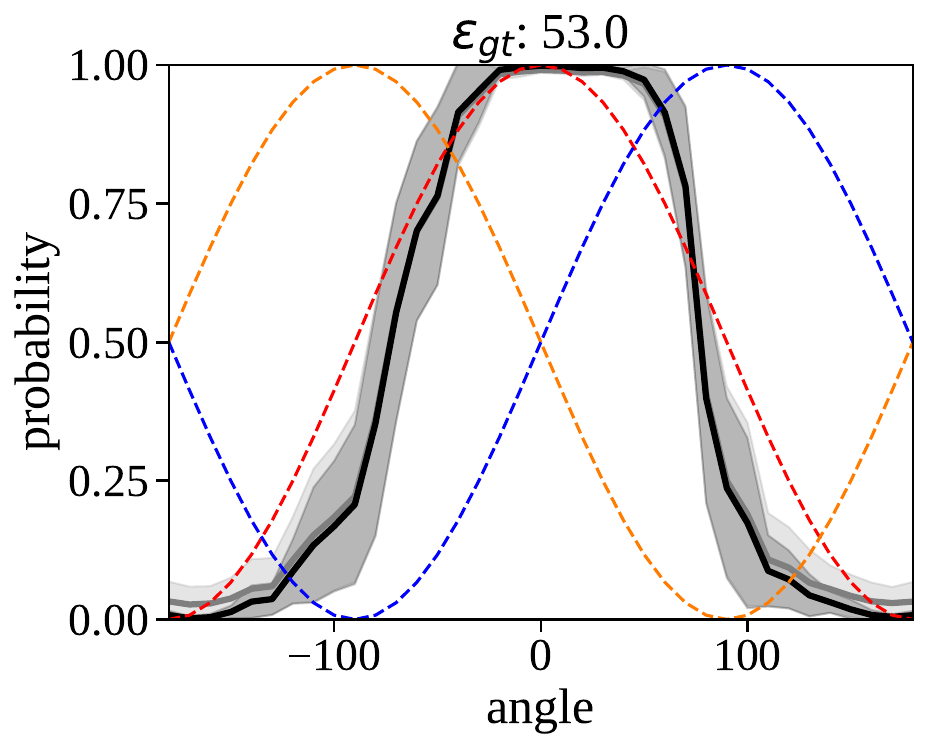} & \includegraphics[width=0.2\linewidth]{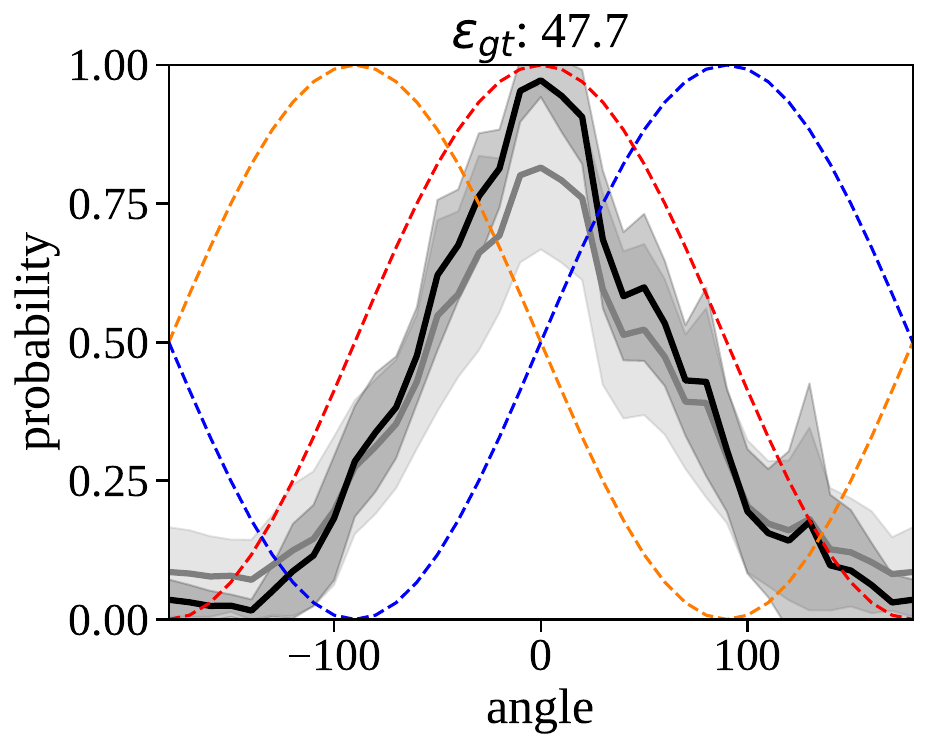} & \includegraphics[width=0.2\linewidth]{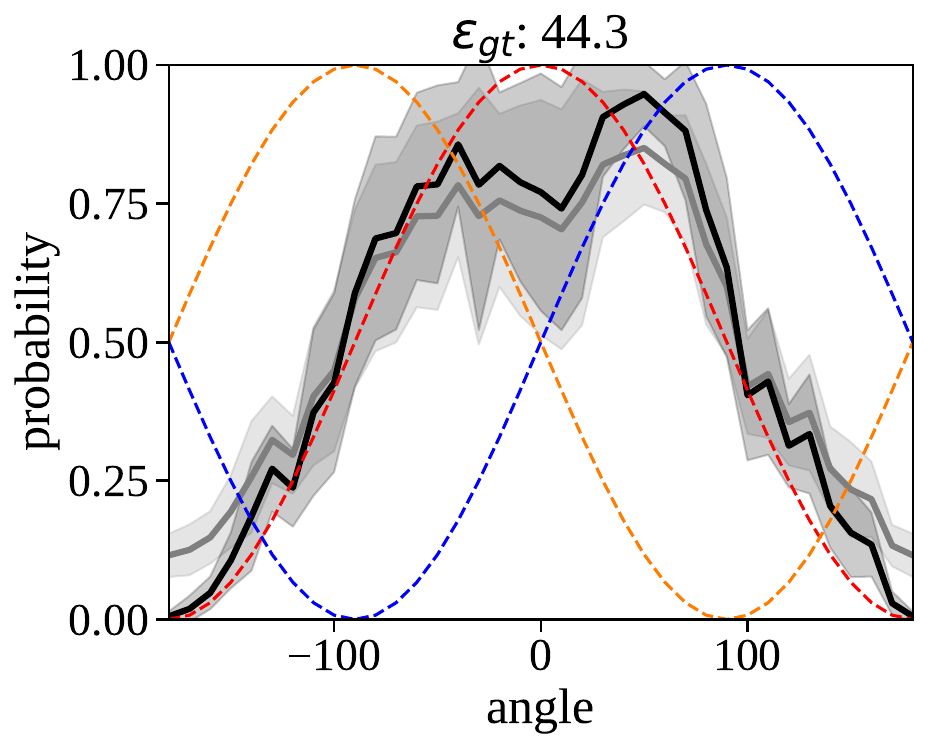} \\
\multirow{-8}{*}{\rotatebox[origin=c]{90}{\scriptsize MiniCPM}} & \includegraphics[width=0.2\linewidth]{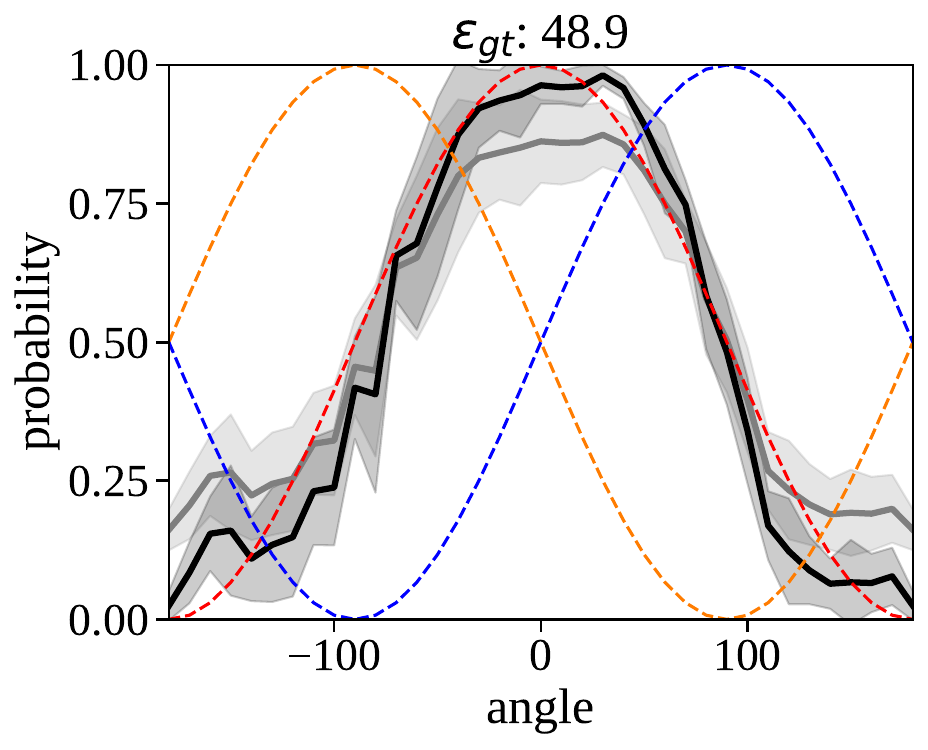} & \includegraphics[width=0.2\linewidth]{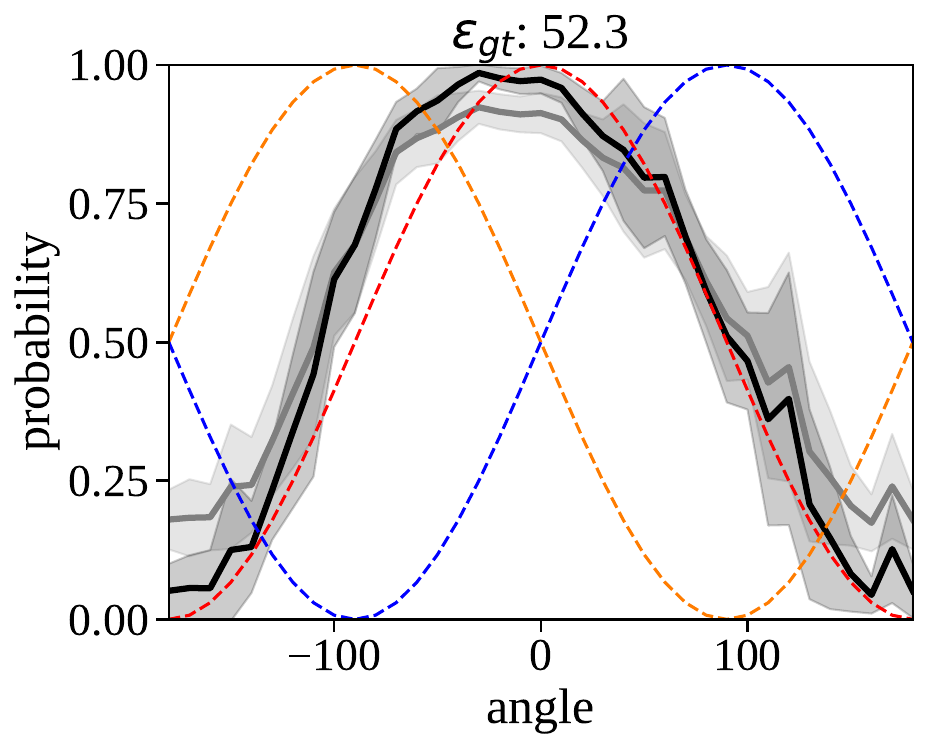} & \includegraphics[width=0.2\linewidth]{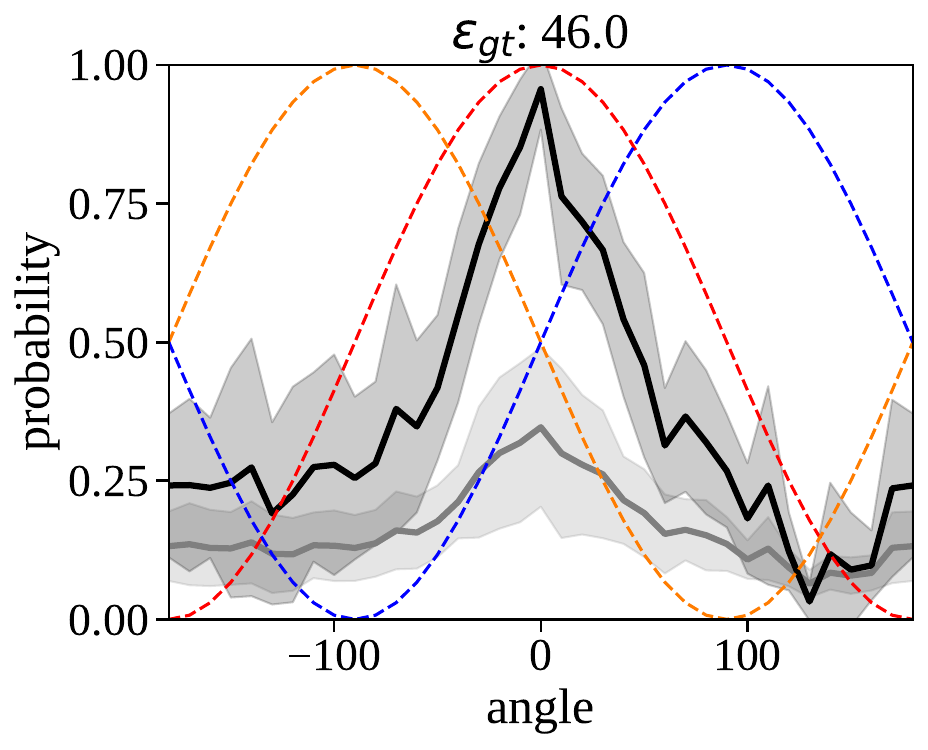} & \includegraphics[width=0.2\linewidth]{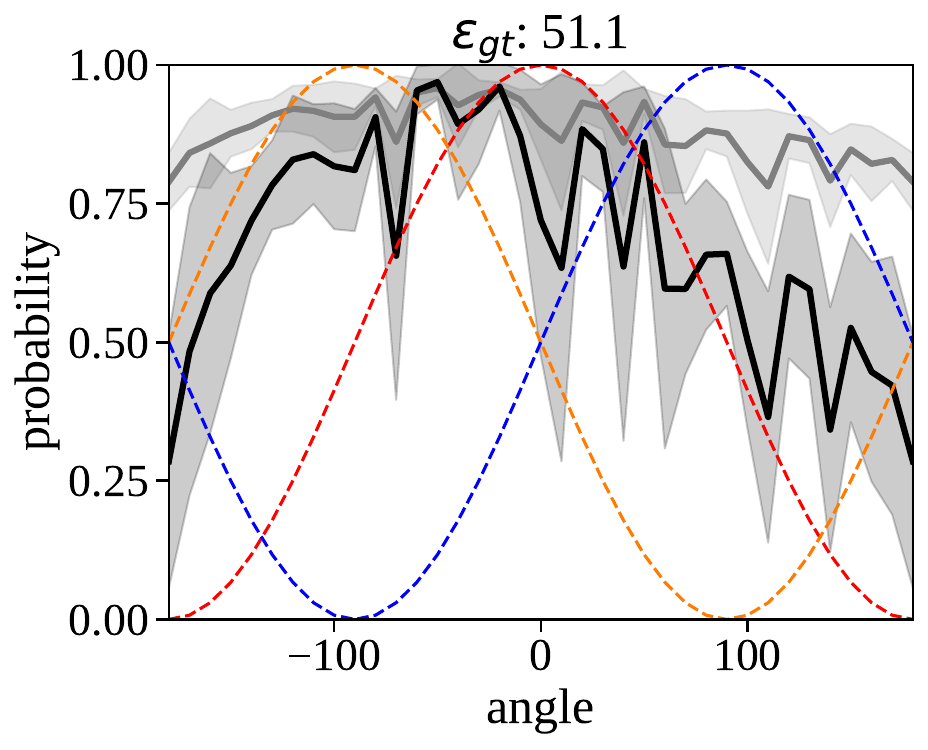} \\
\multirow{-8}{*}{\rotatebox[origin=c]{90}{\scriptsize GPT4o}} & \includegraphics[width=0.2\linewidth]{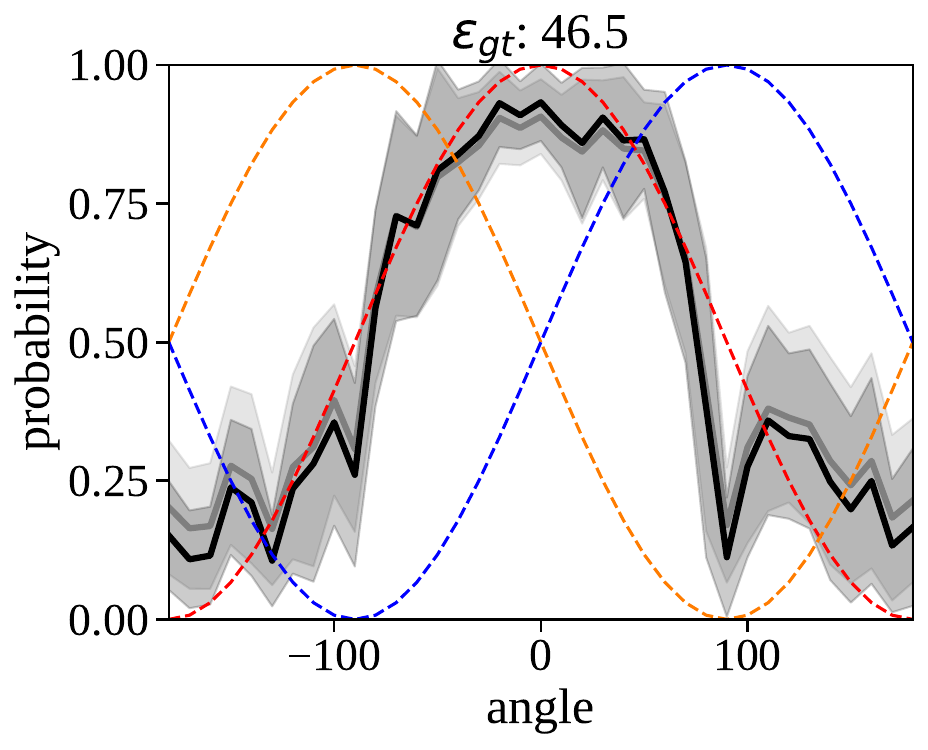} & \includegraphics[width=0.2\linewidth]{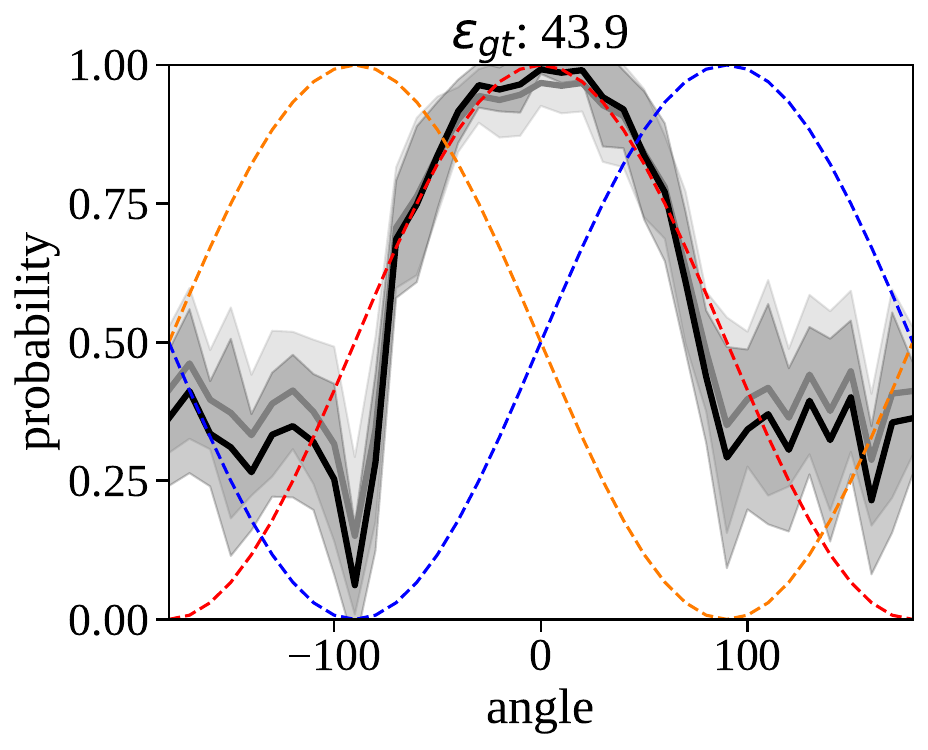} & \includegraphics[width=0.2\linewidth]{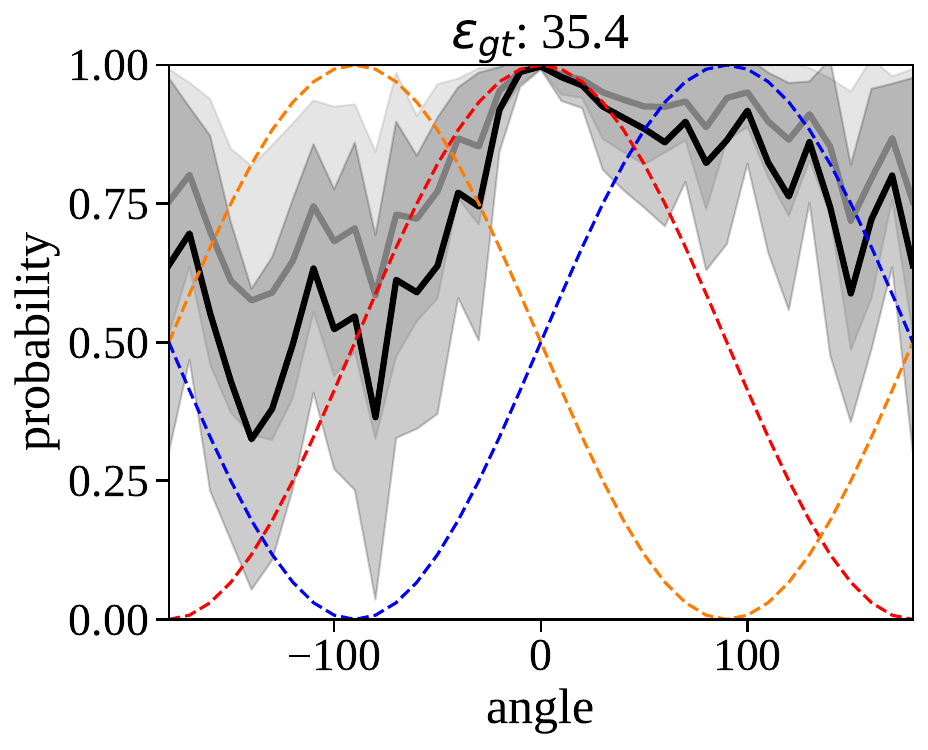} & \includegraphics[width=0.2\linewidth]{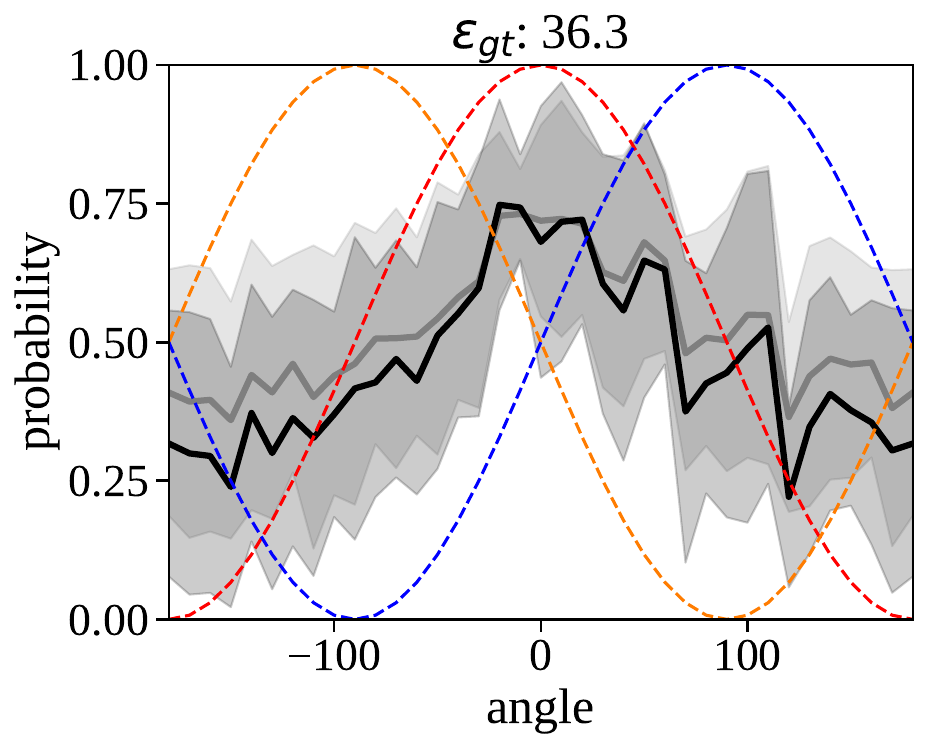} \\
  \end{tabular}
 }
\vspace*{-5pt}
\caption{All prediction plots for each model on \texttt{COMFORT-CAR} using the relatum perspective prompt (\texttt{rel}). The raw probability $p(\theta)$ in gray, normalized probability $\widehat{p}(\theta)$ in black, and the reference probabilities $p_\textrm{cos}(\theta)$ of \texttt{cam} in red, \texttt{add} in orange, \texttt{rel} in blue. To avoid overlapping reference probabilities of \texttt{add} and \texttt{rel}, we use plots on \texttt{COMFORT-CAR} with relatum facing left for left and right relations and \texttt{COMFORT-CAR} with relatum facing right for front and behind relations.}
\label{fig:car-rel-all}
\end{figure*}
\begin{figure*}[ht!]
  \centering
  \vspace*{-30pt}
  \makebox[\textwidth][c]{
  \begin{tabular}{ccccc}
  & \hphantom{aa} Left & \hphantom{aa} Right & \hphantom{aa} Front & \hphantom{aa} Back \\
\multirow{-8}{*}{\rotatebox[origin=c]{90}{\scriptsize InstructBLIP7B}} & \includegraphics[width=0.2\linewidth]{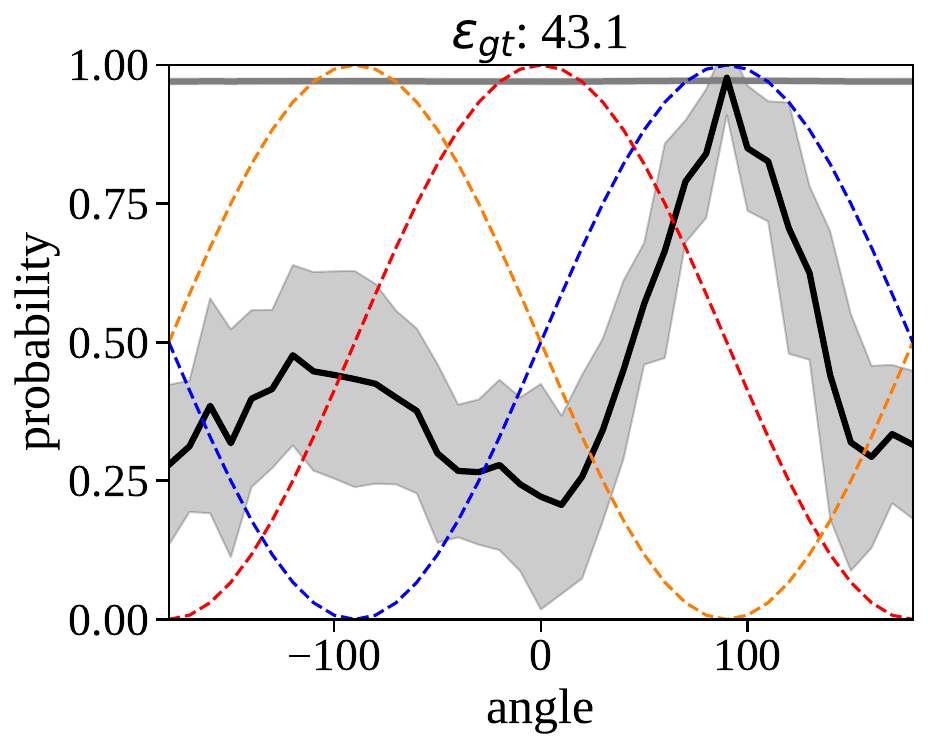} & \includegraphics[width=0.2\linewidth]{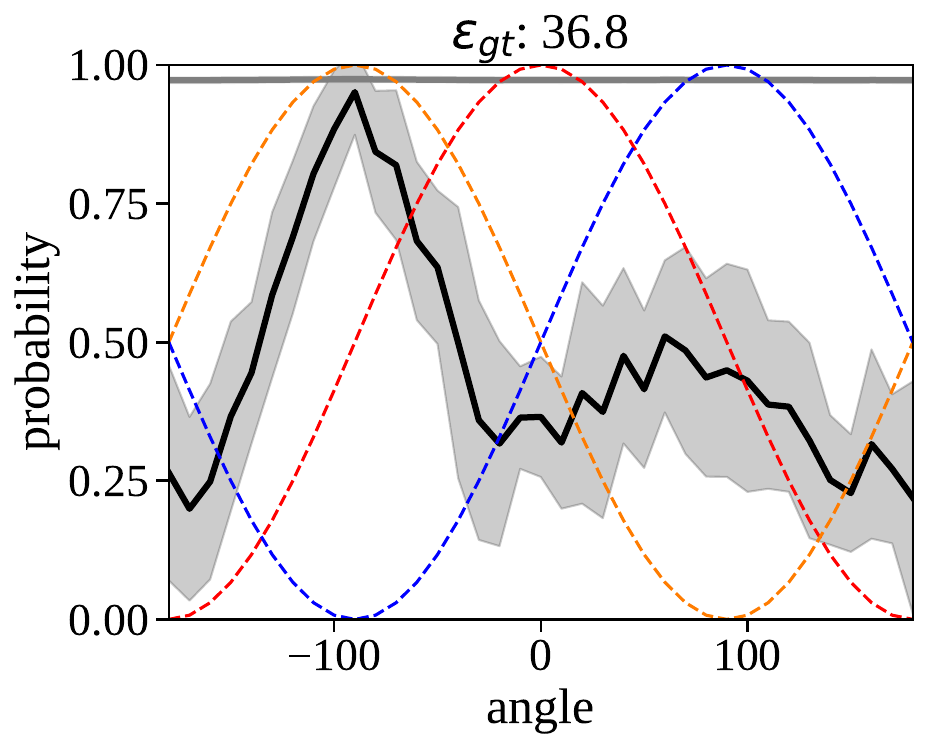} & \includegraphics[width=0.2\linewidth]{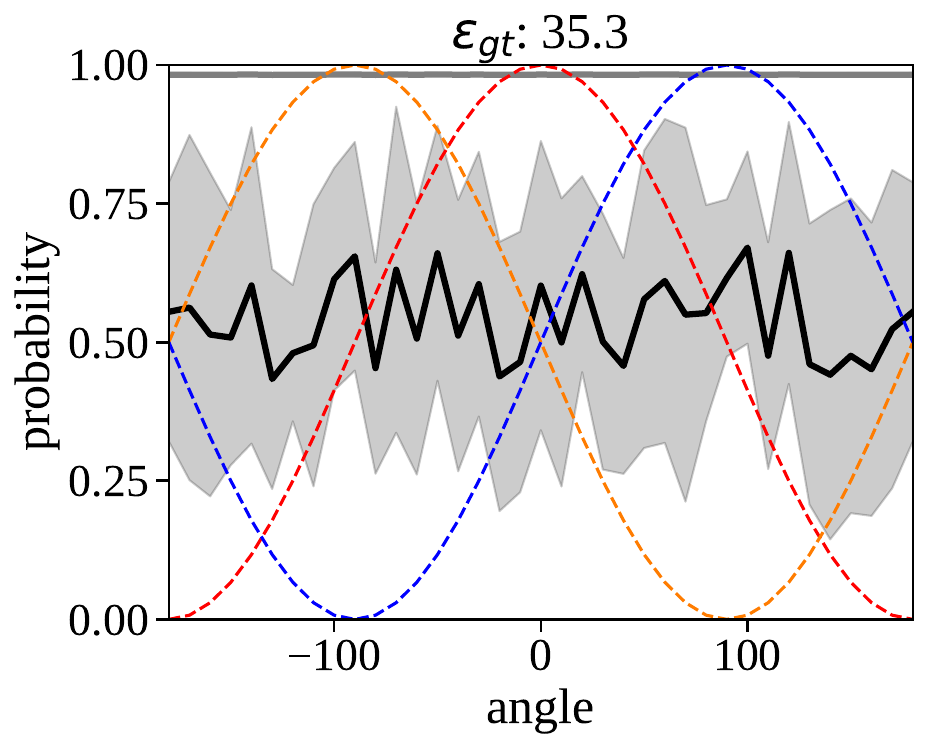} & \includegraphics[width=0.2\linewidth]{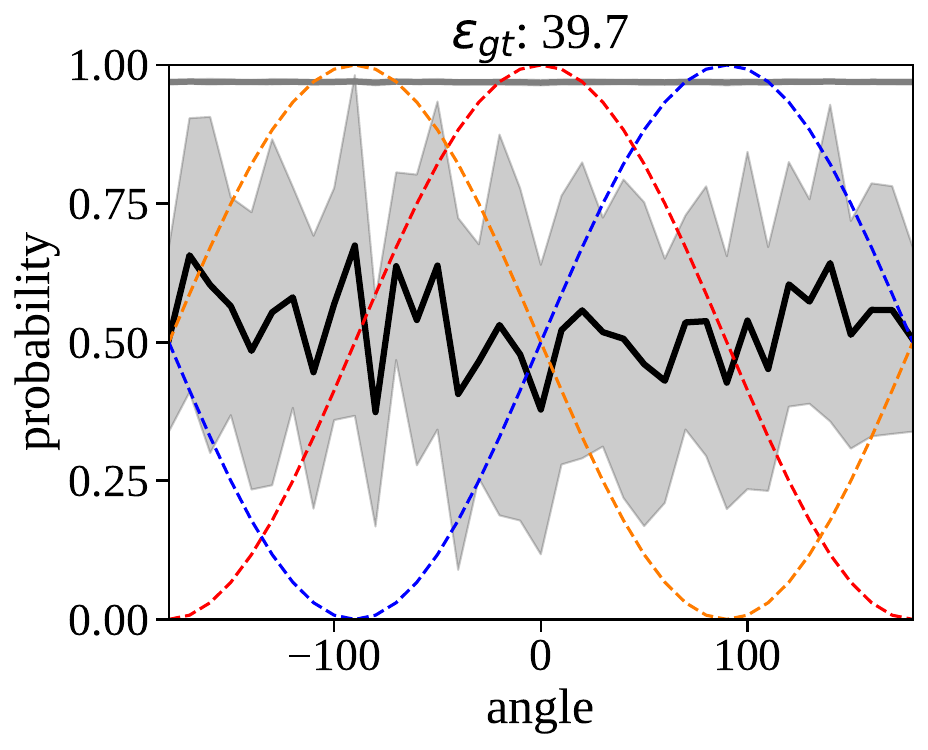} \\
\multirow{-8}{*}{\rotatebox[origin=c]{90}{\scriptsize InstructBLIP13B}} & \includegraphics[width=0.2\linewidth]{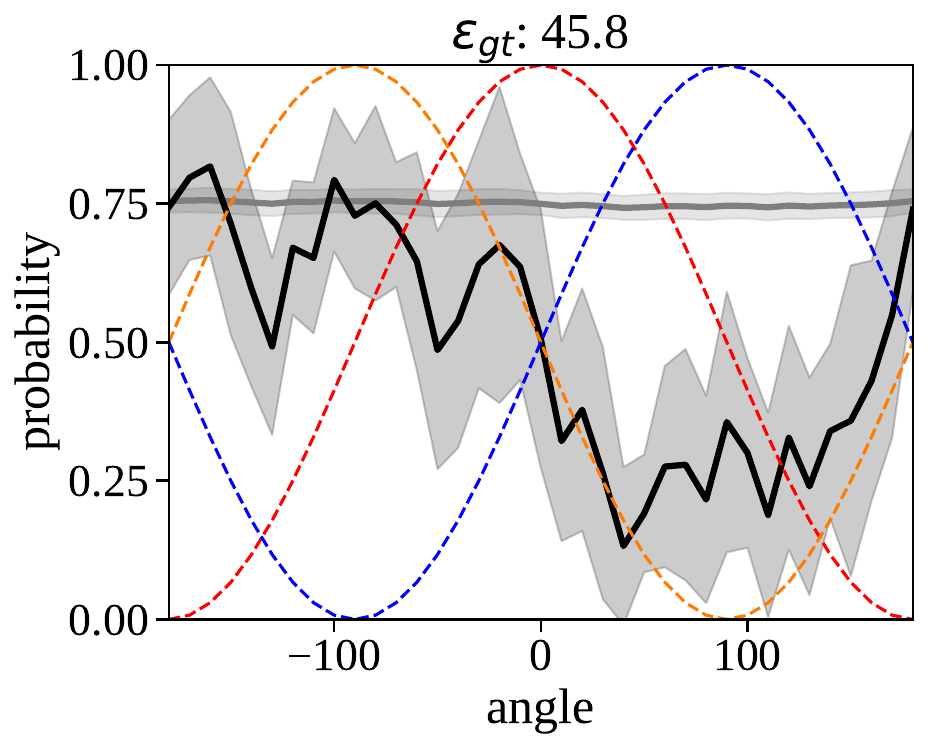} & \includegraphics[width=0.2\linewidth]{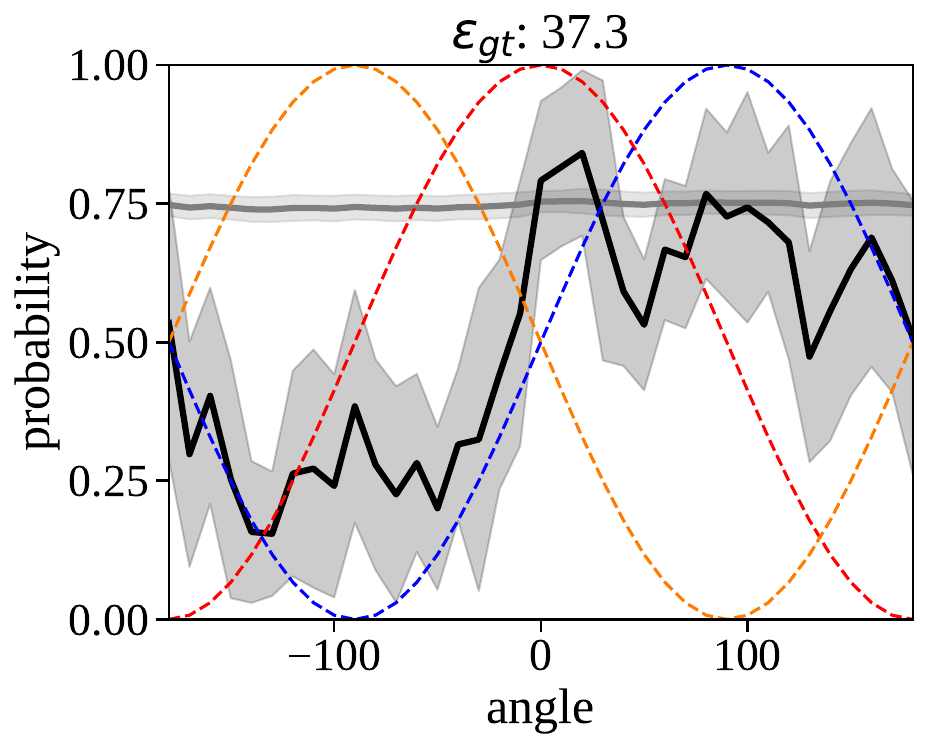} & \includegraphics[width=0.2\linewidth]{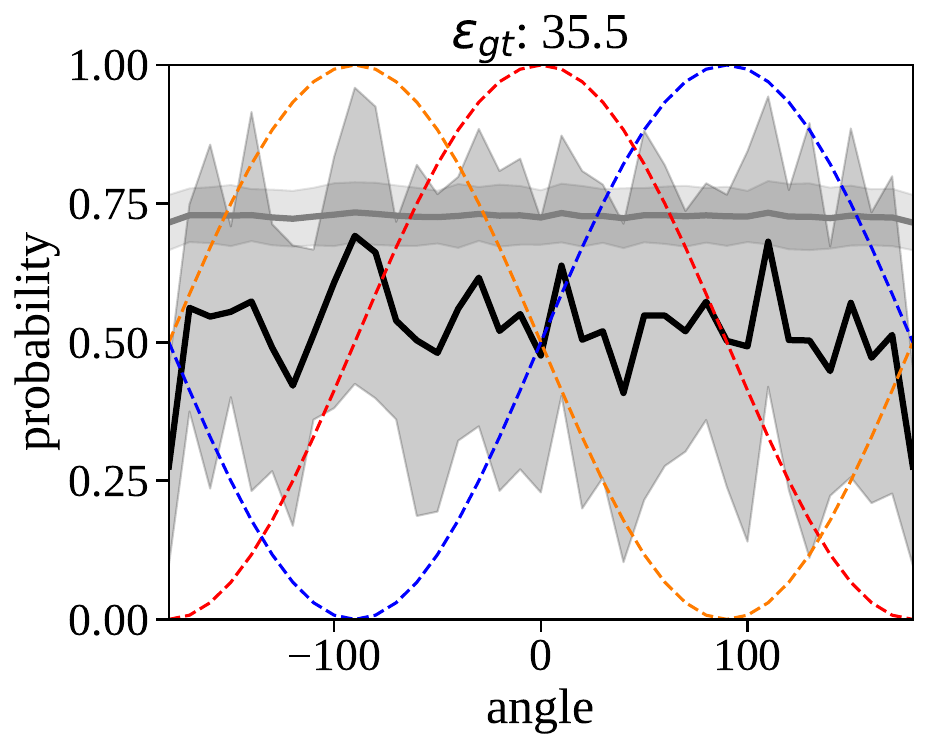} & \includegraphics[width=0.2\linewidth]{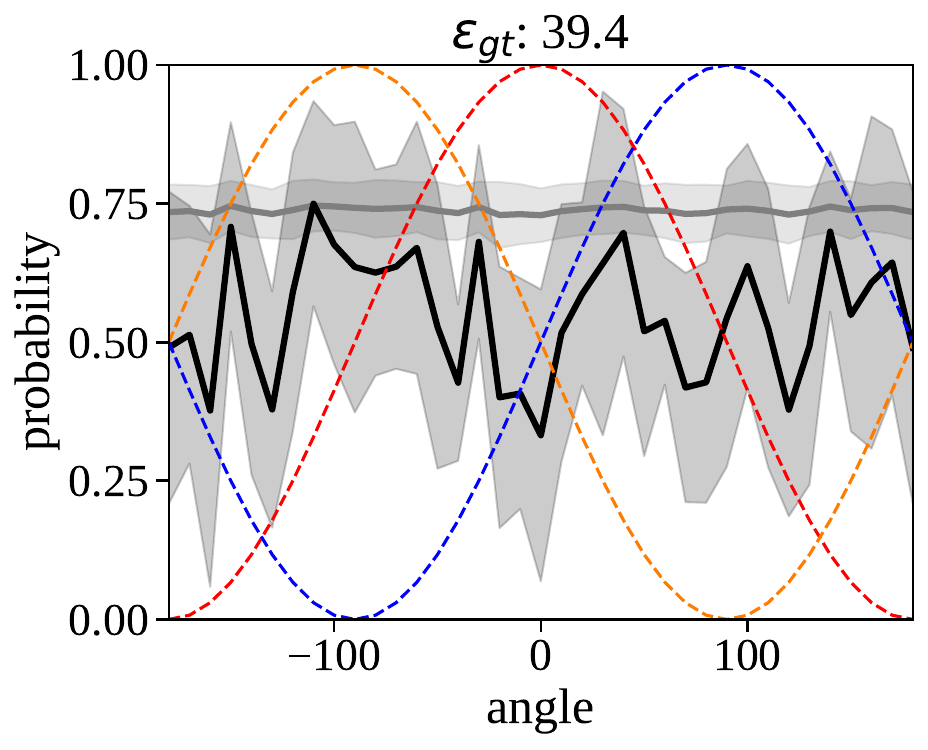} \\
\multirow{-8}{*}{\rotatebox[origin=c]{90}{\scriptsize MBLIPBLOOMZ7B}} & \includegraphics[width=0.2\linewidth]{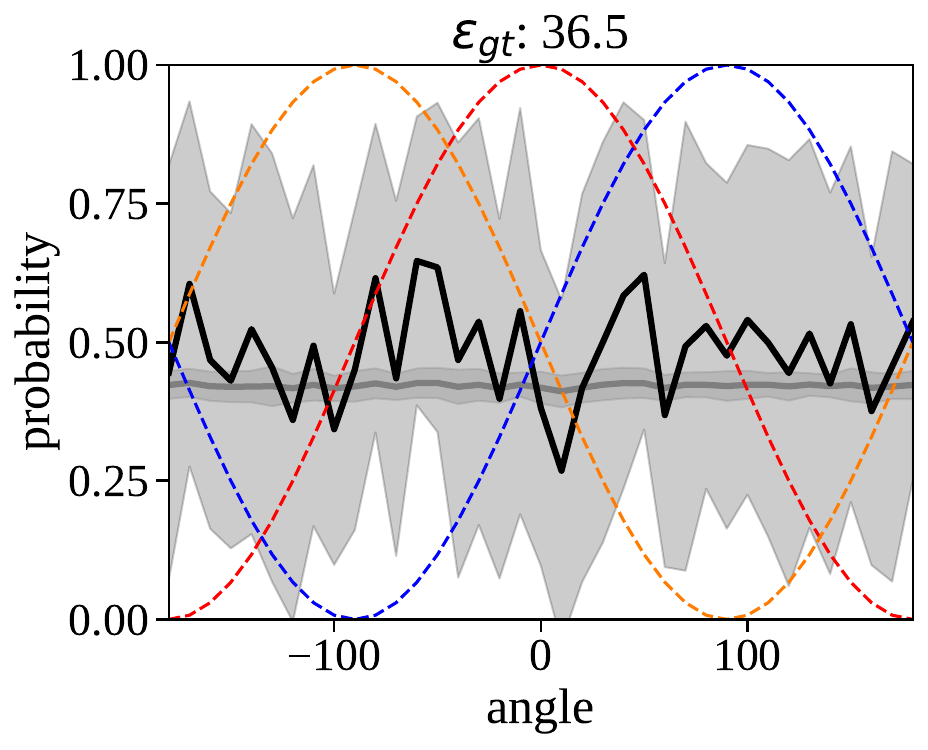} & \includegraphics[width=0.2\linewidth]{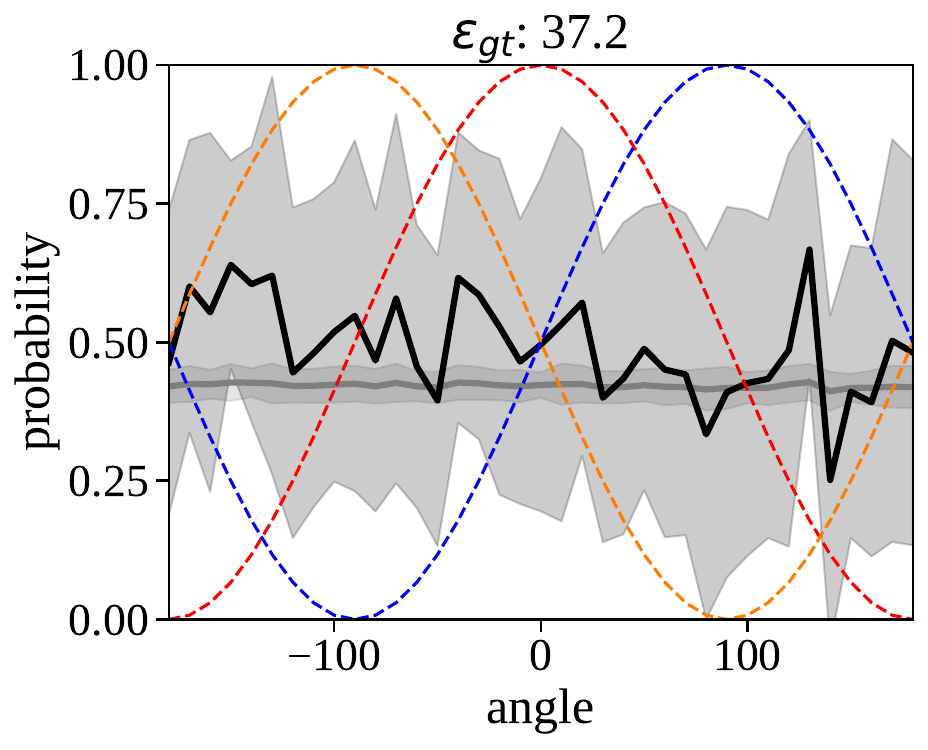} & \includegraphics[width=0.2\linewidth]{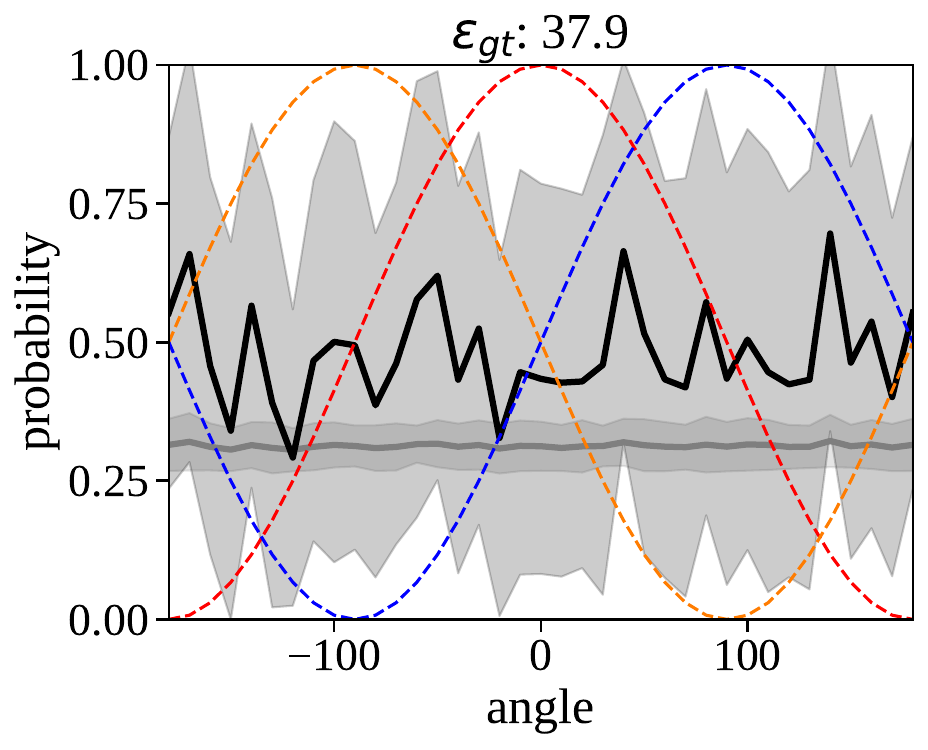} & \includegraphics[width=0.2\linewidth]{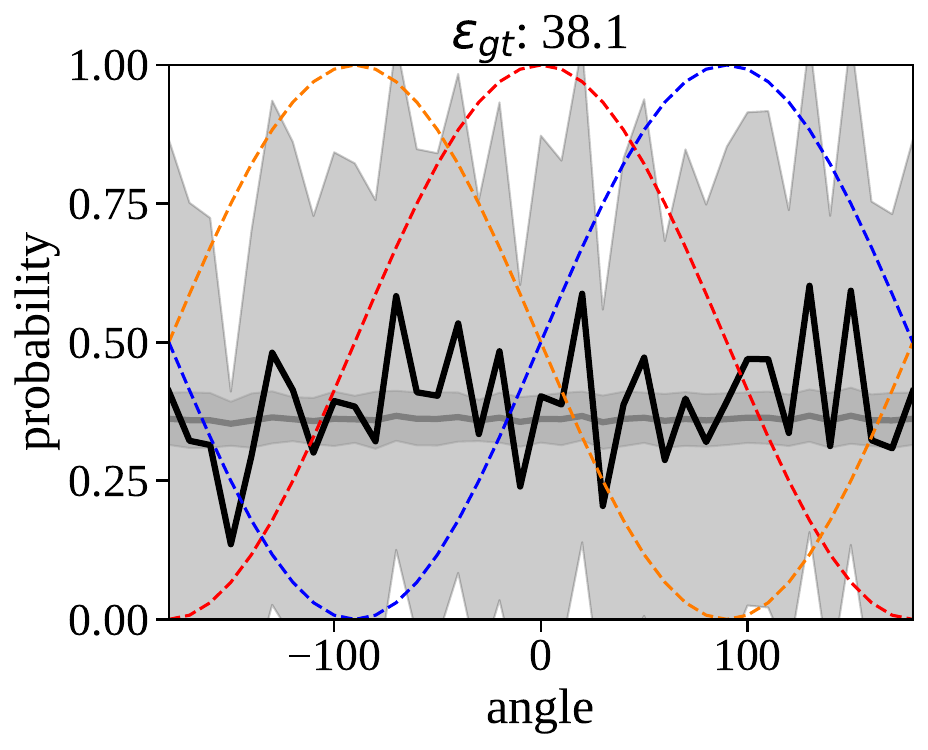} \\
\multirow{-8}{*}{\rotatebox[origin=c]{90}{\scriptsize GLaMM}} & \includegraphics[width=0.2\linewidth]{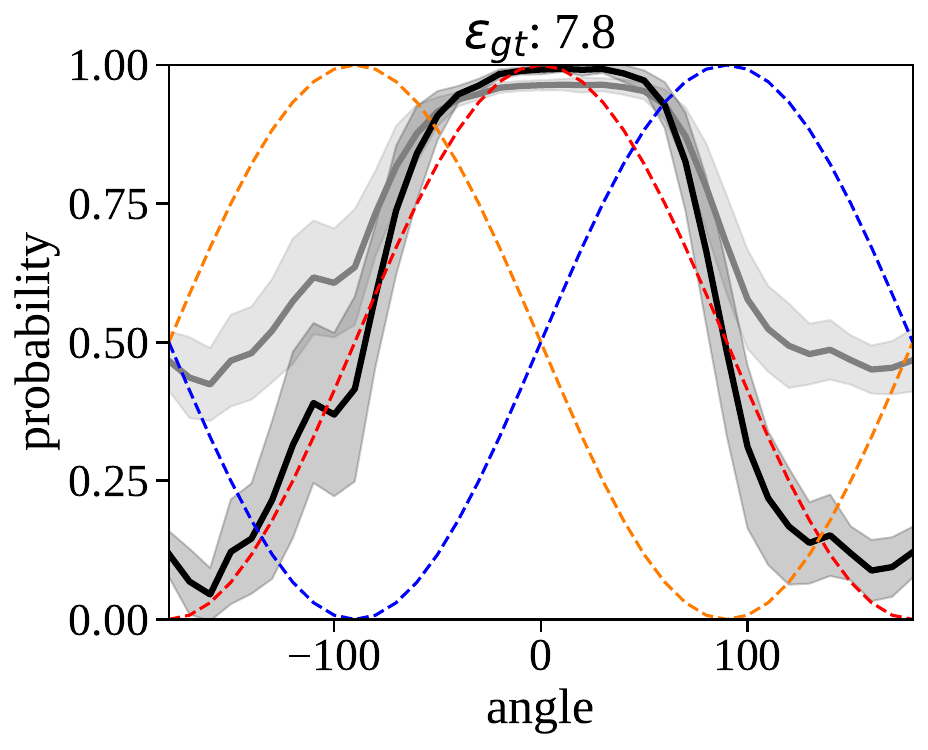} & \includegraphics[width=0.2\linewidth]{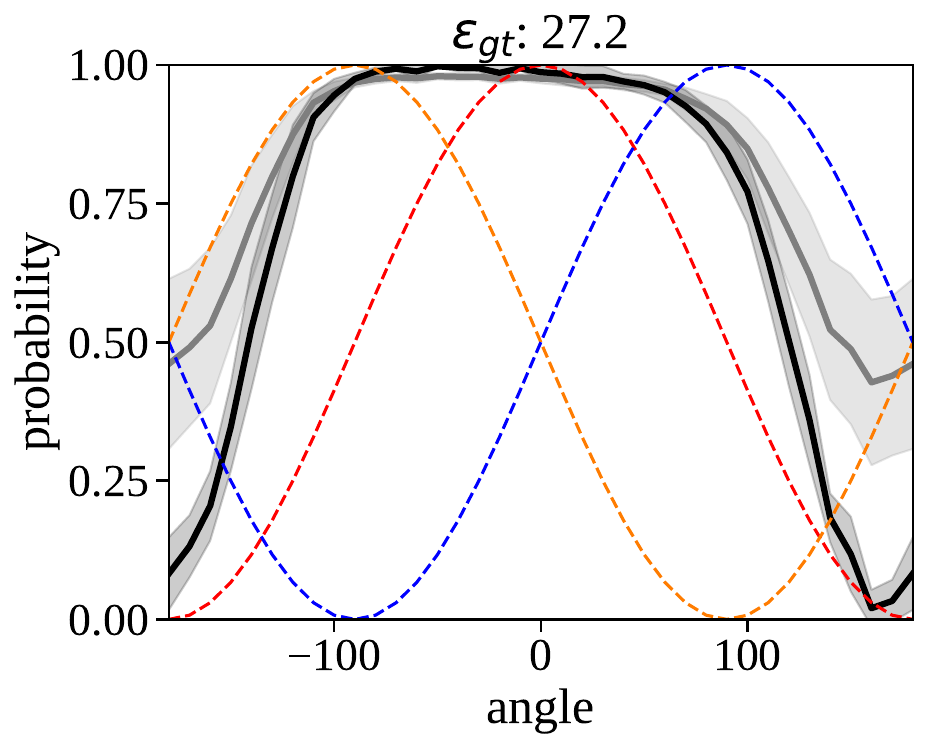} & \includegraphics[width=0.2\linewidth]{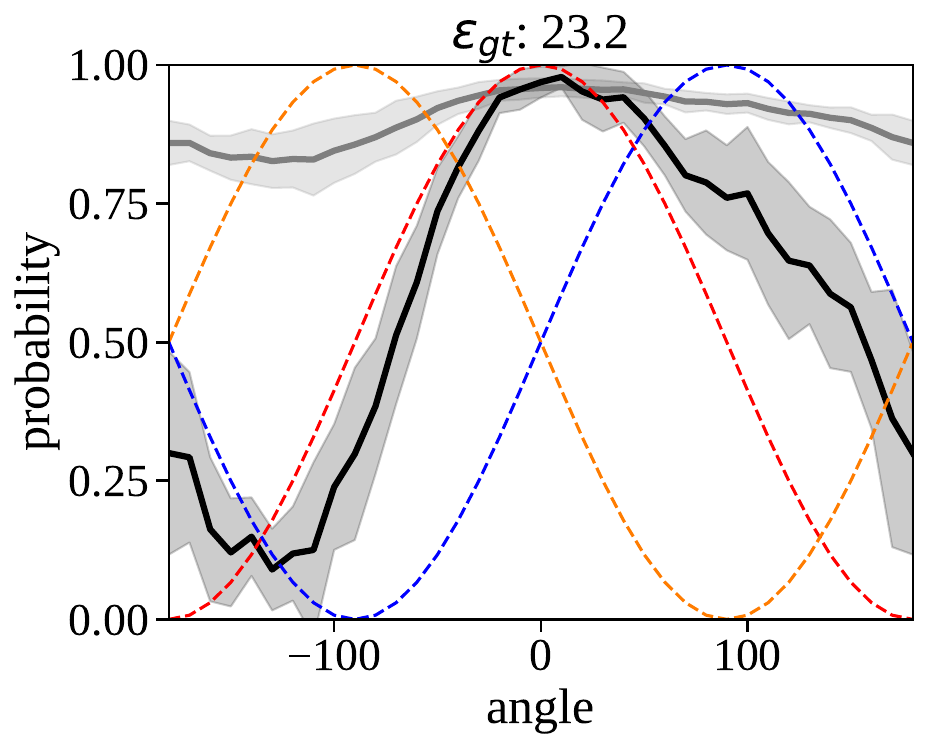} & \includegraphics[width=0.2\linewidth]{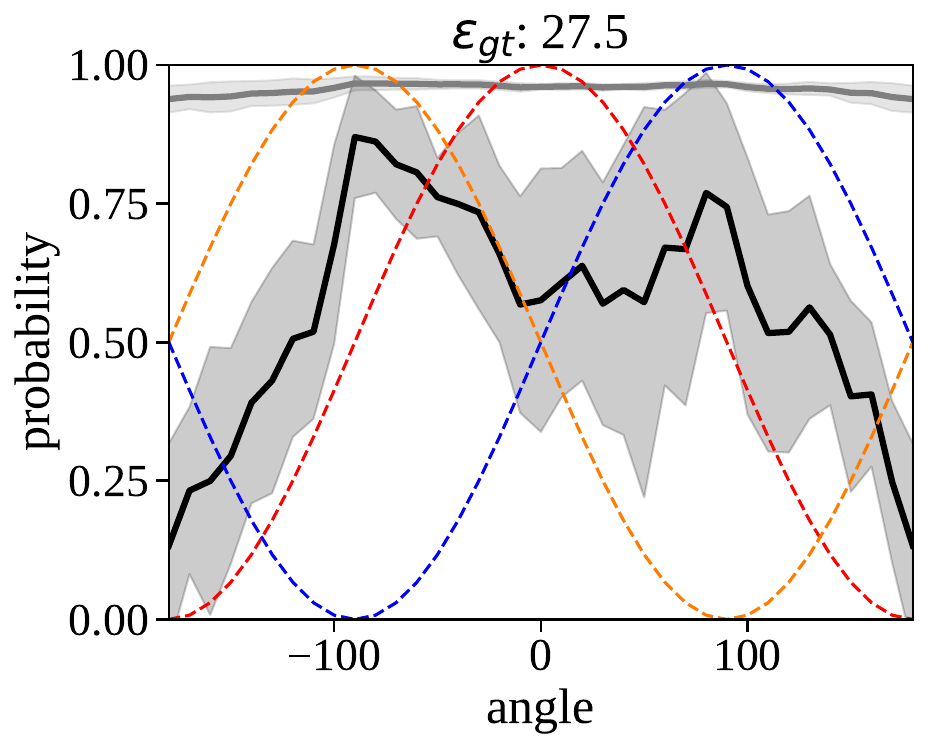} \\
\multirow{-8}{*}{\rotatebox[origin=c]{90}{\scriptsize LLaVA1.57B}} & \includegraphics[width=0.2\linewidth]{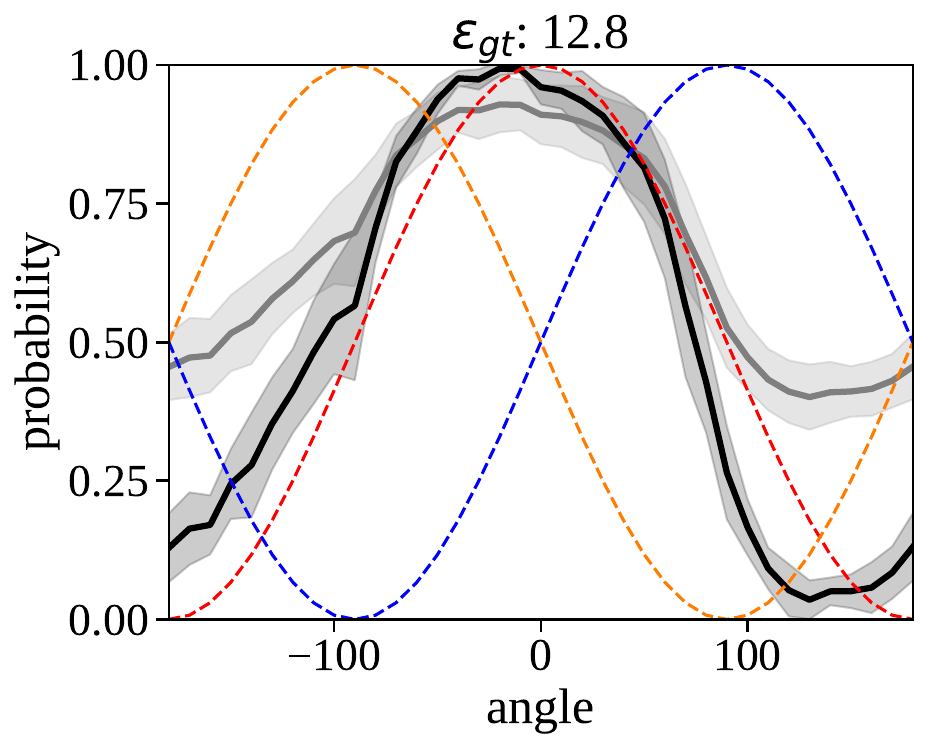} & \includegraphics[width=0.2\linewidth]{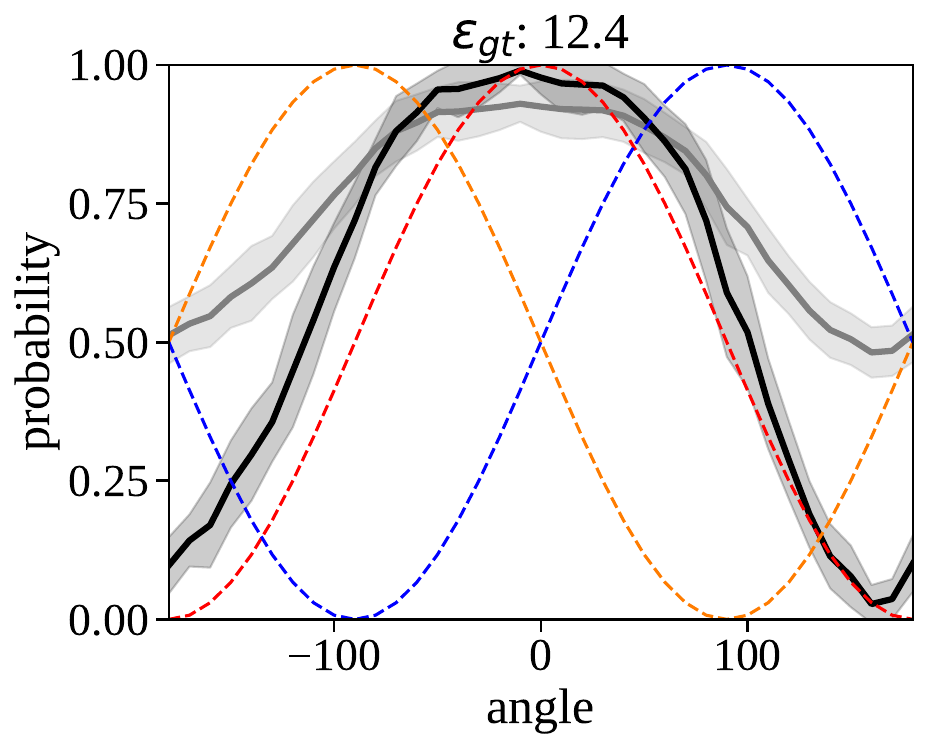} & \includegraphics[width=0.2\linewidth]{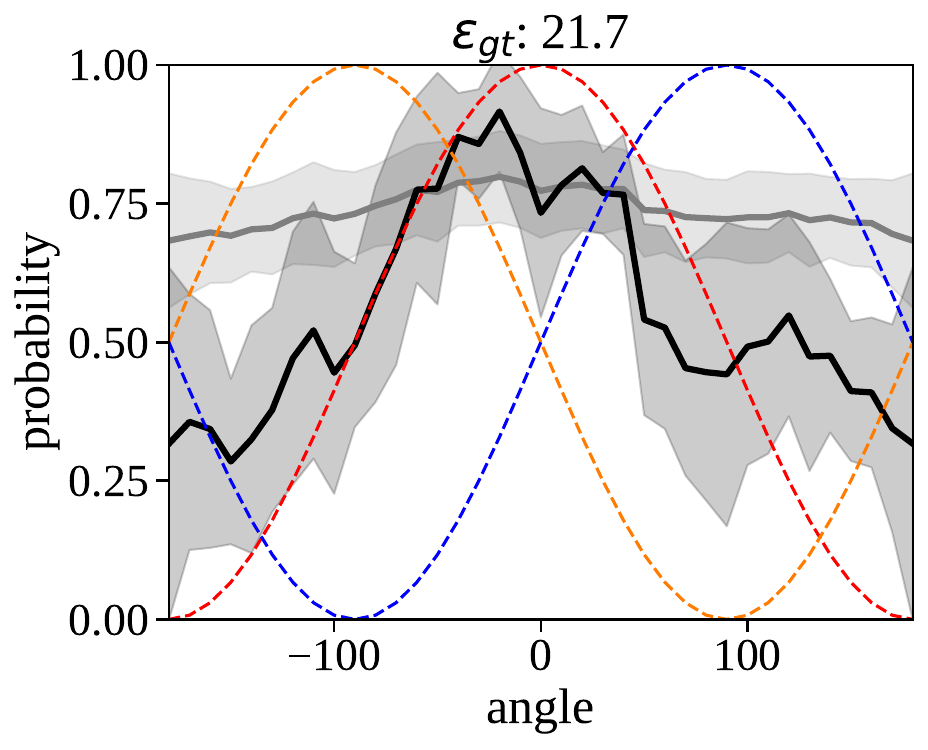} & \includegraphics[width=0.2\linewidth]{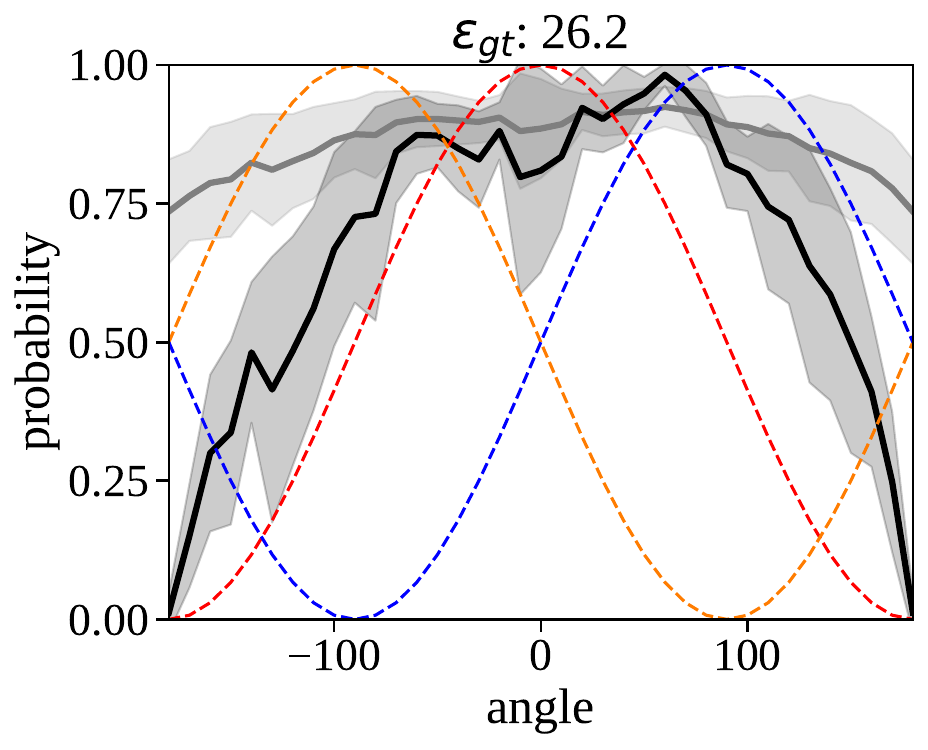} \\
\multirow{-8}{*}{\rotatebox[origin=c]{90}{\scriptsize LLaVA1.513B}} & \includegraphics[width=0.2\linewidth]{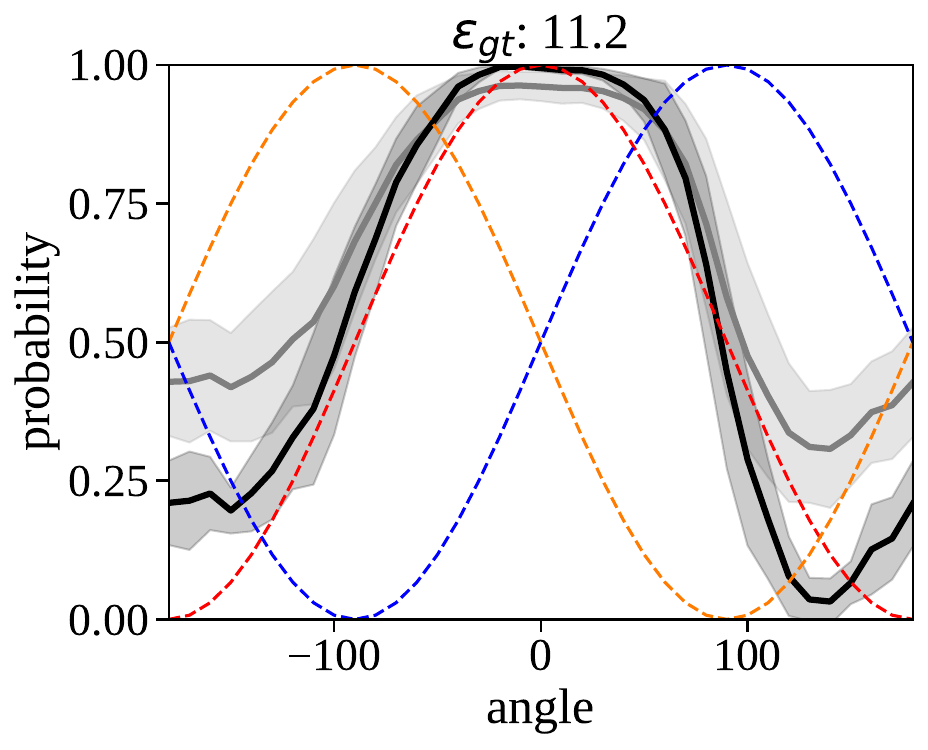} & \includegraphics[width=0.2\linewidth]{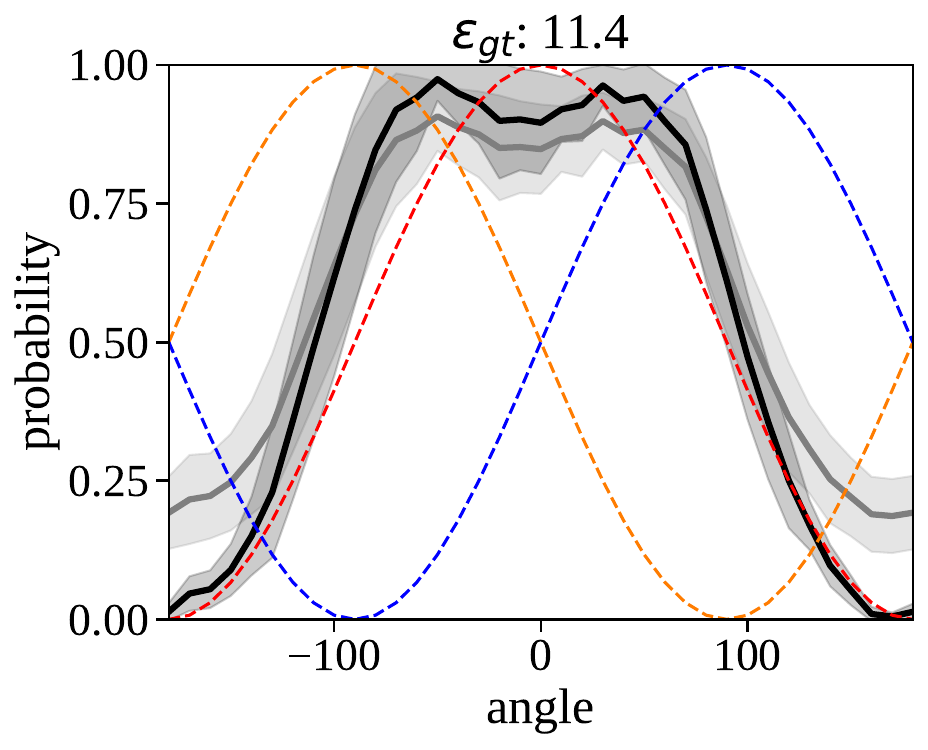} & \includegraphics[width=0.2\linewidth]{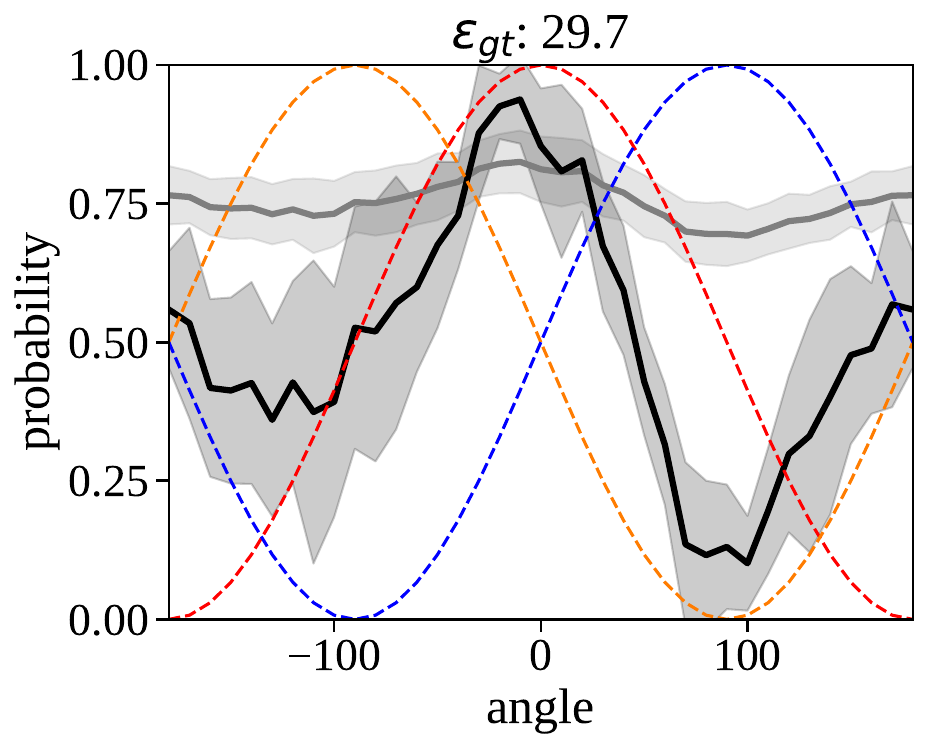} & \includegraphics[width=0.2\linewidth]{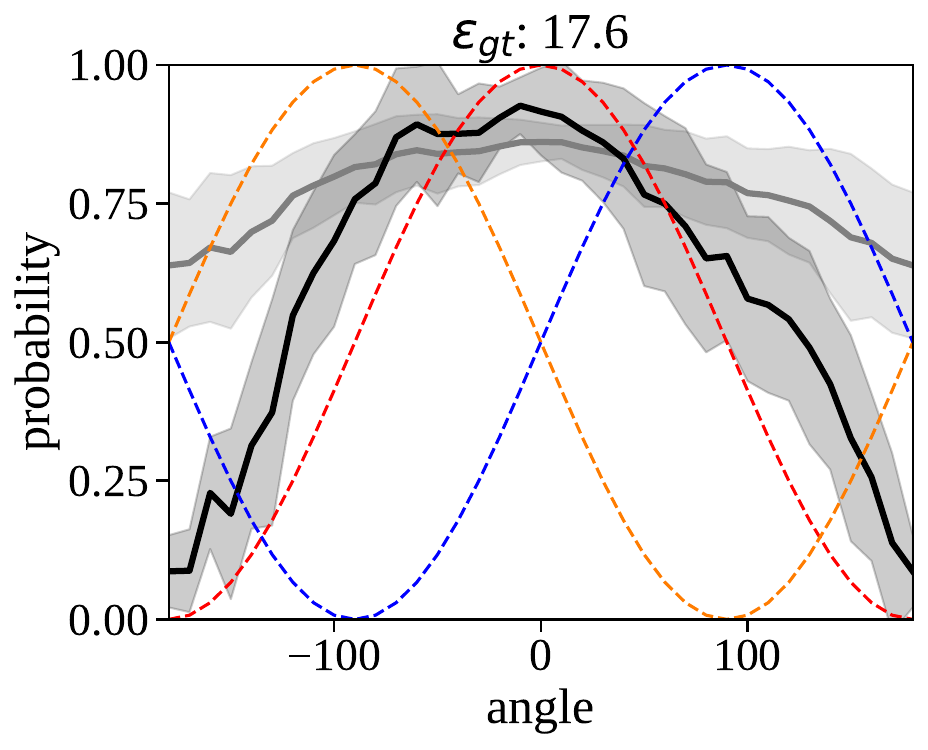} \\
\multirow{-8}{*}{\rotatebox[origin=c]{90}{\scriptsize XComposer2}} & \includegraphics[width=0.2\linewidth]{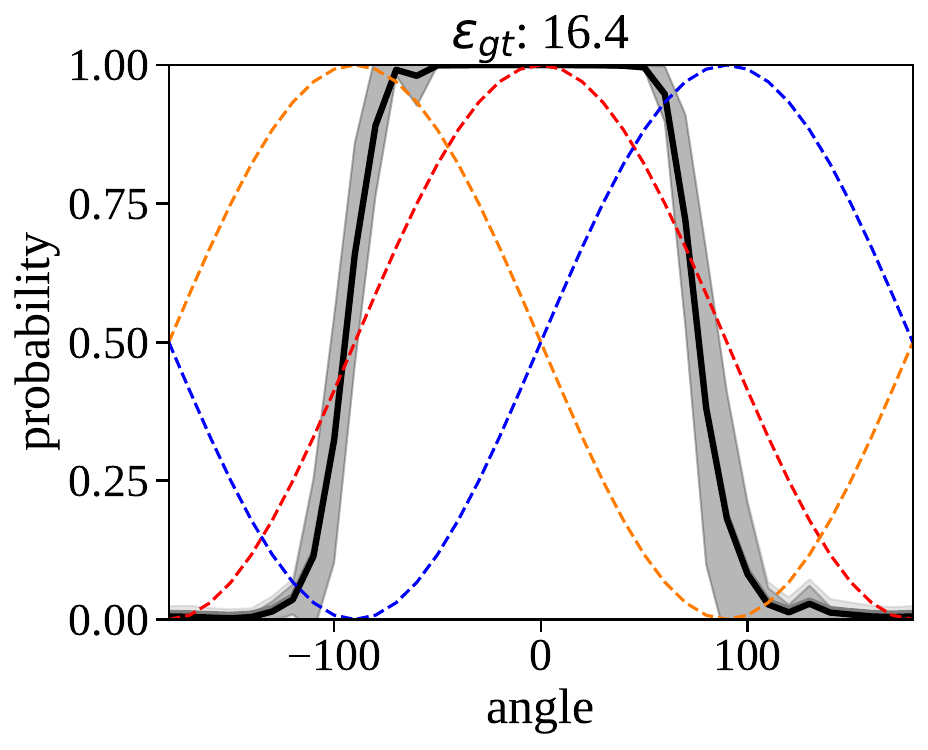} & \includegraphics[width=0.2\linewidth]{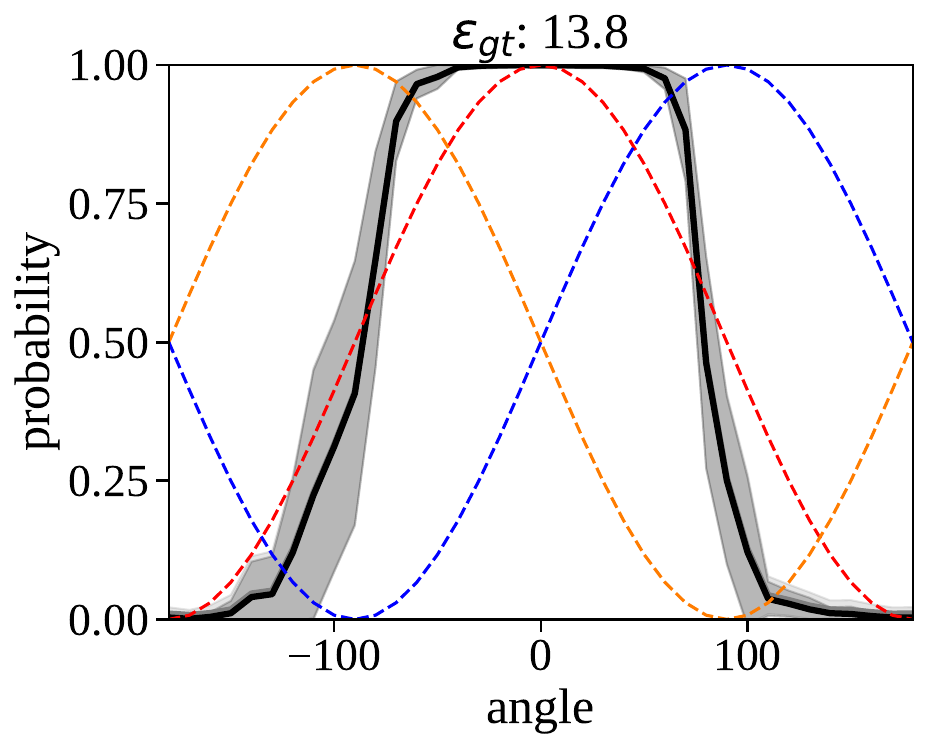} & \includegraphics[width=0.2\linewidth]{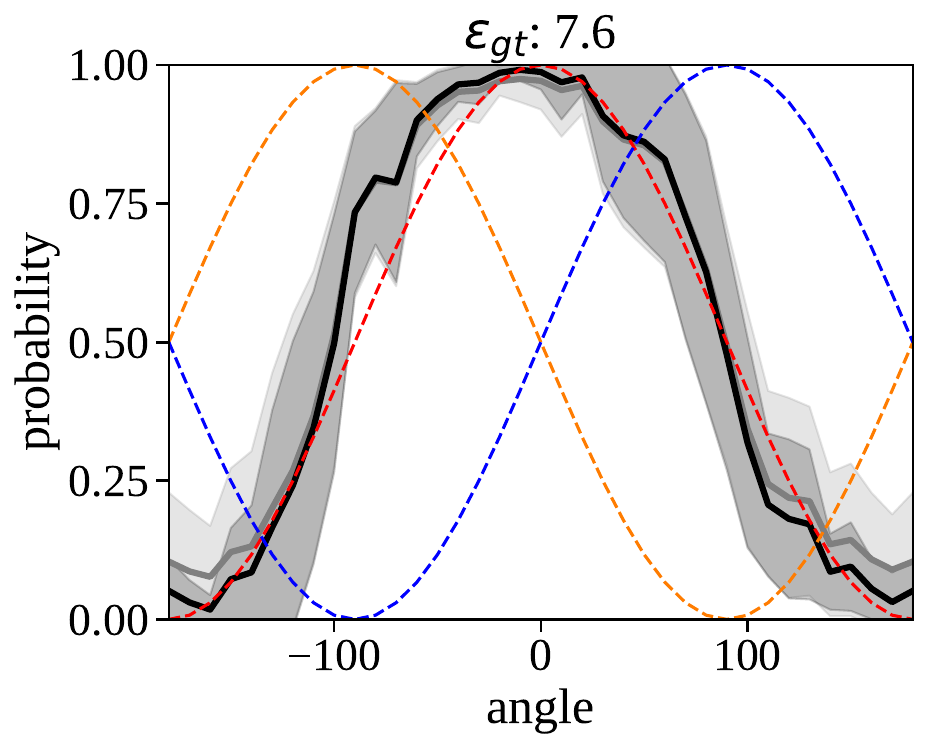} & \includegraphics[width=0.2\linewidth]{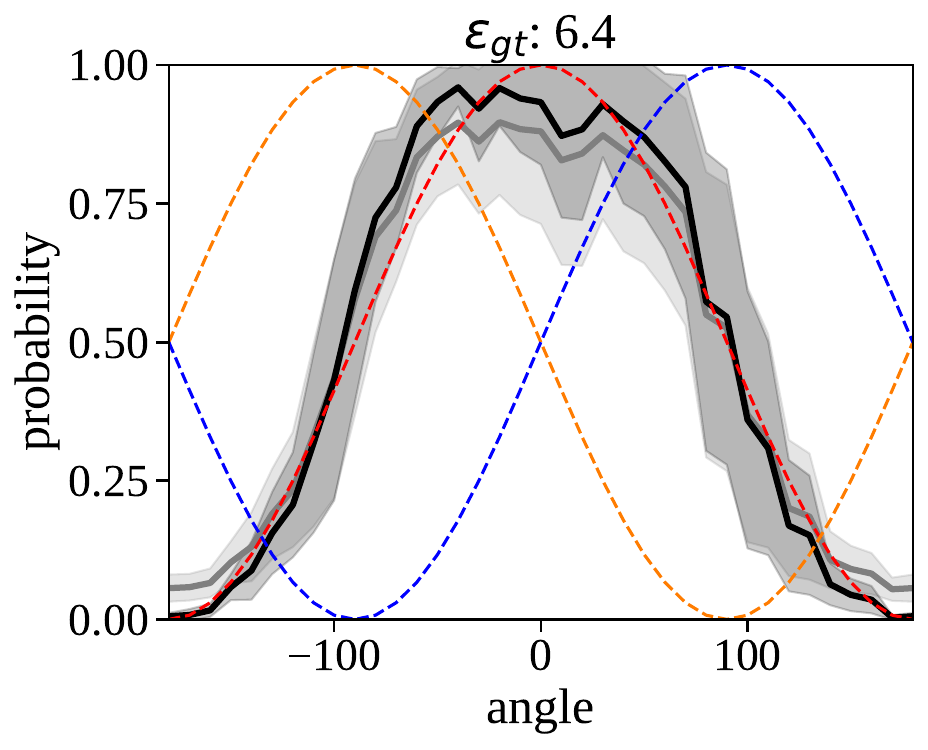} \\
\multirow{-8}{*}{\rotatebox[origin=c]{90}{\scriptsize MiniCPM}} & \includegraphics[width=0.2\linewidth]{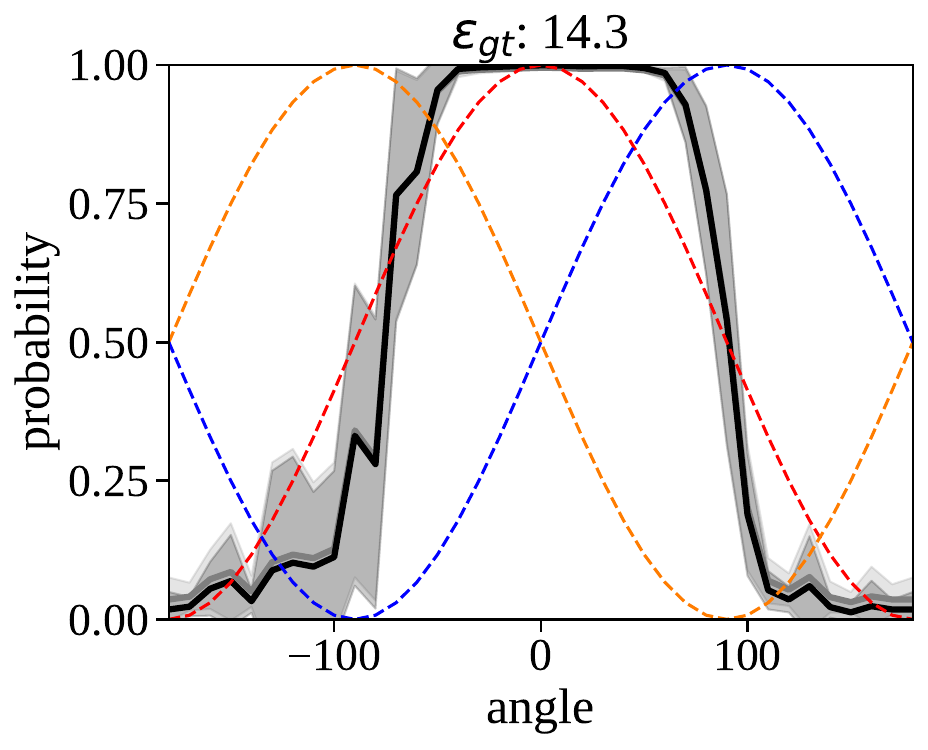} & \includegraphics[width=0.2\linewidth]{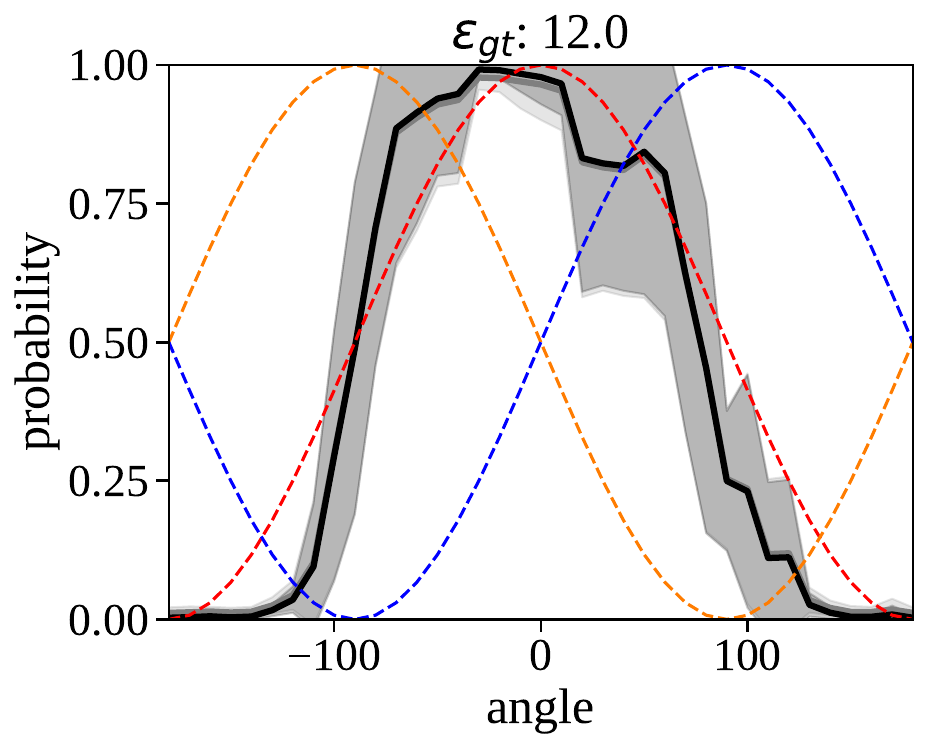} & \includegraphics[width=0.2\linewidth]{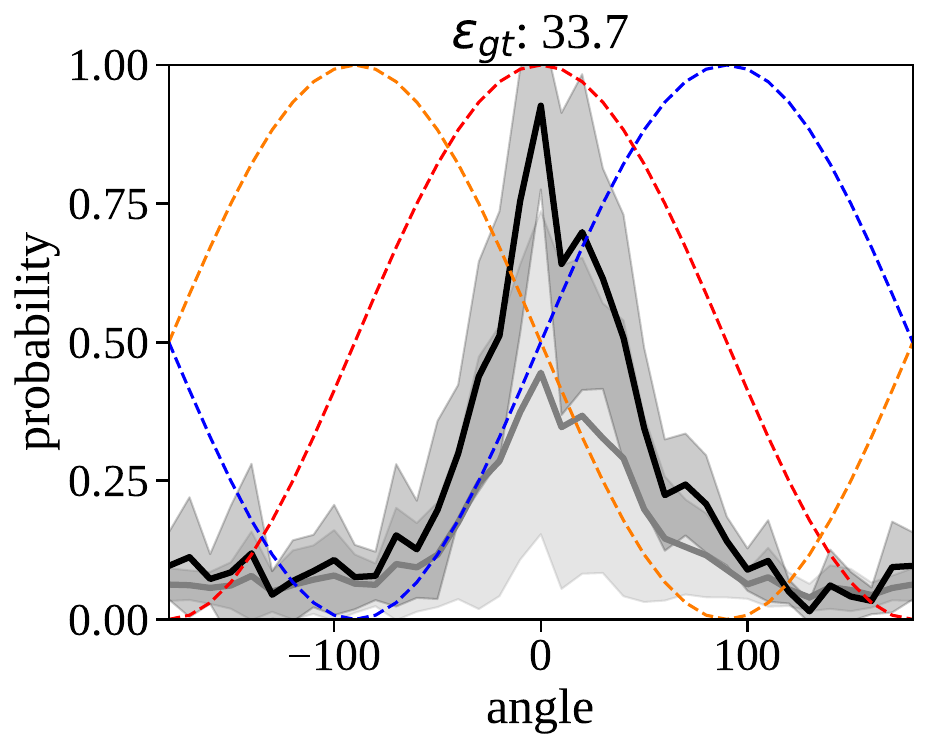} & \includegraphics[width=0.2\linewidth]{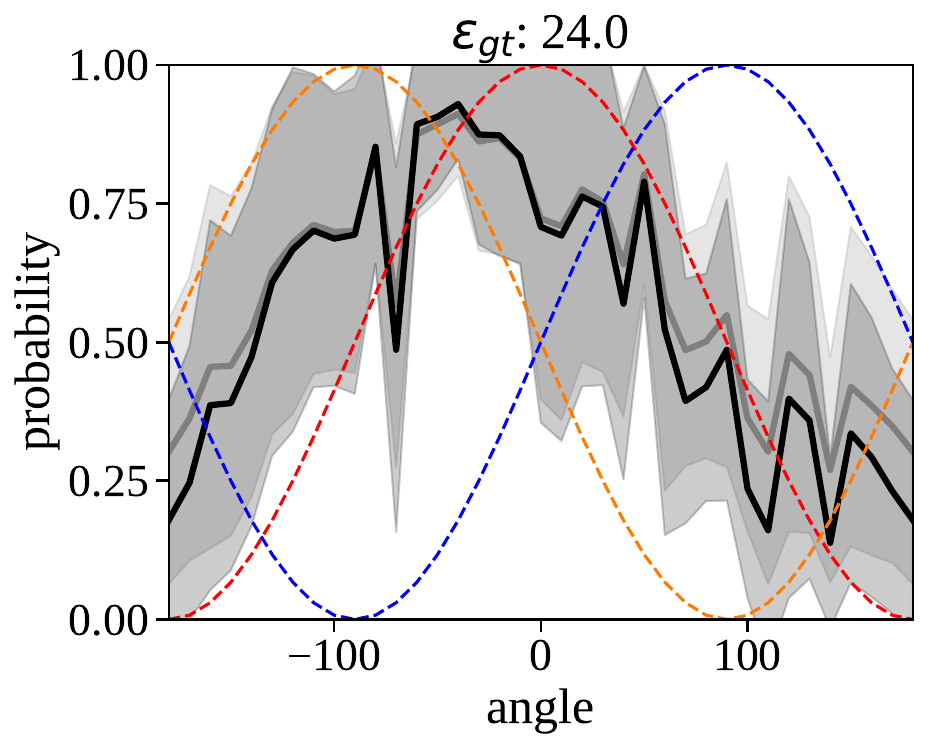} \\
\multirow{-8}{*}{\rotatebox[origin=c]{90}{\scriptsize GPT4o}} & \includegraphics[width=0.2\linewidth]{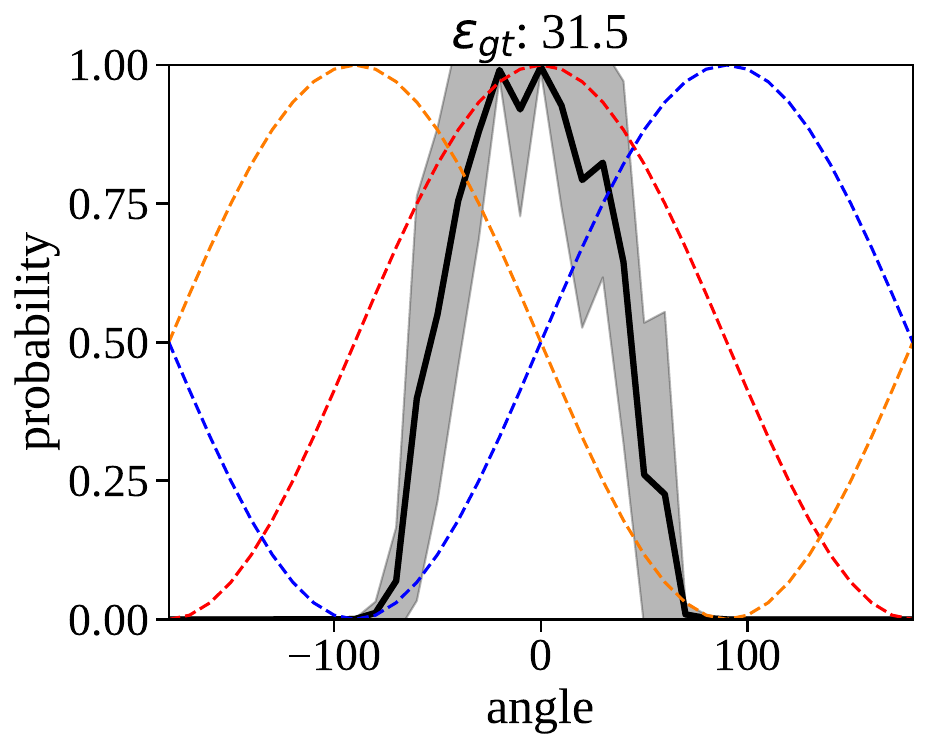} & \includegraphics[width=0.2\linewidth]{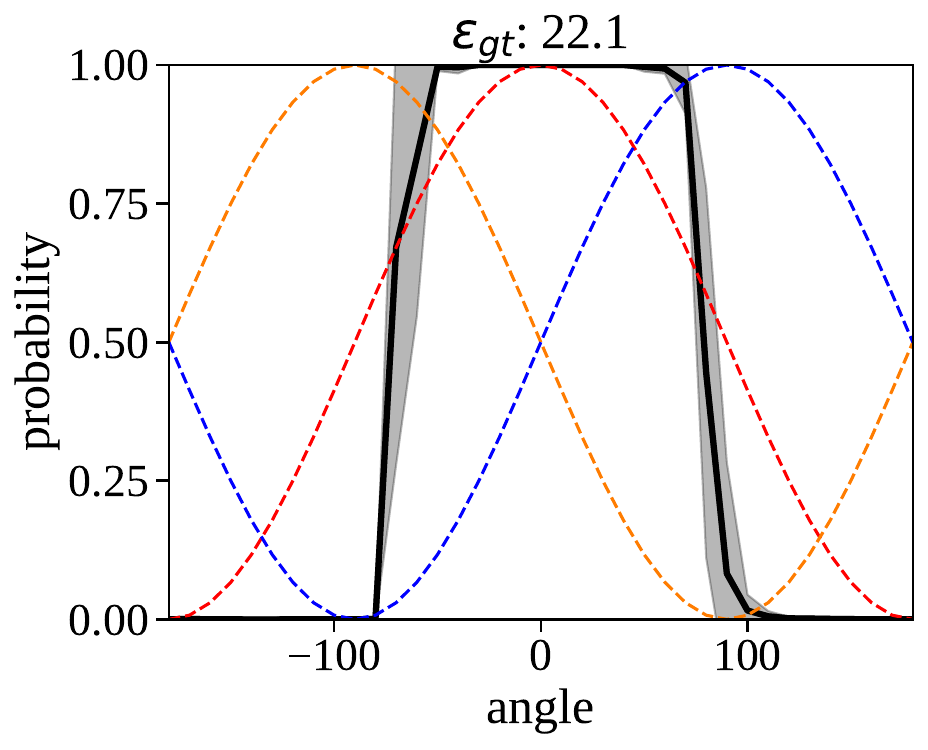} & \includegraphics[width=0.2\linewidth]{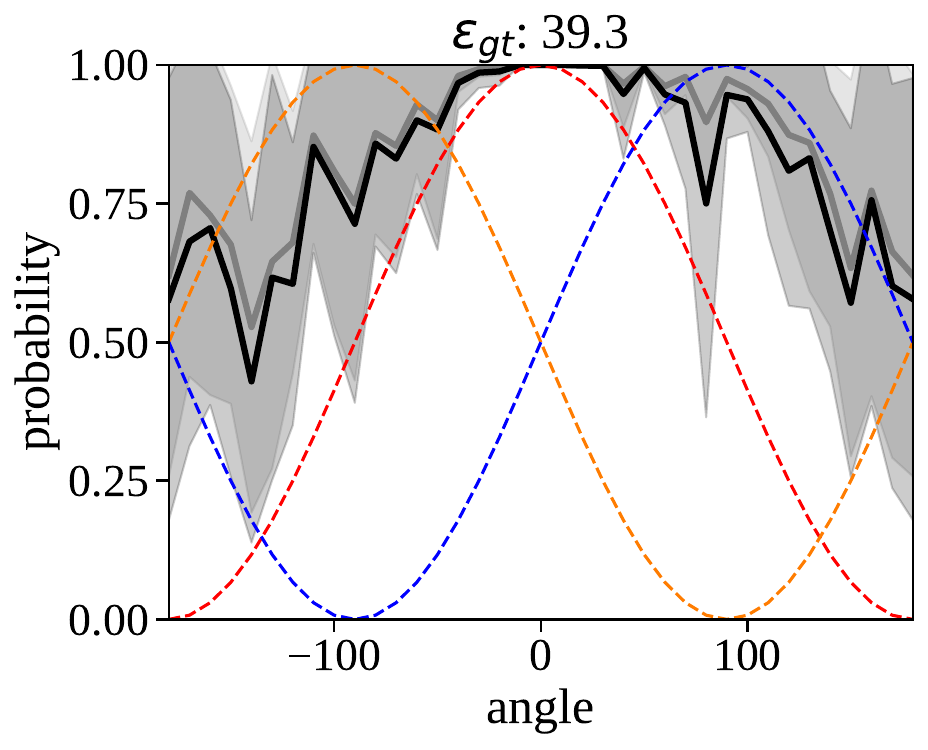} & \includegraphics[width=0.2\linewidth]{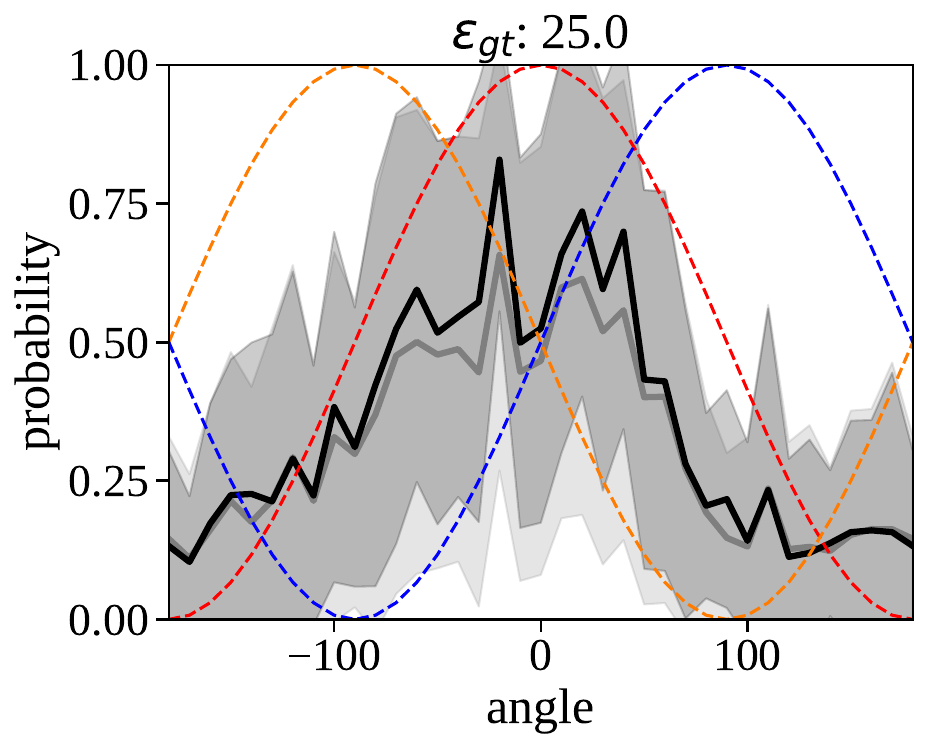} \\
  \end{tabular}
 }
\vspace*{-5pt}
\caption{All prediction plots for each model on \texttt{COMFORT-CAR} without perspective prompt (\texttt{nop}). The raw probability $p(\theta)$ in gray, normalized probability $\widehat{p}(\theta)$ in black, and the reference probabilities $p_\textrm{cos}(\theta)$ of \texttt{cam} in red, \texttt{add} in orange, \texttt{rel} in blue. To avoid overlapping reference probabilities of \texttt{add} and \texttt{rel}, we use plots on \texttt{COMFORT-CAR} with relatum facing left for left and right relations and \texttt{COMFORT-CAR} with relatum facing right for front and behind relations.}
\label{fig:car-nop-all}
\end{figure*}

\end{document}